\documentclass[10pt,twocolumn,letterpaper]{article}

\usepackage[pagenumbers]{cvpr} 

\usepackage{wrapfig}
\usepackage{overpic}
\usepackage{enumitem} 
\usepackage{color}
\usepackage{comment}
\usepackage{booktabs} 
\usepackage[accsupp]{axessibility} 

\usepackage[ruled,linesnumbered]{algorithm2e}
\usepackage{amsfonts}
\usepackage{threeparttable}
\usepackage{tabularx}
\usepackage{mathtools}

\usepackage{caption}
\usepackage{subcaption}
\usepackage{multirow}
\usepackage{algpseudocode}
\usepackage{soul}
\usepackage{multirow}
\usepackage{arydshln} 
\usepackage{stmaryrd}
\usepackage{makecell}
\usepackage{tabularx}
\newcolumntype{L}{>{\raggedright\arraybackslash}X}
\usepackage[hang,flushmargin]{footmisc}

\usepackage{amsmath,amsfonts,bm,mathtools}

\def\eqref#1{equation~\ref{#1}}

\def\1{\bm{1}}

\newcommand{\bR}{\mathbf{R}}

\newcommand{\by}{\mathbf{y}}
\newcommand{\bz}{\mathbf{z}}

\newcommand{\btheta}{{\boldsymbol{\theta}}}

\newcommand{\bpsi}{{\boldsymbol{\psi}}}

\DeclareMathAlphabet{\mathsfit}{\encodingdefault}{\sfdefault}{m}{sl}
\SetMathAlphabet{\mathsfit}{bold}{\encodingdefault}{\sfdefault}{bx}{n}

\newcommand{\R}{\mathbb{R}}  

\newcommand{\z}{\mathbf{z}}

\definecolor{turquoise}{cmyk}{0.65,0,0.1,0.3}
\definecolor{purple}{rgb}{0.65,0,0.65}
\definecolor{dark_green}{rgb}{0, 0.5, 0}
\definecolor{orange}{rgb}{0.8, 0.6, 0.2}
\definecolor{red}{rgb}{0.9, 0.1, 0.1}
\definecolor{darkred}{rgb}{0.6, 0.1, 0.05}
\definecolor{blueish}{rgb}{0.0, 0.3, .6}
\definecolor{light_gray}{rgb}{0.7, 0.7, .7}
\definecolor{pink}{rgb}{1, 0, 1}
\definecolor{greyblue}{rgb}{0.25, 0.25, 1}

\newcommand{\methodname}{NIVeL\xspace}

\newcommand{\bc}{\mathbf{c}}

\newcommand{\bp}{\mathbf{p}}

\newcommand{\bs}{\mathbf{s}}

\def\R{{\cal R}}

\newcommand\blfootnote[1]{%
  \begingroup
  \renewcommand\thefootnote{}\footnote{#1}%
  \addtocounter{footnote}{-1}%
  \endgroup
}

\newcommand*{\dictchar}[1]{
    \twocolumn[
    \centerline{\parbox[c][3cm][c]{2cm}{%
            \fontsize{16}{16}
            \selectfont
            \textbf{{#1}}}}]
}

\usepackage{blindtext}

\ifpdf

\else
\fi

\ifpdf

\else
\fi

\ifpdf

\else
\fi

\usepackage{color}
\definecolor{cyan}{cmyk}{1,0,0,0}
\definecolor{darkgreen}{rgb}{0,0.5,0}
\definecolor{orange}{rgb}{1,0.5,0}
\definecolor{magenta}{cmyk}{0,1,0,0}
\definecolor{darkyellow}{cmyk}{0,0,0.75,0}
\definecolor{gray}{rgb}{0.8,0.8,0.8}
\definecolor{good}{rgb}{0.75,0.9,0.75}
\definecolor{decent}{rgb}{0.9,0.93,0.75}
\definecolor{bad}{rgb}{0.9,0.75,0.75}
\definecolor{na}{rgb}{0.8,0.8,0.8}
\definecolor{textprompts}{rgb}{0.0,0.705,0.313}

\definecolor{fontgreen}{RGB}{0,190,0}
\definecolor{glyphblue}{RGB}{0,0,220}

\usepackage{graphicx}
\usepackage{amsmath}
\usepackage{amssymb}
\usepackage{booktabs}

\definecolor{cvprblue}{rgb}{0.21,0.49,0.74}
\usepackage[pagebackref,breaklinks,colorlinks,citecolor=cvprblue]{hyperref}

\usepackage[capitalize]{cleveref}
\crefname{section}{Sec.}{Secs.}
\Crefname{section}{Section}{Sections}
\Crefname{table}{Table}{Tables}
\crefname{table}{Tab.}{Tabs.}

\begin{document}

\title{\methodname: Neural Implicit Vector Layers for Text-to-Vector Generation}

\author{
Vikas Thamizharasan$^{*1,2}$~~~ 
Difan Liu$^2$~~~
Matthew Fisher$^{2}$~~~\\
Nanxuan Zhao$^2$~~~
Evangelos Kalogerakis$^1$~~~
Michal Lukáč$^2$~~~
\vspace{0.3cm} \\
{\small
University of Massachusetts Amherst$^1$~~~
Adobe Research$^2$~~~
}
}

\twocolumn[{
\vspace{-1.2cm}
\maketitle
\vspace{-1.2cm}

\begin{center}
    \includegraphics[width=1.0\linewidth]
    {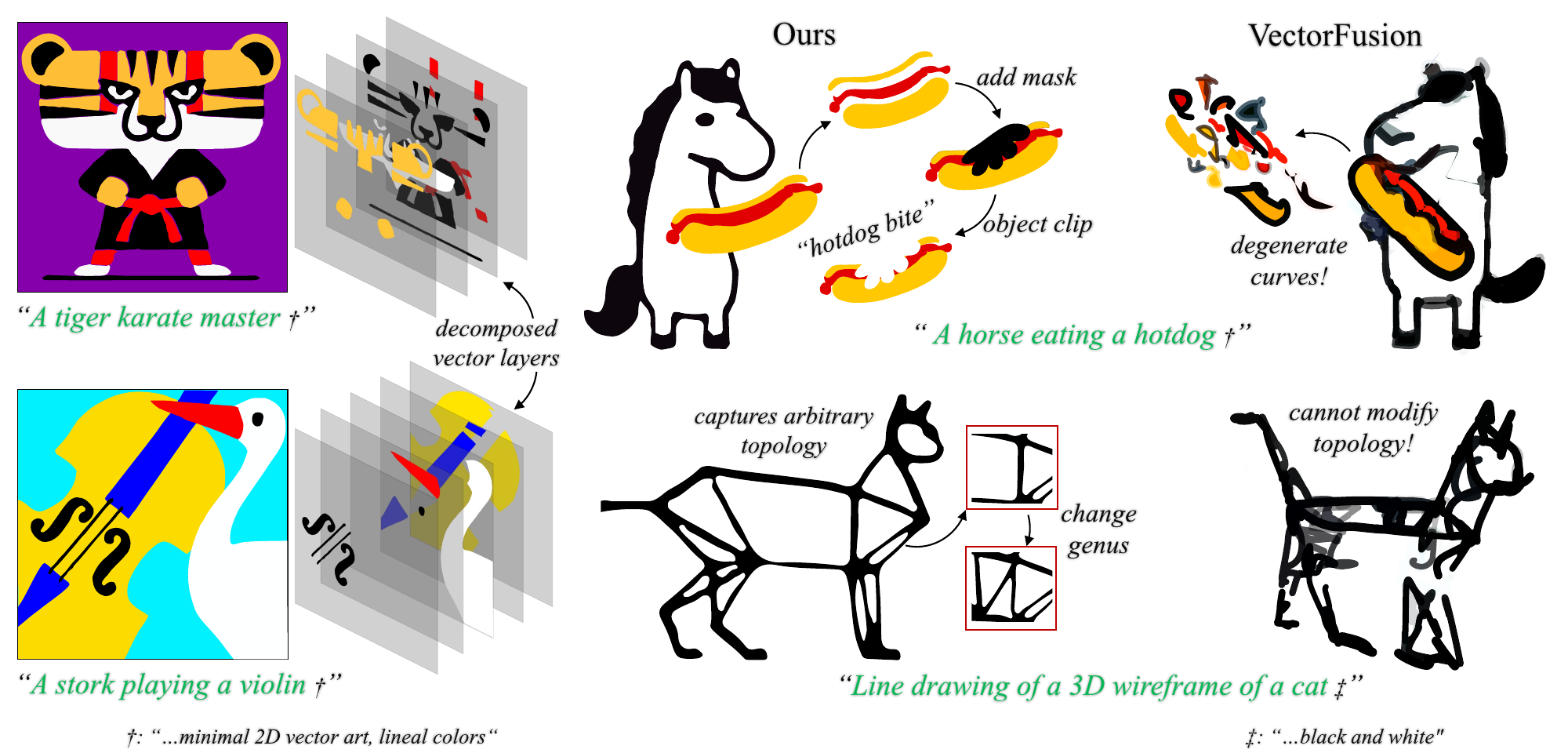}  
        \vspace{-4mm}
        \captionof{figure}{We present \methodname, a neural implicit vector layer representation for generative text-to-vector graphics. Given an input \textcolor{textprompts}{text prompt}, \methodname outputs resolution-independent  
        2D shapes and colors in editable layers. Our neural representation can capture plausible shapes of arbitrary topology and genus, which were challenges of previous text-to-vector works (VectorFusion\cite{jain2022vectorfusion}) that directly operated  on classical Bézier curves. Importantly, our results can easily be imported into traditional vector graphics software and make intuitive edits. 
        }
    \label{fig:teaser}  
\end{center}
}]


\section*{\centering Abstract}
\vspace{-0.1em}
\emph{The success of denoising diffusion models in representing rich data distributions over 2D raster images
has prompted research on extending them to other data representations, such as vector graphics. Unfortunately due to their variable structure and scarcity of vector training data, directly applying diffusion models on this domain remains a challenging problem. Using workarounds like optimization via Score Distillation Sampling (SDS) is also fraught with difficulty, as vector representations are non-trivial to directly optimize and tend to result in implausible geometries such as redundant or self-intersecting shapes. \methodname addresses these challenges by reinterpreting the problem on an alternative, intermediate domain which preserves the desirable properties of vector graphics -- mainly sparsity of representation and resolution-independence. This alternative domain is based on neural implicit fields expressed in a set of decomposable, editable layers. 
Based on our experiments, \methodname produces text-to-vector graphics results of significantly better quality than the state-of-the-art.}

\section{Introduction}
\label{sec:intro}
\blfootnote{
~$^*$Work was done during an internship at Adobe Research.
\\
~Project website: \textcolor{red}{https://vikastmz.github.io/NIVeL/}
}
Vector Graphics is a widely used 
representation to express visual concepts as a compact collection of primitives, such as Bézier curves, polygons, circles, lines and colors in a resolution-independent manner. 
Synthesizing vector graphics through generative models and  high-level guidance, such as text prompts, 
would be highly desirable for automating modeling pipelines. 

By leveraging massive scale, denoising diffusion has become the gold standard in generative raster imaging. 
In vector graphics however, no equivalent exists. The variable structure of the vector representation (e.g., varying number and types of primitives)
means that we could only apply diffusion to a fixed subset of this domain at a time (i.e. the subdomain of vector graphics with one particular structure); and if we did so, we would not be able to find training data at requisite scale.

\begin{figure}[t]
    \centering
    \includegraphics[width=\linewidth]{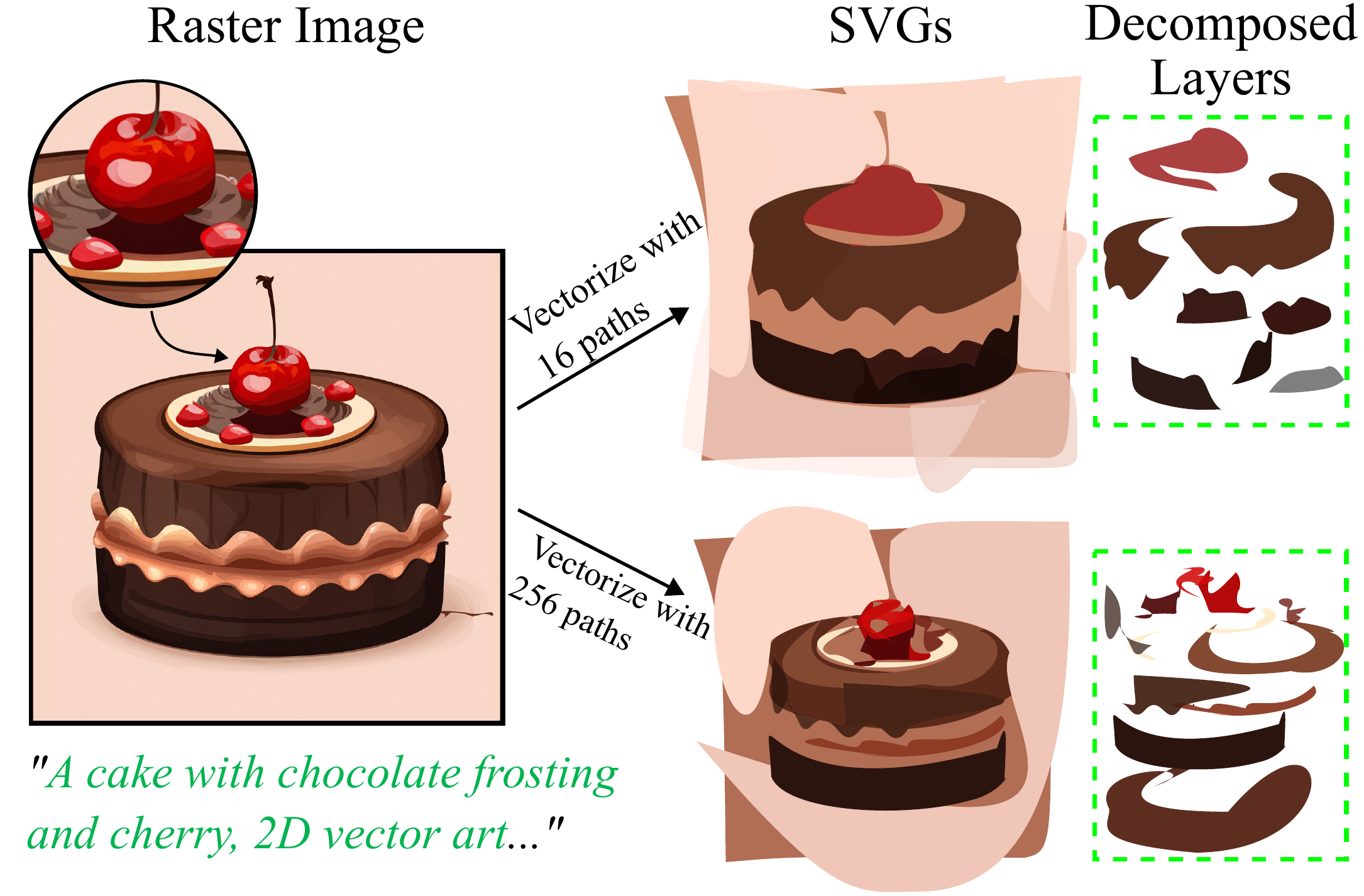}
    \vspace{-2mm}
    \caption{Sampling raster images from diffusion models, then applying vectorization leads to implausible geometries, redundant curves, and semantically meaningless layers. Here we show the results of sampling a text-to-image diffusion model DeepFloyd \cite{deepfloyd}, then applying LIVE \cite{Du:2023:IVE}, a vectorizer that produces layer-decomposed SVGs. The sampled raster images often contain complex signals that are difficult to vectorize and interpet.} 
    \label{fig:ddpm_vectorizer}
\end{figure}

As a result, to enable generative vector applications, such as text-to-vector, one 
may turn to various workarounds, such as ex post facto vectorization of raster diffusion outputs. This is not optimal because these generated samples are not close to vector graphics; they typically contain shapes and appearance that cannot be modelled with classical vector graphic primitives without unnecessarily increasing complexity and largely degrading quality
(Figure \ref{fig:ddpm_vectorizer}).
Even if we accept this cost, optimal vectorization still requires a human to exhaustively search over various settings and hyper-parameters present in these systems.

Score Distillation Sampling \cite{poole2022dreamfusion} offers an alternative procedure, where a raster diffusion model can guide the optimization of a vector graphics representation, as proposed  in VectorFusion \cite{jain2022vectorfusion}. 
But this quickly offers a different problem; in traditional vector representations, the structure of the representation depends on the content of the graphic. These changes in content are discrete, even changing the number and meaning of parameters being optimized, and so cannot be modeled by smooth optimization like SDS. Additionally, optimizing traditional vector representations on the image domain is often not stable, easily leading to degenerate geometries (Figure \ref{fig:teaser}).

In this paper, we propose \methodname, which makes use of an intermediate, vector-like representation as the optimization domain for Score Distillation Sampling. Derived from neural implicit functions, this representation maintains the resolution independence and simplicity of primitives we expect from vector graphics, while allowing for smooth arbitrary changes of shape topology without discrete jumps. Additionally, we show how to modify the SDS pipeline to reliably produce stable vector-like results in this representation, freeing us from the difficult problem of vectorizing arbitrary diffusion outputs.

\paragraph{Contributions.}
We introduce the following contributions:
\begin{itemize}
    \item A new neural vector graphic representation based on a decomposable set of implicit fields, which  captures arbitrary topology and is easier to optimize.
    \item A generative text-to-vector model that produces editable, layer-decomposed shapes, without any explicit  prior, or supervision.
    \item Significant improvement in vector quality outputs over the current state-of-the-art (VectorFusion \cite{jain2022vectorfusion}). 
\end{itemize}

\section{Related Work}
\label{sec:relwork}

\begin{figure*}[t!]
\begin{center}
\includegraphics[width=1.0\linewidth]{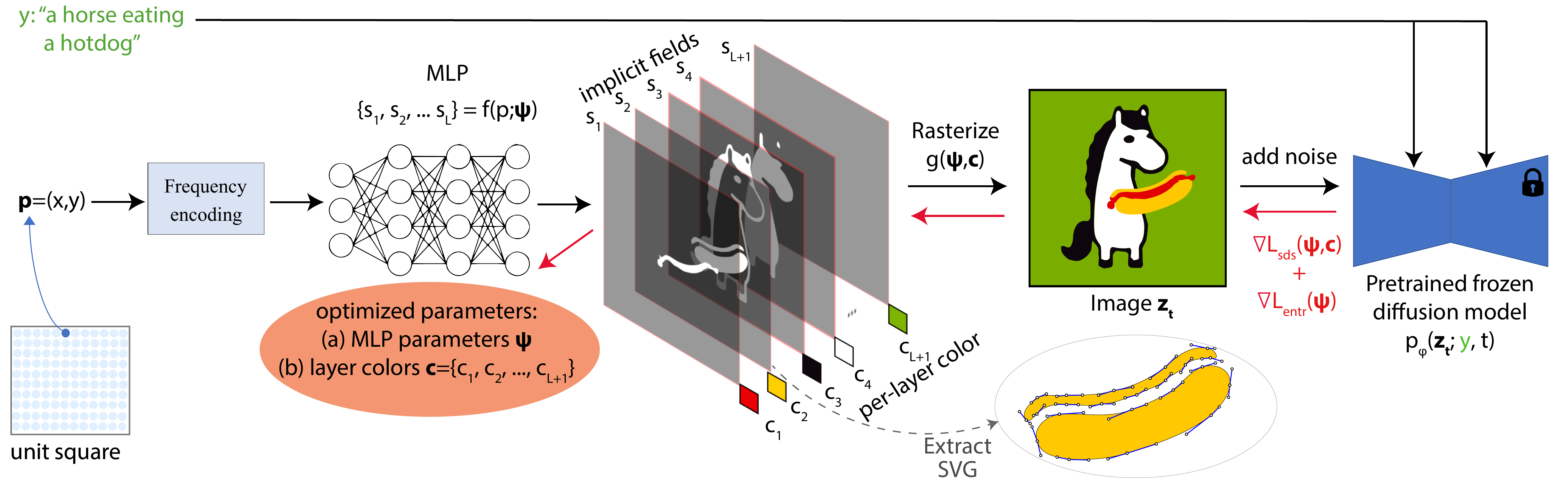}
\vspace{-6mm}
\caption{\methodname's architecture: given points $\bp$ on a 2D unit square domain, our method incorporates a MLP-based network with parameters 
$\bpsi$ that predicts a set of implicit fields on this domain, each representing a geometric shape. In addition, it predicts per-shape colors $\bc$. The representation is continuous, resolution-independent, and can easily converted to parametric curve formats (e.g, Bézier curves). 
To estimate the parameters, our method uses SDS-based optimization driven by a raster diffusion model conditioned on an input text prompt.
\label{fig:our_pipeline}}
\end{center}
\vspace{-5mm}
\end{figure*} 

For decades now, vector graphics has been synonymous in both industrial and artistic practice with parametric primitives, such as B\'{e}zier curves \cite{bezier1986courbes}, as also codified in one of several industrial standards (e.g. Scalable Vector Graphics \cite{svg12}). Recent addition of some more expressive fill primitives \cite{OBWBTS08} allow for richer appearance control where such is called for. 
Crucially, with visual features represented by primitives, the structure of the representation changes with the content, which causes challenges in machine learning scenarios \cite{smirnov2020dps}.

\paragraph{Sequence-generating methods.} A first family of ML approaches attempt to address these challenges by applying sequence-generating learning techniques to the vector representation itself, e.g. by generating drawing commands \cite{carlier2020deepsvg,sketchRNN,Lopes2019ALR,SPIRAL, wang2021deepvecfont, wang2023deepvecfont, wu2023iconshop}. Since the rendering process is non-smooth and non-trivial, it becomes difficult to maintain visual coherence and so these approaches struggle to scale.

\paragraph{Differentiable rasterizers.} The alternative approach is to do the learning in the image domain by making use of a differentiable rasterization process \cite{Li:2020:DVG, reddy2020discovering} as part of the pipeline. Such methods receive supervisory signal on the image domain to achieve a variety of effects \cite{reddy2021im2vec,CLIPdraw}. The difficulty here is that these methods need to fix the structure of the graphic, since backpropagating image-space gradients only gives a gradient signal for the existing continuous parameters of the graphic, but not for how the structure of the representation should be  changed. Moreover, image-space supervision provides no accurate regularization signal on the geometry, and these optimization approaches thus often result in degenerate geometries that are difficult for users to interpret or edit.

\paragraph{Neural implicits.} We are directly inspired by methods which address the above problem by adapting more flexible, vector-like structures, such as neural implicit fields \cite{park2019deepsdf, OccupancyNetworks, reddy2021multi, chen2021learning, paschalidou2021neural} proposed in 3D shape generation. These methods have the flexibility to make drastic changes to the shape topology without requiring any changes to structure. We seek to extend these to represent vector graphics and apply them to diffusion in particular.

\paragraph{Diffusion models for vector graphics.} In the space of diffusion models \cite{ho2020denoising}, attempts to apply them to vector graphics have included diffusion directly on vector parameters \cite{das2023chirodiff, wangsketchknitter, thamizharasan2023vecfusion}. Since using large foundational models often yields better results in general applications, Score Distillation Sampling \cite{poole2022dreamfusion, wang2022score} has been the key enabler allowing the extraction of information from large-scale image-based models and application thereof to non-raster imaging. Combining this with differentiable rasterization has led to VectorFusion \cite{jain2022vectorfusion}, a technique which allows for creating of vector images from text prompt as if by regular diffusion; however, like other rasterization-based methods, this one suffers from producing redundant and degraded geometry.
In contrast, our method leverages the idea of neural implicit functions 
expressed in
a set of decomposable layers
to create an inherently more stable and interpretable representation, sharing some key vector properties (resolution-independence, compactness in the representation). In addition, it is easy to reinterpret the representation as a layered vector graphic. It also simultaneously renders in a differentiable manner for image-space supervision. As discussed in our experiments section, our method produces text-to-vector graphics results
of significantly better fidelity than the state-of-the-art.

\section{Method}

Given an input text prompt, \methodname's goal is to synthesize 2D vector graphics representations (Figure \ref{fig:our_pipeline}). In the following section, we\ discuss our  representation of geometric shapes and layers, which is used to mediate the synthesis of vector graphics in our approach. Then we discuss the optimization procedure for learning the parameters of this  representation such that it synthesizes desirable vector images conditioned on text.  

\subsection{Representation}

\paragraph{Shape Representation.} Traditional vector graphics represent images as a set of geometric shapes defined on the 2D Cartesian plane, through parametric primitives, such as polylines and polygons, open or closed parametric paths (e.g., Bezier curves), and analytic primitives (e.g., arcs, circles, rectangles). These geometric shapes can be stacked on top of each other as different layers to create a target vector image. Inferring these geometric shapes by directly predicting these parametric primitives from high-level input is cumbersome due to their variable structure, as discussed in Section \ref{sec:intro}.
We instead propose to represent geometric shapes in different layers as a  2D continuous implicit function $f$ with learnable parameters $\bpsi$:

\begin{equation}
f(\bp;\bpsi): \R^2 \rightarrow [0,1]^L. \nonumber
\end{equation}
The implicit takes as input a 2D point $\bp$ in the unit square ${\bp=(x,y) \in [0,1]^2}$ and outputs a probability  whether the point is present in a shape (classification as $1$) or absent from the shape (classification as $0$) in each of the layers. The number of layers $L$ is a hyper-parameter and represents an \emph{upper bound} to the actual number of layers used in our representation i.e., some predicted layers can be empty (i.e. zero output for all their points), thus are unused in the final image.
Similar to other neural implicit representations proposed in vision and graphics \cite{neuralfields}, we model $f$  with the help of a MLP neural network (Figure ~\ref{fig:our_pipeline}). 

\paragraph{Positional encoding.} 
The network takes as input a positional encoding $\gamma(\bp)$ of the input $(x,y)$ coordinates that maps each point into a higher dimensional space. The space is useful to capture shape variations at different frequencies, as also used in other neural implicit representations, such as NeRFs \cite{mildenhall2020nerf}:
\begin{equation}
\begin{split}
    \gamma(\bp) & = 
    \begin{bmatrix}
        \sin( O \pi \bp) &
        \cos( O \pi \bp)
    \end{bmatrix}
    \\    
    \mathrm{where}
    \;
    &O=
    \begin{bmatrix}
        2^0 & 2^1 & \dots & 2^{F-1}
    \end{bmatrix}^\top
\end{split}
\label{eq:enc}
\end{equation}
with $F$  being a hyper-parameter in the representation denoting number of octaves.

\paragraph{Network architecture.}  The network processing these positional encodings is a sequence of  fully connected layers with residual connections. More details about the MLP\ architecture and its hyperparameters are provided in our suppmentary material.  The MLP outputs a set of continuous values, or probabilities: ${\bs=\{s_l\}_{l=1}^{L}}$, where each  value $s_l$ is bounded in $[0,1]$  through the use of a per-layer sigmoid non-linearity. In short, our network implements the following transformation parameterized by the MLP parameters $\bpsi$:
\begin{equation}
\bs=sigmoid\Big(MLP_{}\big(\gamma(\bp);\bpsi\big)\Big)
\label{eq:mlp}
\end{equation}

\paragraph{Color Representation.} Traditional vector graphics also represent color per primitive. We  represent per-layer color as a set of $L+1$ learnable parameters ${\bc=\{\bc_1,\bc_2,\cdots,\bc_{L+1}\}}$ taking RGB\ values: $\bc_l \in [0,1]^3$. The inclusion of the last color parameter $\bc_{L+1}$ is associated with the background color that a vector image can have, or in other words, the color of a dummy layer representing background.

\paragraph{Layer composition.} The final vector image is created by stacking the predicted layer outputs on top of each other. The last dummy layer $L+1$ is placed as the last layer in the stack, while the first layer created by the MLP outputs $\bs_1$ is the front-most one. Any shape at layer $l$ occludes shapes at layers $[l+1,\cdots,L+1]$. 
The output continuous\ image is created as a layer compositing function $g(\bp;\bpsi, \bc)$:

\begin{equation}
g(\bp;\bpsi, \bc)= \sum^{L+1}_{l=1} k_l \cdot c_l,
\end{equation}
where $k_l = s_l  \prod_{m < l} (1-s_m)$ is a mixing coefficient for the layer at that point, computed from opacities of layers on top of it. We note that the final image is a continuous representation with ``infinite'' resolution, as in the case of vector images i.e, one can zoom into a region of the unit square without pixelization artifacts.
We also note that the sequence of layers in the stack matters -- the network should learn to distribute shapes to layers such that their composition with the above function yields a desired target image. Fortunately, for learning purposes our compositing function is differentiable wrt both color parameters $\bc$ as well as $\bpsi$ 
i.e., ${\partial g/\partial \bpsi= \sum_l(\partial g/\partial k_l)\cdot(\partial k_l/\partial s_l)\cdot(\partial s_l/\partial \bpsi)}$ -- here, we omit input points $\bp$ for clarity. 

\subsection{Parameter estimation}

As discussed in our introduction, compared to raster images, there is a lack of datasets including SVGs and text pairs. One possibility to circumvent this problem is to leverage generative models of raster images trained on massive datasets of generic images for vector image synthesis in a zero-shot generation setting.  Most recent powerful generative models of raster images are based on diffusion models. One problem, however, is that these models output raster images with incompatible data types, style, and dimensionality compared to vector or implicit representations, such as ours. A common strategy to deal with this incompatibility is to perform parameter estimation and sample synthesis via optimization based on the score distillation approach  \cite{poole2022dreamfusion,jain2022vectorfusion}.

\paragraph{Score distillation from image-based diffusion models.} The goal of score distillation in our setting is to estimate parameters of our implicit representation such that it synthesizes a sample output with high probability according to a pre-trained image-based diffusion model conditioned on text prompts related to vector styles 
(e.g., ``minimal 2D vector art, lineal colors, line drawing'', see also Figure \ref{fig:teaser}). The loss penalizes the KL-divergence of a unimodal Gaussian distribution centered at a learned sample produced by our model $g(\bpsi, \bc)$ and the data distribution $\bp_\phi(\bz;y,t)$ captured by the frozen diffusion model conditioned on text embeddings $\by$. The loss is averaged over several  time steps $t$ sampled throughout the diffusion process:
\begin{align}
& \mathcal{L}_\text{sds}(\bpsi, \bc) = \nonumber
\\ &\mathbb{E}_{t} \left[ \frac{\sigma_t}{\alpha_t} w(t) \text{KL}(q(\z_t|g(\bpsi, \bc); y, t) \| p_\phi(\z_t ; y,t ))\right]
\label{eq:sds}
\end{align}
where $t$ is a timestep, $w(t)$ is a weighting function and $\alpha_t, \sigma_t$ are coefficients of the diffusion model depending on the timestep $t$. 
The loss perturbs the image produced by our model  $g(\bpsi, \bc)$  with a random amount of noise corresponding to the timestep, and estimates an update direction that moves it towards a higher probability density region dictated  by the diffusion model, while being constrained on the implicit data representation imposed by our model.

\begin{figure}[t]
    \centering
    \setlength{\tabcolsep}{2pt}
    \begin{tabular}{c | c c}
       Input &  w/o $\mathcal{L}_{entr}$ & with $\mathcal{L}_{entr}$\\
        \includegraphics[width=0.3\linewidth]{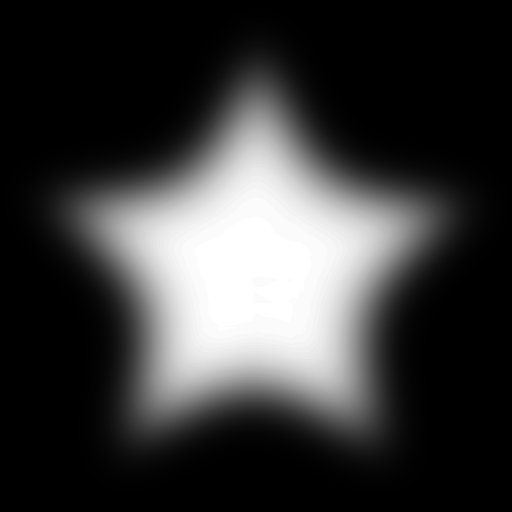}
        &
        \includegraphics[width=0.3\linewidth]{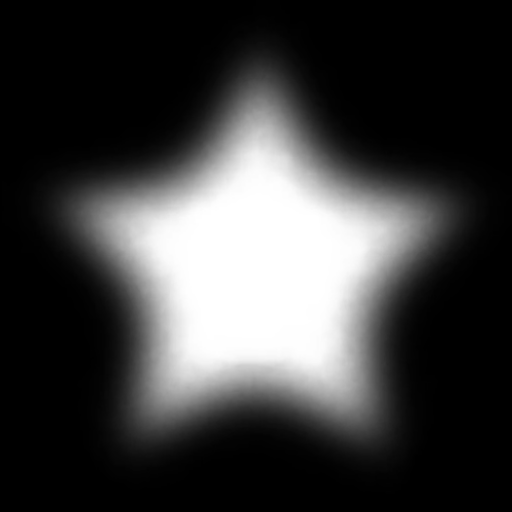}
        &
        \includegraphics[width=0.3\linewidth]{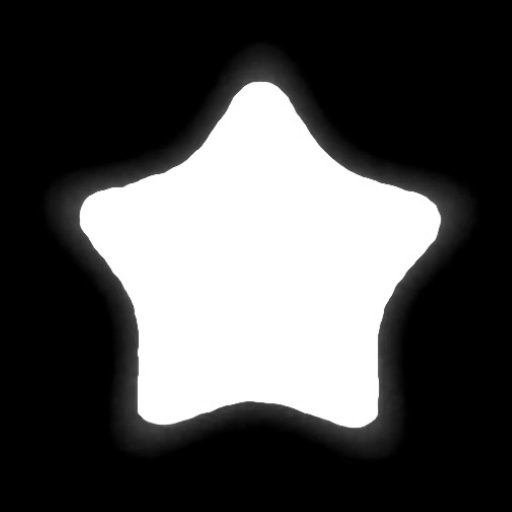}
        \\
    \end{tabular}
    \vspace{-3mm}
    \caption{Given an input sampled image (left), we estimate the \methodname's parameters through L2 reconstruction loss without entropy $L_{entr}$ (middle), or with entropy (right). The entropy results in a cleaner shape with delineated boundaries.
    }
    \label{fig:entropy}
    \vspace{-3mm}
\end{figure}

In addition to the above SDS loss, we found useful to include a regularization term penalizing uncertainty in our model's mixing coefficients. Since we do not model translucent shapes, we prefer these mixing coefficients to be skewed towards either 0 or 1 (see also Figure \ref{fig:entropy}). The preference can be expressed through the following entropy loss applied to the  
per-layer mixing coefficients $k_l$, involving only the MLP parameters $\bpsi$:
\begin{align}
    \mathcal{L}_{entr}(\bpsi) &= -\sum_l k_l \log(k_l)
\end{align}

To learn the parameters of our model, we minimize 
a weighted sum of the SDS loss and the above entropy loss:

\begin{equation}
\mathcal{L}_{total} = \mathcal{L}_\text{sds}(\bpsi, \bc) + \lambda \mathcal{L}_{entr}(\bpsi)
\label{eq:final_loss}
\end{equation}
where $\lambda$ is a hyperparameter serving as weighting term for the entropy loss.  

\paragraph{Initialization.} To initialize the above SDS-based optimization, one possibility is to start with random values for parameters $\{\bpsi, \bc\}$. Unfortunately, this strategy is suboptimal. The initial random value of parameters often correspond to areas of low probability density according to the image-based diffusion model, where the SDS-based gradients tend to be noisy, slowing convergence or even leading to undesirable local maxima causing implausible results (see also Section \ref{sec:results} for an ablation wrt this strategy).

An alternative strategy for initialization is to sample a raster image from the diffusion model, then initialize our parameters such that the output of our model reconstructs the sample image as closely as possible.
More specifically, given a sample RGB raster image $\hat{\bz}$ produced by executing the reverse diffusion of an existing pretrained diffusion model conditioned on the input text prompt, the parameters can be pre-trained to minimize the loss:
\begin{equation}
\mathcal{L}_{rec}=
\mathcal{L}_\text{2}(\bpsi, \bc) + \lambda' \mathcal{L}_{entr}(\bpsi)
\label{eq:rec}
\end{equation}
where $\mathcal{L}_2$ is summed over the pixel locations of the raster image: ${{\mathcal{L}_2(\bpsi, \bc)  = \sum_{\bp \in \bR} \lVert g(\bp;\bpsi, \bc) - \hat{\bz}(\bp) \rVert^2}}$ (the effect of including the entropy loss is highlighted in Fig. \ref{fig:entropy}).
Still, this initialization strategy
was also suboptimal, as shown in our experiments. The reason is that the sample raster images tend to incorporate a photographic style with high-frequency texture variation, details and natural background that are not typically present in illustrations modeled by artists in vector graphics, even with text prompts related to vector style, as also noted in VectorFusion \cite{jain2022vectorfusion}. 

To overcome these challenges, we employ to an approach where the initialization of our model is not based on fitting it to a sample raster image, but instead a low-frequency image representation better matching the output parameterizaton of our model. More specifically, during our initialization phase, we train a model mapping continuous image coordinates $\bp=(x,y) \in [0,1]^2$ to RGB values using the same MLP-based architecture as in Eq. \ref{eq:mlp}. Its output is a predicted per-point RGB color $\bz(\bp; \btheta)$  in $[0,1]^3$, and $\btheta$ are the MLP parameters estimated via SDS optimization guided by the same image-based diffusion model i.e., we use Eq. \ref{eq:sds} with the above implicit RGB image generator instead. An important design choice of this generator is that the used positional encoding $\gamma(\bp)$ incorporates a limited set of octaves (up to $F=6$ bands in our implementation) to encourage low-frequency texture and background variation in the output image. 

After estimating the parameters $\btheta$ of the above implicit RGB generator, we initialize the parameters $\{\bpsi, \bc\}$ of our model using the reconstruction-based loss $\mathcal{L}_{rec}$.
Finally, we fine-tune the parameters with our main loss $\mathcal{L}$
(Equation \ref{eq:final_loss}). 
This combination of initialization and fine-tuning offered the best results in our experiments. 

\begin{figure*}[t]
\centering

\includegraphics[width=0.9\linewidth]{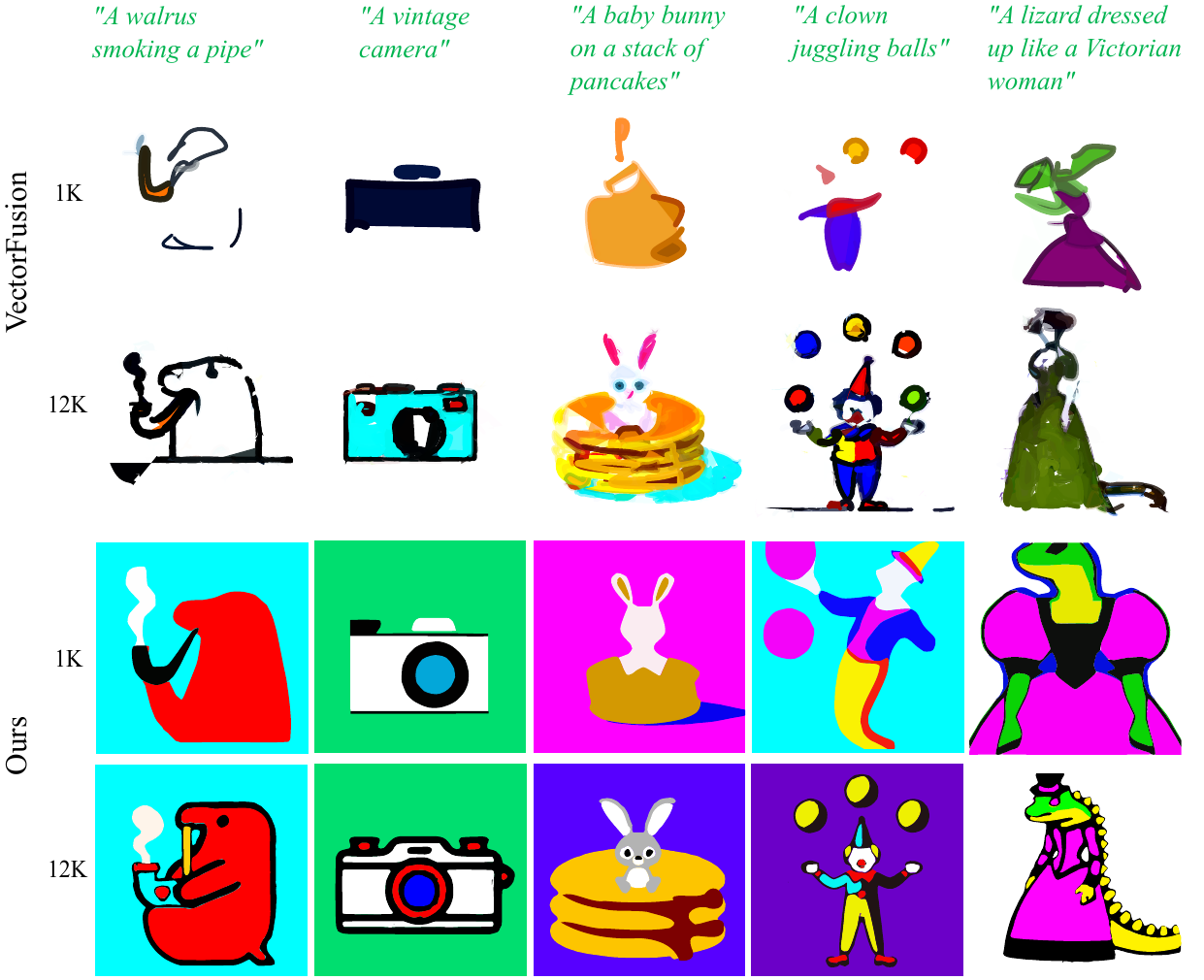}

\vspace{-2mm}
\caption{Text-to-Vector Graphics generation results. We compare generated SVG results for the \textcolor{textprompts}{input text prompt} between \methodname (ours) vs VectorFusion, at two settings involving $1K$ or $12K$ number of parameters. Our vector results contain much cleaner shape geometry across diverse topologies while VectorFusion's SVGs contain redundant, degenerate curves, and self-intersecting shapes. Our method  also remains robust at a low capacity (1K parameters), faithfully capturing the abstraction of the concepts in the input text prompt.}
\label{fig:ours_vs_vectorfusion}
\vspace{-3mm}
\end{figure*}

\subsection{Implementation details}
\label{ref:implementation_details}

First, we note that our implementation \emph{will become publicly available upon acceptance.} Below we discuss important implementation details. 

\paragraph{Image-based diffusion model.} We use the open sourced pre-trained DeepFloyd model \cite{deepfloyd} to compute our SDS gradients during the initialization and fine-tuning phase. DeepFloyd performs denoising diffusion in pixel-space, unlike Stable Diffusion which performs it in latent space. This avoids the computational bottleneck of back-propagating through the image encoder and, in practice, we observe better results and faster convergence with DeepFloyd. 

\paragraph{Network architecture} We trained two versions of our model for evaluation: a 12,000 (12K) parameter MLP with 64 hidden nodes and 4 layers, and a smaller 1,000 (1K) parameter MLP with 32 hidden nodes and 3 layers. We set $F$ (number of octaves) to 6 for the former and 2 for the latter. Note, the frequency encoding is a fixed function without additional learnable parameters. Both these models contain LeakyReLU activations and Sigmoid at the final layer. We set $L$ (the number of layers) to 5 for all experiments to have a comparable trainable parameter count to VectorFusion. We discard unused layers (all zero outputs) after optimization. We jitter the query points during training for additional robustness to evaluation points during inference. We implement our model using the tiny-cuda-nn library \cite{tiny-cuda-nn}.

\vspace{-0.5em}
\paragraph{SDS scheduler and optimization.} We use the vanilla timestep scheduler proposed in DreamFusion \cite{poole2022dreamfusion}. We note that we explored with alternative timestep schedulers (e.g., Dreamtime \cite{huang2023dreamtime}), yet it did not offer any improvements in our setting . We perform the SDS-based optimization using AdamW \cite{loshchilov2019decoupled}. We set the loss hyperparameters $\lambda=10^{-5}$, $\lambda'=10^{-4}$,  batch size to 3, and total optimization iterations to $8000$ for all our experiments. Our final model takes 5 minutes to converge on a single NVIDIA A100 gpu. We provide a model card in table \ref{tab:model_details}.

\paragraph{Converting implicit layers to Bézier curves.}
We extract iso-curves from our layered implicit shapes through marching squares at any prescribed resolution. We query $f_\bpsi$ at $N \times N$ grid points $\bp$ to produce $L$ raster shapes ($N=2048$ in our implementation). For our experiments, we set $L=5$ as the upper bound on the number of layers. 
We analyse the effect of different $L$ on the generated results for a given text prompt in section \ref{supp:ablation_layers}. We then fit cubic Bézier curves to each of our layers using an open source image tracing software (Inkscape \cite{Inkscape}).

\section{Experiments}
\label{sec:exps}
\label{sec:results}

\begin{figure}[t!]
  \centering  
  \begin{tabular}{@{}lccc|cc@{}}
    \toprule
    & & \multicolumn{4}{c}{CLIP L/14} \\ 
    \textbf{Method} & Parameters & \multicolumn{2}{c}{R-Prec $\uparrow$} & \multicolumn{2}{c}{Sim $\uparrow$} \\ 
    & & mean & std & mean & std\\ \midrule
    \multirow{2}{*}{VectorFusion} &  1K & 59.5 & 1.51 & 21.4 & 0.78\\
     & 12K & 71.3 & 2.37 & 26.7 & 0.73 \\ \midrule
    \multirow{2}{*}{\methodname} & 1K & \textbf{68.2} & 1.26 & \textbf{25.1} & 0.81\\
      & 12K &  \textbf{78.5} & 1.40 & \textbf{32.0} & 0.72
    \\ \bottomrule
  \end{tabular}%
  \vspace{-1mm}\captionof{table}{CLIP metrics computed with the \emph{clip-vit-large-patch14} pre-trained model on rasterized SVG results of \methodname and VectorFusion \cite{jain2022vectorfusion} under our two parameter settings. Both methods are optimized with DeepFloyd \cite{deepfloyd}. We also show the official reported results from VectorFusion when optimized with Stable Diffusion (row ``with SD''). We note their implementation is not available, thus, the number of parameters in their experiments is unknown. 
  \vspace{-3mm} 
  } \label{tab:ours_vs_vectorfusion_clip}
\end{figure}

\paragraph{Dataset.} We use the prompt dataset curated by VectorFusion \cite{svgprompts} to generate our qualitative results and compute our quantitative metrics. This dataset consists of two prompt families: (i) line drawings with prefix and suffix \emph{"Line drawing of ..., minimal 2d line drawing, on a white background, black and white"}, and (ii)  minimal lineal colors with suffix \emph{"..., minimal flat 2d vector art, lineal color, on a white background, trending on artstation"}. We include all prompts in section \hyperref[supp:prompts]{E}.

\paragraph{Comparison.} We compare with VectorFusion \cite{jain2022vectorfusion}, which shares the same goal of creating vector art from text prompt as our method. We note that we use our own implementation of VectorFusion, since it is not publicly available. 
Their method
optimizes randomly initialized cubic Bezier curves with SDS. For a fair comparison, we plugged the same diffusion model DeepFloyd \cite{deepfloyd} (DF for shot) as ours. In addition, we adjust both VectorFusion and our method to use a comparable number of parameters for optimization. More specifically, we use the following two settings:
\begin{enumerate}
    \item \textbf{1K parameters.} \emph{VectorFusion} uses 16 paths each containing up to $5$ cubic Bézier curves. Each Bézier curve has 17 parameters (control points, stroke width and color, fill color) $^{\curlyvee}$\blfootnote{$^{\curlyvee}$ 4 control points * 2 (x,y) + 1 (stroke width) + 8 (rgba for both stroke and fill color) = 17}. \emph{\methodname} uses an MLP with 32 hidden nodes and 3 layers and $F=2$ (number of octaves).
    \item \textbf{12K parameters}: \emph{VectorFusion} uses 256 paths each containing up to $5$ cubic Bézier curves (17 parameters each) . \emph{\methodname} uses an MLP with 64 hidden nodes and 4 layers with $F=6$ octaves.
\end{enumerate}

For both settings, we extract Bezier curves for our method, using the procedure discussed in Section \ref{ref:implementation_details}. We note that our output number of Bezier paths is not the same with VectorFusion. For our method, their number is automatically selected according to Inkscape's implementation operating on our output. We argue that automatically adjusting the number of paths depending on the target output is more desirable than imposing a hard constraint on their number. 

\begin{figure}[t!]
  \centering
\includegraphics[width=1.0\linewidth]{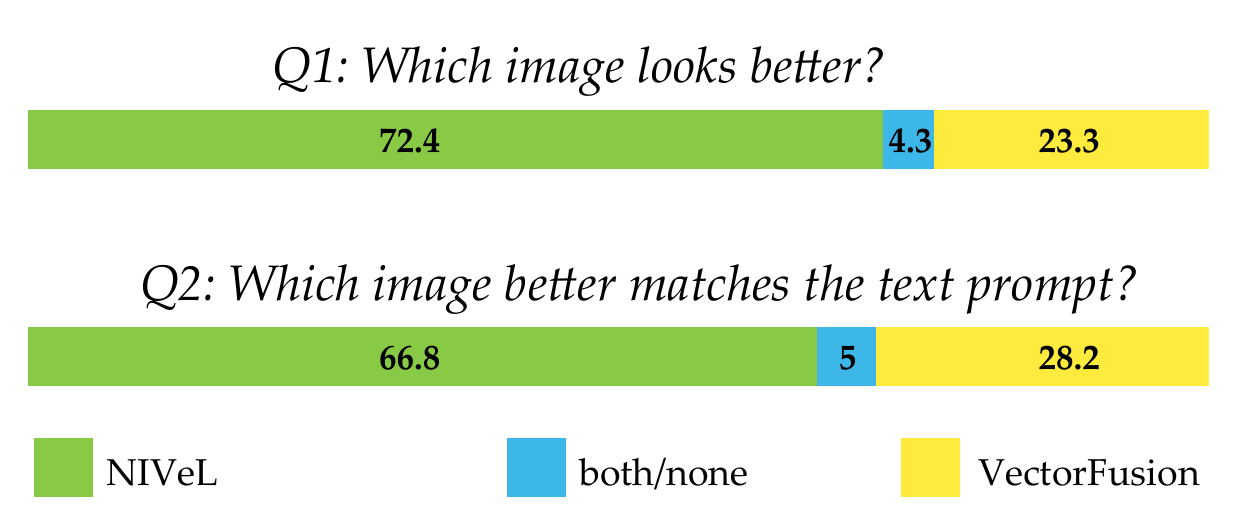}
  \caption{User study voting results.}
\vspace{-4mm}  \label{tab:ours_vs_vectorfusion_userstudy}
\end{figure}

\begin{figure*}[t]
\centering
\includegraphics[width=0.95\linewidth]{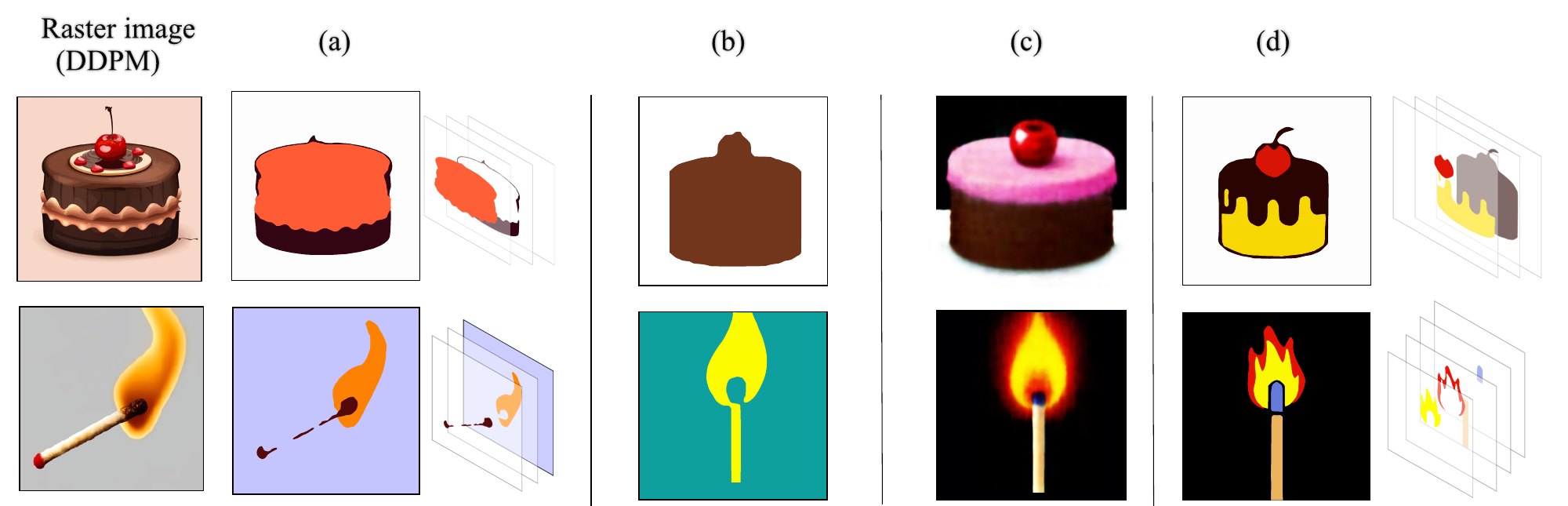}
\vspace{-1.0mm}
\caption{ For the prompts \textcolor{textprompts}{\emph{``A cake with chocolate frosting
and cherry, 2D vector art''}} and
\textcolor{textprompts}{\emph{``A match stick on fire, 2D vector art''} }
we sample raster images from a pre-trained text-to-image Diffusion Model via DDPM
(a) We optimize \methodname by using the reconstruction-based loss $\mathcal{L}_{rec}$ on the sampled image (no SDS optimization). (b) We  optimize \methodname by using our SDS-based loss $\mathcal{L}_{total}$ with random initialization. (c) We optimize \methodname by using the reconstruction-based loss $\mathcal{L}_{rec}$ on the RGB generator network (no SDS optimization), (d) We optimize \methodname by fine-tuning with SDS-based loss after initialization from the RGB generator network. The last strategy offers the most visually compelling vector art outputs.
}
\label{fig:ablation}
\vspace{-2mm}
\end{figure*}

\vspace{-0.2em}
\paragraph{Evaluation metrics.} First, we report the CLIP-based score (denoted as ``Sim'')  \cite{clipscore} also used in VectorFusion \cite{jain2022vectorfusion}. The score measures the cosine similarity between the embeddings for a raster image and the embeddings for a text caption. To this end, we rasterize the Bézier  curves extracted from VectorFusion and \methodname in the same resolution ($224 \times 224$) using the same rasterizer \cite{Inkscape}.
The CLIP score is averaged over all prompts. We also report the R-Precision, as also used in VectorFusion (``R-Prec''). This is the percent of output images having maximal CLIP similarity with the correct input caption among all text prompts. Finally, we conduct a perceptual evaluation based on a user study asking participants to compare results between our method and VectorFusion in terms of their  plausibility and degree of matching with the input text.

\paragraph{Quantitative comparisons.} Table \ref{tab:ours_vs_vectorfusion_clip} reports results for the CLIP-based similarity and precision measures, comparing our method and VectorFusion. We outperform VectorFusion for both parameter setting, demonstrating the parameter efficiency of our representation and faithfulness of our generation to the input prompt. The table row ``VectorFusion (with SD)'' represents their originally reported metrics when optimizing randomly initialized curves with Stable Diffusion (SD for short) on the same dataset \cite{jain2022vectorfusion}. 
We note that their implementation is not available, thus, the number of parameters used in their experiments is unknown. 

\paragraph{Qualitative comparisons.} Figures \ref{fig:teaser}, \ref{fig:ours_vs_vectorfusion}, and \ref{fig:ours_vs_vectorfusion_2} compare the results of our method and VectorFusion qualitatively. Compared to VectorFusion, our results are devoid of artifacts (floating curves, degenerate geometry) and capture cleaner shapes and appearance 
(i.e., properties commonly associated with vector graphics). In addition, our representation is parameter-efficient: with just 1K parameters, our generated results are faithful to the prompt and capture the input concepts. On the other hand, VectorFusion cannot faithfully capture the geometry of objects with low parameter count. With 12K parameters, VectorFusion's generated results include redundant curves and suffers from the aforementioned artifacts, making their results much harder  to edit intuitively (Figure \ref{fig:teaser}). Section \hyperref[supp:ours_vs_vectorfusion_linedrawing]{B} details results on the line drawing style.

\paragraph{User study.} 
We also conducted two Amazon MTurk studies as additional perceptual evaluations. In the first study, each questionnaire page showed participants a randomly
ordered pair of raster outputs from our method and VectorFusion.
We asked participants 
``which image looked better'', explaining to them in our instructions that they should choose based on which image had fewer artifacts and looked more aesthetically plausible. 
Participants could pick either image,  specify ``none'' or ''both'' images looked good enough. We asked questions twice in a random order to verify participants' reliability. We had $100$ reliable participants, each comparing $10$ unique pairs generated from our pool of text prompts (total $1000$ comparisons). 

In our second study, we showed participants randomly ordered pairs of outputs from \methodname and VectorFusion, along with the input text prompt, and asked them ``which image better matches the input text prompt''. Participants could again pick either image, specify ``none'' or ''both'' images matched the input text  well. We also asked questions twice in a random order to verify participants' reliability. We again had $100$ reliable participants, each comparing $10$ unique pairs as before (total $1000$ comparisons). 

Figure \ref{tab:ours_vs_vectorfusion_userstudy} summarizes the percentage of votes for the above two types of questions. We observe that our method's outputs were preferred by a significantly larger proportions of participants for both questions. 

\paragraph{Ablation.} In Figure \ref{fig:ablation} we demonstrate the importance of a good initialization strategy. A randomly initialized network is often prone to failure modes, wherein it allocates entire shapes into the initial layers (Figure \ref{fig:ablation}b).
This prevents the generation of semantically meaningful decomposed shapes. Another possibility is to initialize our network by fitting it to reconstruct a sampled image from the diffusion model (Figure \ref{fig:ablation}a). This initialization yields a rather coarse, oversimplified  layers. In contrast, initializing the network by fitting it to reconstruct the implicit RGB generator's output provides a more appealing starting point (Figure \ref{fig:ablation}c). Fine-tuning the network through SDS-based optimization 
after this initialization provides the best results
(Figure \ref{fig:ablation}d). Effects of using different random seeds is detailed in section \hyperref[supp:ablation_seed]{A}. We additionally visualize the output of SDS optimization iterations in section \hyperref[supp:sds_gradients_visualized]{C}.

\section{Conclusion}
\label{sec:conclusion}

We presented a method that proposes a layered implicit field representation as a mediator for generating vector graphics. We demonstrated significantly better results than the state-of-the-art in the case of text-to-vector synthesis. 

\vspace{-0.8em}
\paragraph{Limitations and future work.} Our method still has limitations that can spur new research directions for generative vector graphics. First, our representation is currently bounded by an upper number of layers. Generating layers dynamically would alleviate this issue. Currently, we use an open-source vectorizer to convert our implicit field to parametric curves. It would be worth investigating a differentiable implicit-to-vector module to perform this conversion. Investigating better SDS-based optimization strategies e.g., with more adaptive time-step schedulers and more stability could further improve our results. Finally, extending our representation to the 3D domain for predicting  3D parametric modeling primitives guided by neural implicits is generally a worthwhile research  direction.

\vspace{-1em}
\paragraph{Acknowledgments.} We thank Dmitry Petrov, Matheus Gadelha, and Thibault Groueix for their helpful discussions. Our project was funded by Adobe Research.

{\small
\bibliographystyle{ieee_fullname}
\bibliography{main.bib}
}

\newpage

\dictchar{Appendix}
    
\section*{A. Additional Ablations}

\begin{figure}[h]
\centering
\includegraphics[width=1.0\linewidth]{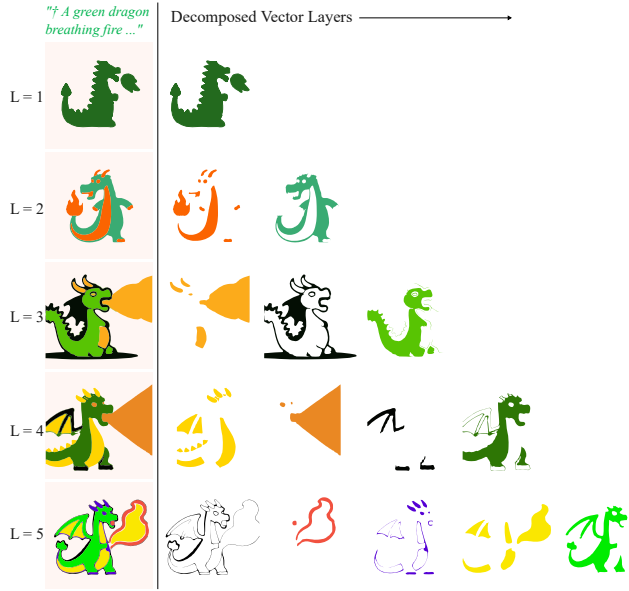}
\vspace{-7mm}
\caption{For the given \textcolor{textprompts}{input text prompt} we show our generated results changing only the value of $L$ (the number of output decomposed layers). $\dagger$: "... minimal 2D vector art, lineal colors"}
\label{fig:num_layers}
\end{figure}

\paragraph{Effect of number of layers $L$.}
\label{supp:ablation_layers}
 The number of decomposed layers generated by our system is set by the hyperparameter $L$. In Figure \ref{fig:num_layers}, we show the effect on the generated results for values of $L$ ranging from 1 (minimum) to 5 (default value in our experiments). We fix the prompt and all other settings for all results. Our method can generate plausible shapes and colors under each constrained setting. We observe presence of distinguishable semantic parts as the number of layers $L$ increase.

\begin{figure}[t]
\centering
\includegraphics[width=0.8\linewidth]{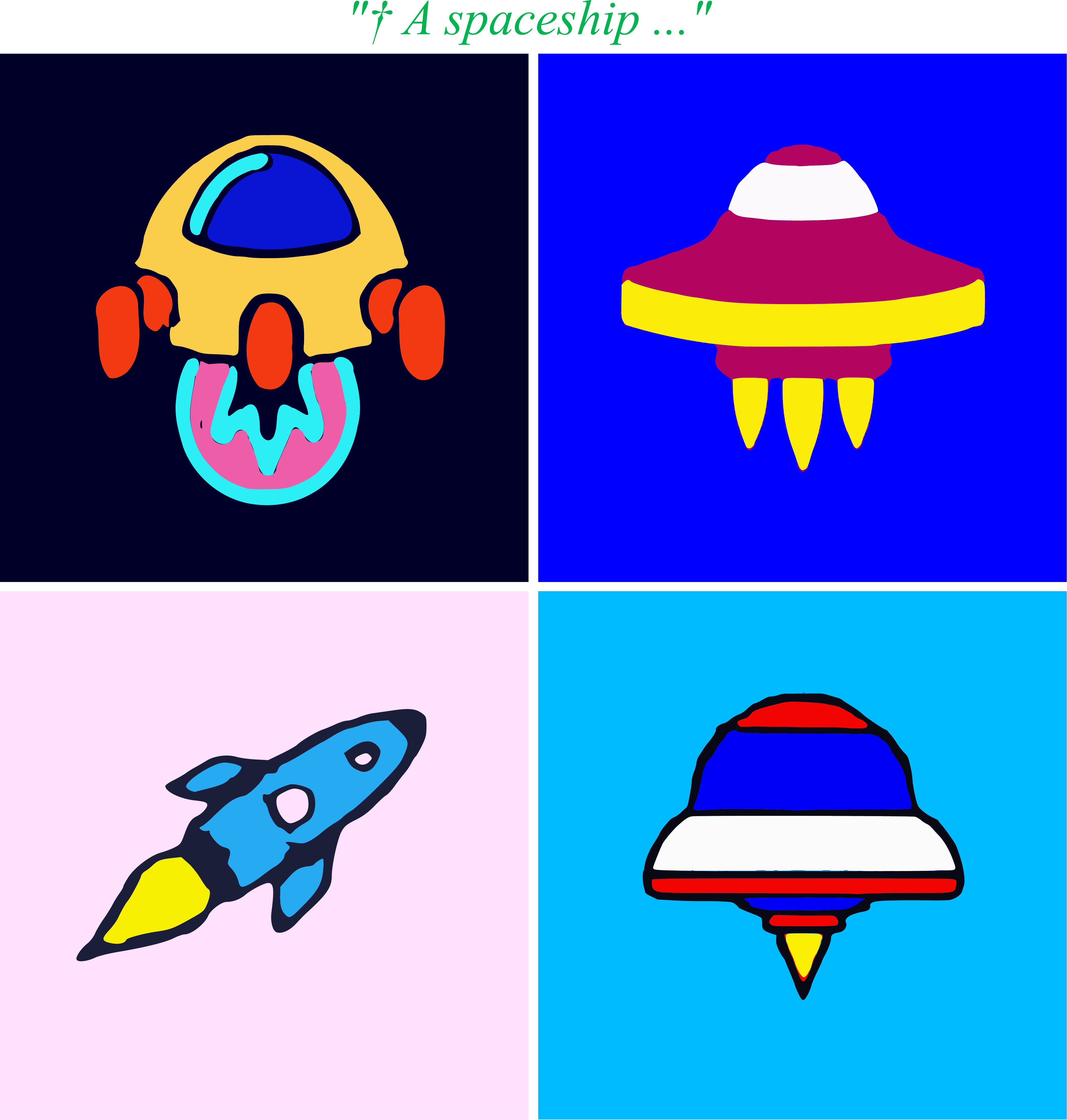}
\vspace{-2mm}
\caption{The effect of changing the random seed for the given \textcolor{textprompts}{input text prompt}. $\dagger$: "... minimal 2D vector art, lineal colors"}
\label{fig:random_seed}
\end{figure}

\begin{figure}[t]
\centering
\includegraphics[width=0.68\linewidth]{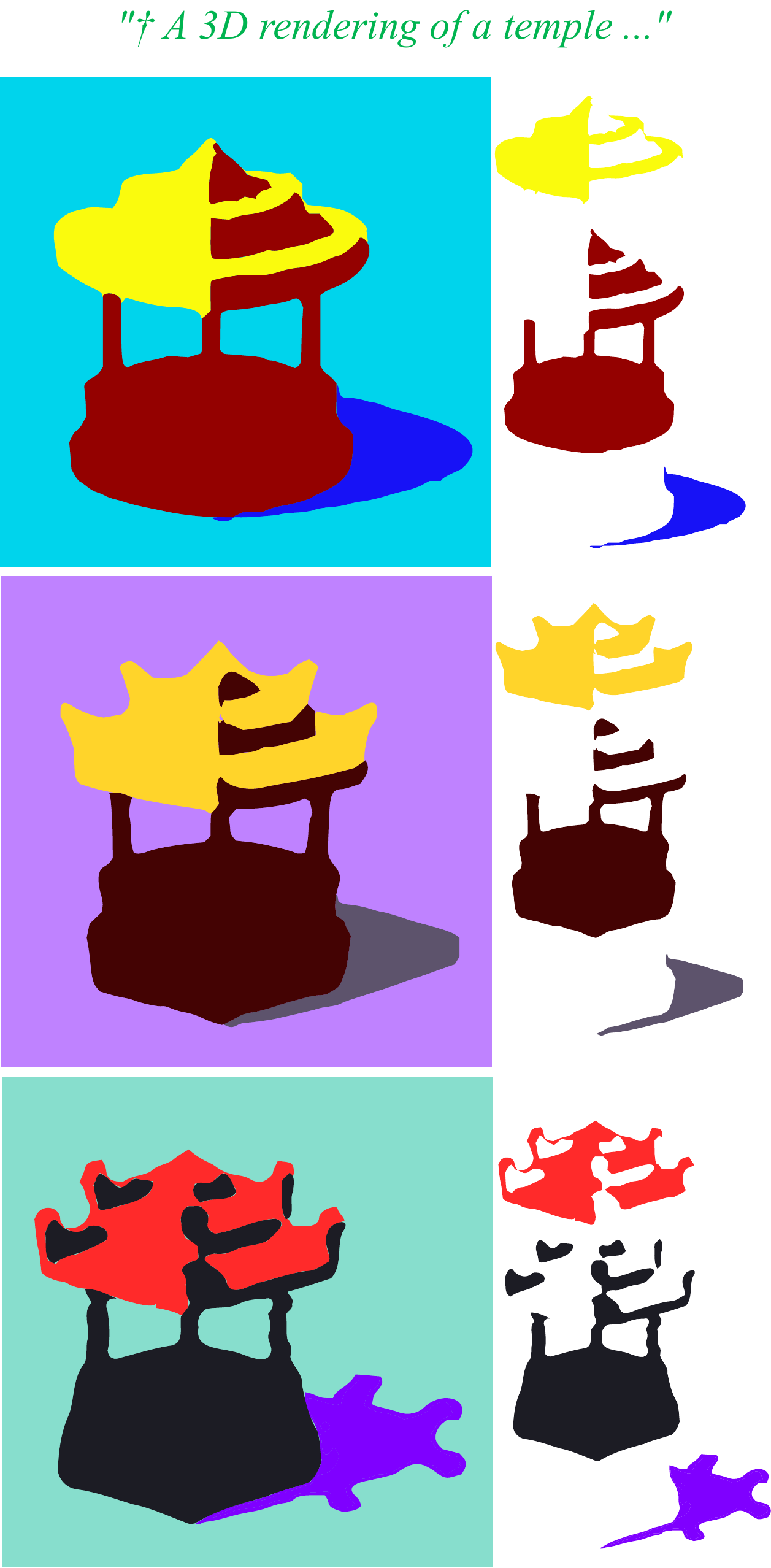}
\vspace{-2mm}
\caption{Rich priors learned by the image diffusion model are preserved in our generated shapes. Specifically, the shape of the shadow matches the style and structure of the roof of the temple across the three shown synthesis runs. Each generation uses a different random seed. $\dagger$: "... minimal 2D vector art, lineal colors"}
\label{fig:sds_priors}
\vspace{-1.0em}
\end{figure}

\paragraph{Effect of random seeds.}
\label{supp:ablation_seed}
Figure \ref{fig:random_seed} and \ref{fig:sds_priors} show the stochasticity of the results achieved by changing the random seed of the image diffusion model.

\begin{figure*}[t!]
    \includegraphics[width=0.9\linewidth]{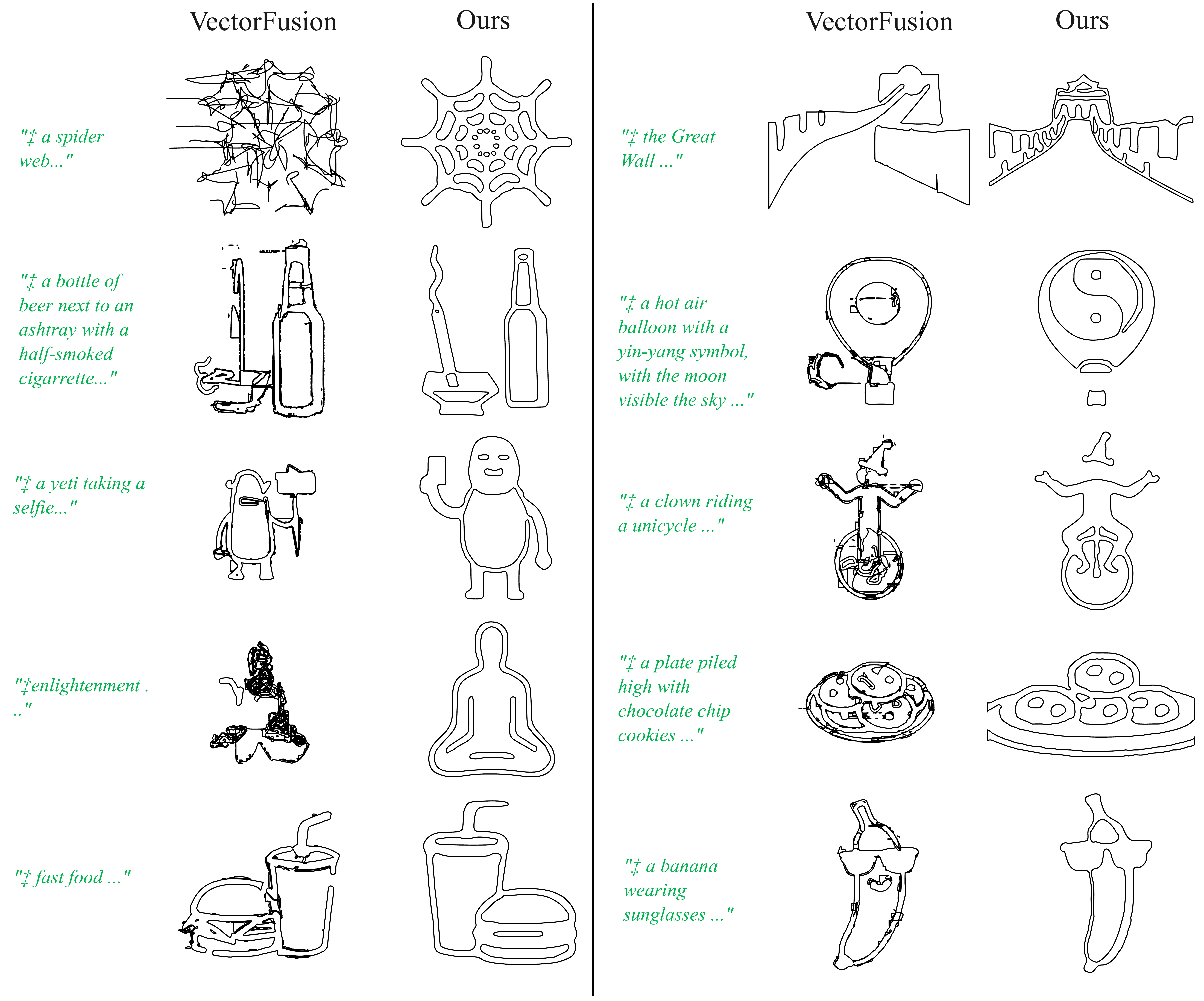} 
    \vspace{-3mm}
    \caption{Text-to-Vector Graphics generation results. We compare generated SVG results for the \textcolor{textprompts}{input text prompt} between \methodname (ours) vs VectorFusion . Our vector results contain much cleaner shape geometry across diverse topology while VectorFusion's SVGs contain redundant, degenerate curves, and self-intersecting shapes. $\ddagger$: "Line drawing of ... black and white"
\label{fig:ours_vs_vectorfusion_2}    
    }    
\end{figure*}

\section*{B. Qualitative Results}
\label{supp:ours_vs_vectorfusion_linedrawing}
In Figure \ref{fig:ours_vs_vectorfusion_2} we compare NIVeL (ours) vs VectorFusion for a  line drawing style. Since the style involves shape outlines or contours, here we use a single layer ($L=1$).
For the rest of the hyperparameters, we set $F=6$ octaves in the positional encoding, and the MLP architecture has $12K$ parameters. For VectorFusion, we use 256 paths each containing upto 5 Bézier curves for a total of also $12K$ parameters. We visualize the generated SVGs without the curve fill attribute to demonstrate the accuracy and smoothness of our produced line drawings of shapes compared to VectorFusion's results which contain redundant overlapping and self-intersecting lines.

\begin{figure*}[t!]
    \centering
    \setlength{\tabcolsep}{0.5pt}

\begin{tabular}{c ccccccccccccccc}
    &
    \multicolumn{7}{c}{NIVeL optimization steps $\xrightarrow{\makebox[2cm]{}}$}
    &
    \multicolumn{8}{c}{}
    \\
    Init & 100 & 300 & 500 & 800 & 1200 & 1500 & 1800 & 2400 & 3000 & 3600 & 4200 & 5000 & 5800 & 6900 & 8000
    \\
    \multicolumn{16}{c}{}
    \\
    \footnotesize{\emph{}}&
    \multicolumn{15}{c}{\emph{\textcolor{textprompts}{"A spider web ..."}}}
    \\
    \includegraphics[width=0.06\linewidth]{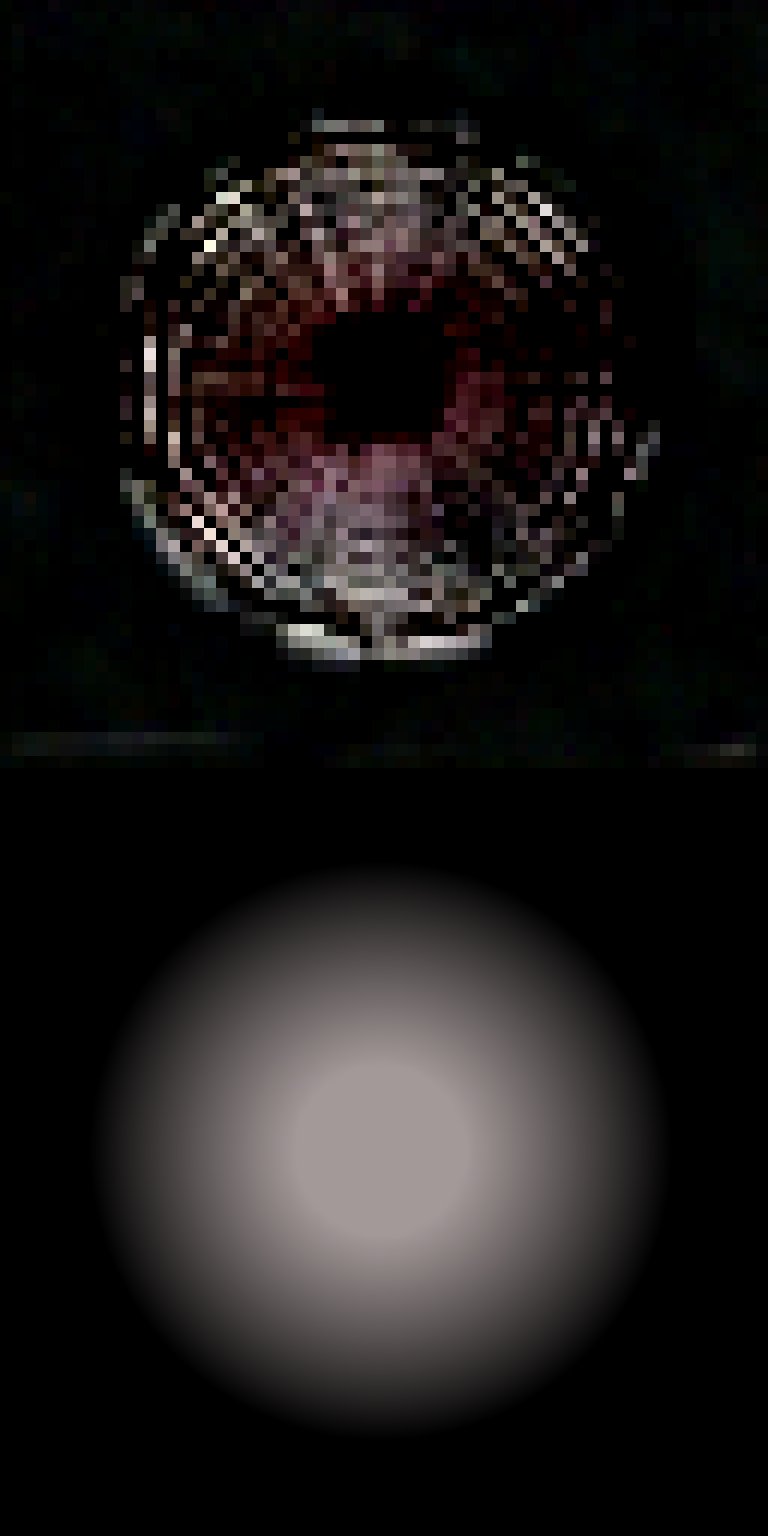} &
    \includegraphics[width=0.06\linewidth]{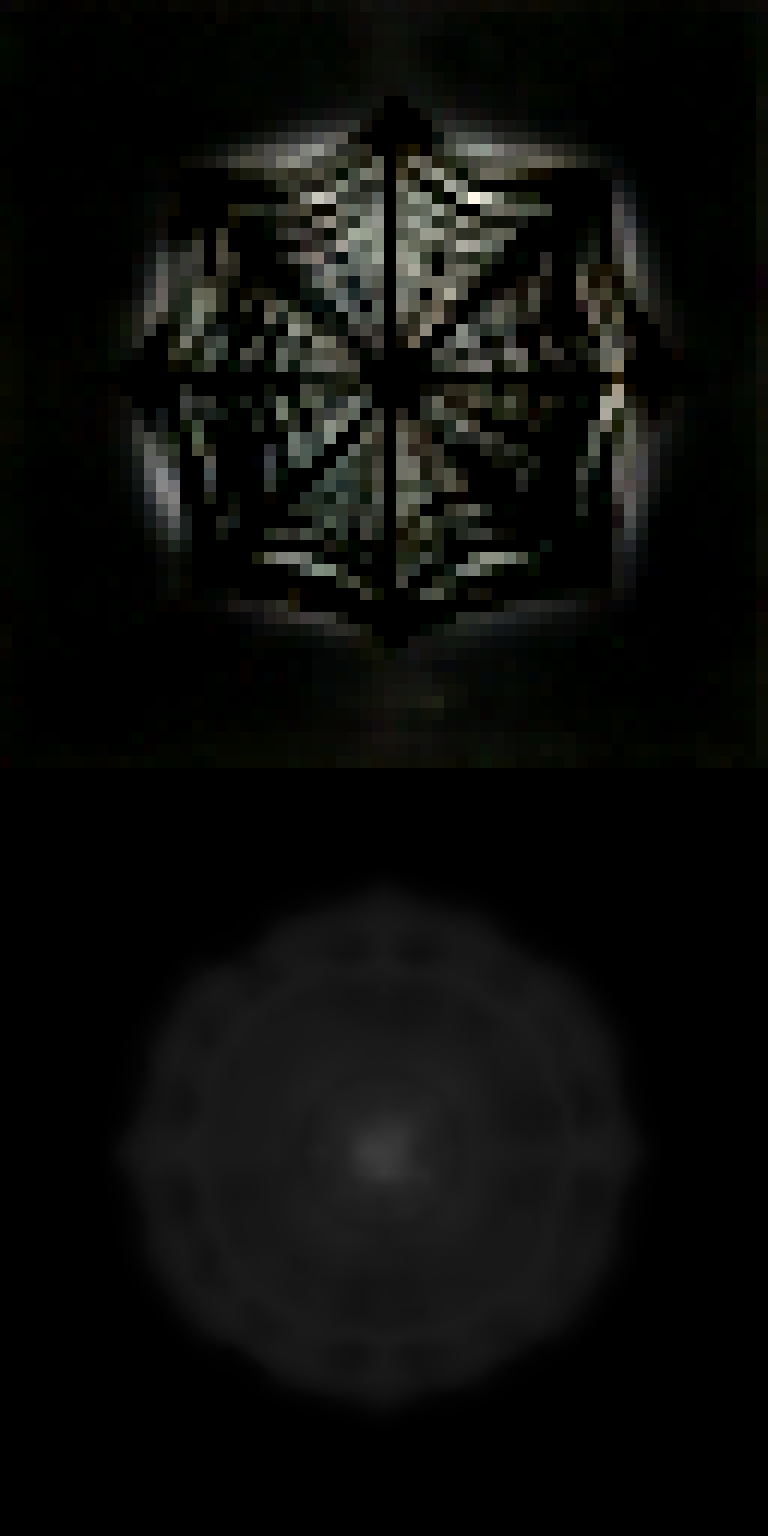} &
    \includegraphics[width=0.06\linewidth]{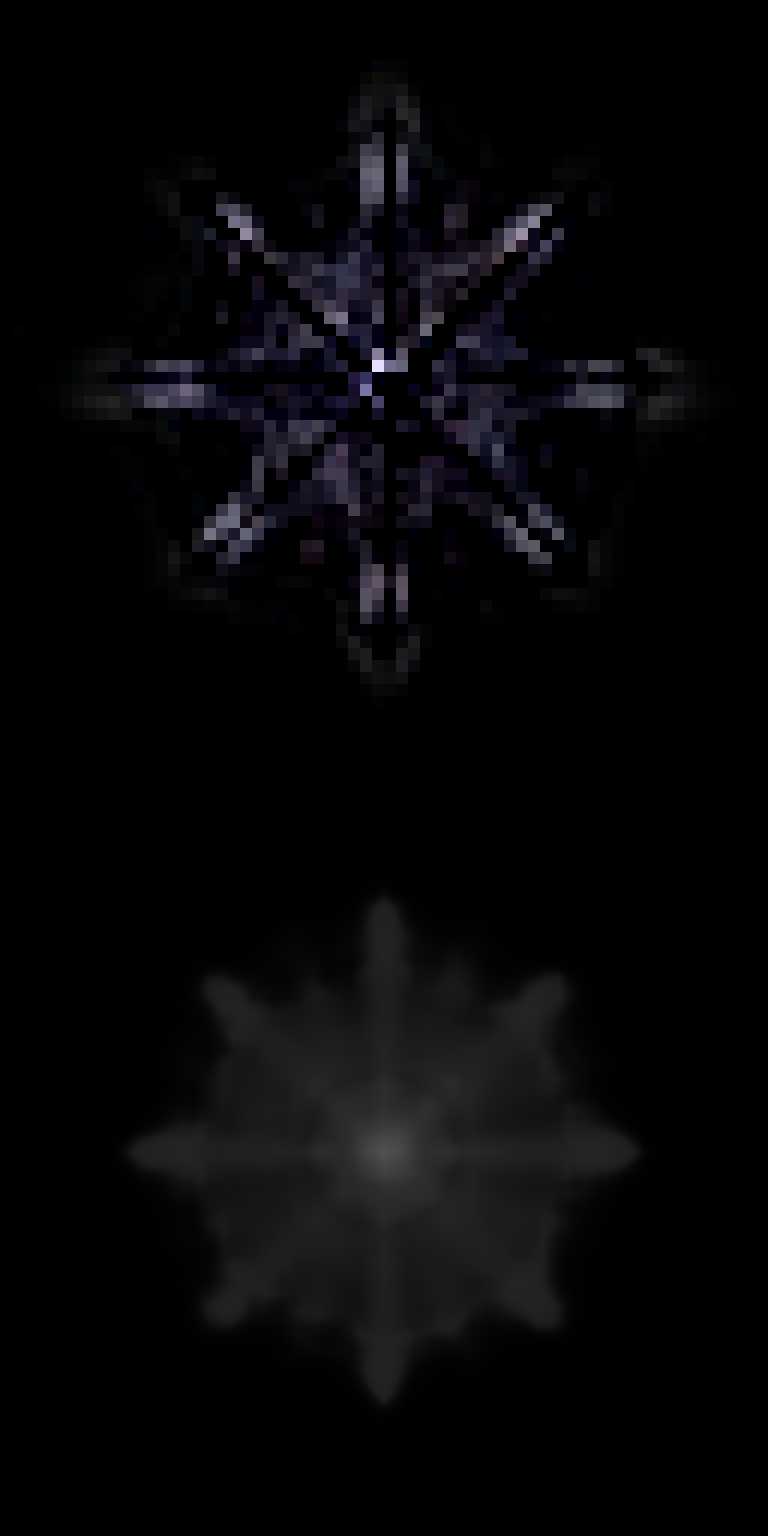} &
    \includegraphics[width=0.06\linewidth]{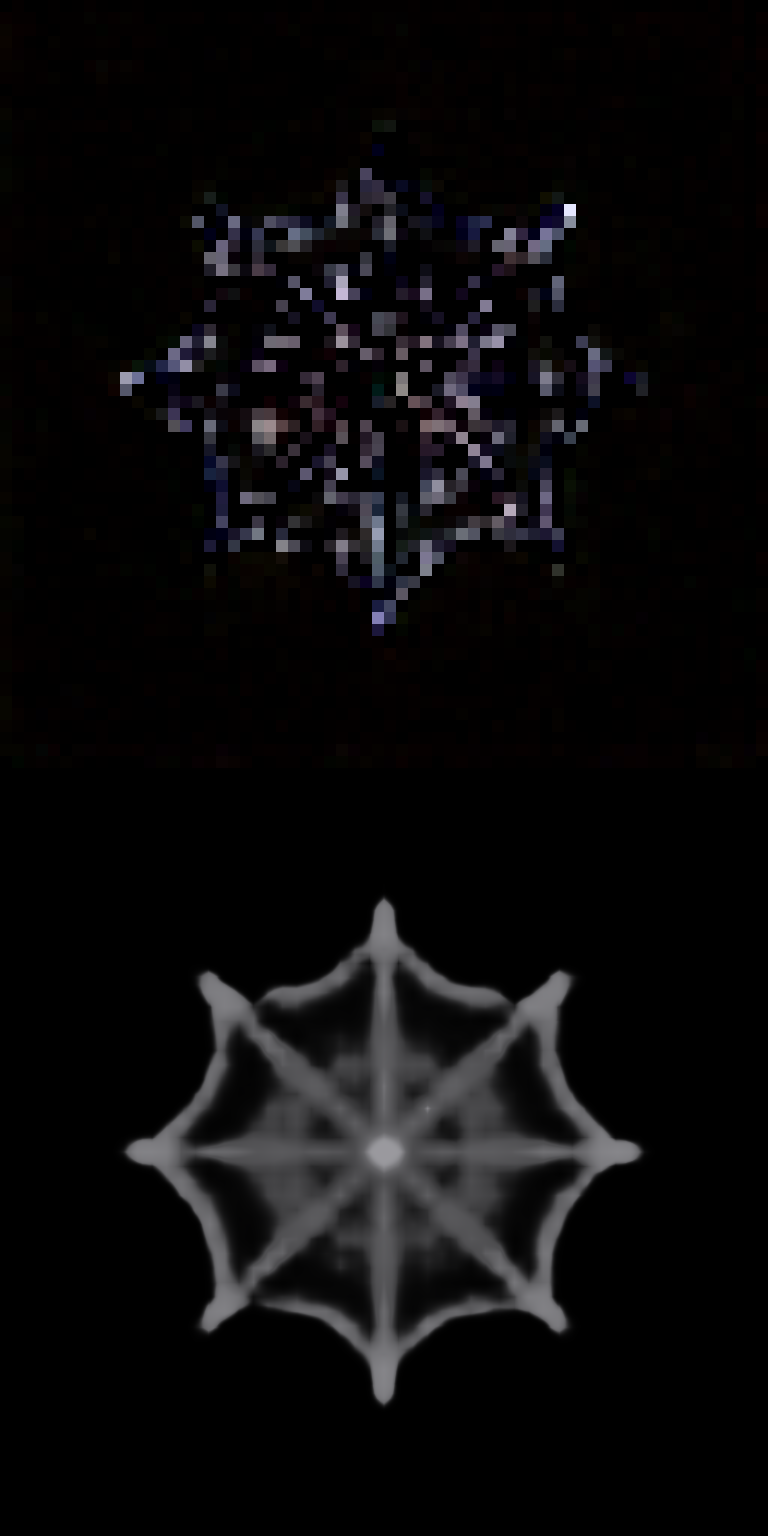} &
    \includegraphics[width=0.06\linewidth]{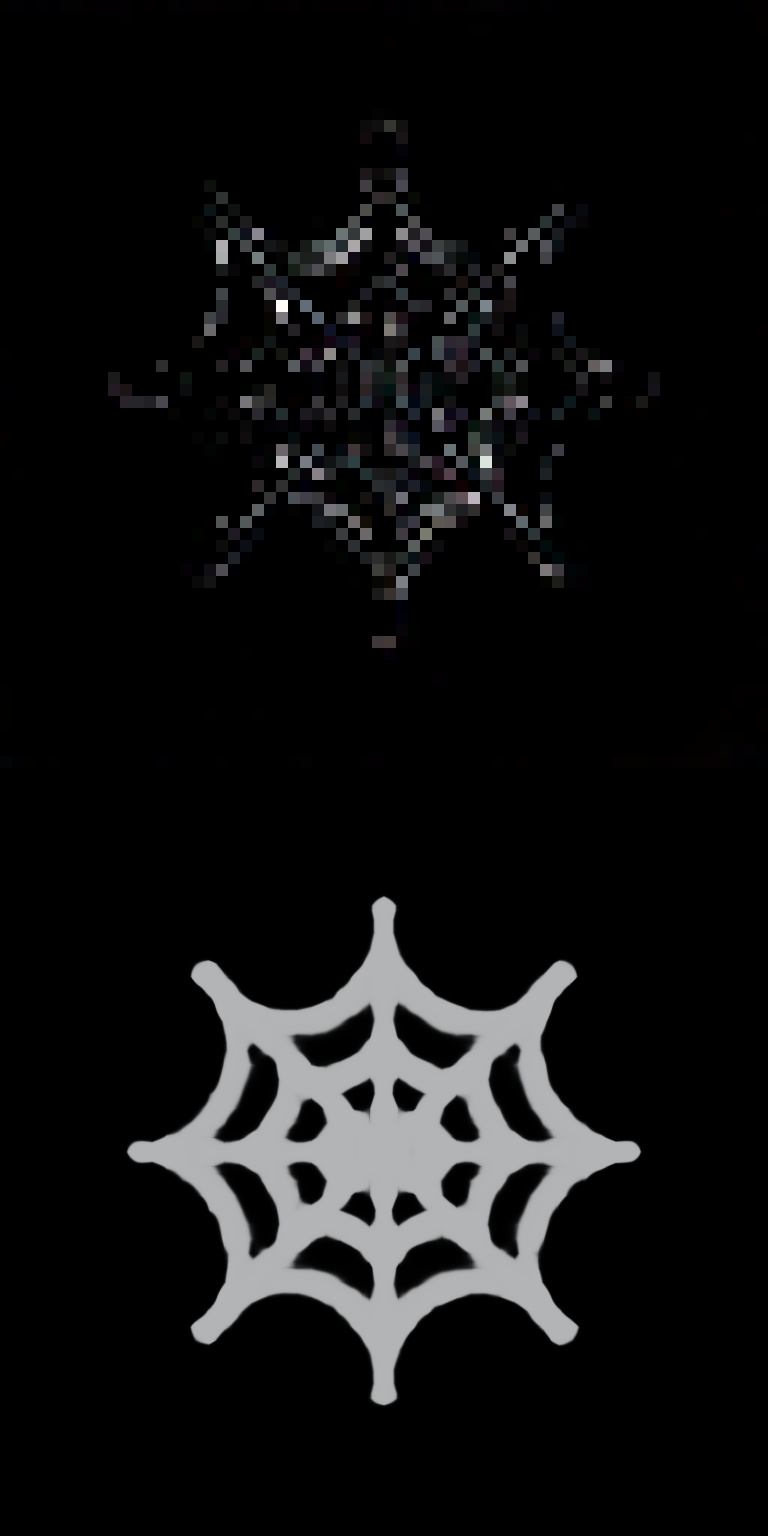} &
    \includegraphics[width=0.06\linewidth]{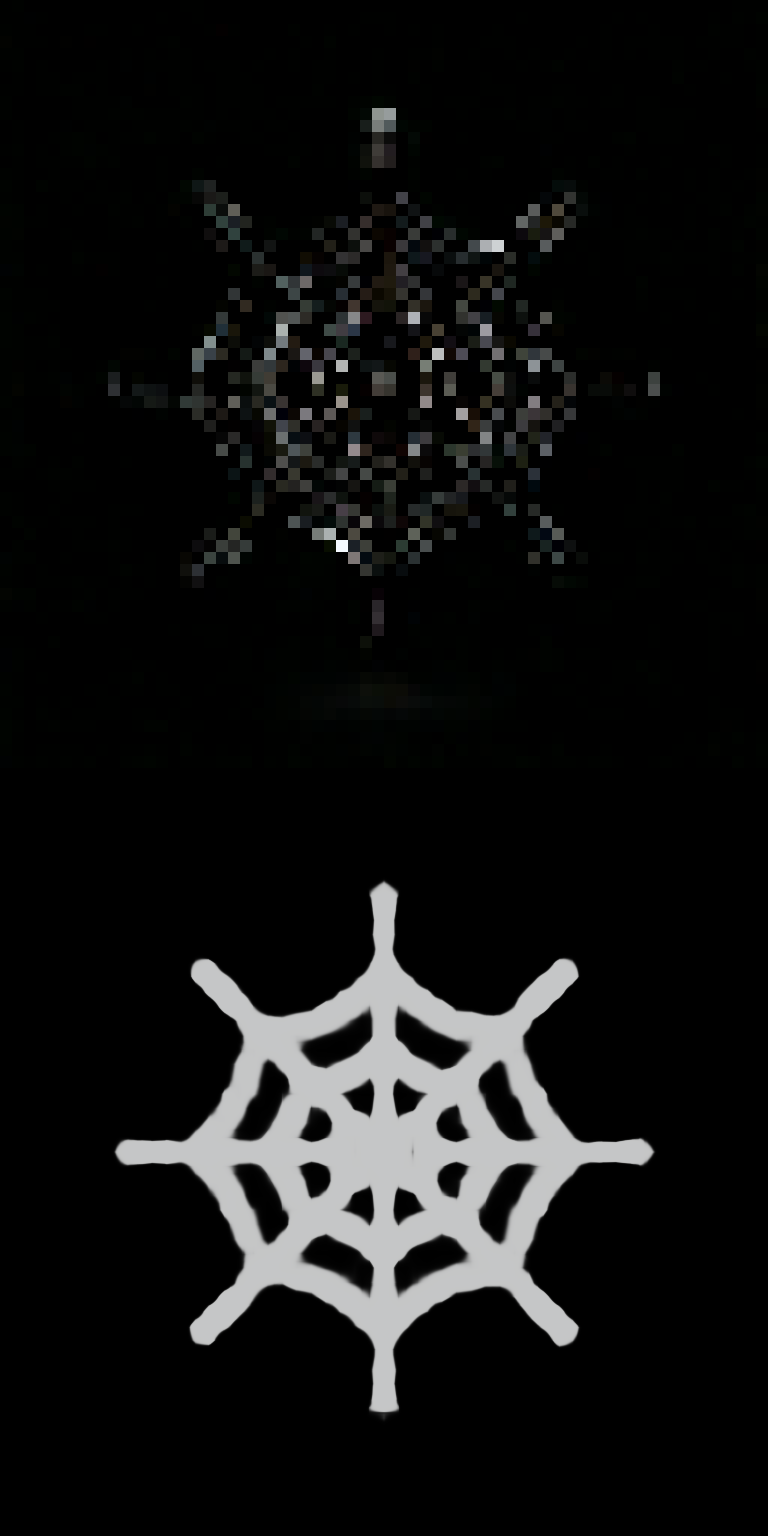} &
    \includegraphics[width=0.06\linewidth]{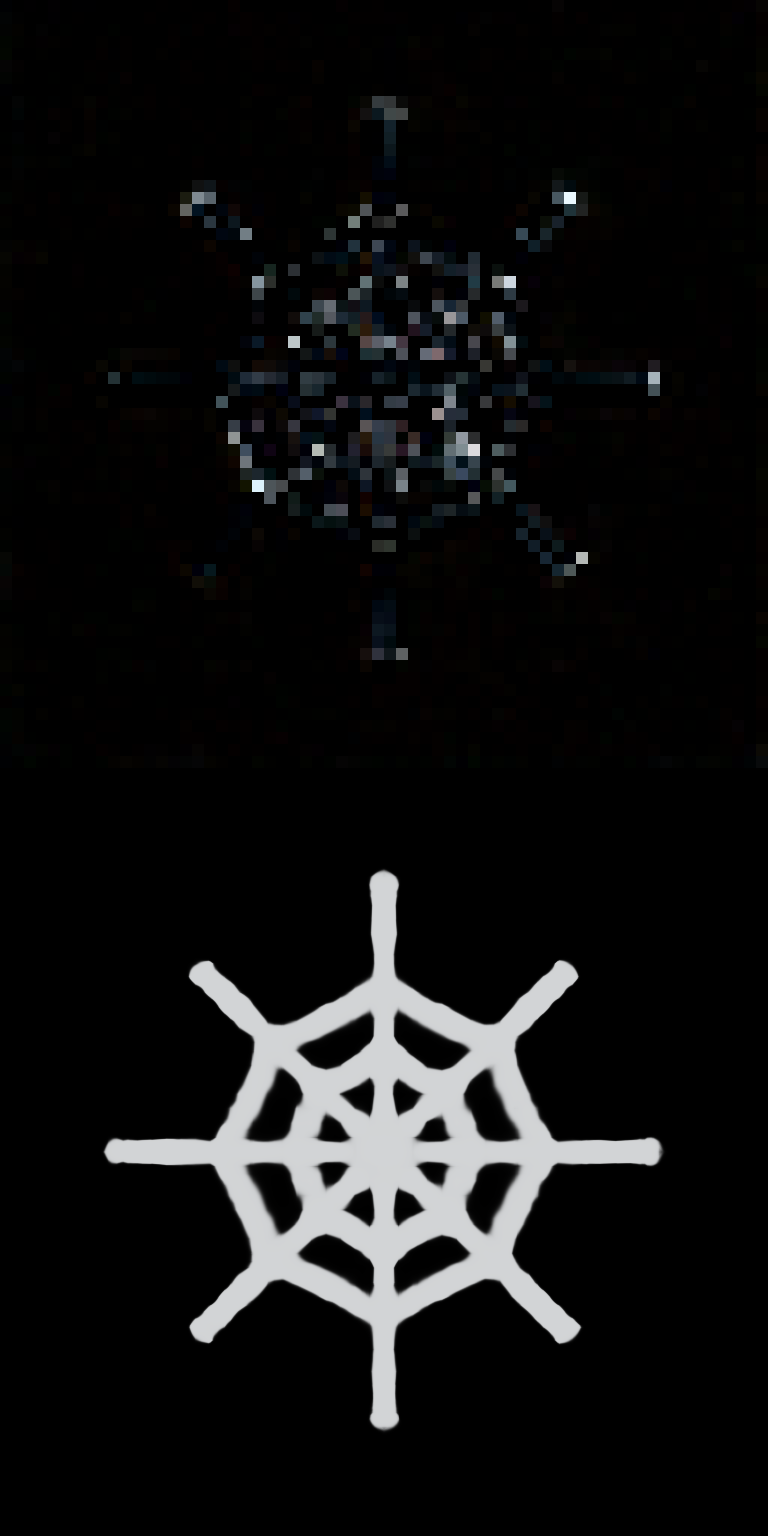} &
    \includegraphics[width=0.06\linewidth]{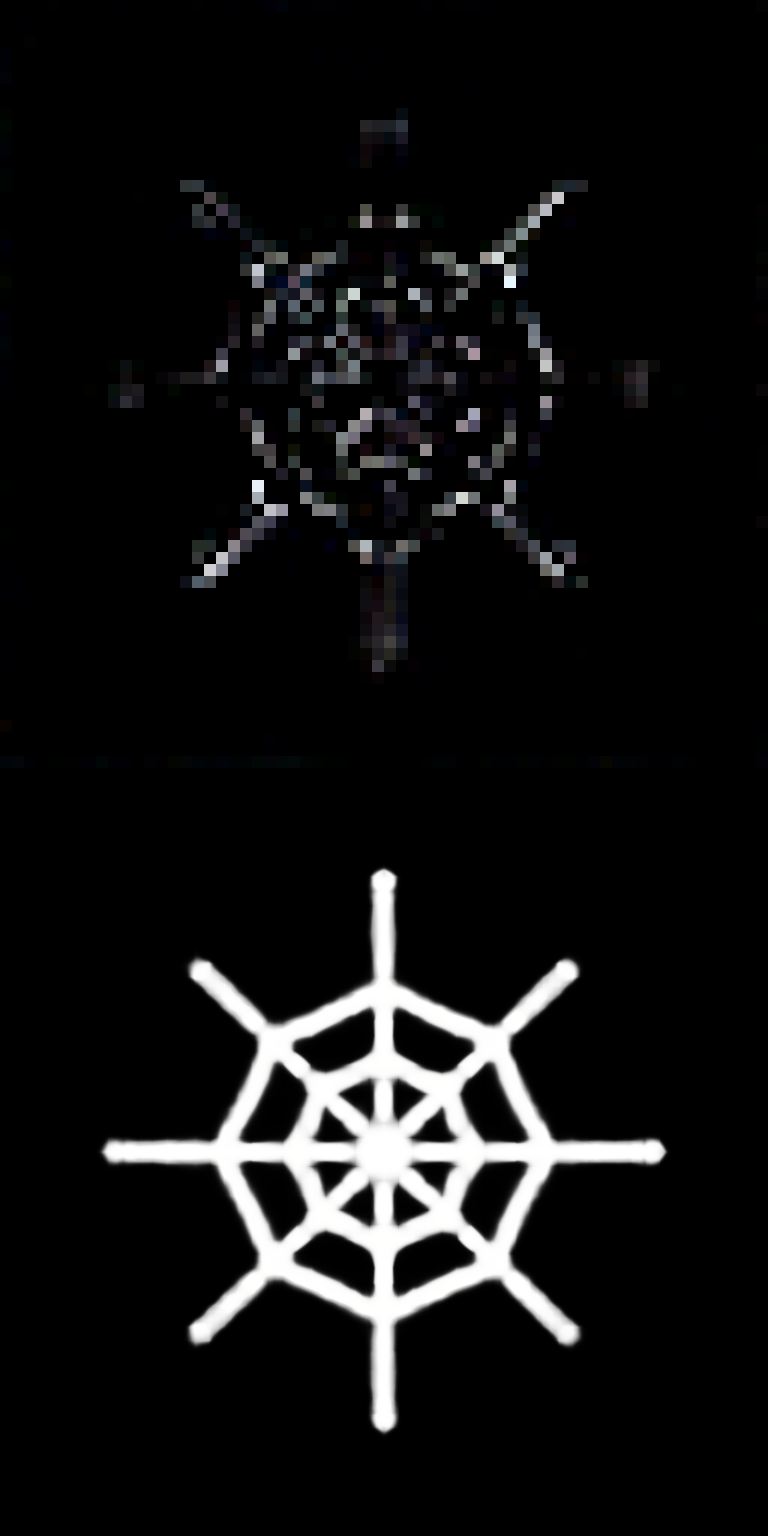} &
    \includegraphics[width=0.06\linewidth]{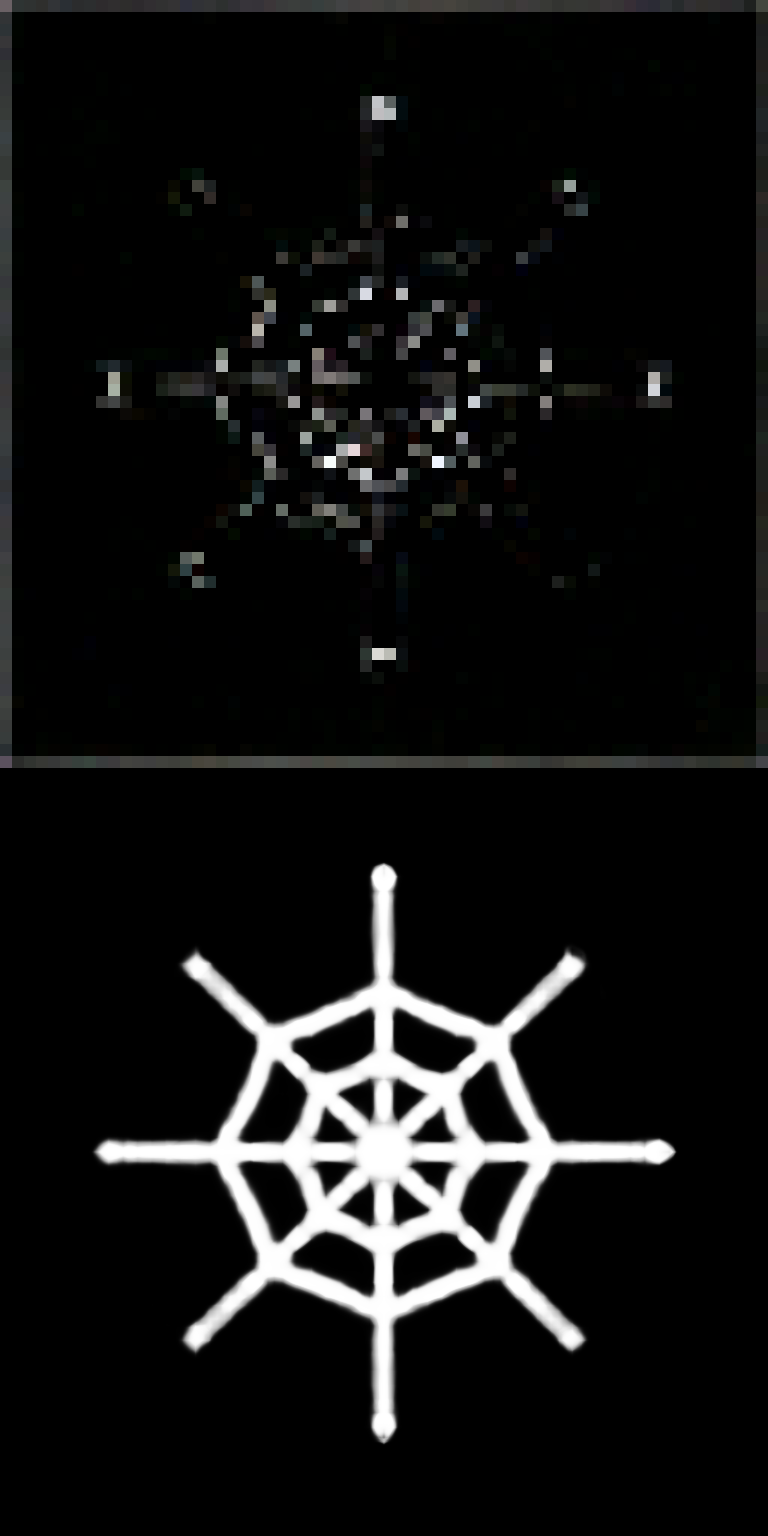} &
    \includegraphics[width=0.06\linewidth]{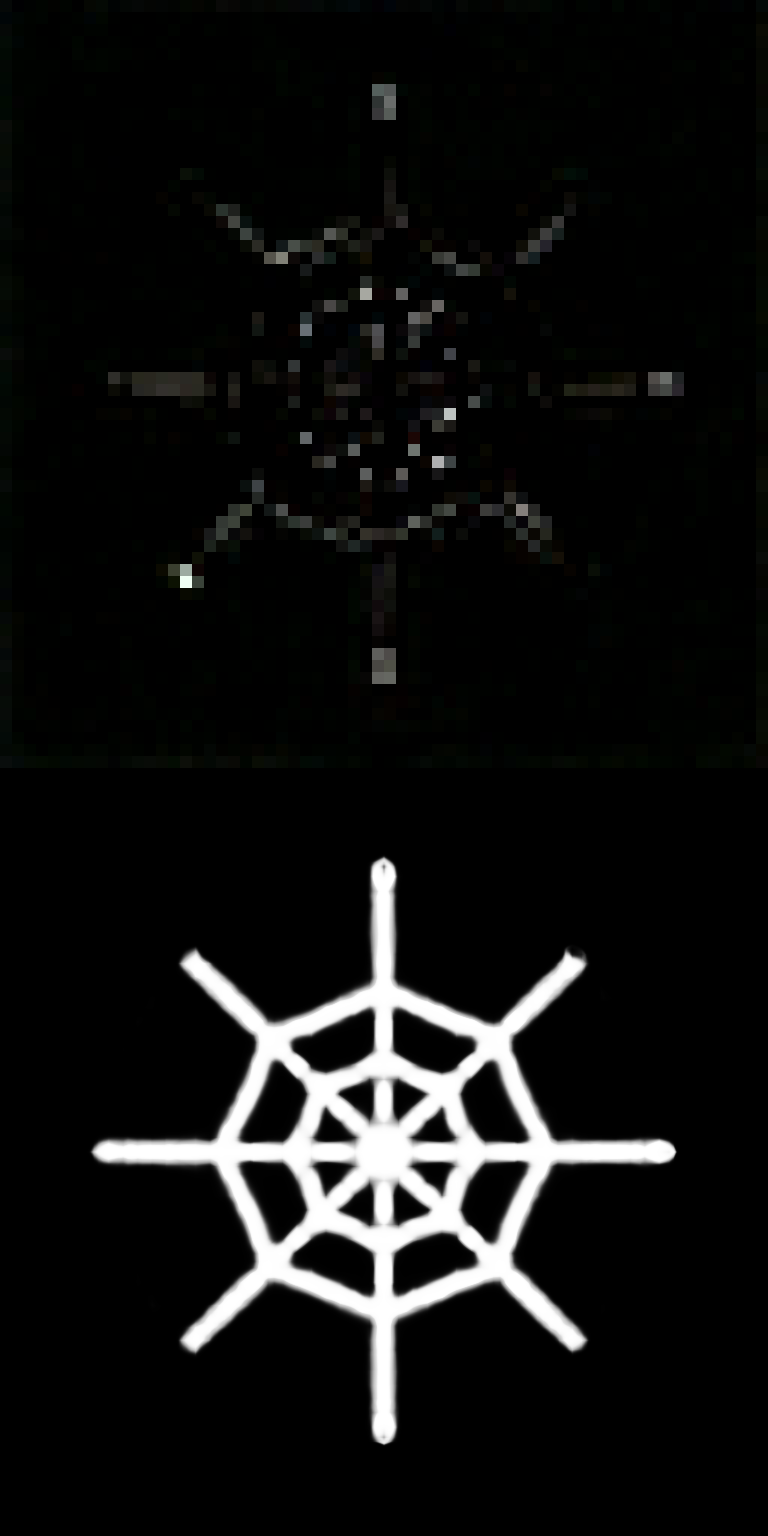} &
    \includegraphics[width=0.06\linewidth]{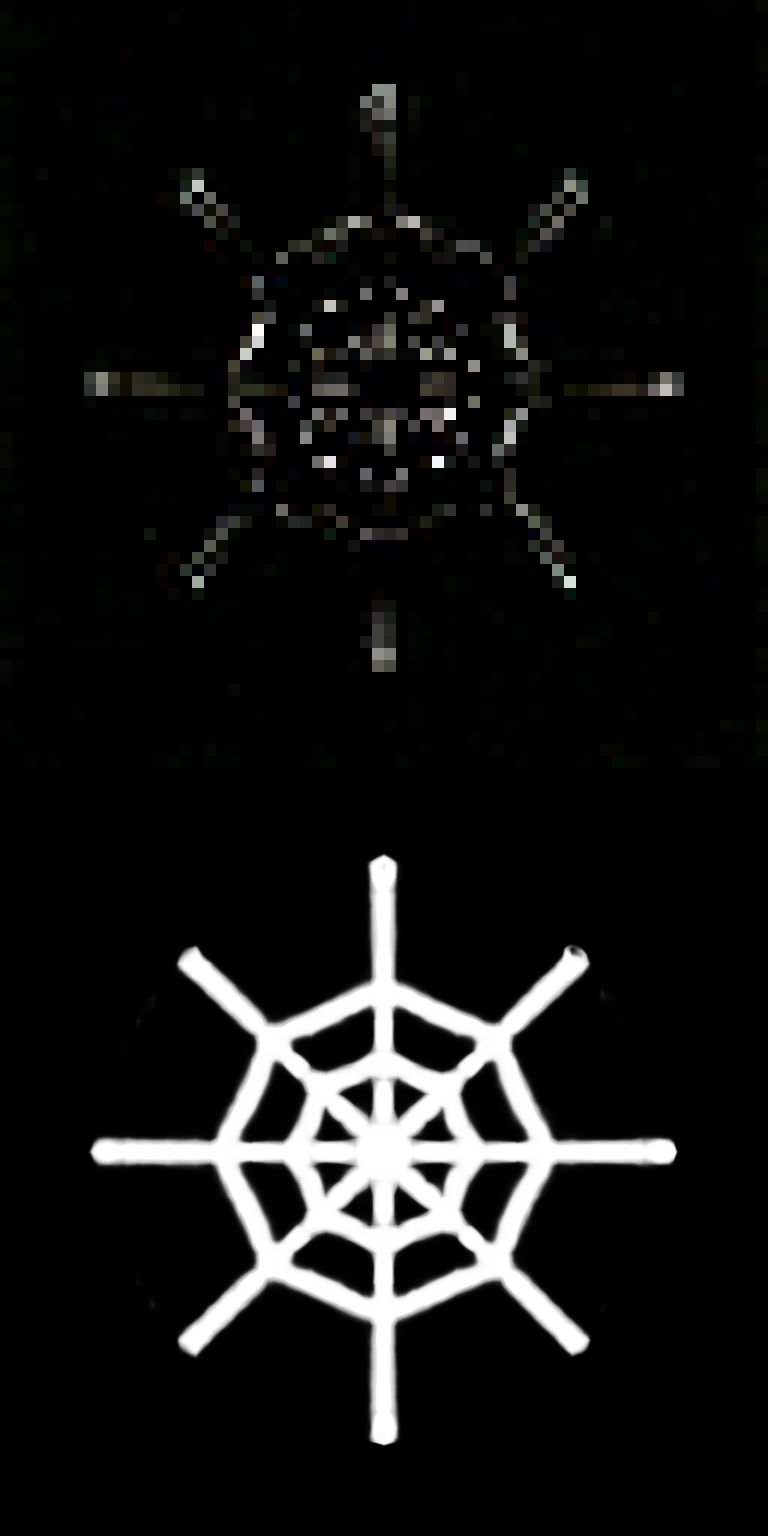} &
    \includegraphics[width=0.06\linewidth]{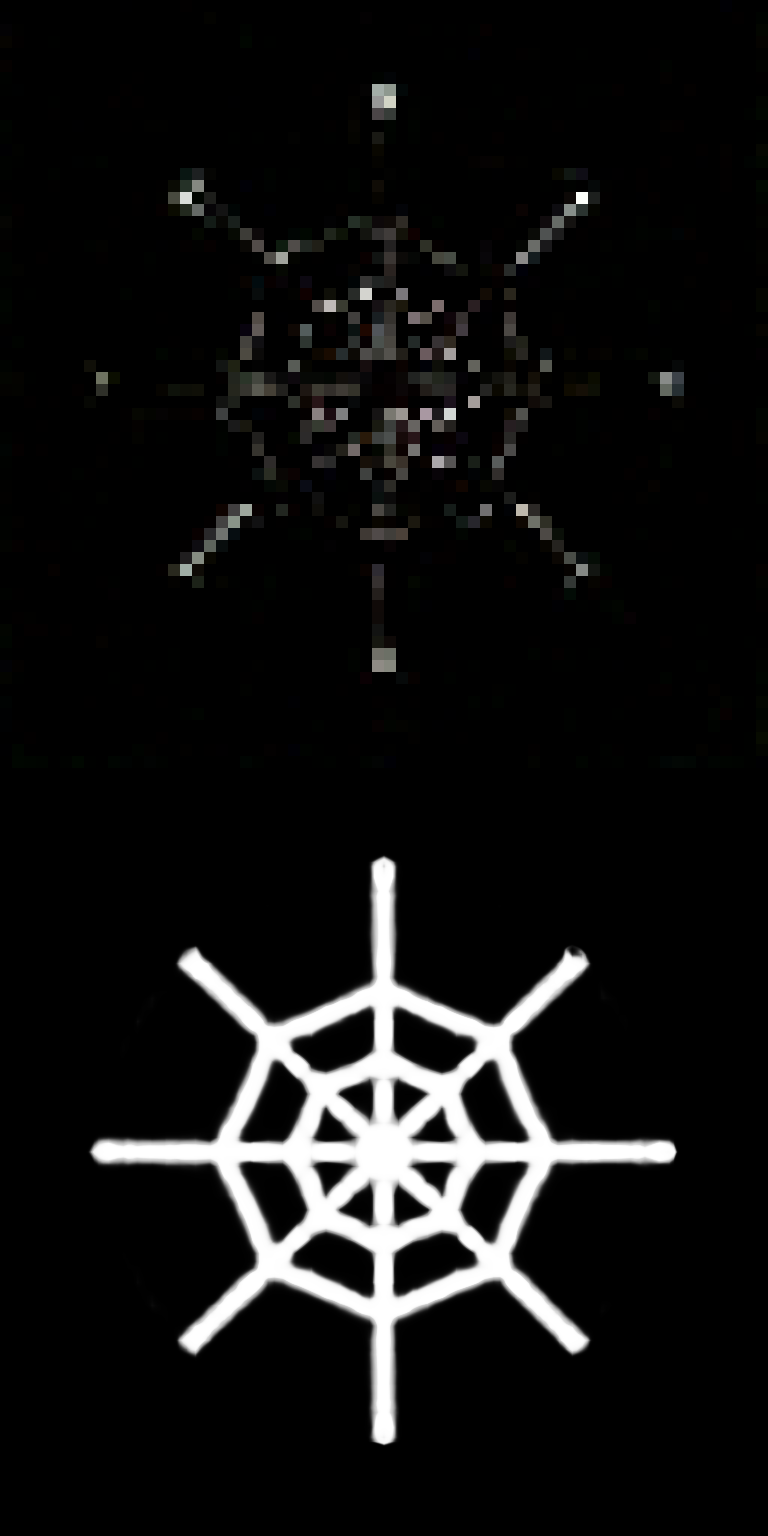} &
    \includegraphics[width=0.06\linewidth]{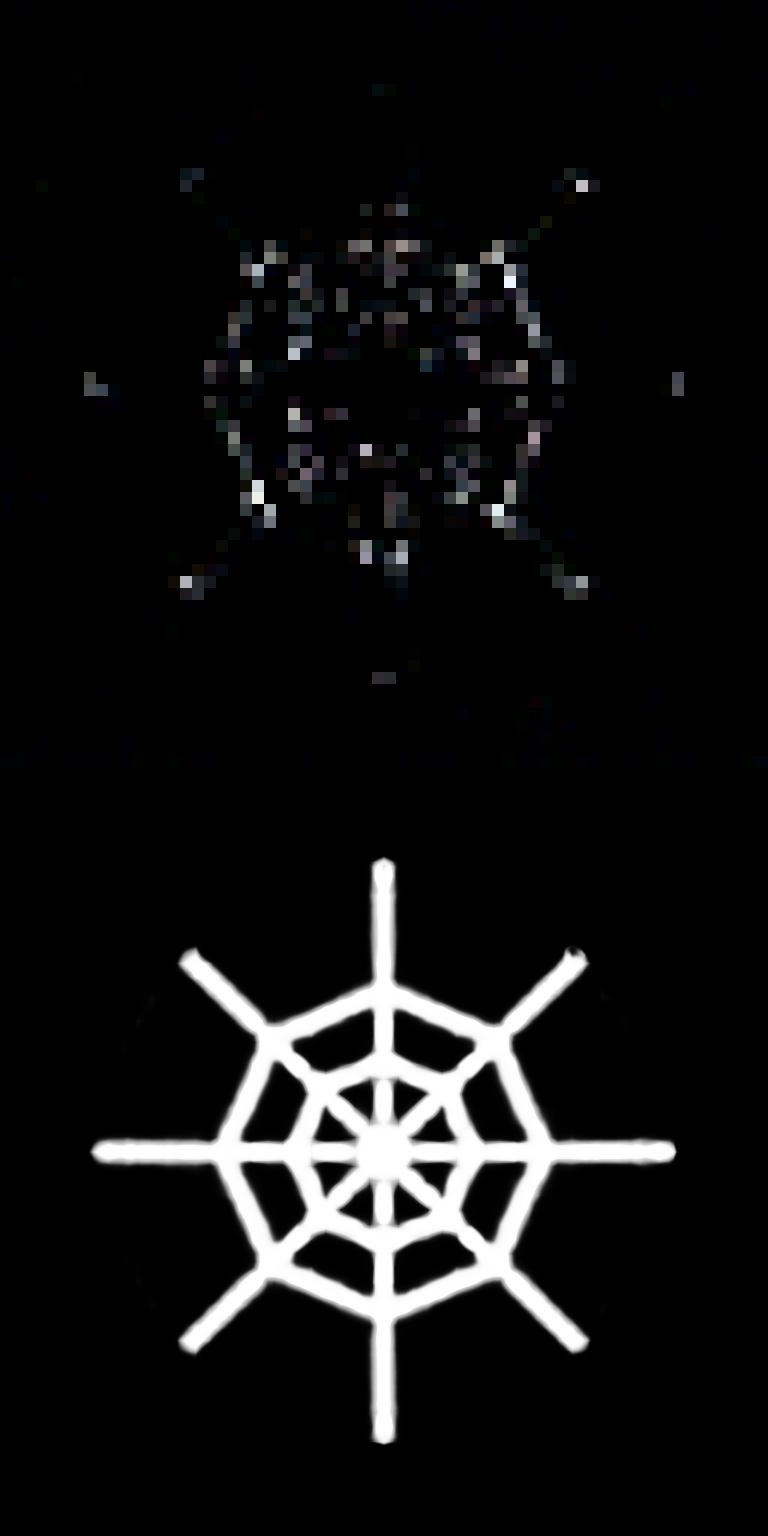} &
    \includegraphics[width=0.06\linewidth]{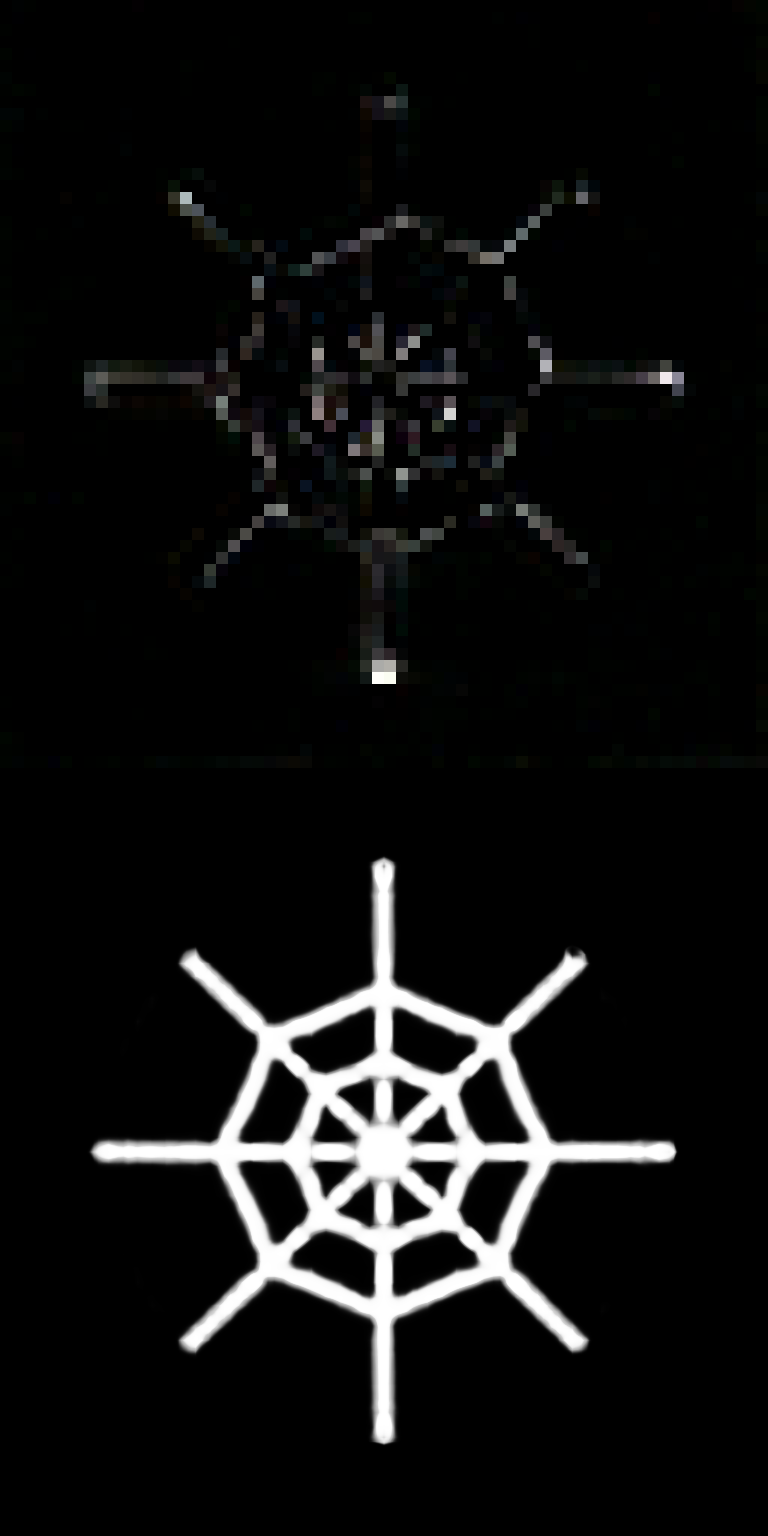} &
    \includegraphics[width=0.06\linewidth]{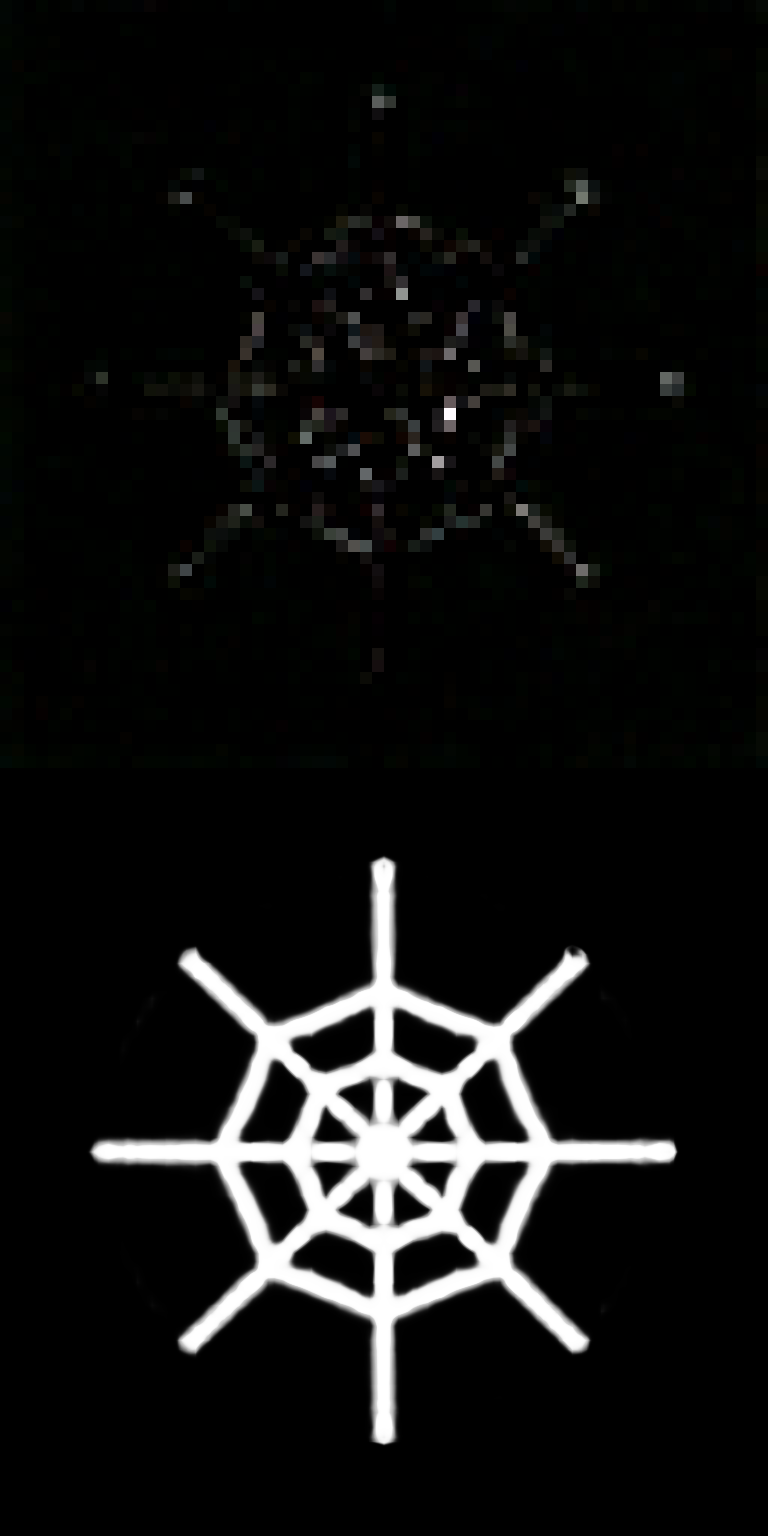} &
    \includegraphics[width=0.06\linewidth]{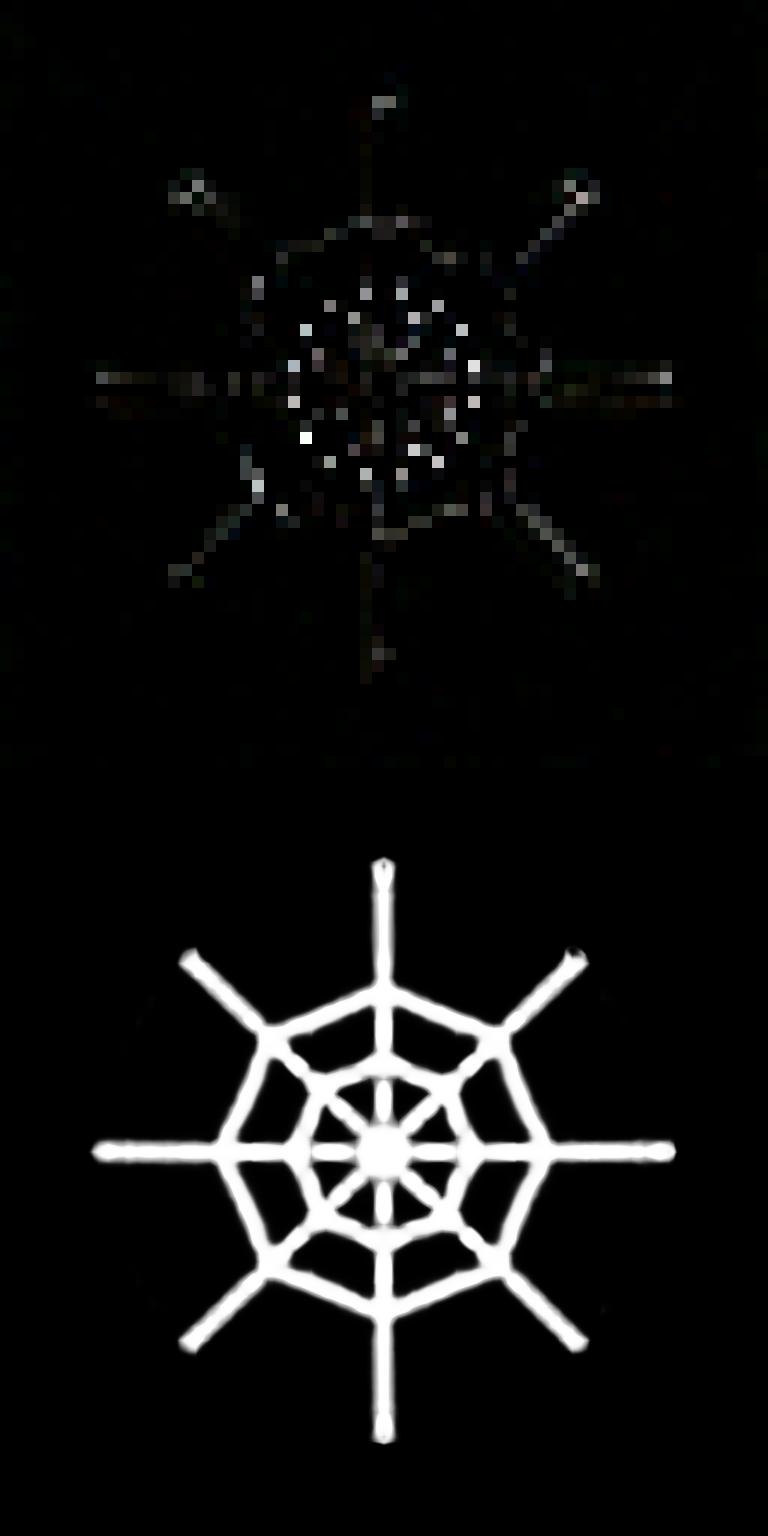} \\
    
    \footnotesize{\emph{}}&
    \multicolumn{15}{c}{\emph{\textcolor{textprompts}{"A yellow swiss cheese with holes ..."}}}
    \\
    \includegraphics[width=0.06\linewidth]{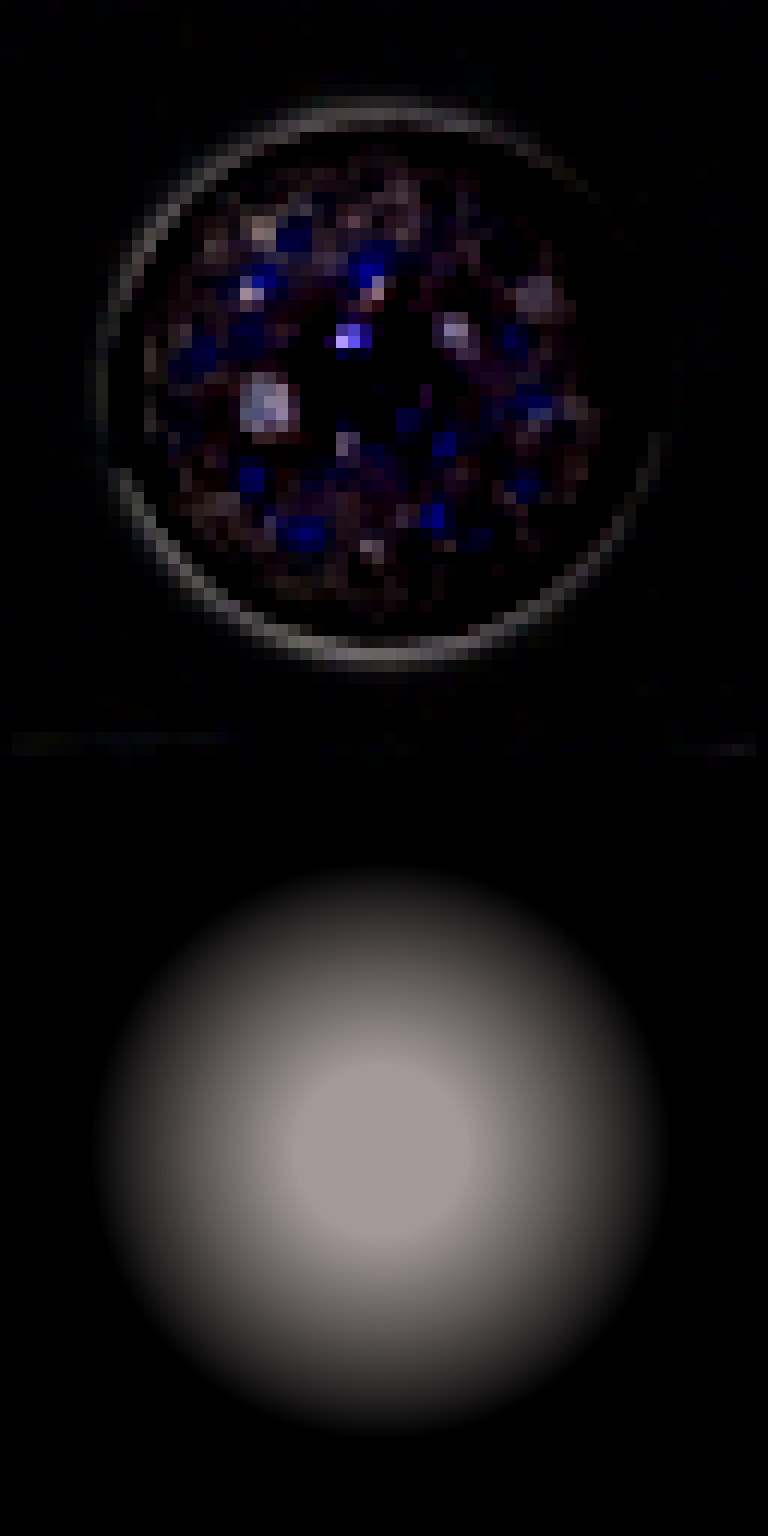} &
    \includegraphics[width=0.06\linewidth]{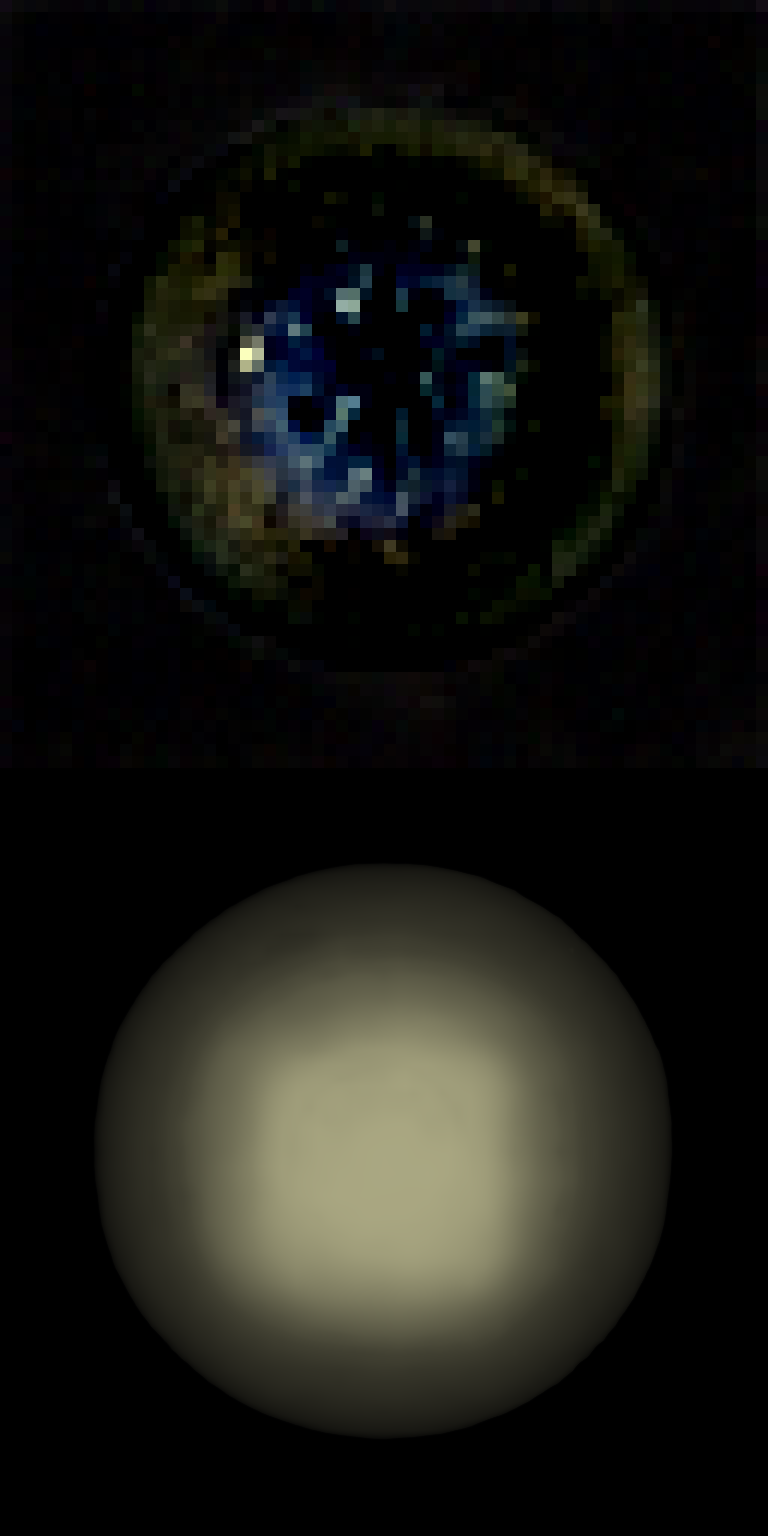} &
    \includegraphics[width=0.06\linewidth]{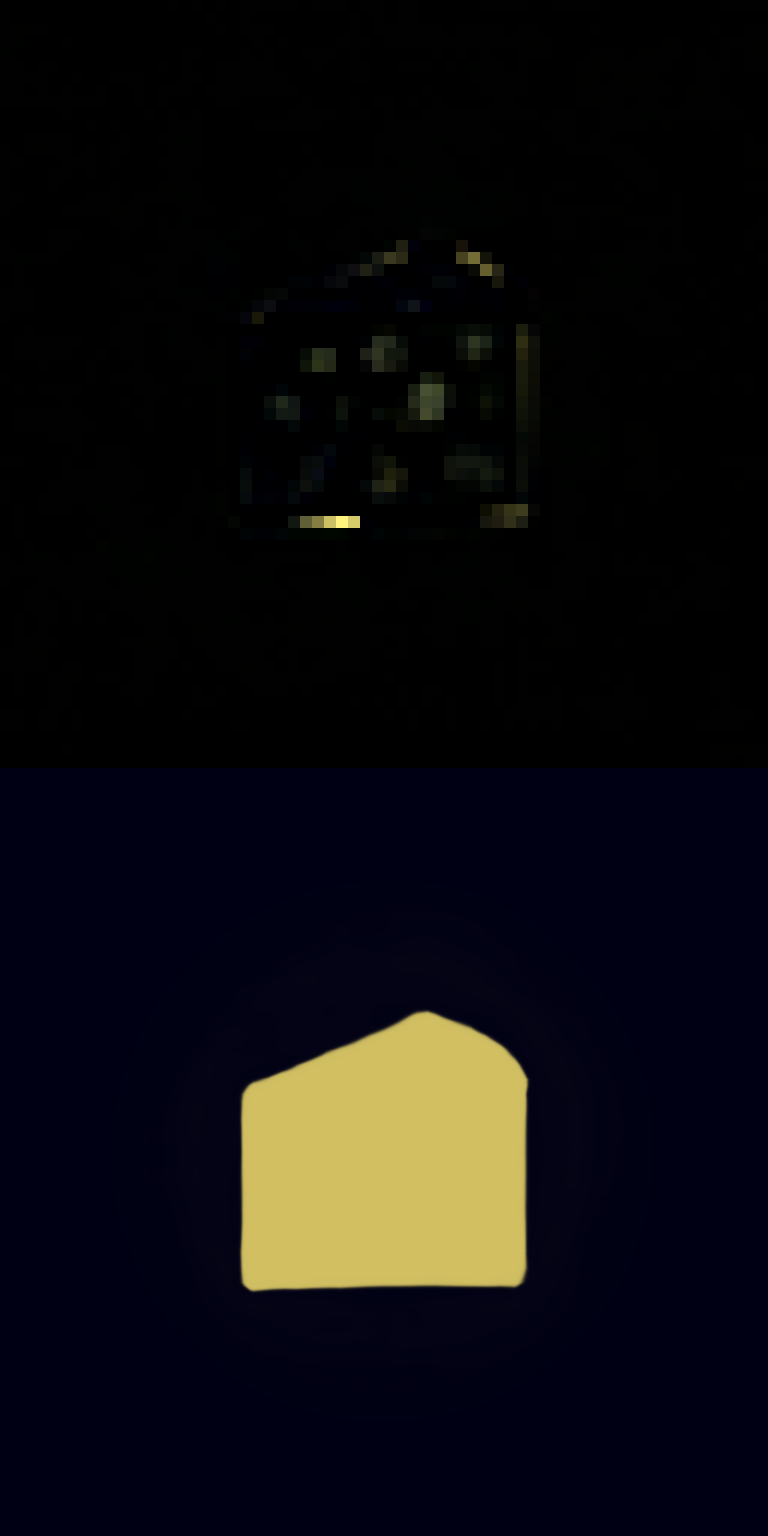} &
    \includegraphics[width=0.06\linewidth]{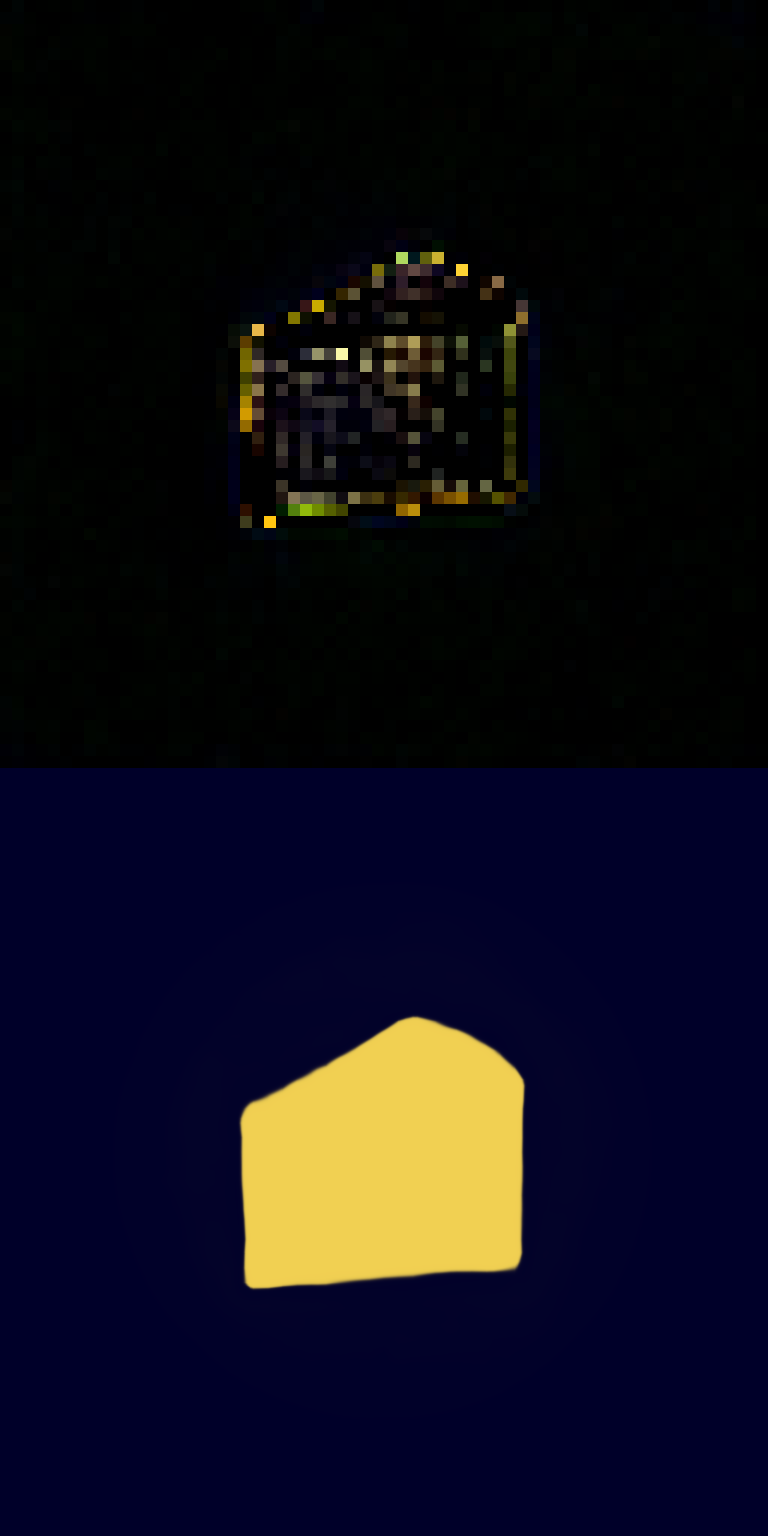} &
    \includegraphics[width=0.06\linewidth]{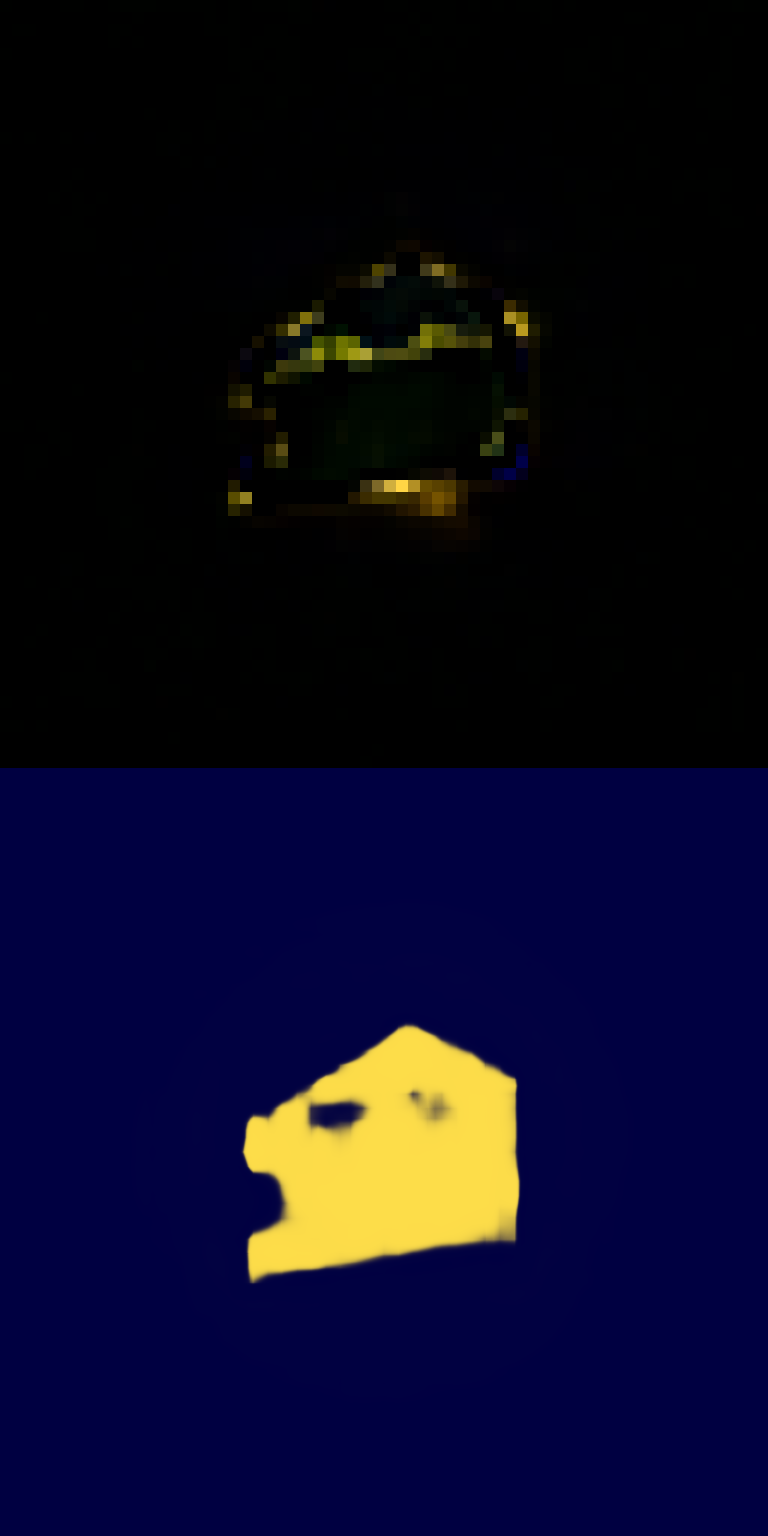} &
    \includegraphics[width=0.06\linewidth]{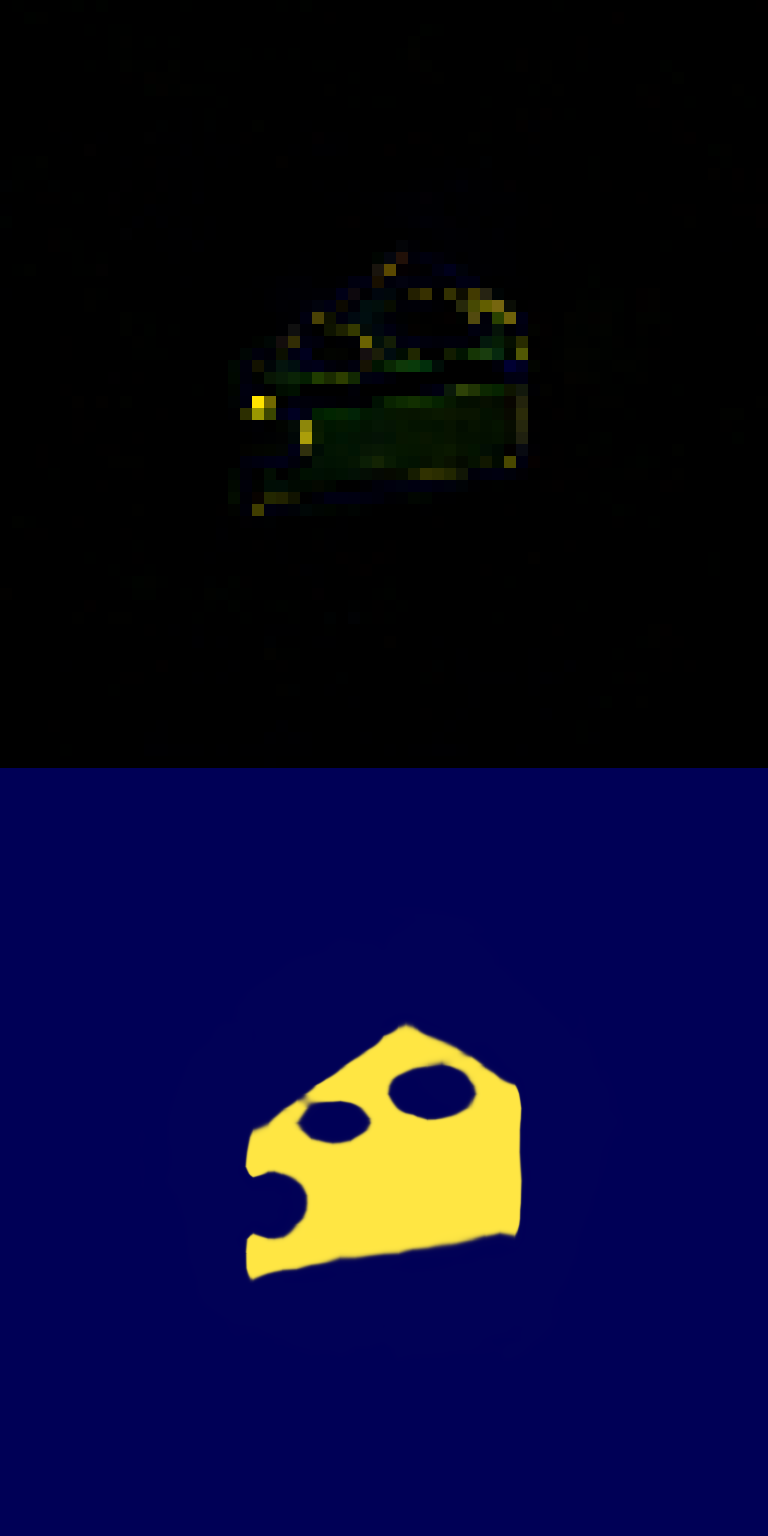} &
    \includegraphics[width=0.06\linewidth]{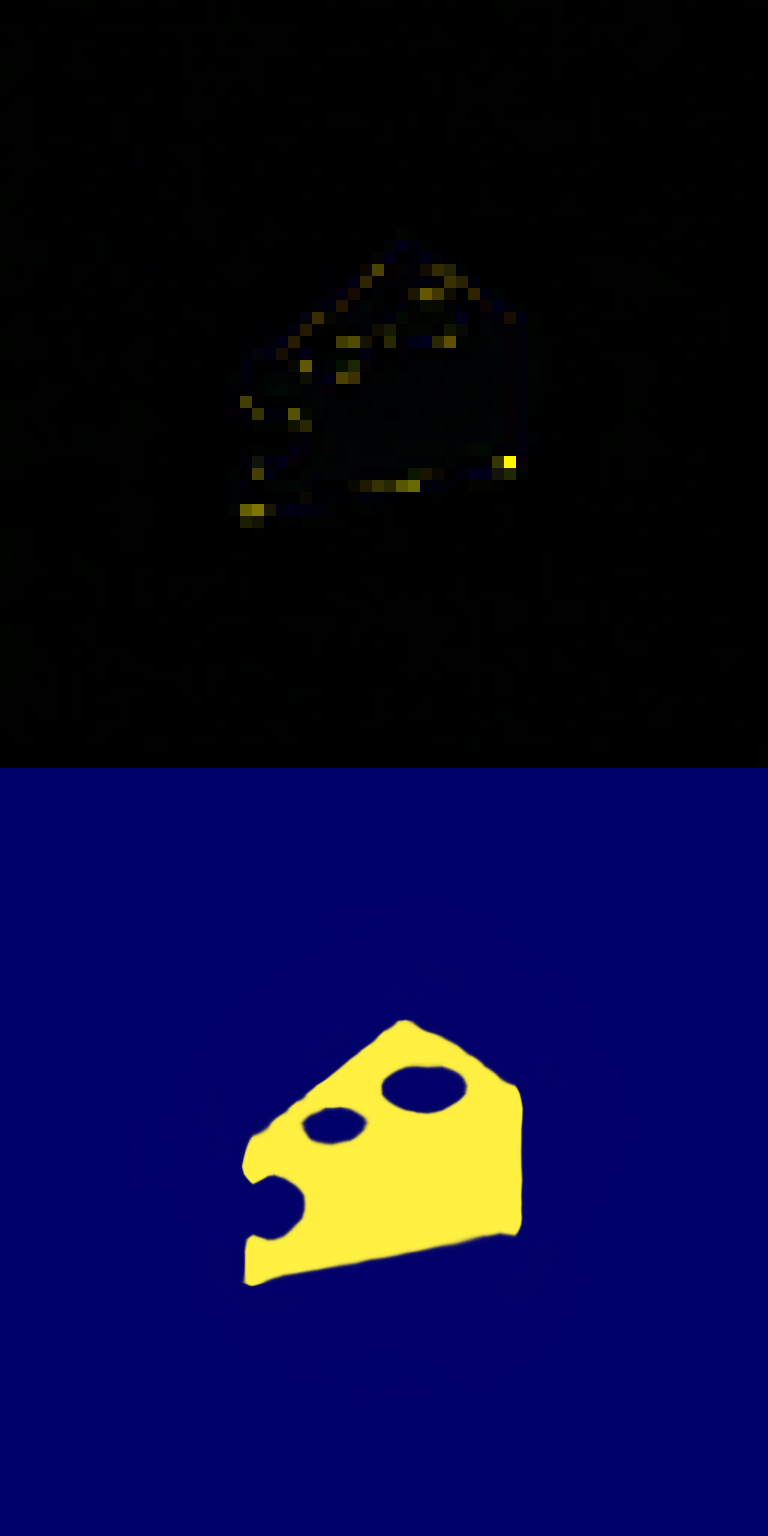} &
    \includegraphics[width=0.06\linewidth]{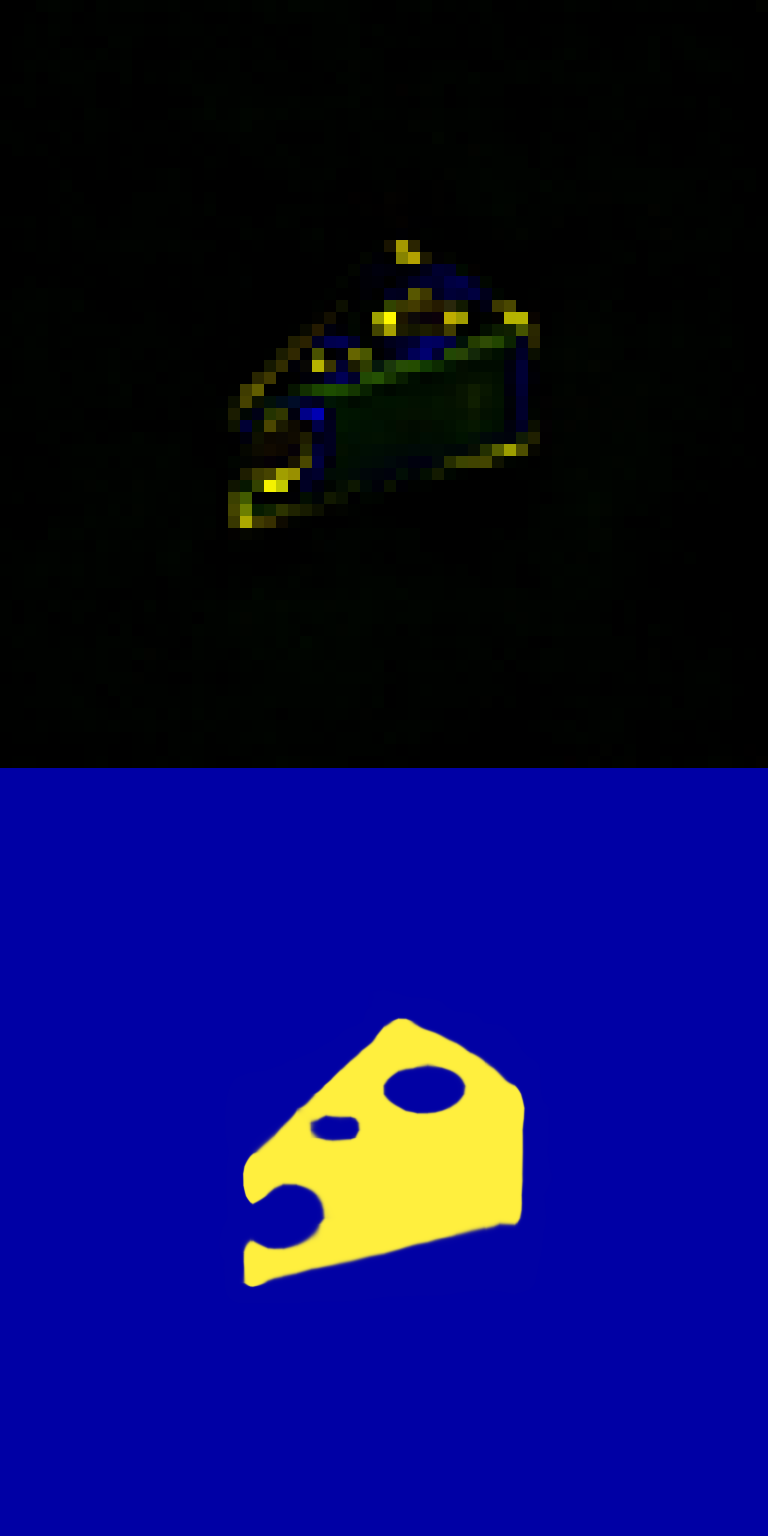} &
    \includegraphics[width=0.06\linewidth]{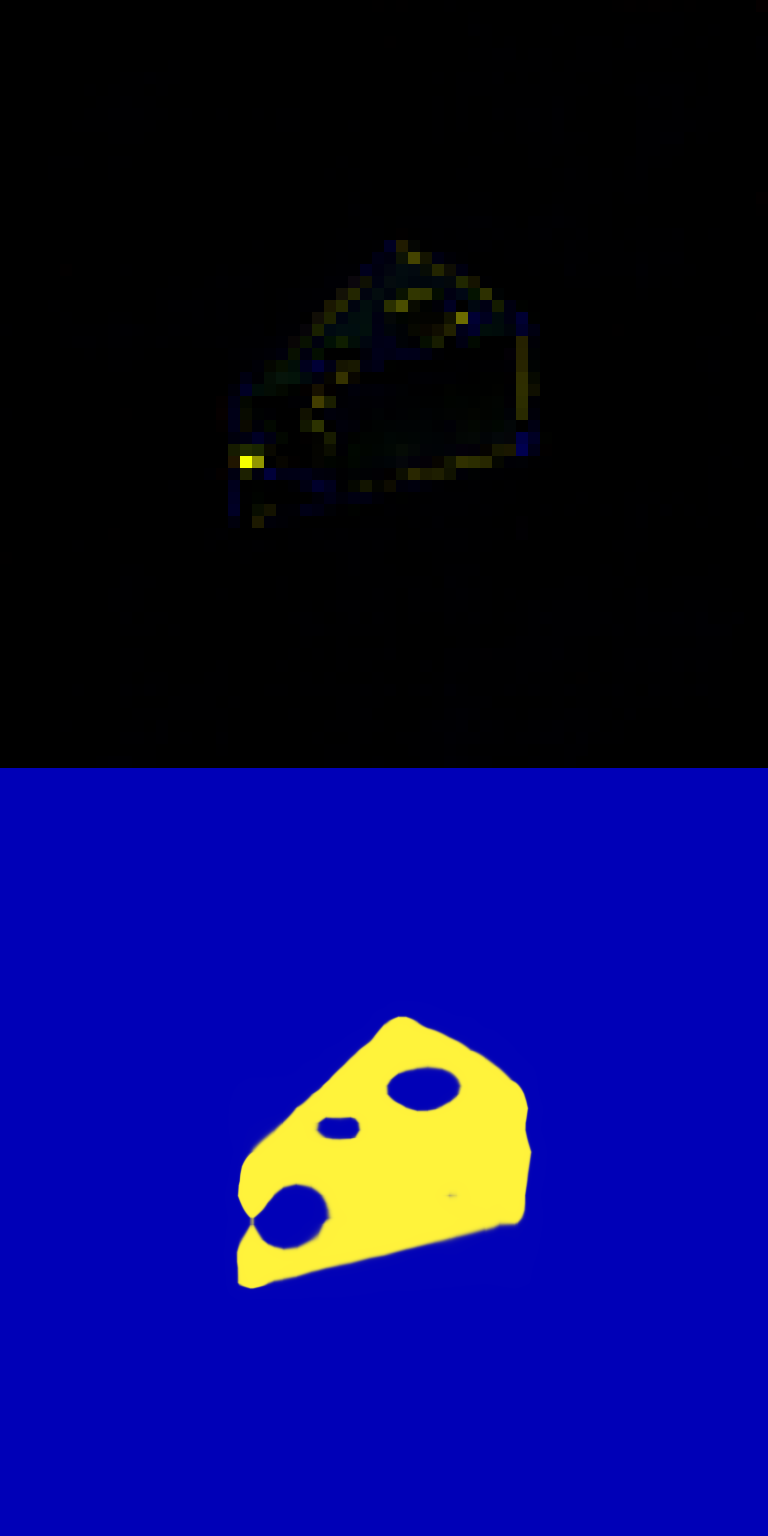} &
    \includegraphics[width=0.06\linewidth]{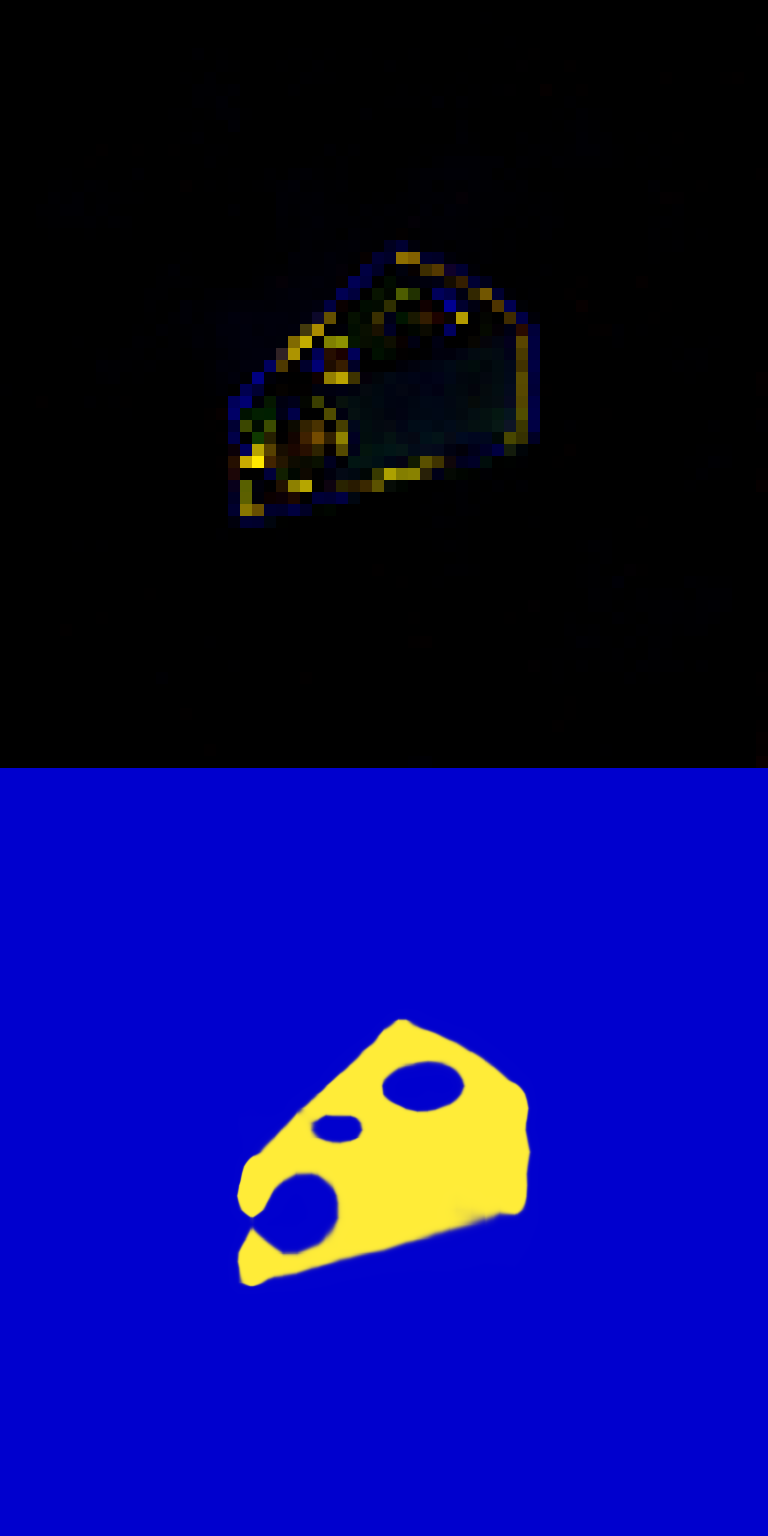} &
    \includegraphics[width=0.06\linewidth]{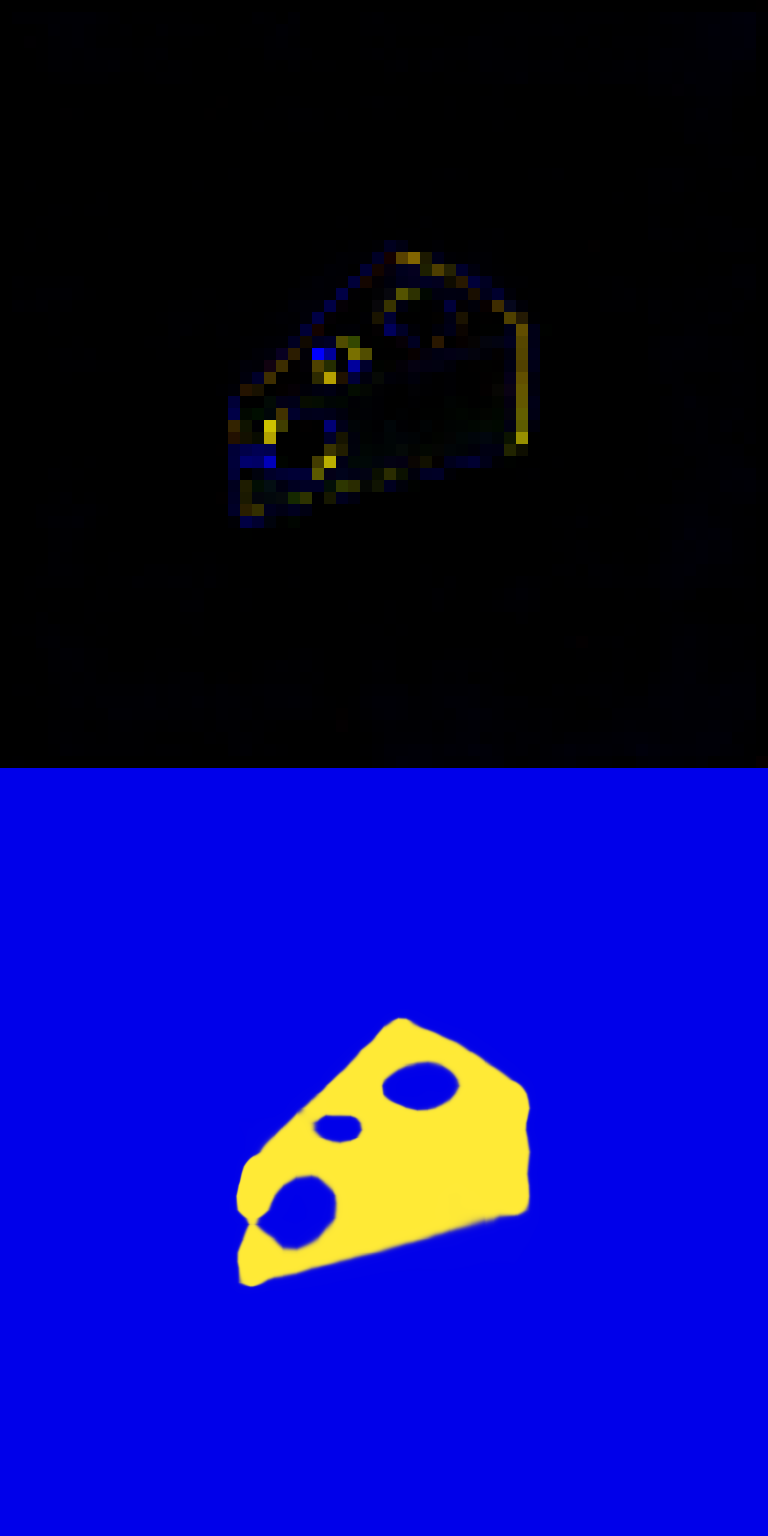} &
    \includegraphics[width=0.06\linewidth]{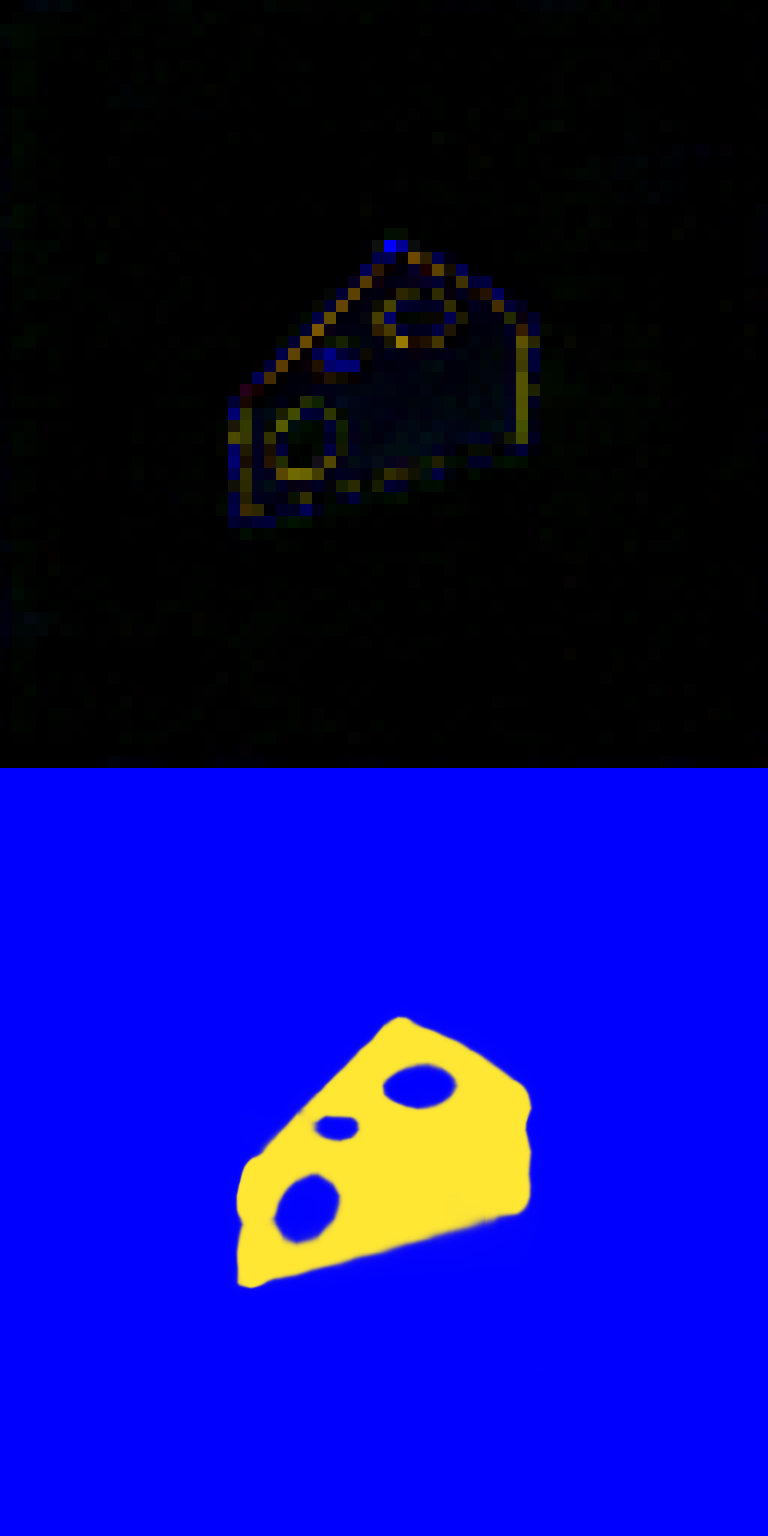} &
    \includegraphics[width=0.06\linewidth]{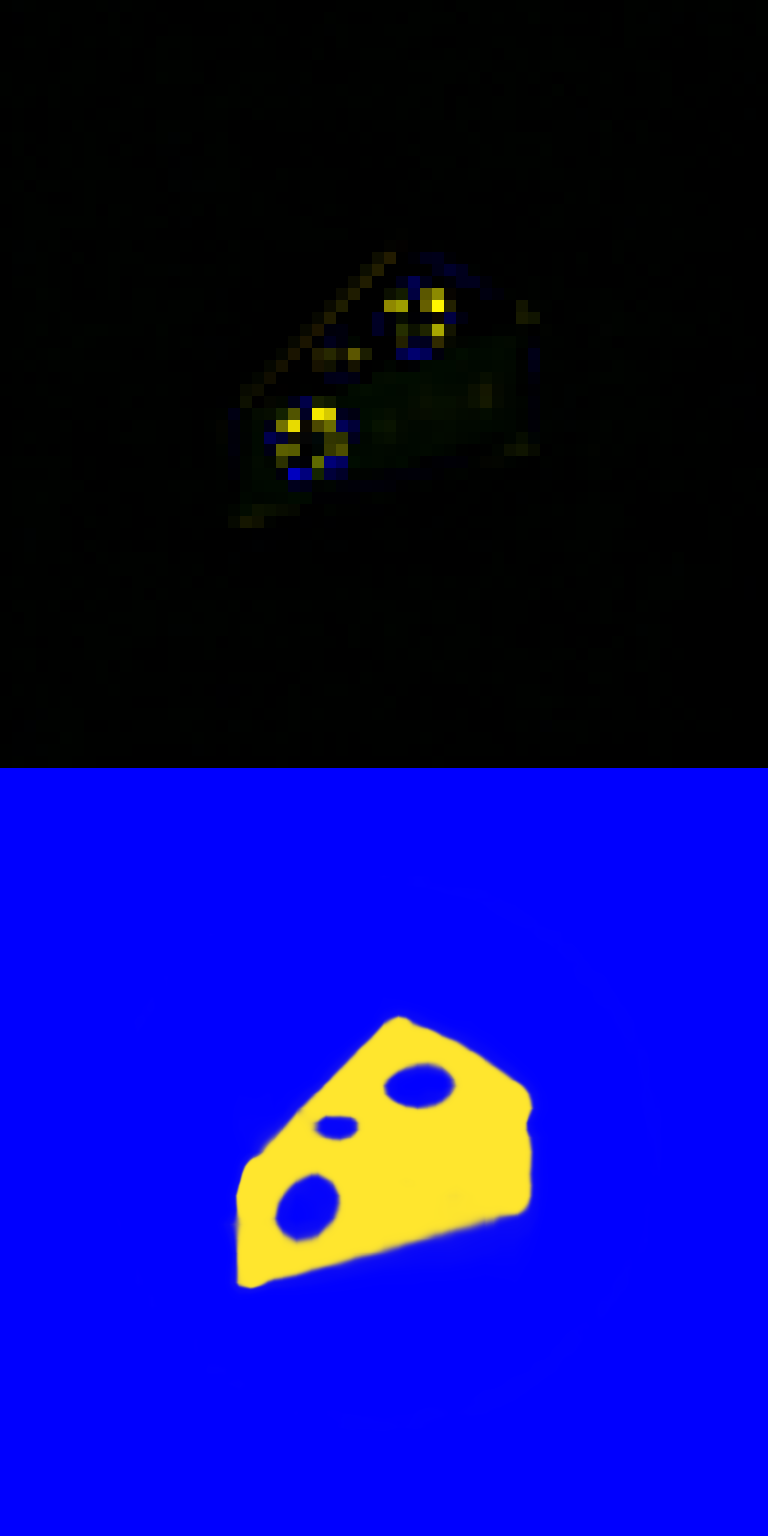} &
    \includegraphics[width=0.06\linewidth]{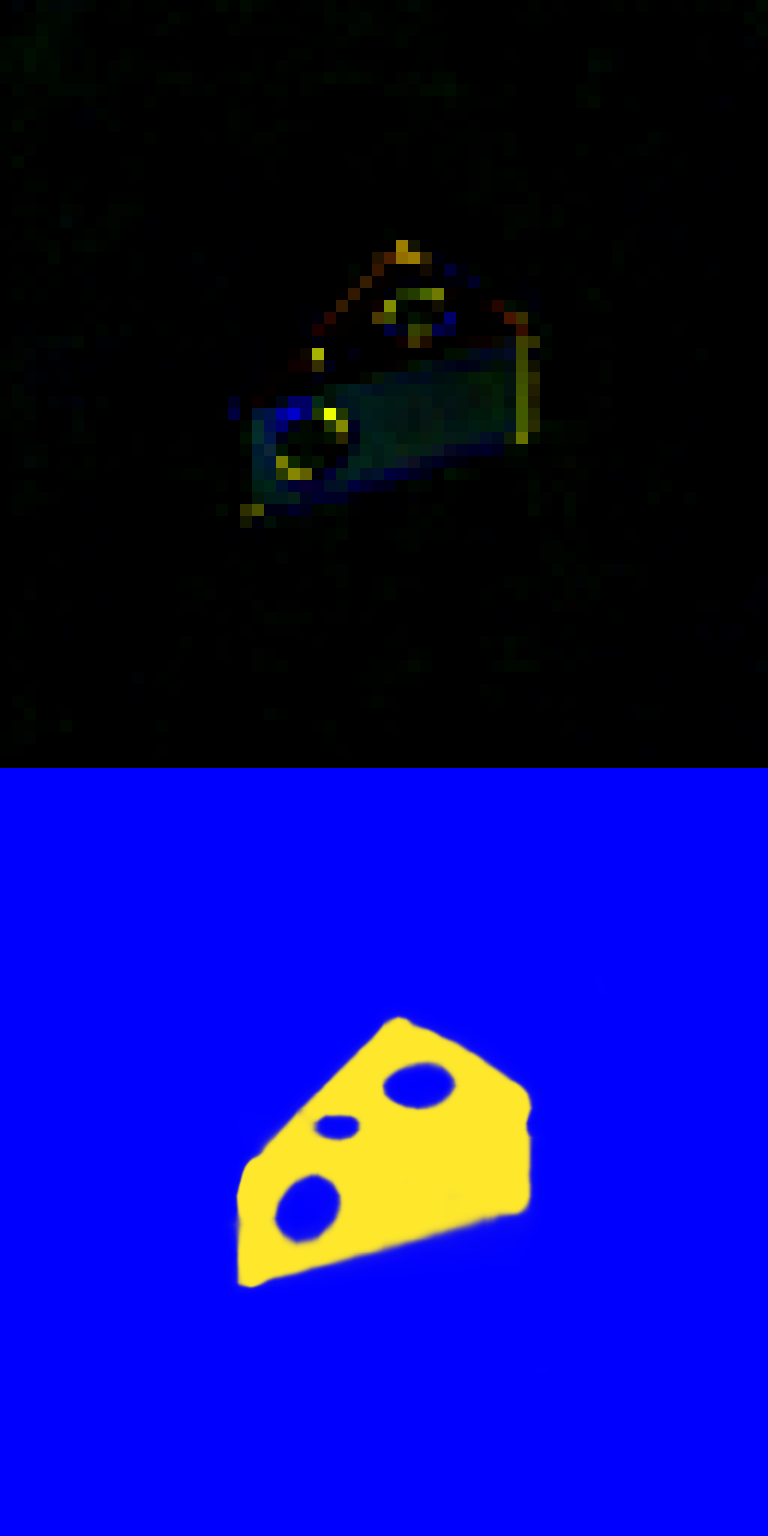} &
    \includegraphics[width=0.06\linewidth]{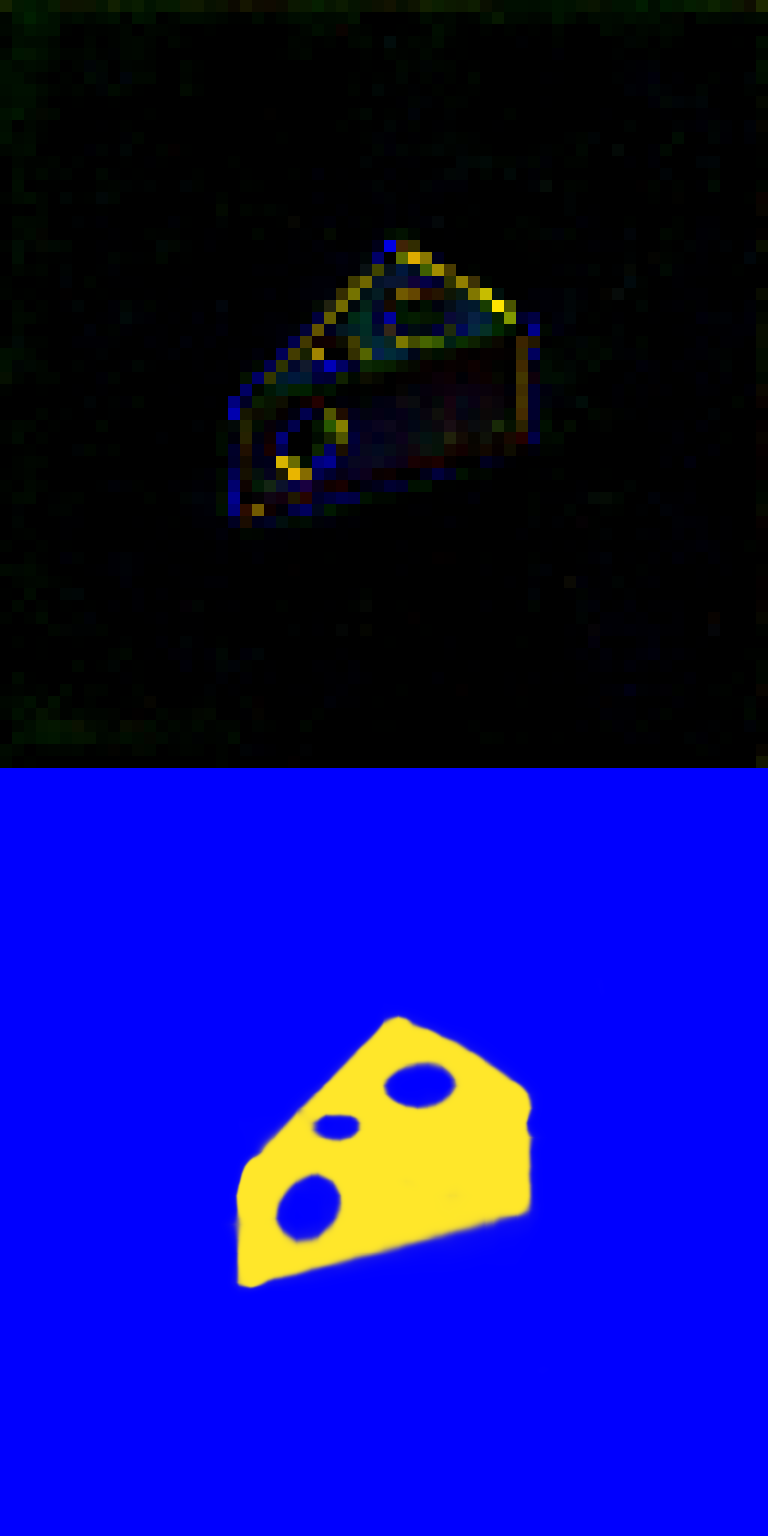} &
    \includegraphics[width=0.06\linewidth]{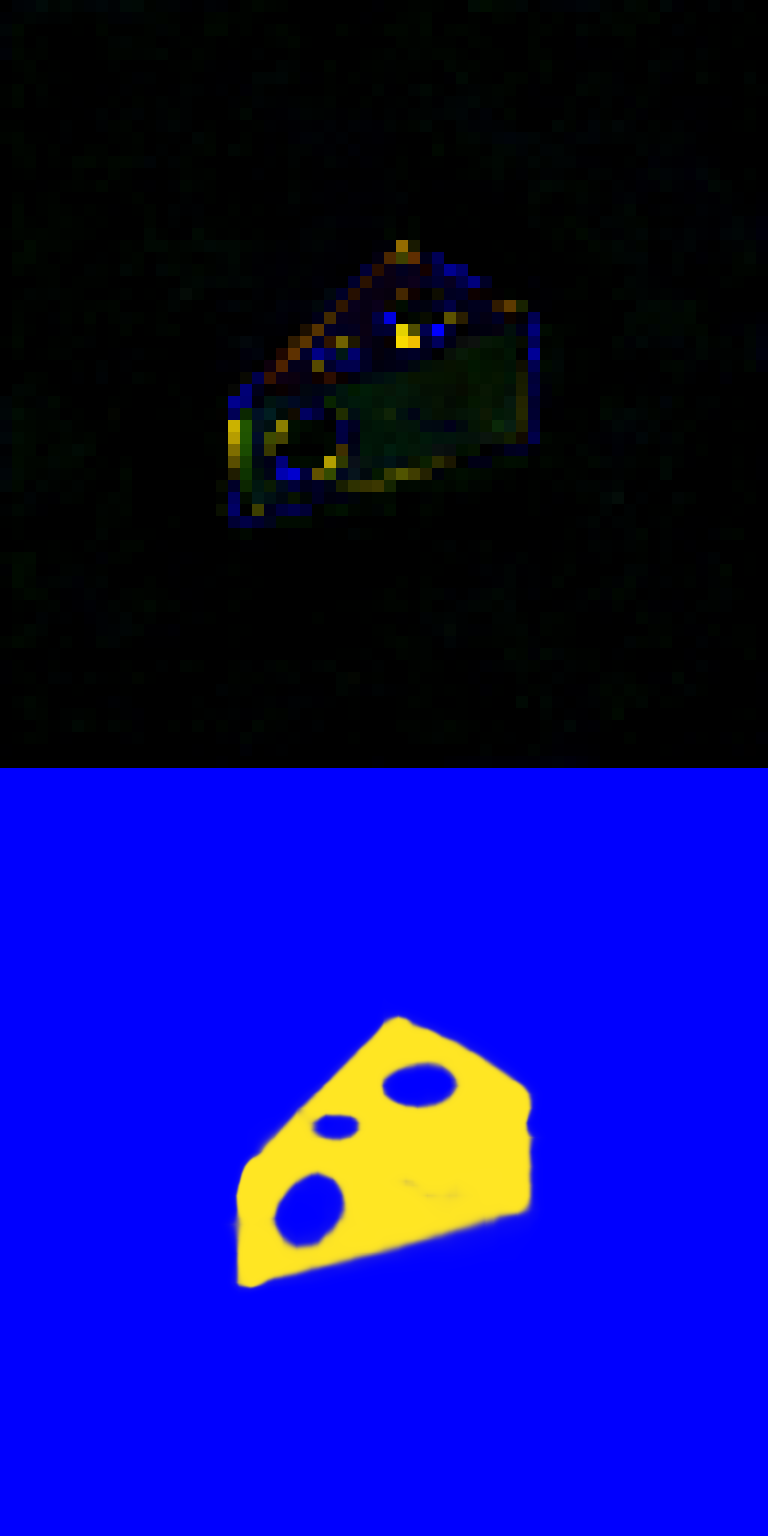} \\
    
\end{tabular}
        \vspace{-3mm}
    \caption{We visualize the SDS gradients in rgb color space ($1^{st}$ row) and generated results ($2^{nd}$ row) during optimization for two prompts. Initialization (Init) is the leftmost column. Columns on its right show the generated result and gradients for the $i^{th}$ optimization step. The number of layers is $L=1$ here.
    \label{fig:sds_grad_viz_single} }
    \vspace{-2mm}
\end{figure*}

\begin{figure*}[t!]
    \centering
    \setlength{\tabcolsep}{0.5pt}

\begin{tabular}{c ccccccccccccccc}
    &
    \multicolumn{7}{c}{NIVeL optimization steps $\xrightarrow{\makebox[2cm]{}}$}
    &
    \multicolumn{8}{c}{}
    \\
    Init & 100 & 300 & 500 & 800 & 1200 & 1500 & 1800 & 2400 & 3000 & 3600 & 4200 & 5000 & 5800 & 6900 & 8000
    \\
    \multicolumn{16}{c}{}
    \\
    \footnotesize{\emph{Box}}&
    \multicolumn{15}{c}{\emph{\textcolor{textprompts}{"A grizzly bear karate master ..."}}}
    \\
    \includegraphics[width=0.06\linewidth]{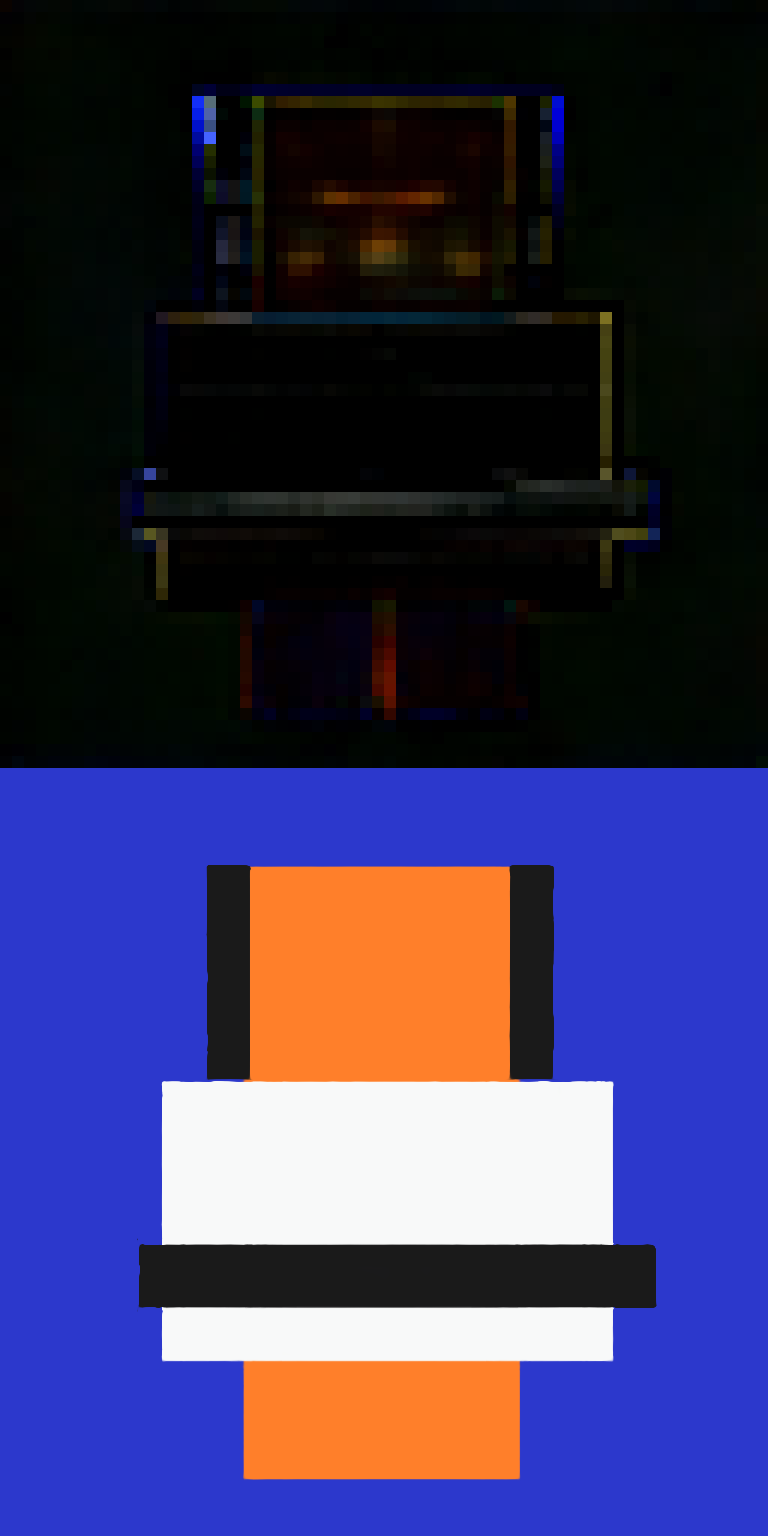} &
    \includegraphics[width=0.06\linewidth]{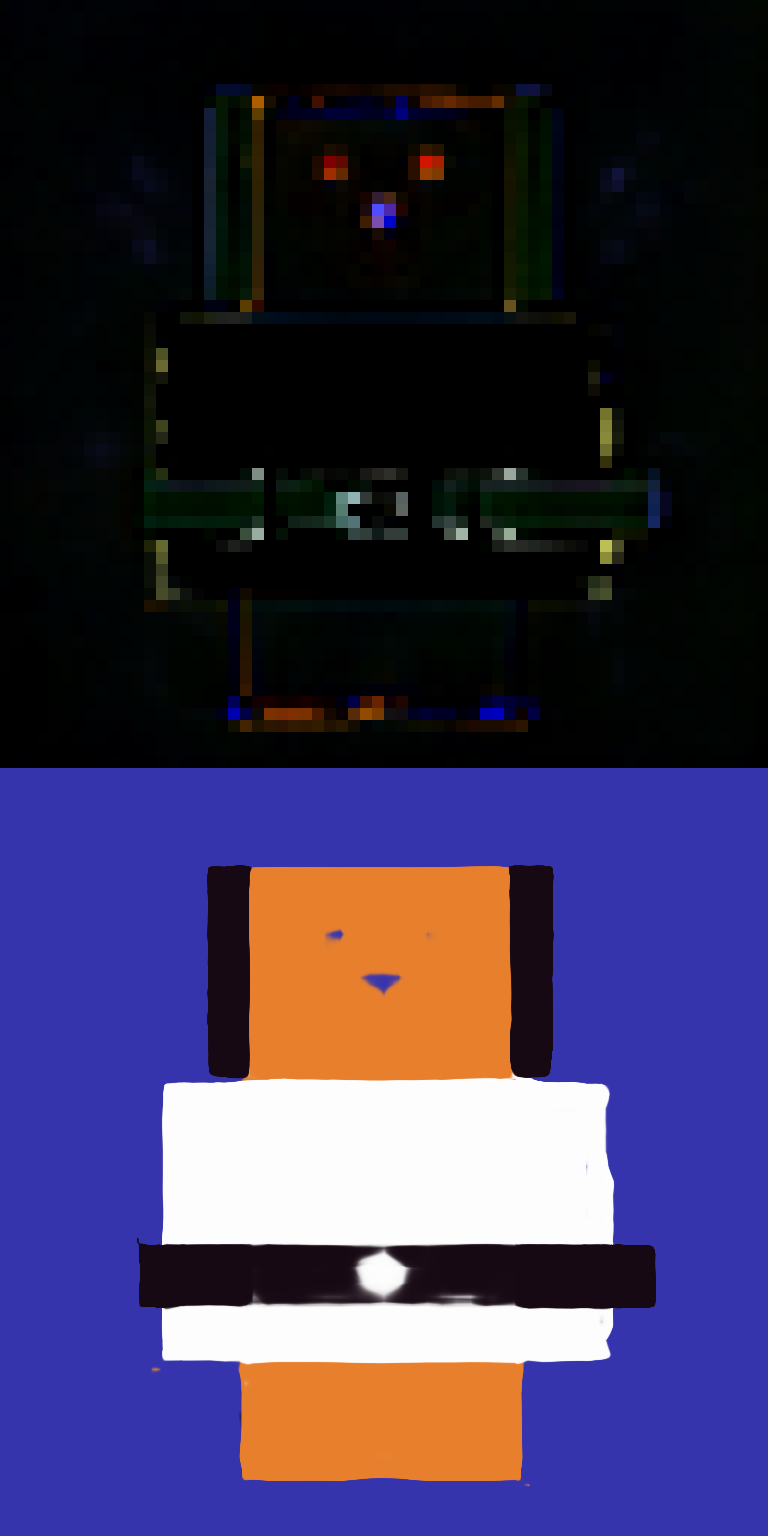} &
    \includegraphics[width=0.06\linewidth]{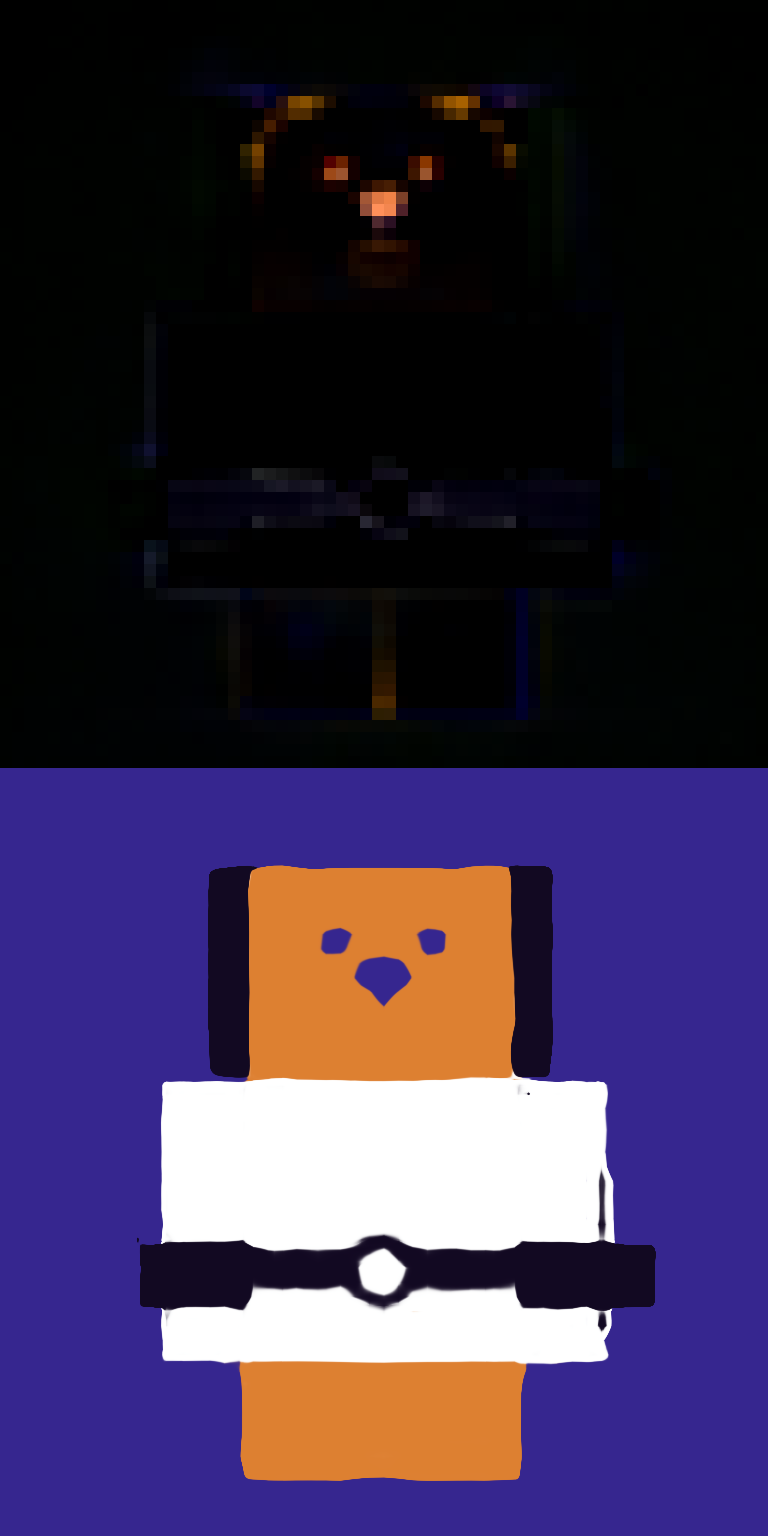} &
    \includegraphics[width=0.06\linewidth]{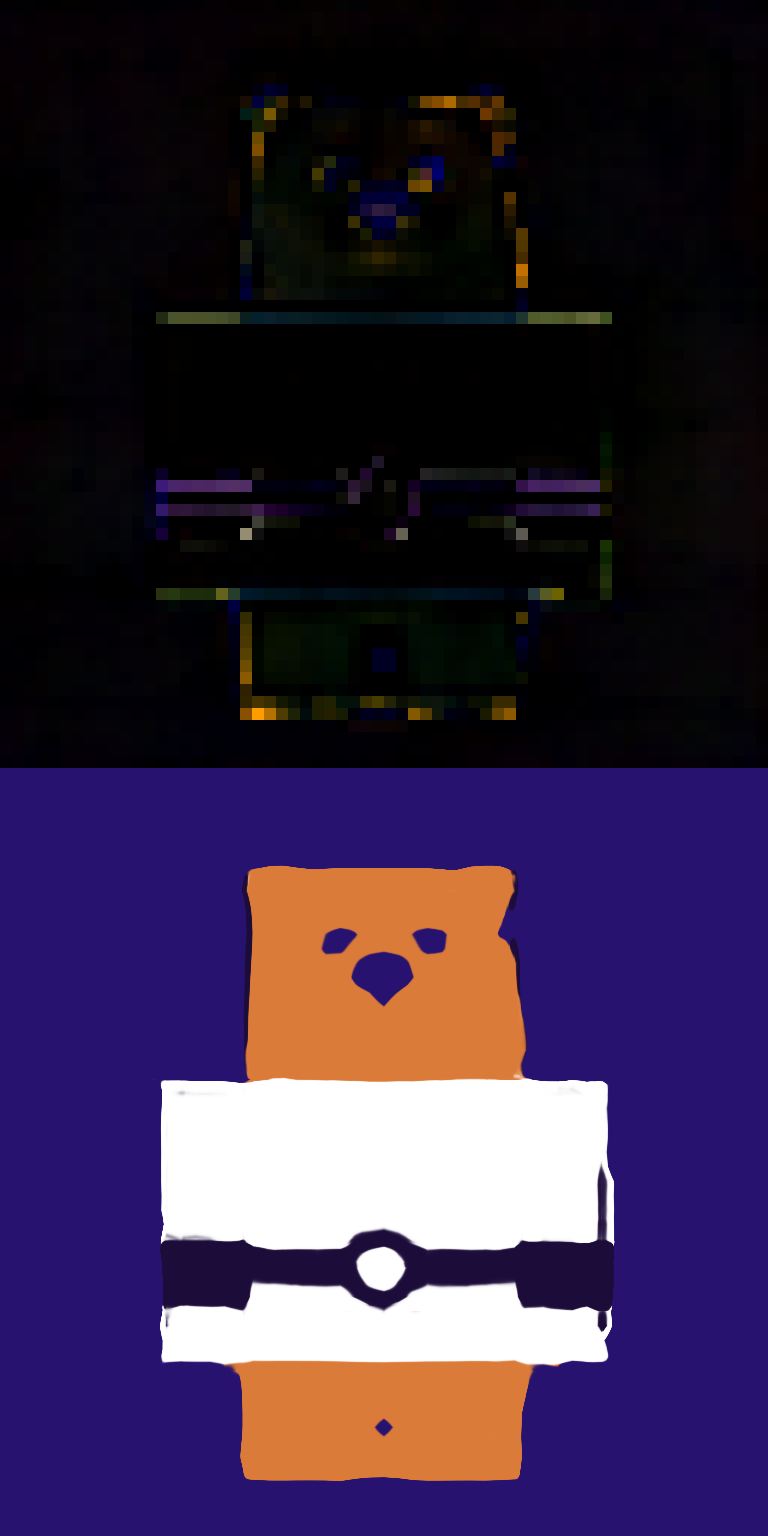} &
    \includegraphics[width=0.06\linewidth]{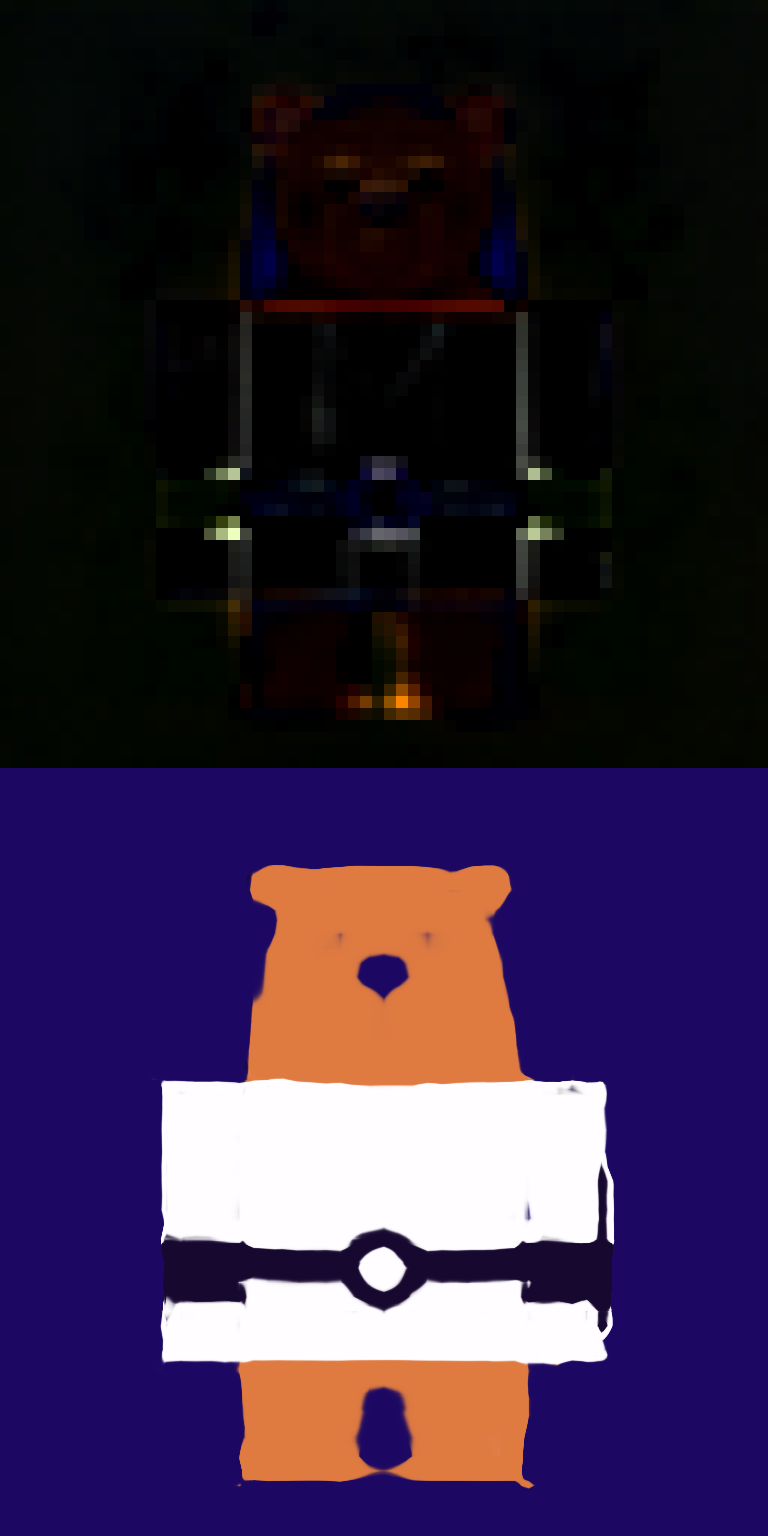} &
    \includegraphics[width=0.06\linewidth]{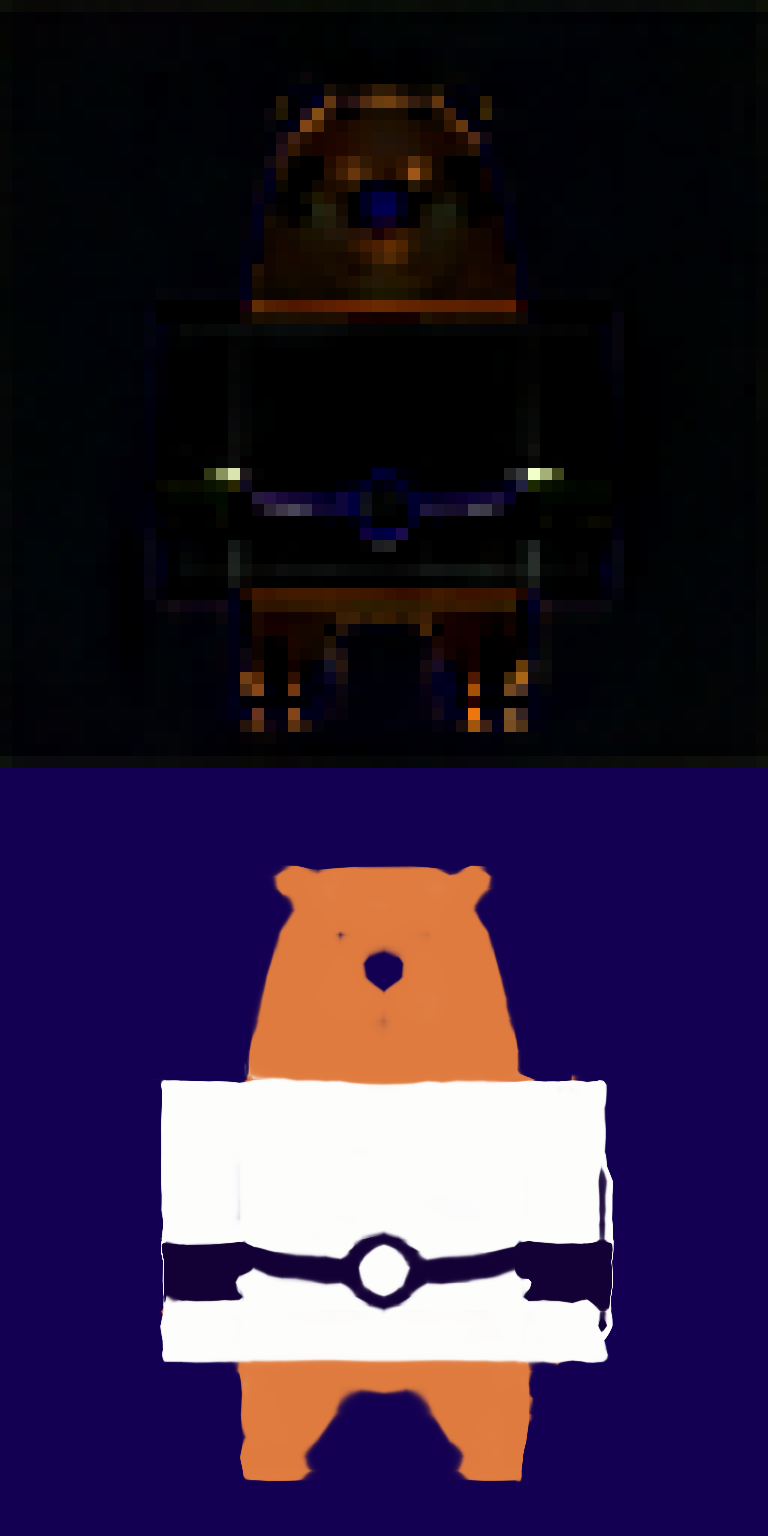} &
    \includegraphics[width=0.06\linewidth]{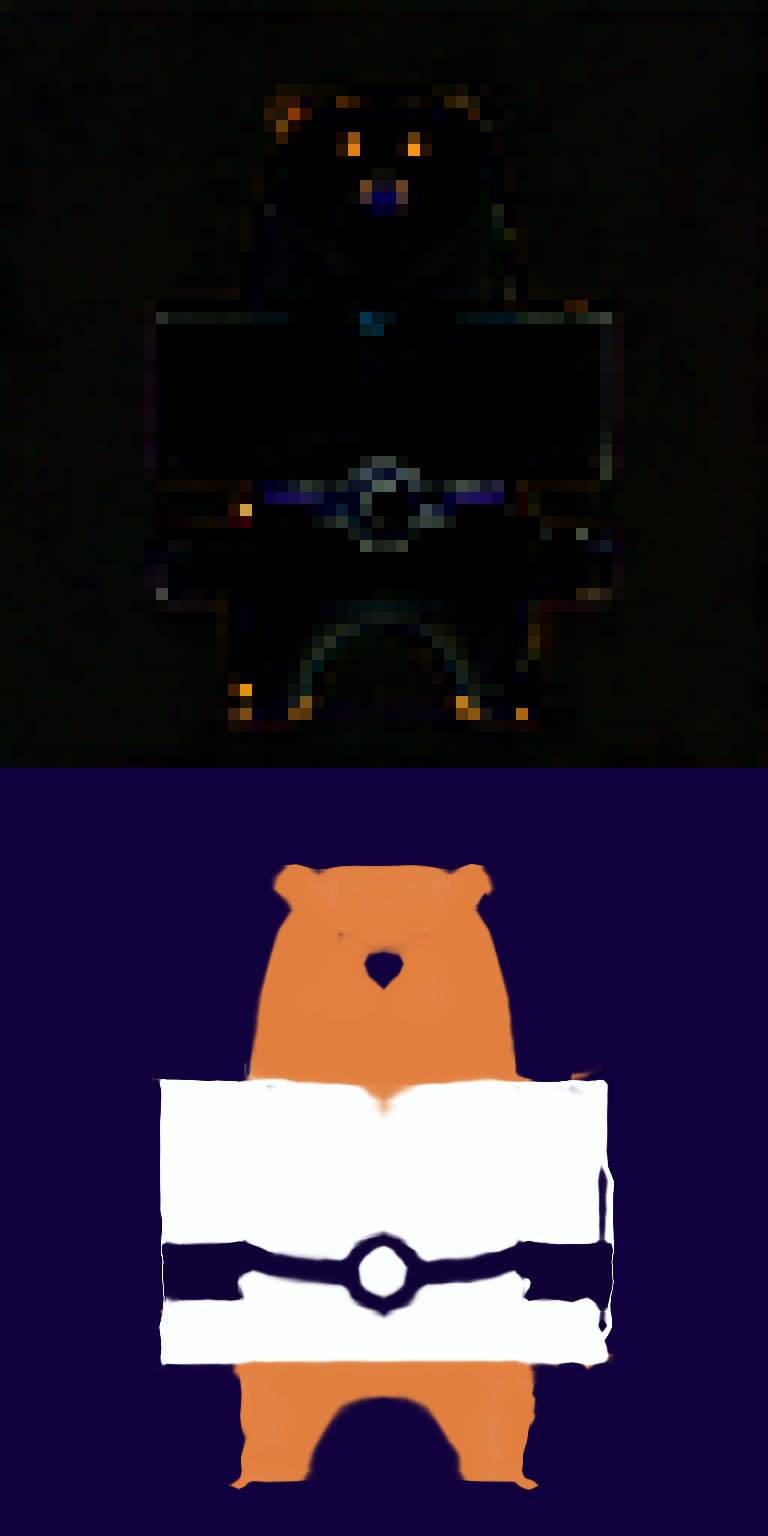} &
    \includegraphics[width=0.06\linewidth]{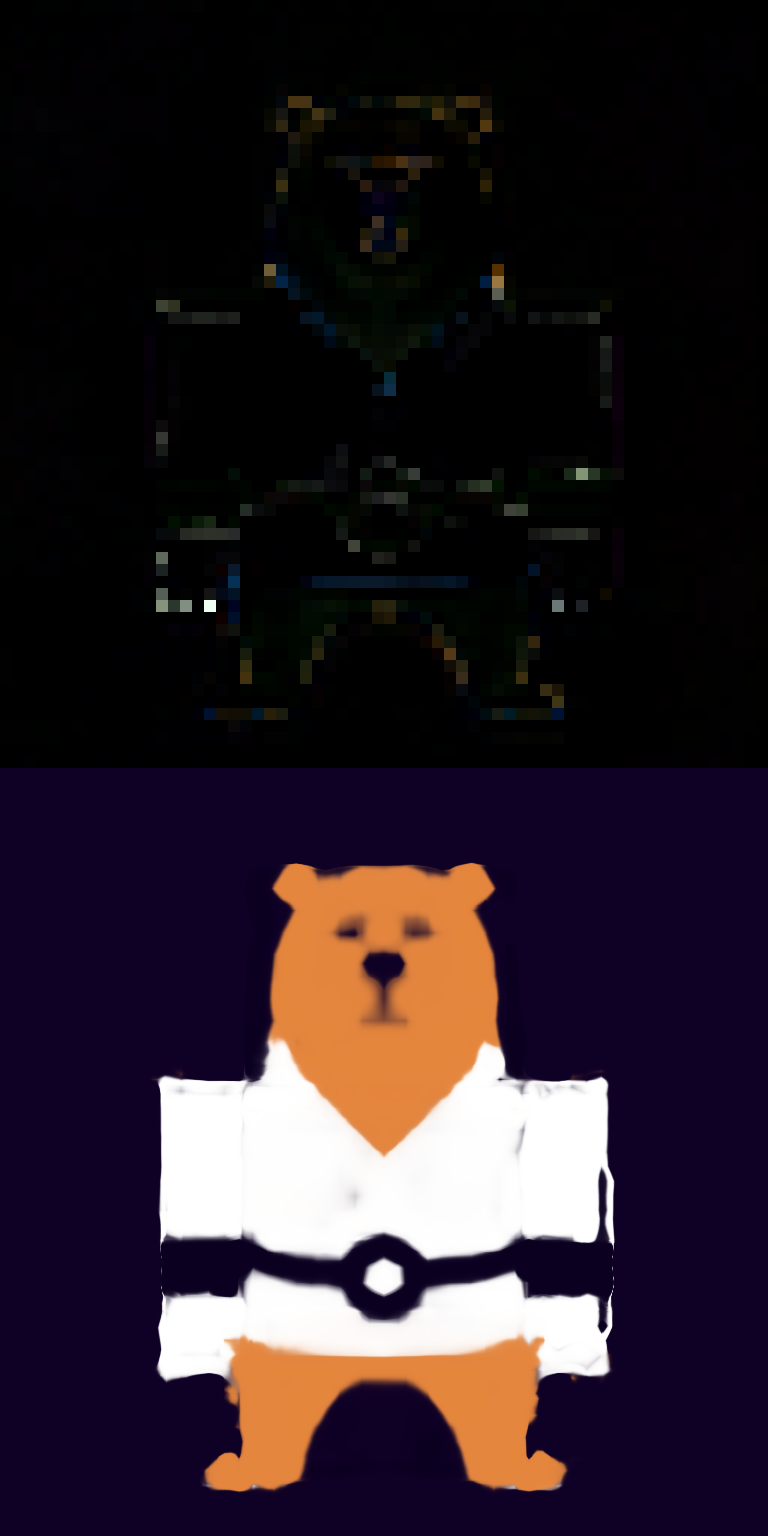} &
    \includegraphics[width=0.06\linewidth]{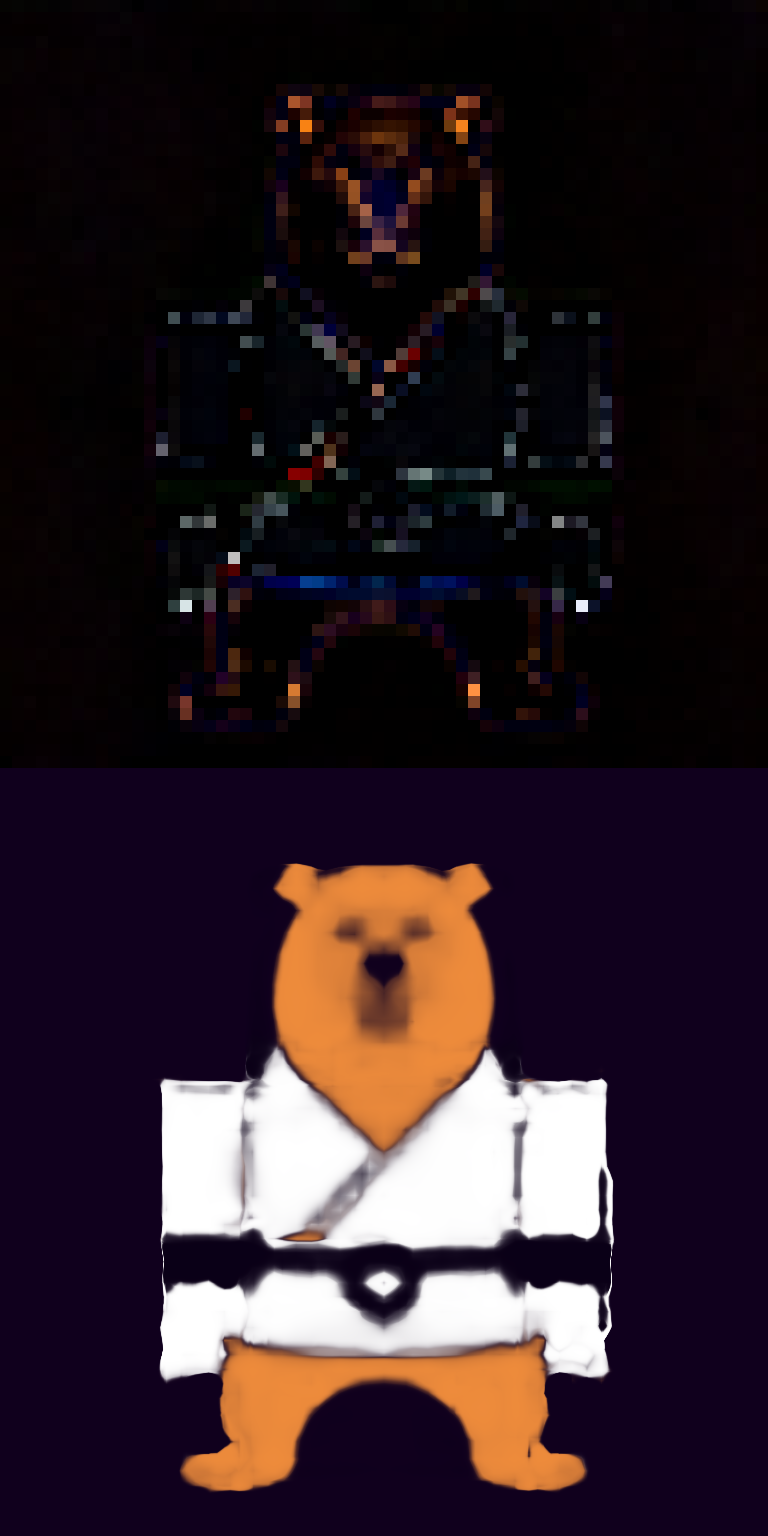} &
    \includegraphics[width=0.06\linewidth]{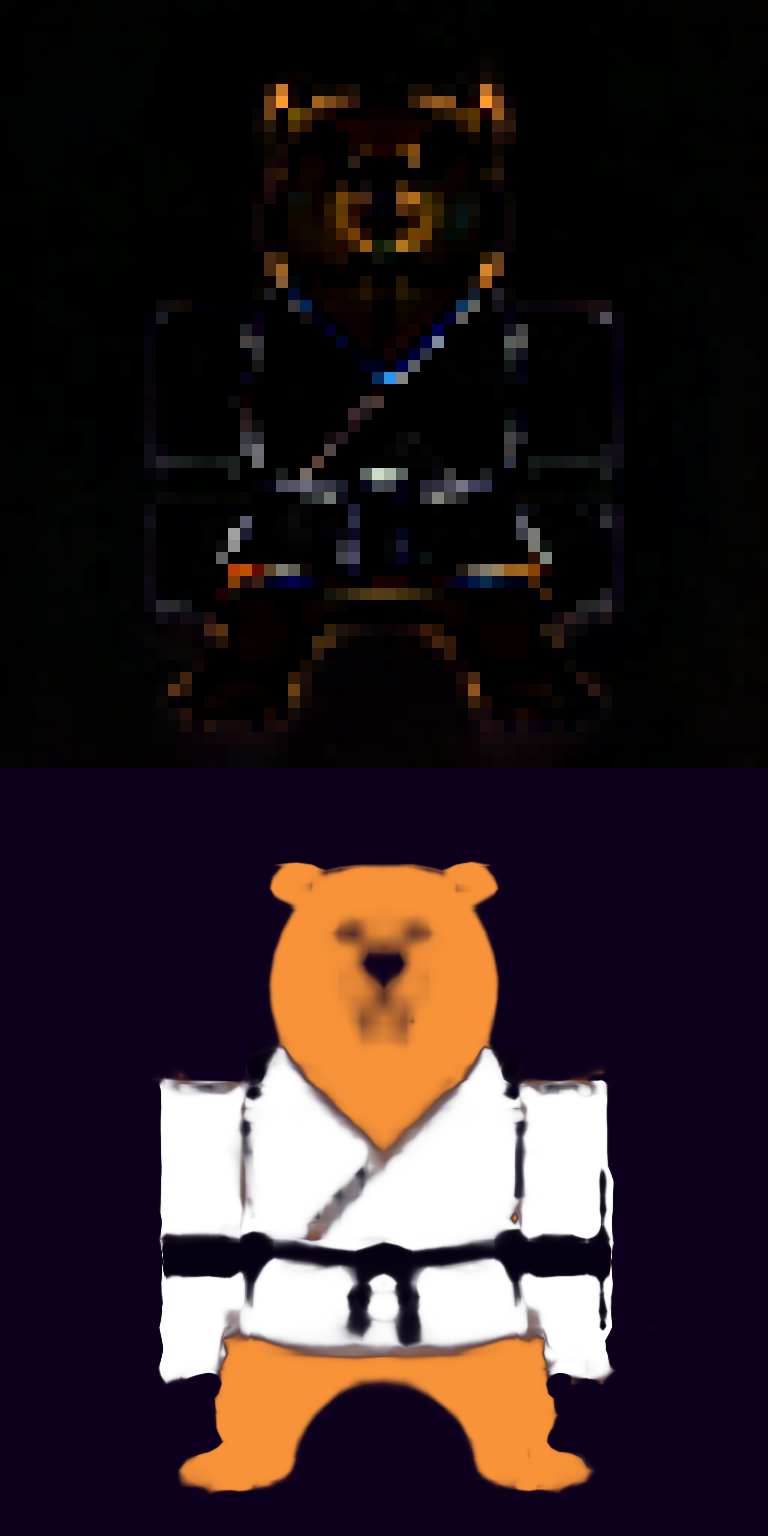} &
    \includegraphics[width=0.06\linewidth]{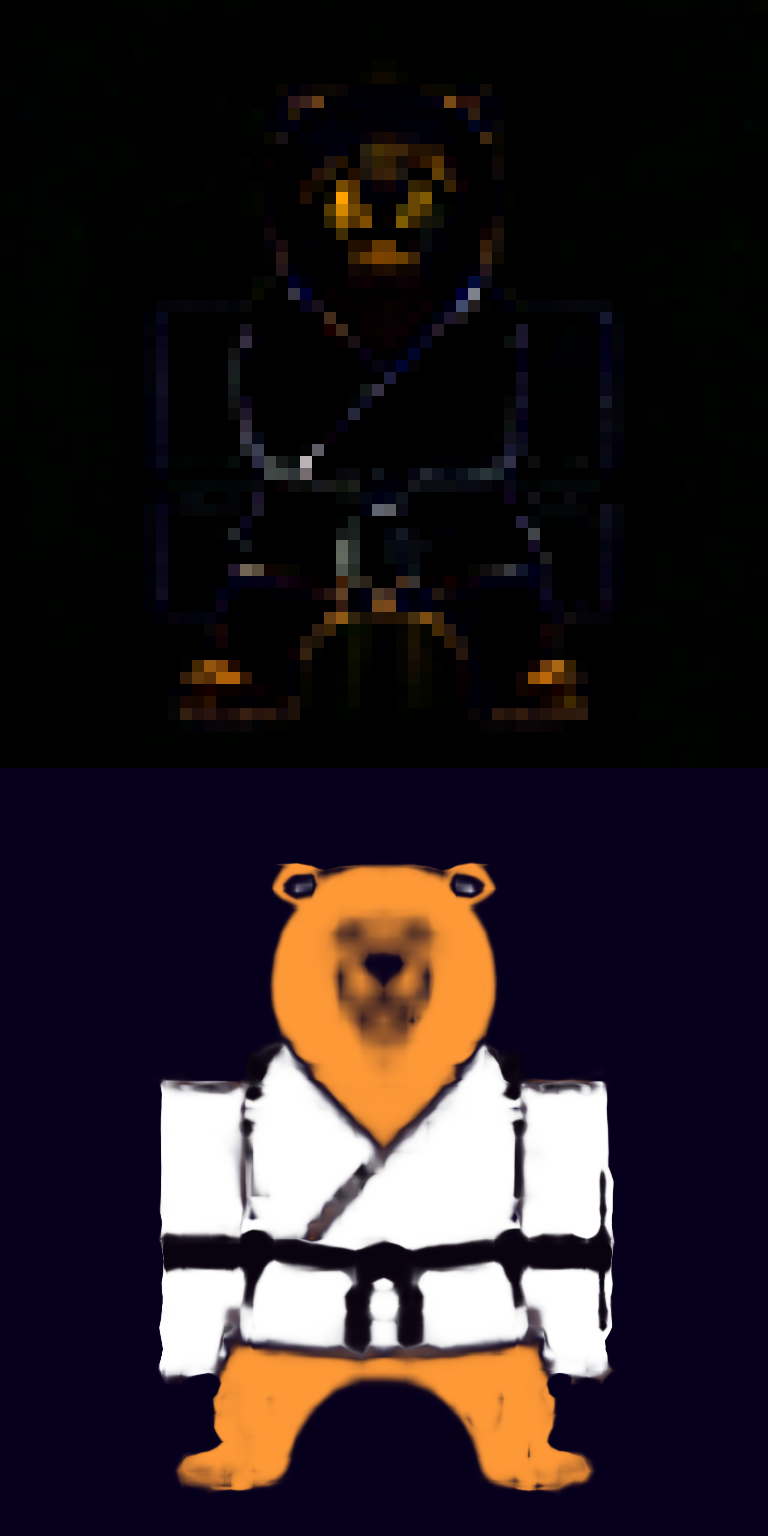} &
    \includegraphics[width=0.06\linewidth]{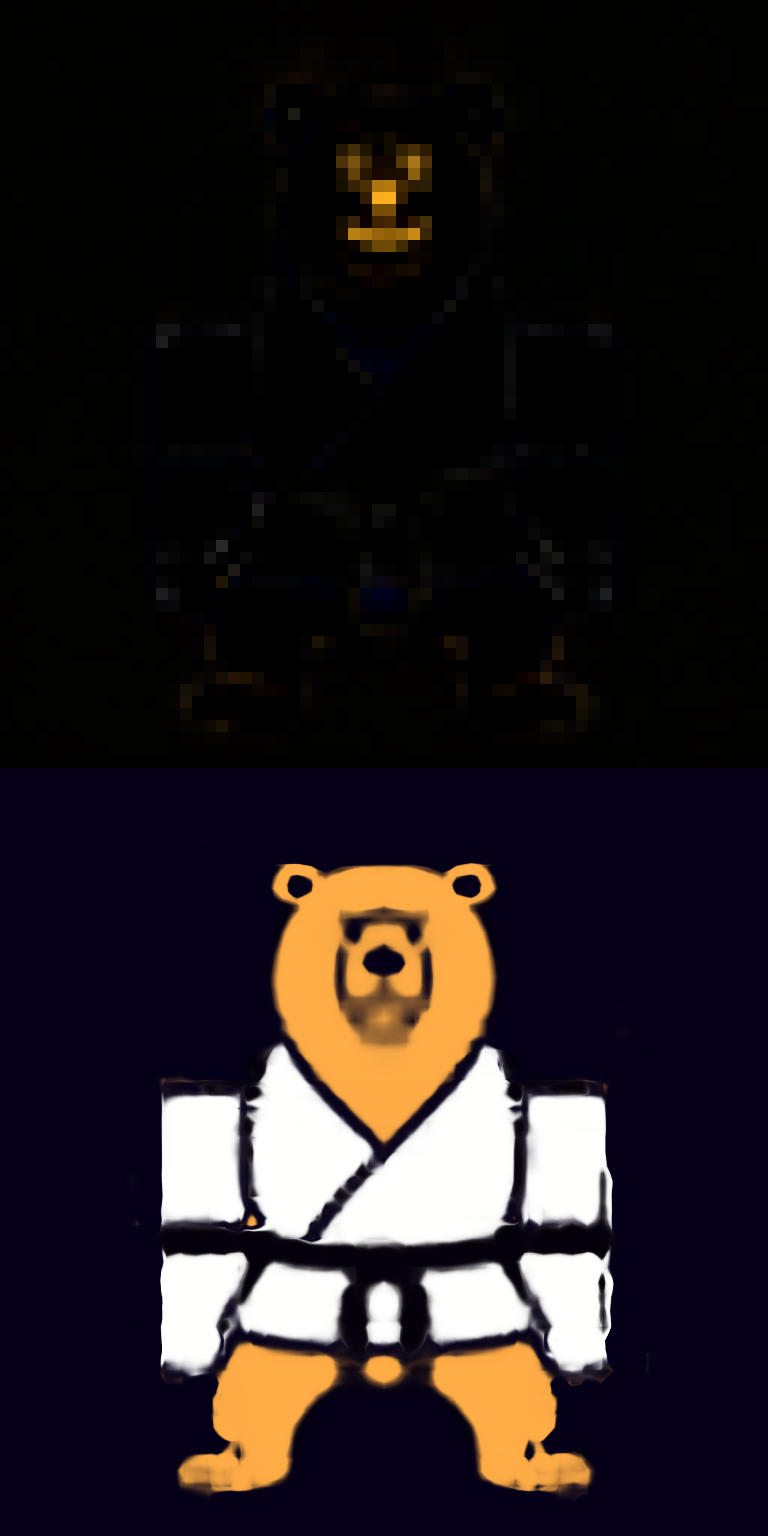} &
    \includegraphics[width=0.06\linewidth]{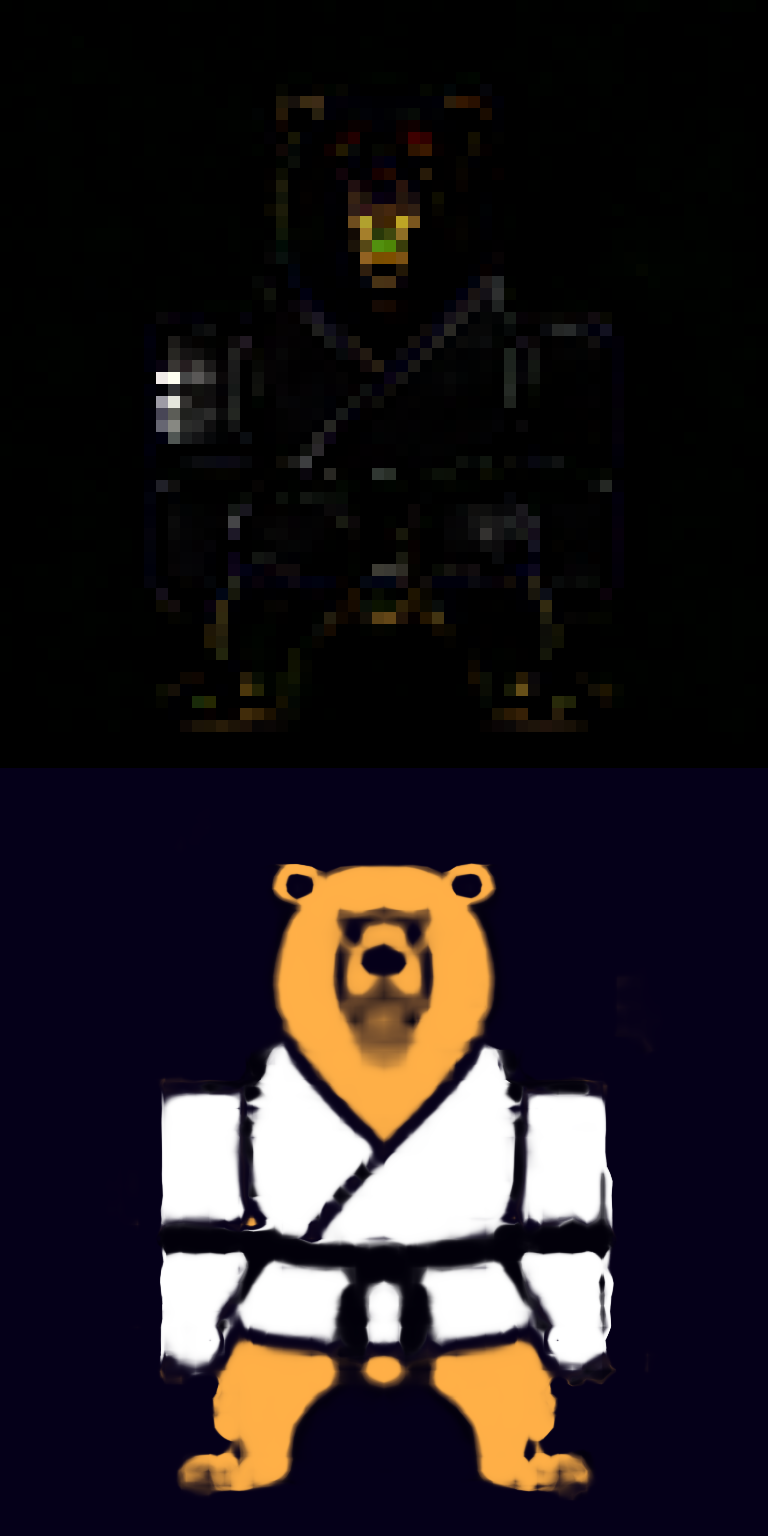} &
    \includegraphics[width=0.06\linewidth]{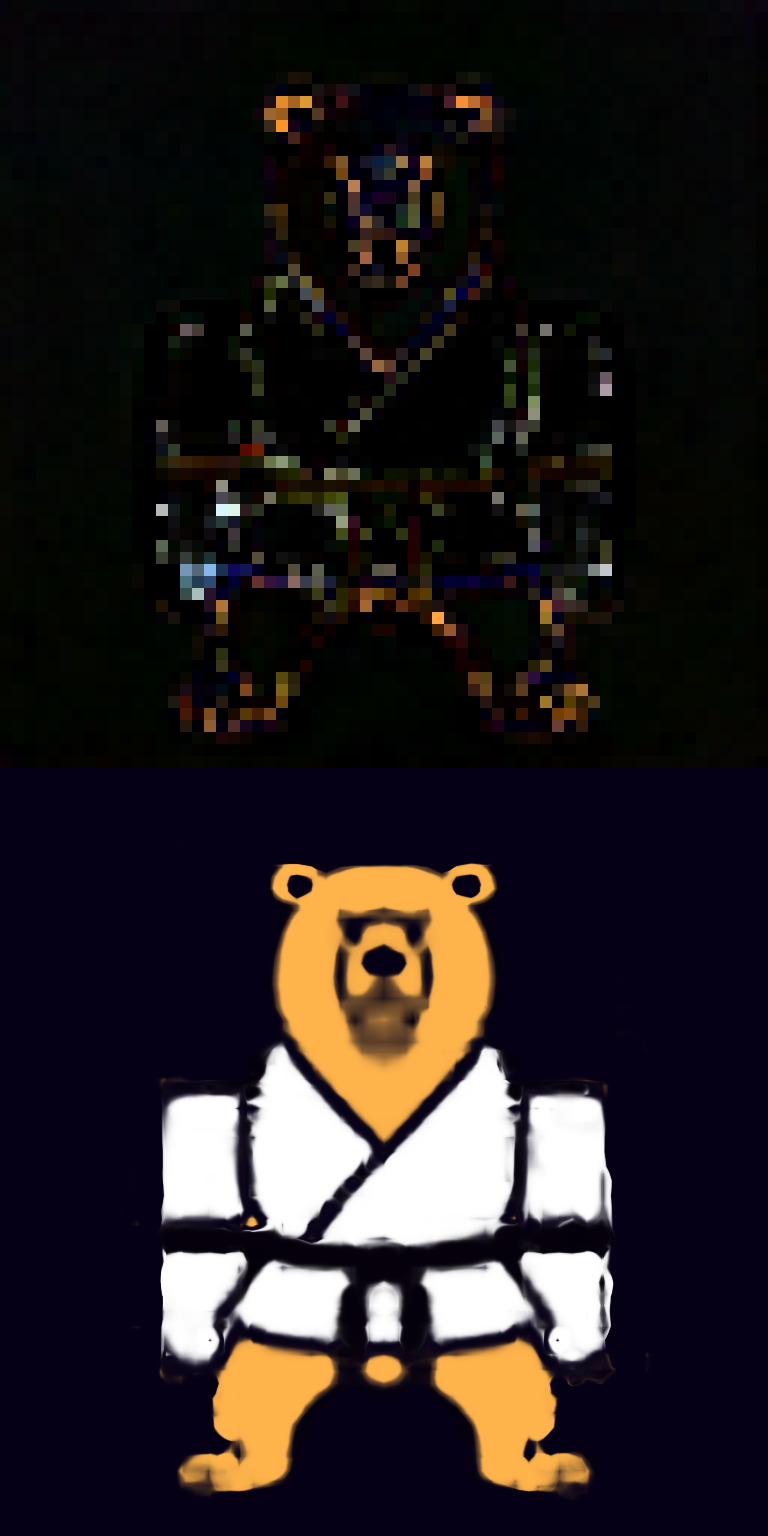} &
    \includegraphics[width=0.06\linewidth]{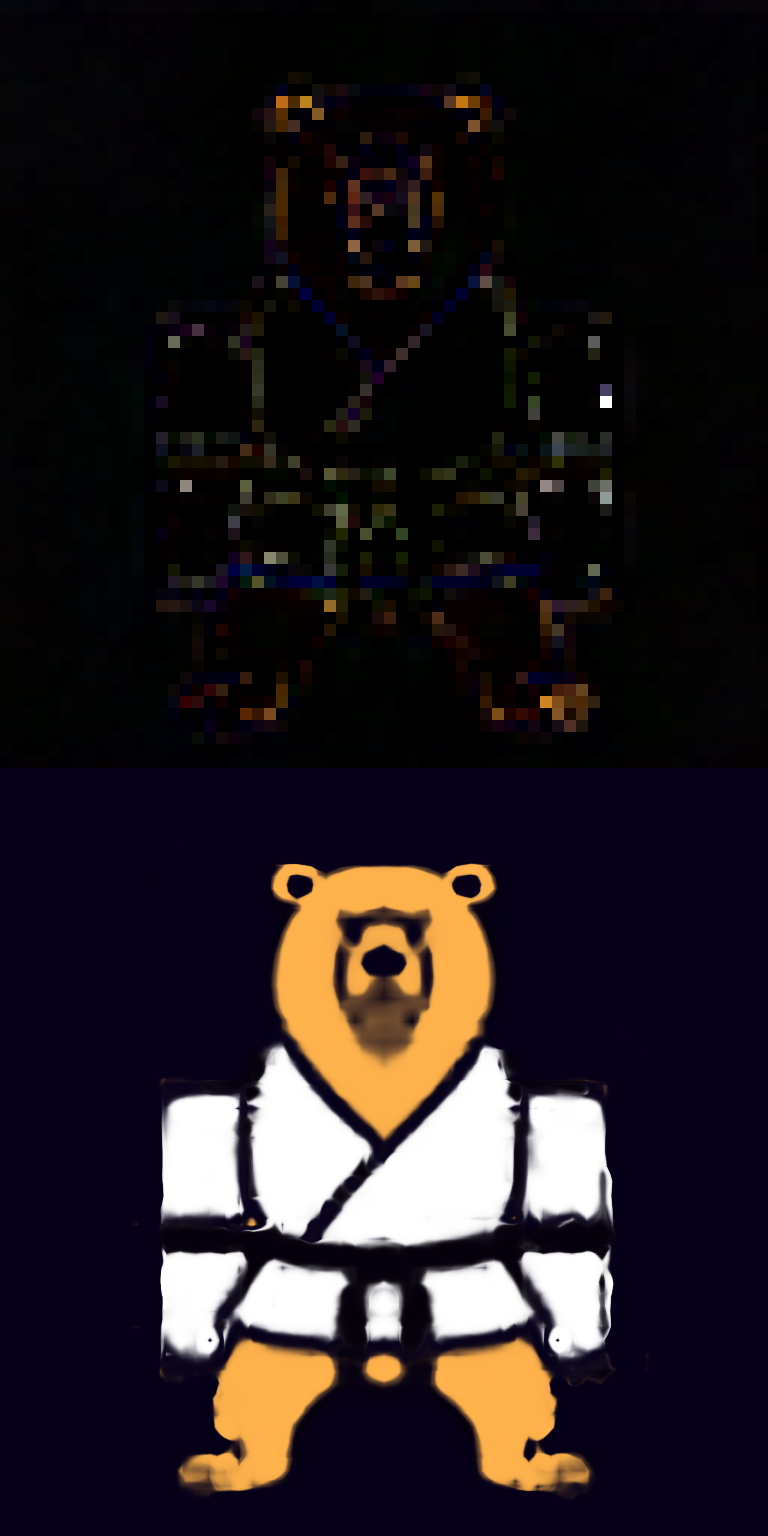} &
    \includegraphics[width=0.06\linewidth]{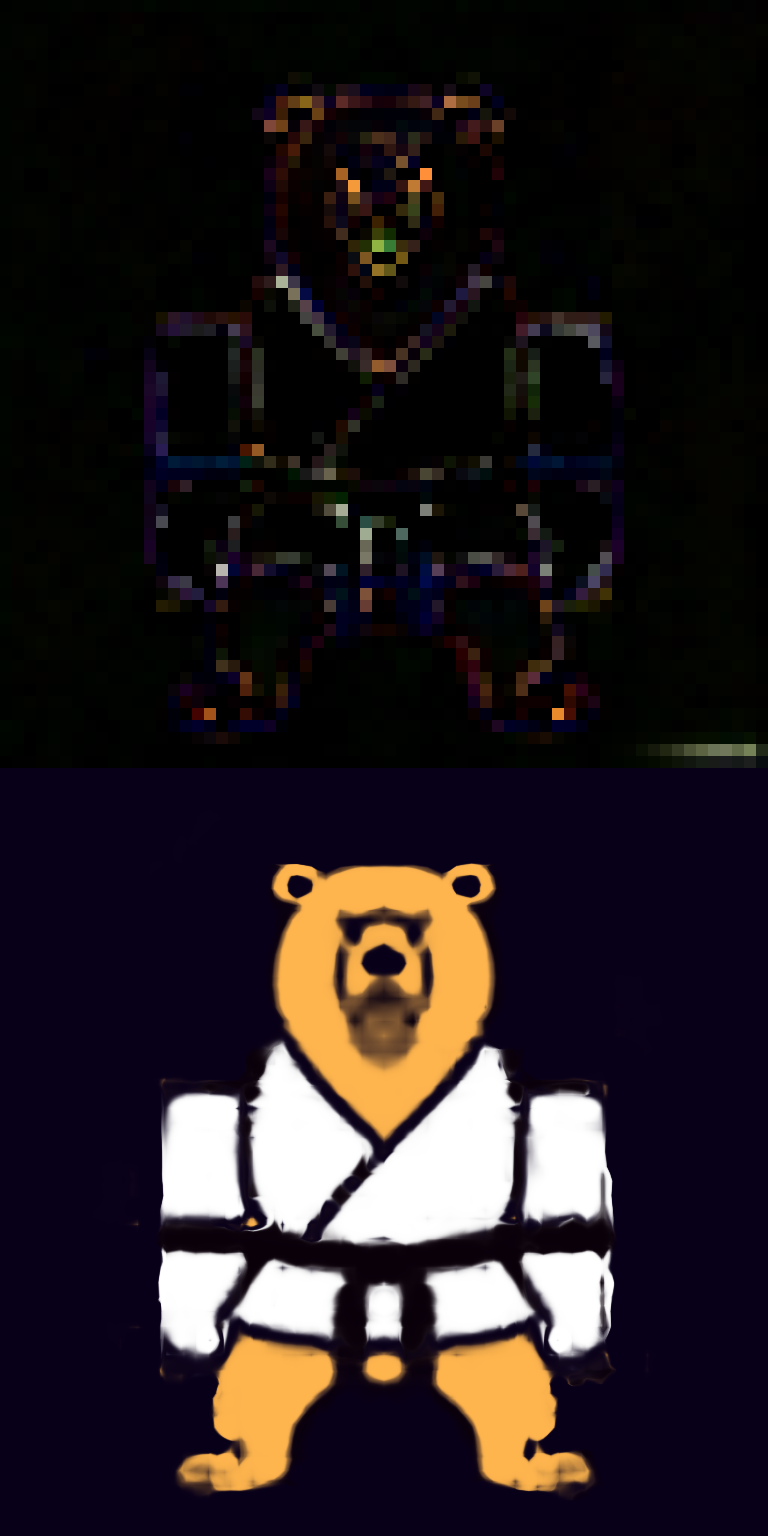} \\
    \footnotesize{\emph{Ellipse}}&
    \multicolumn{15}{c}{}
    \\
    \includegraphics[width=0.06\linewidth]{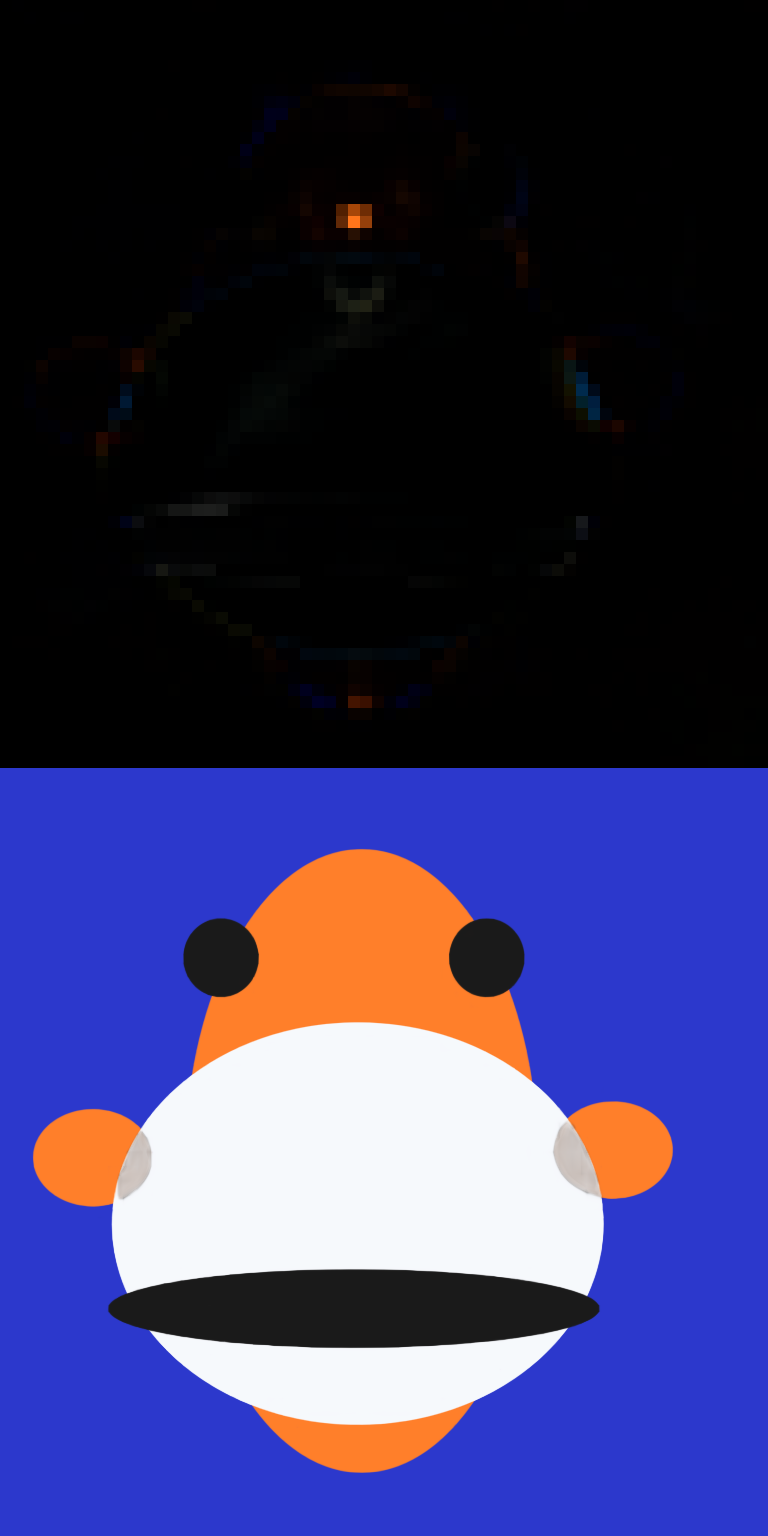} &
    \includegraphics[width=0.06\linewidth]{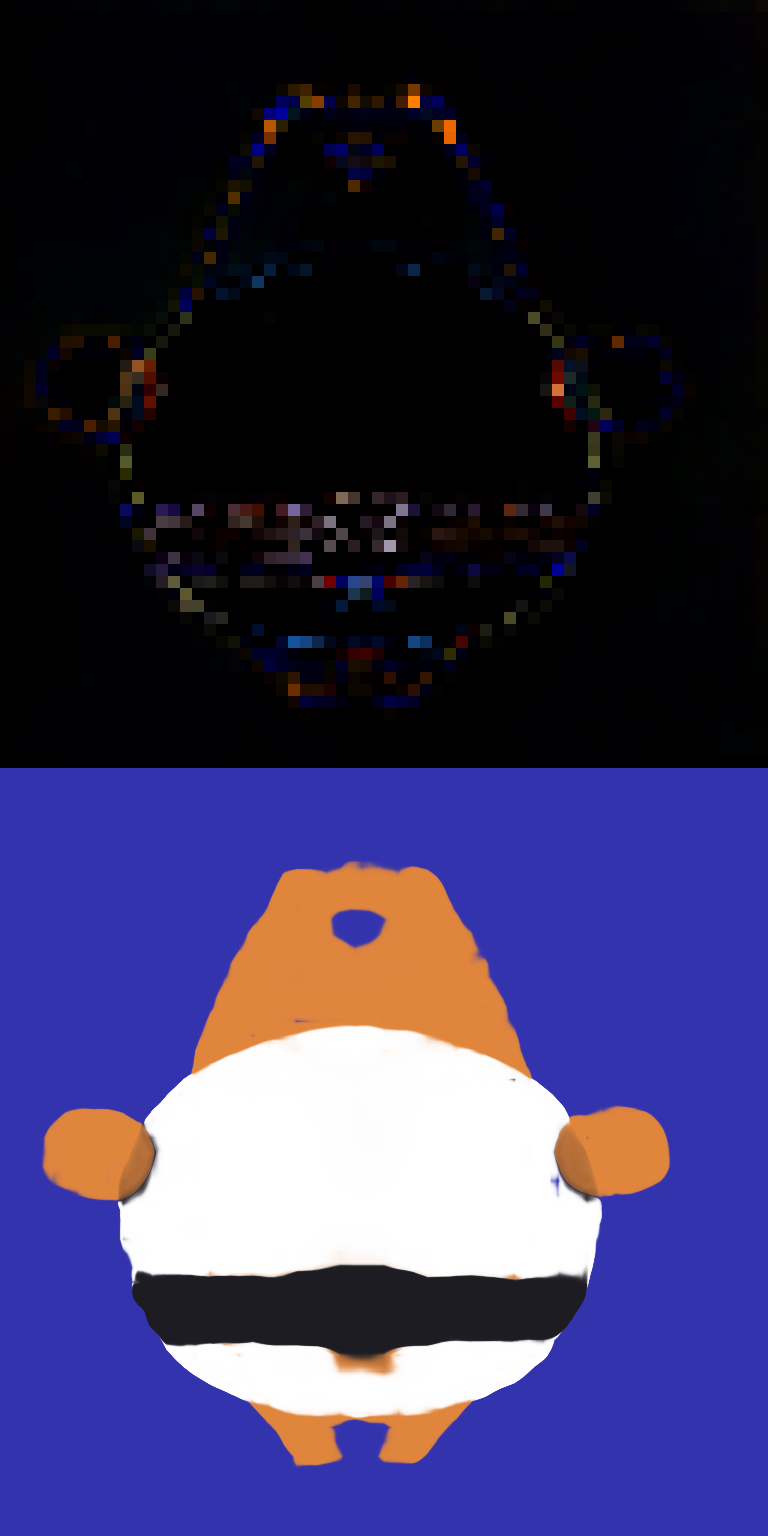} &
    \includegraphics[width=0.06\linewidth]{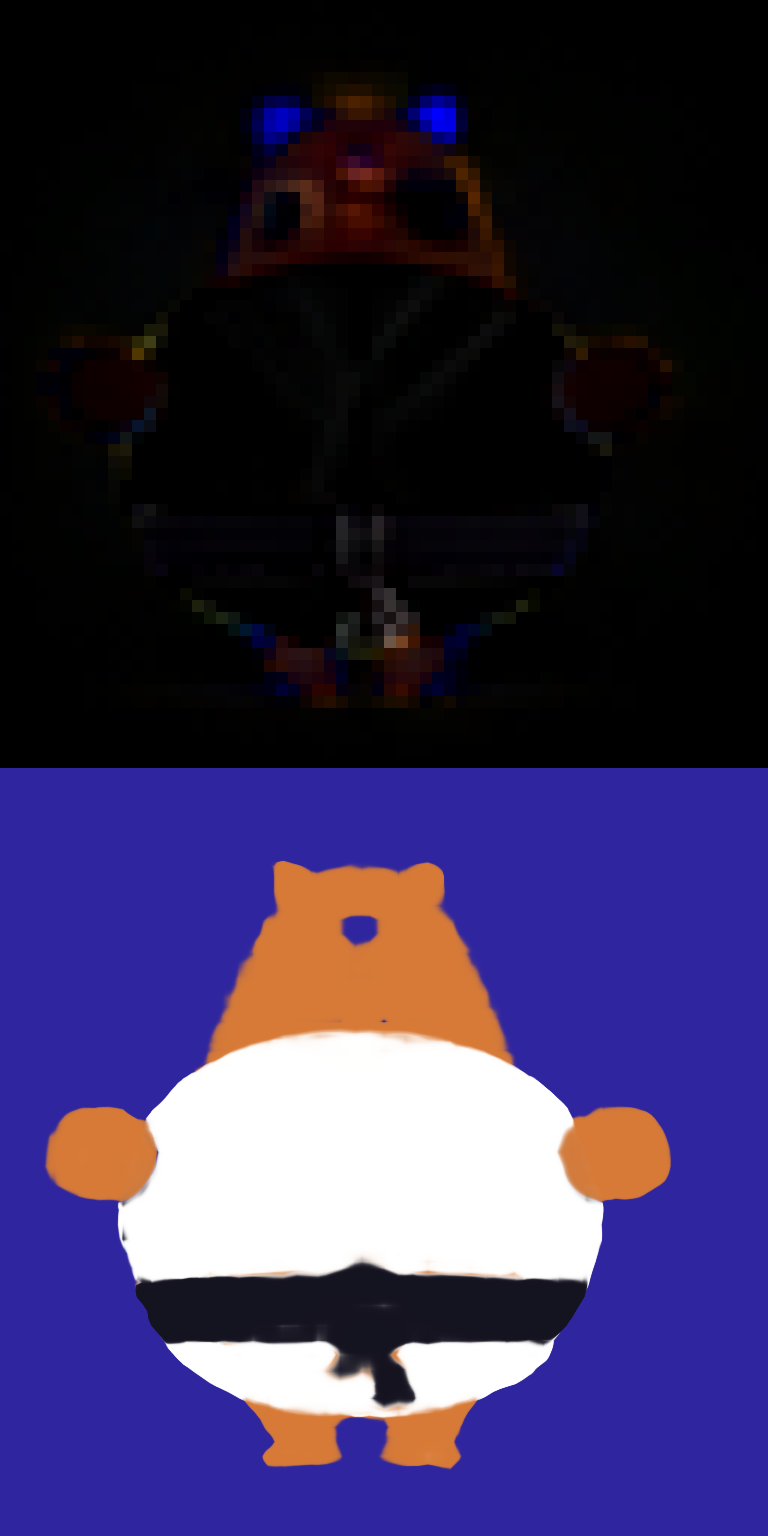} &
    \includegraphics[width=0.06\linewidth]{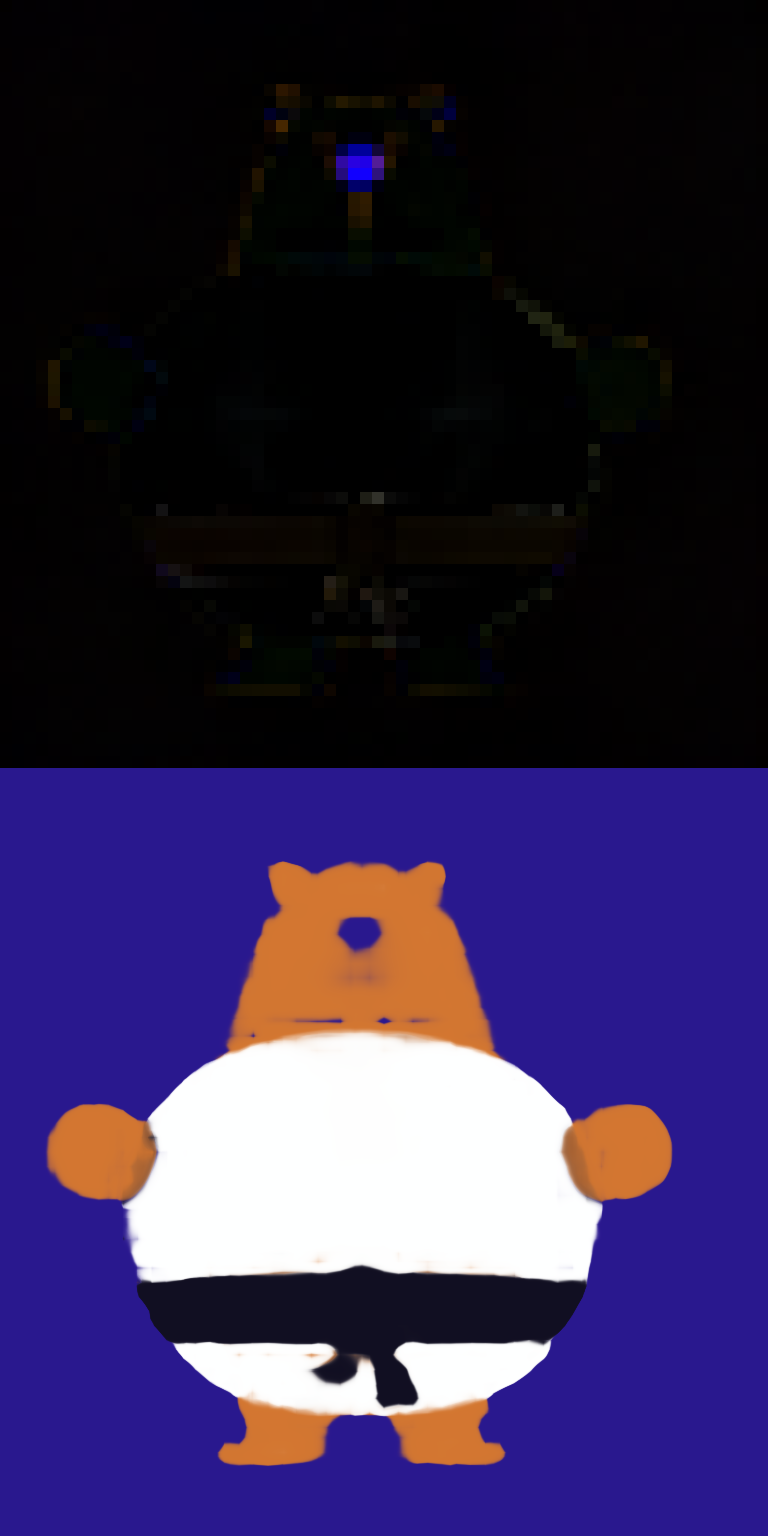} &
    \includegraphics[width=0.06\linewidth]{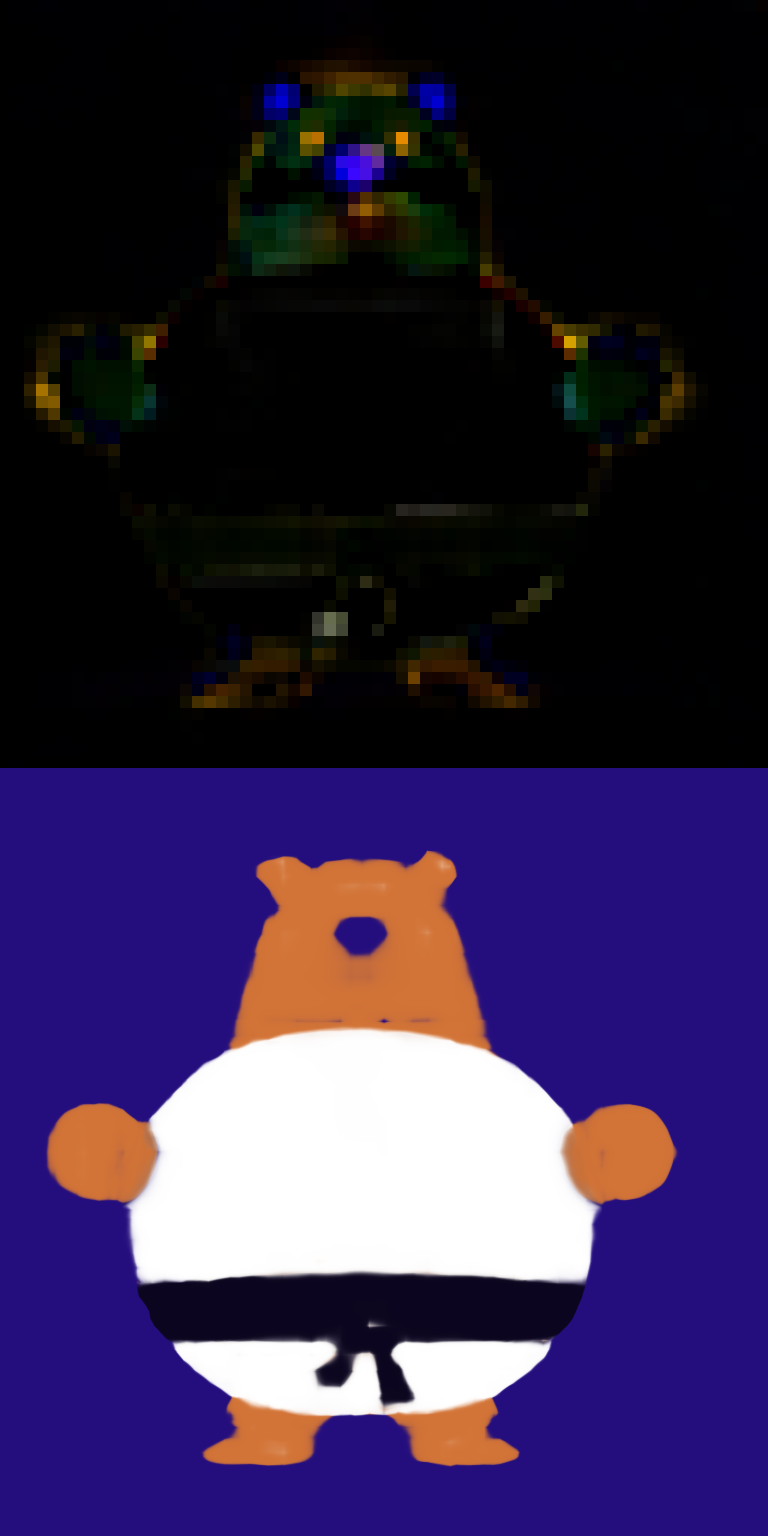} &
    \includegraphics[width=0.06\linewidth]{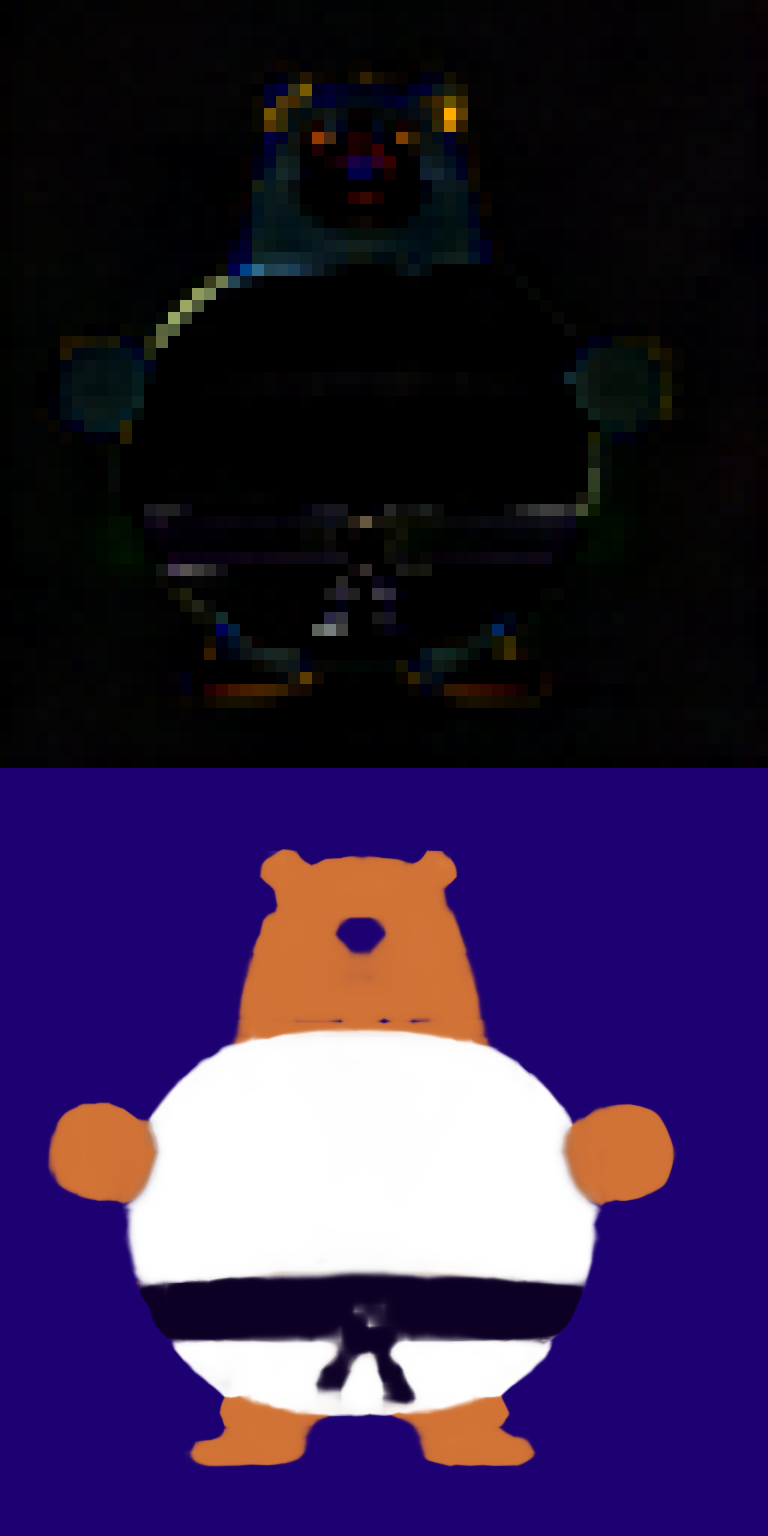} &
    \includegraphics[width=0.06\linewidth]{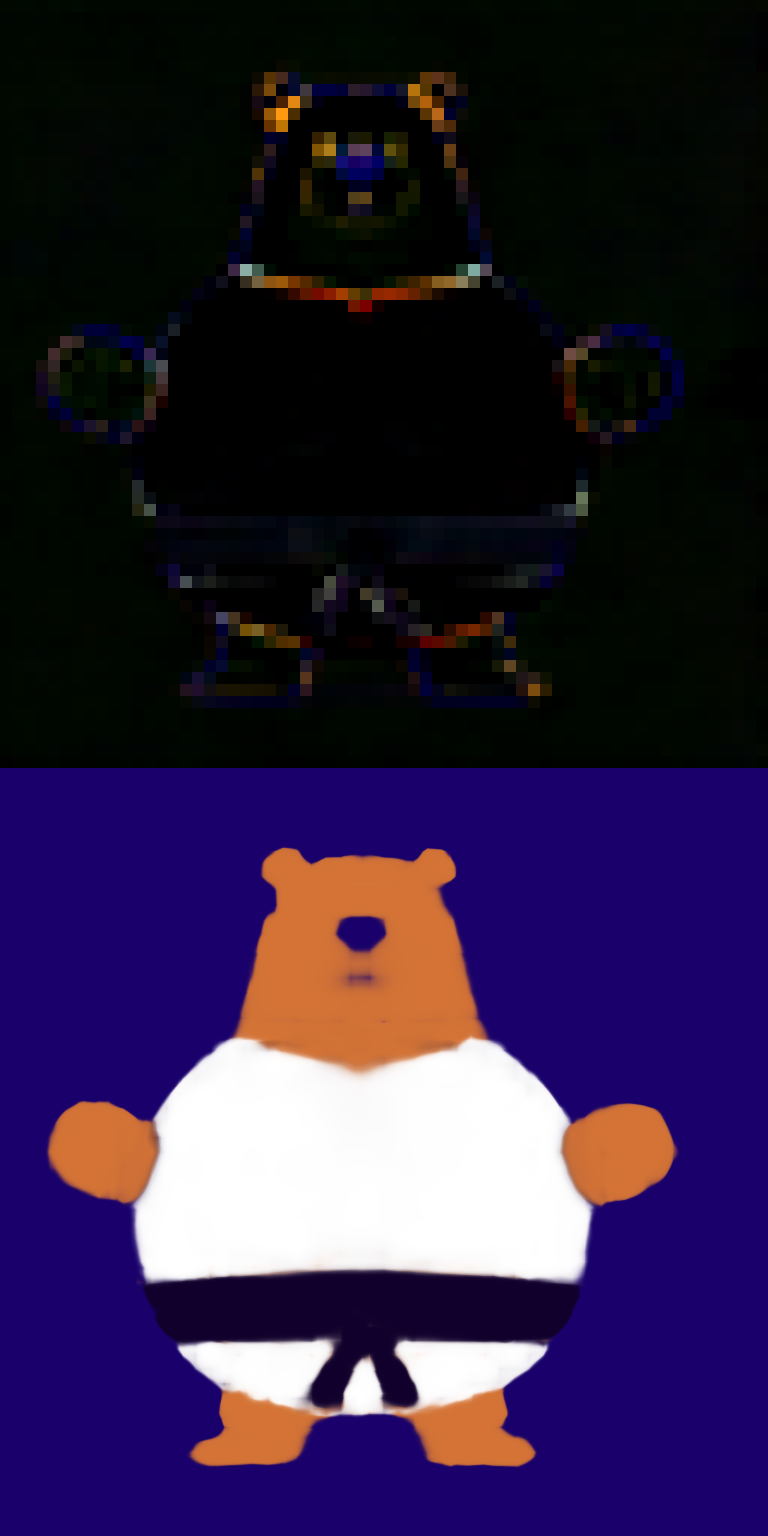} &
    \includegraphics[width=0.06\linewidth]{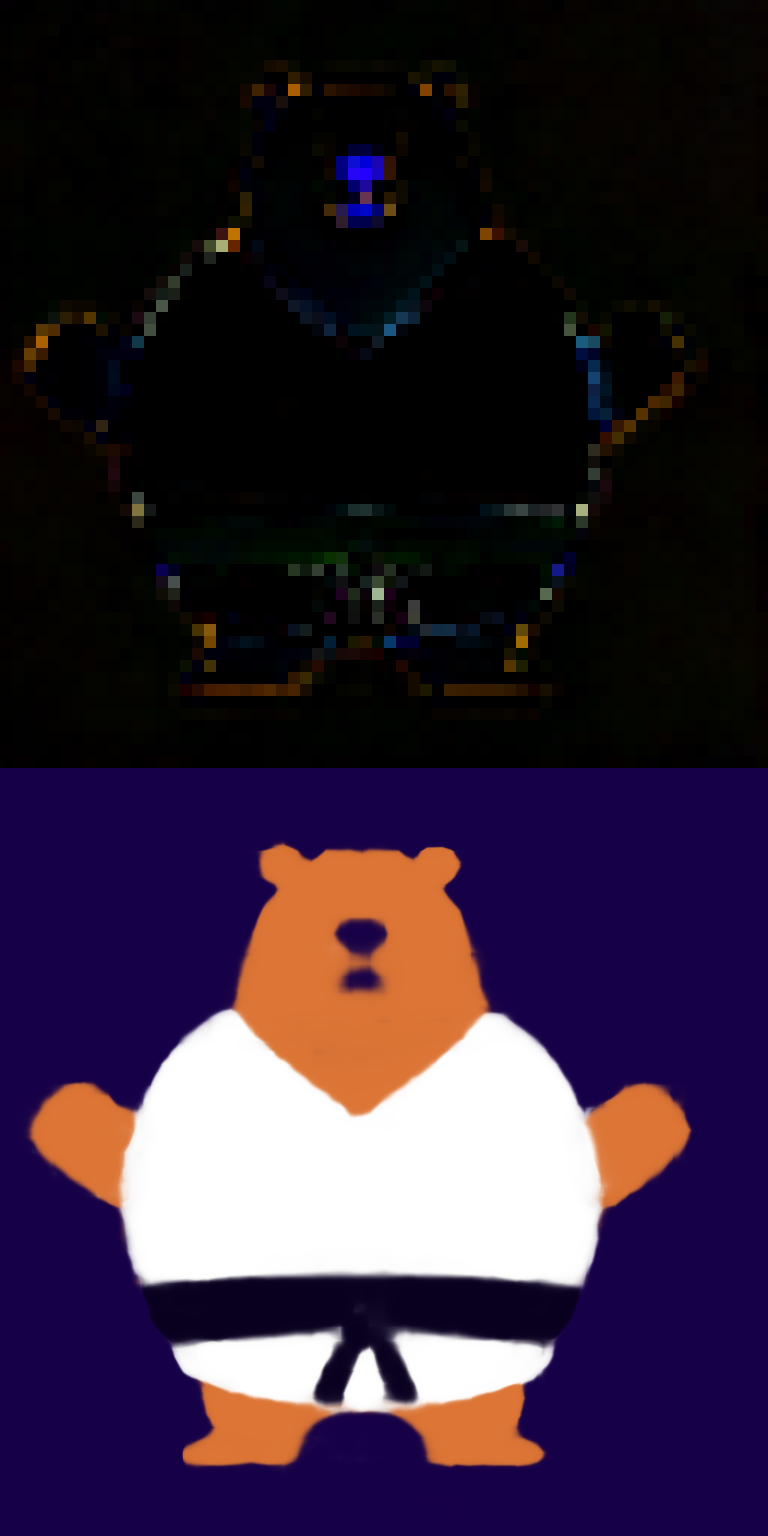} &
    \includegraphics[width=0.06\linewidth]{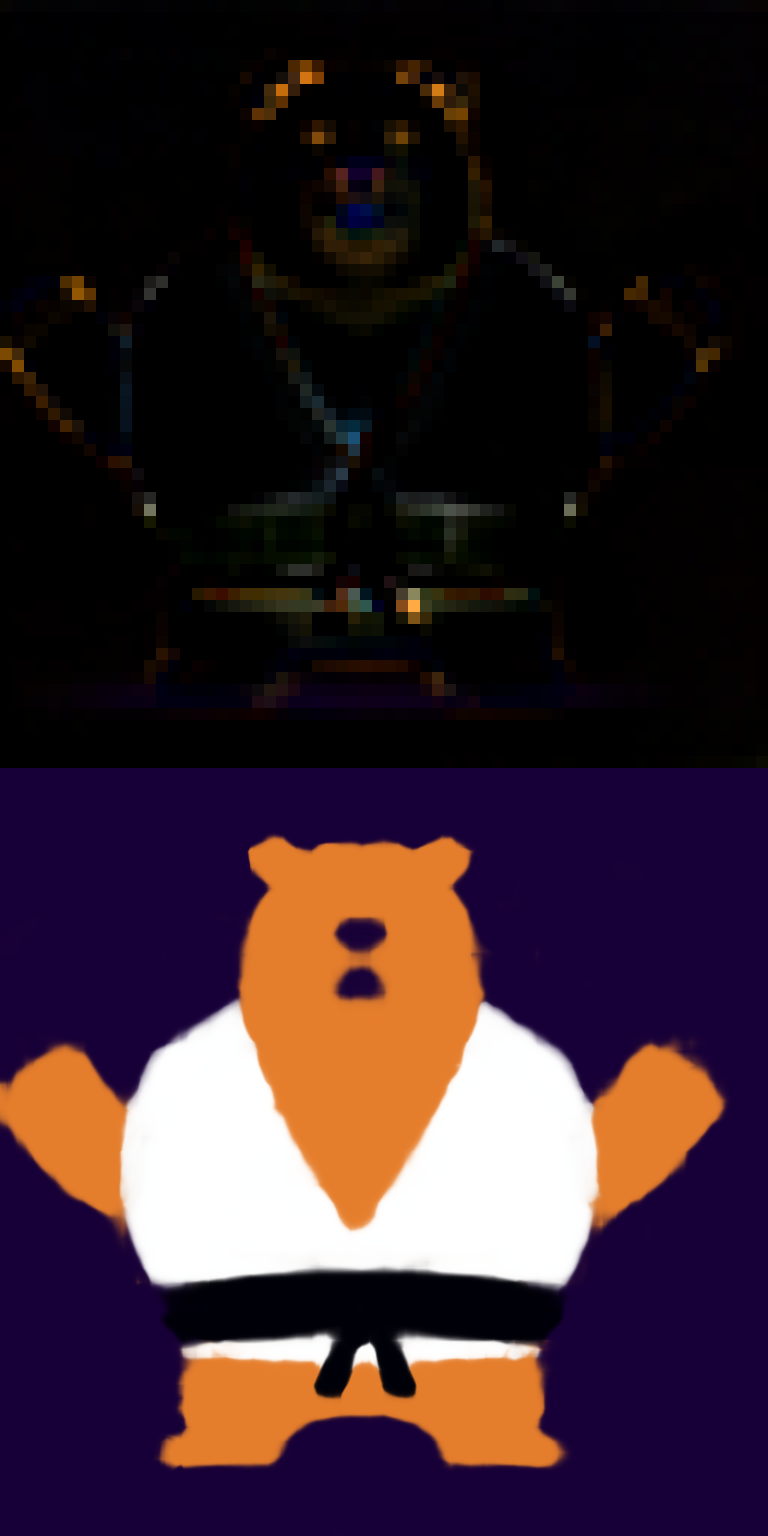} &
    \includegraphics[width=0.06\linewidth]{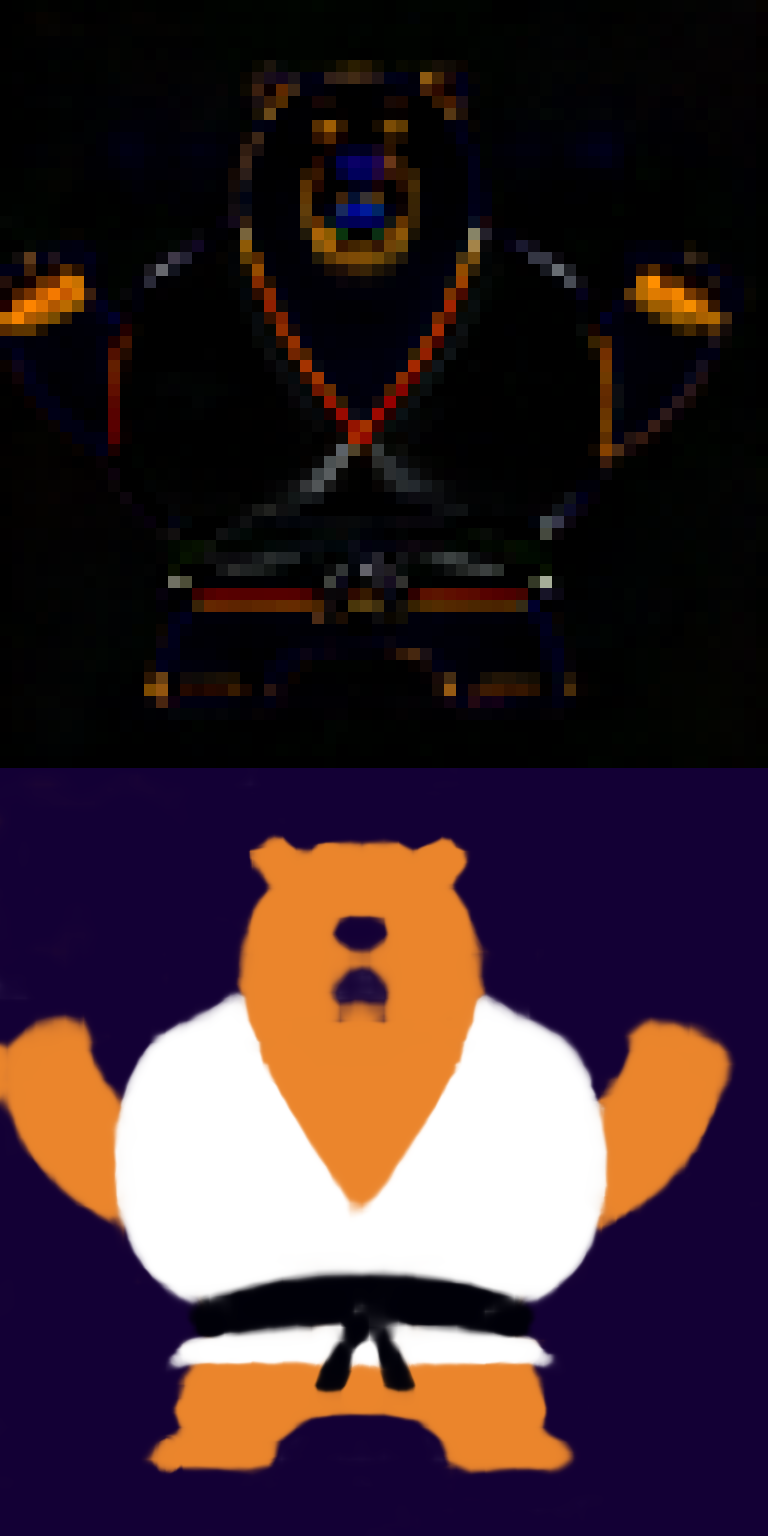} &
    \includegraphics[width=0.06\linewidth]{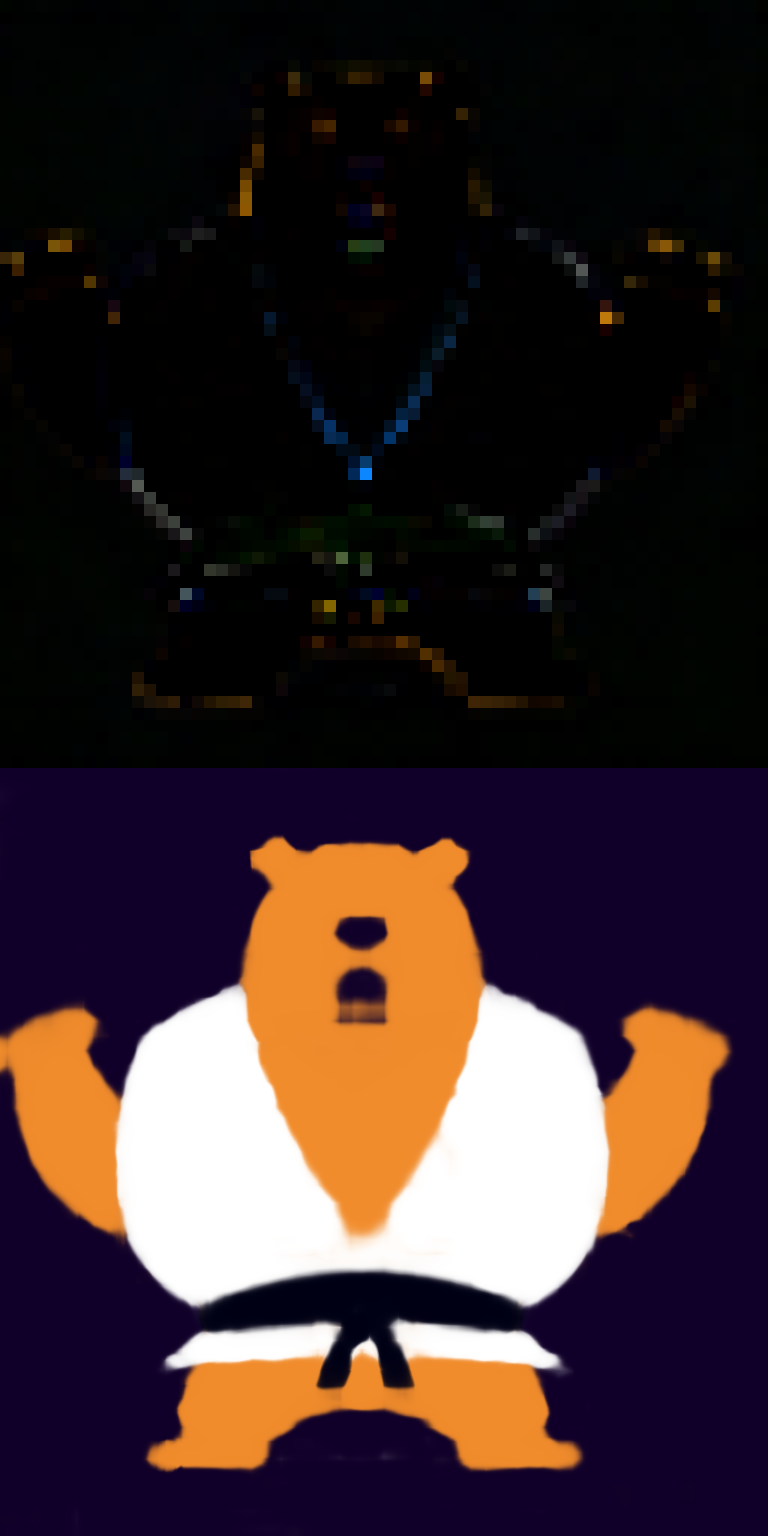} &
    \includegraphics[width=0.06\linewidth]{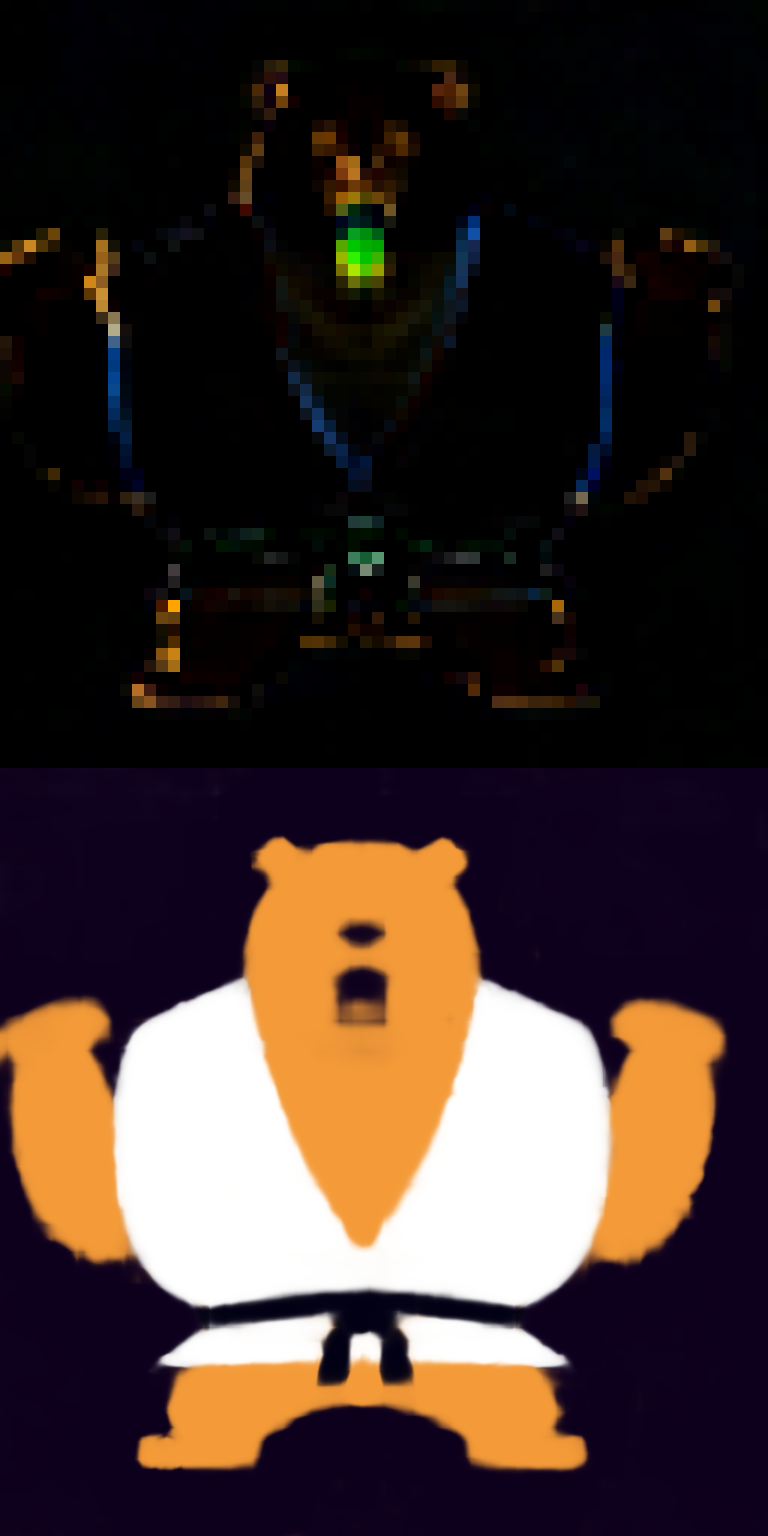} &
    \includegraphics[width=0.06\linewidth]{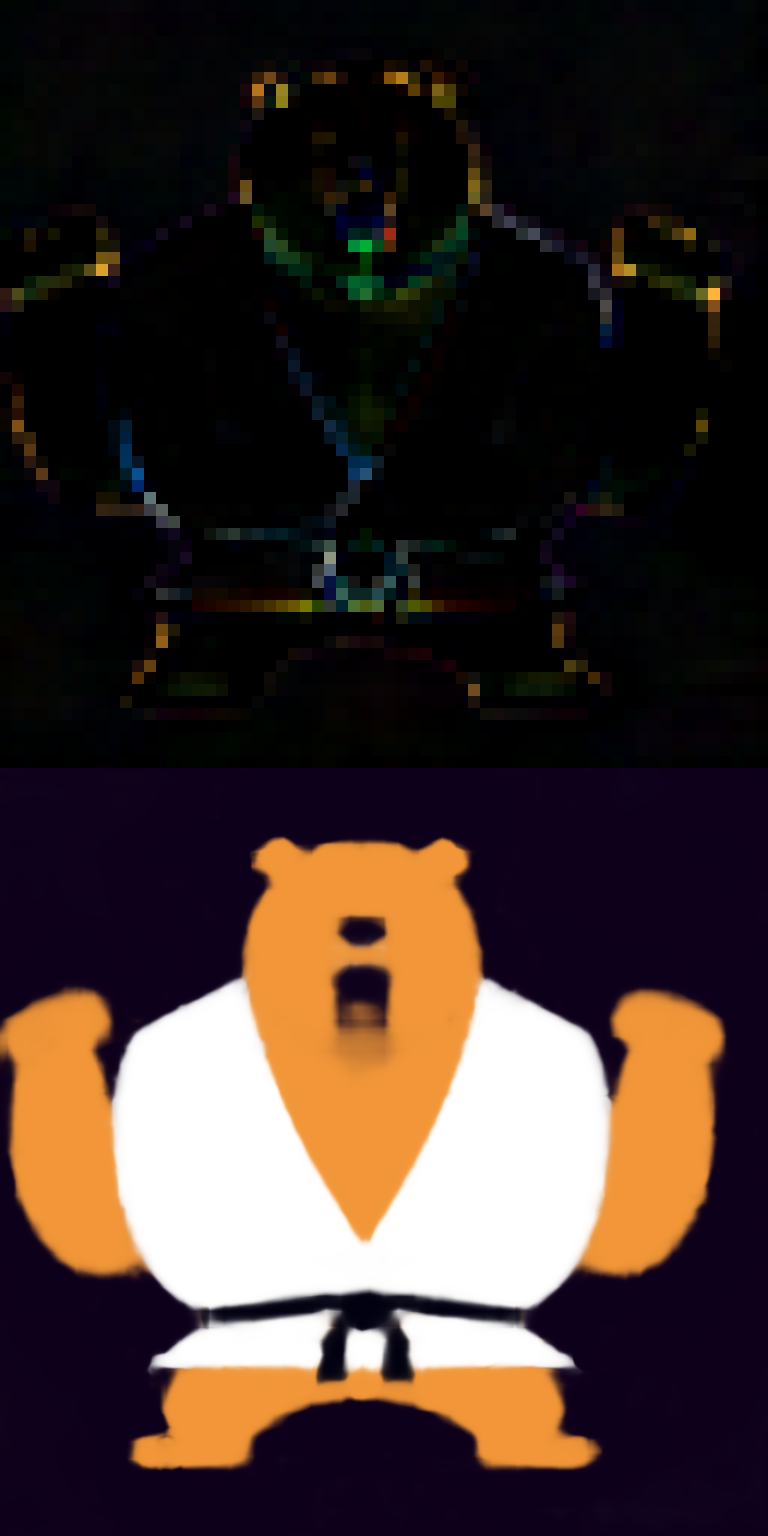} &
    \includegraphics[width=0.06\linewidth]{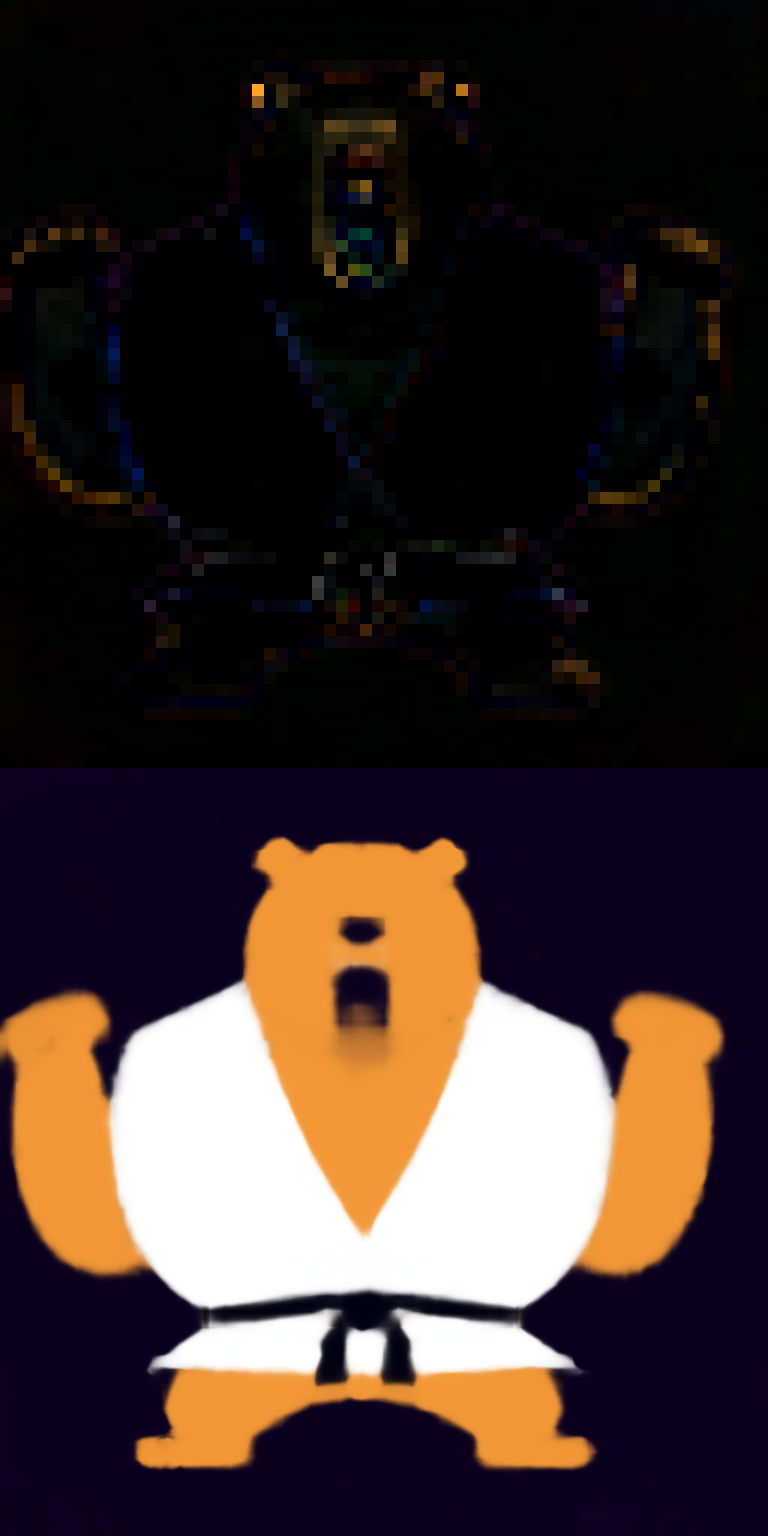} &
    \includegraphics[width=0.06\linewidth]{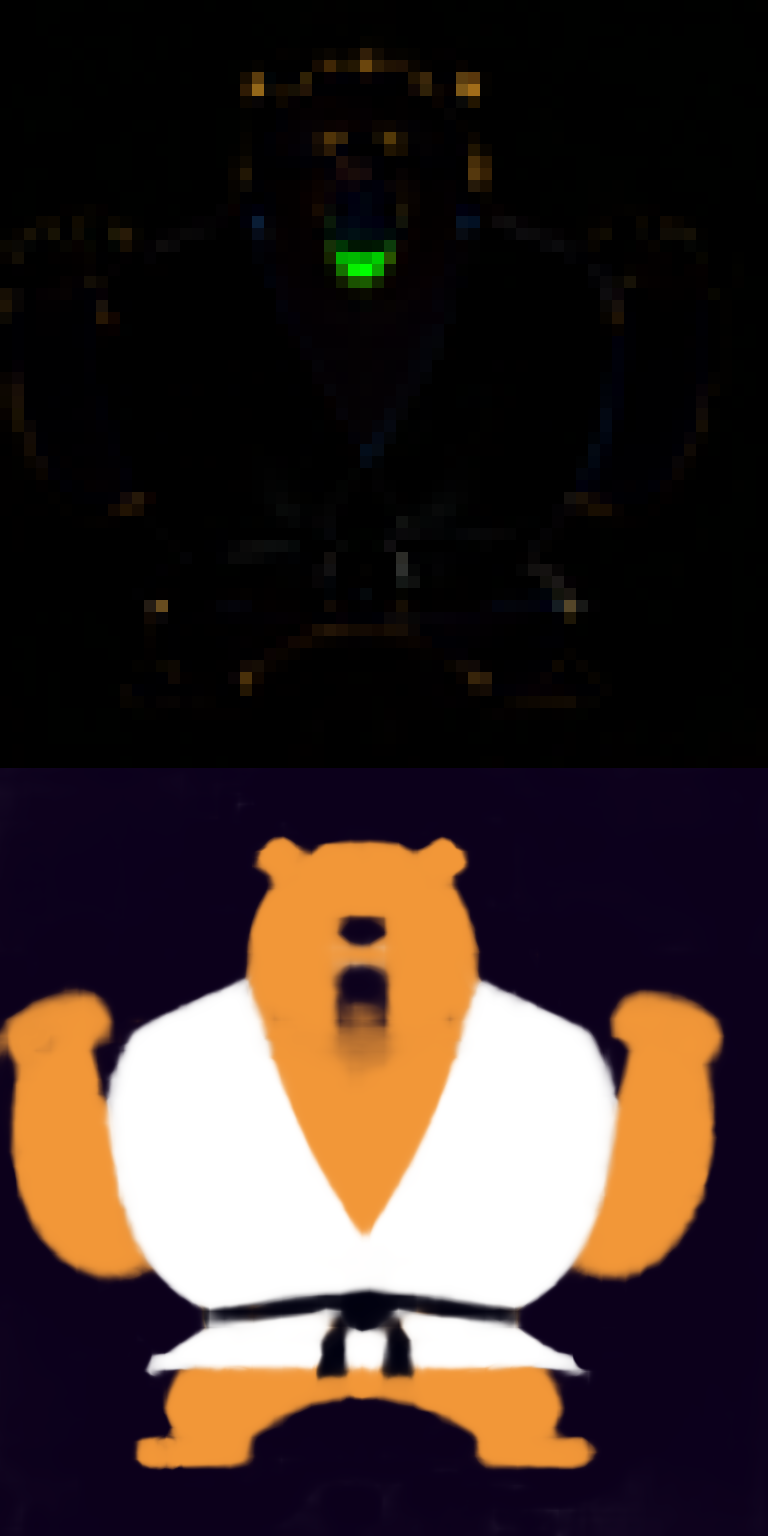} &
    \includegraphics[width=0.06\linewidth]{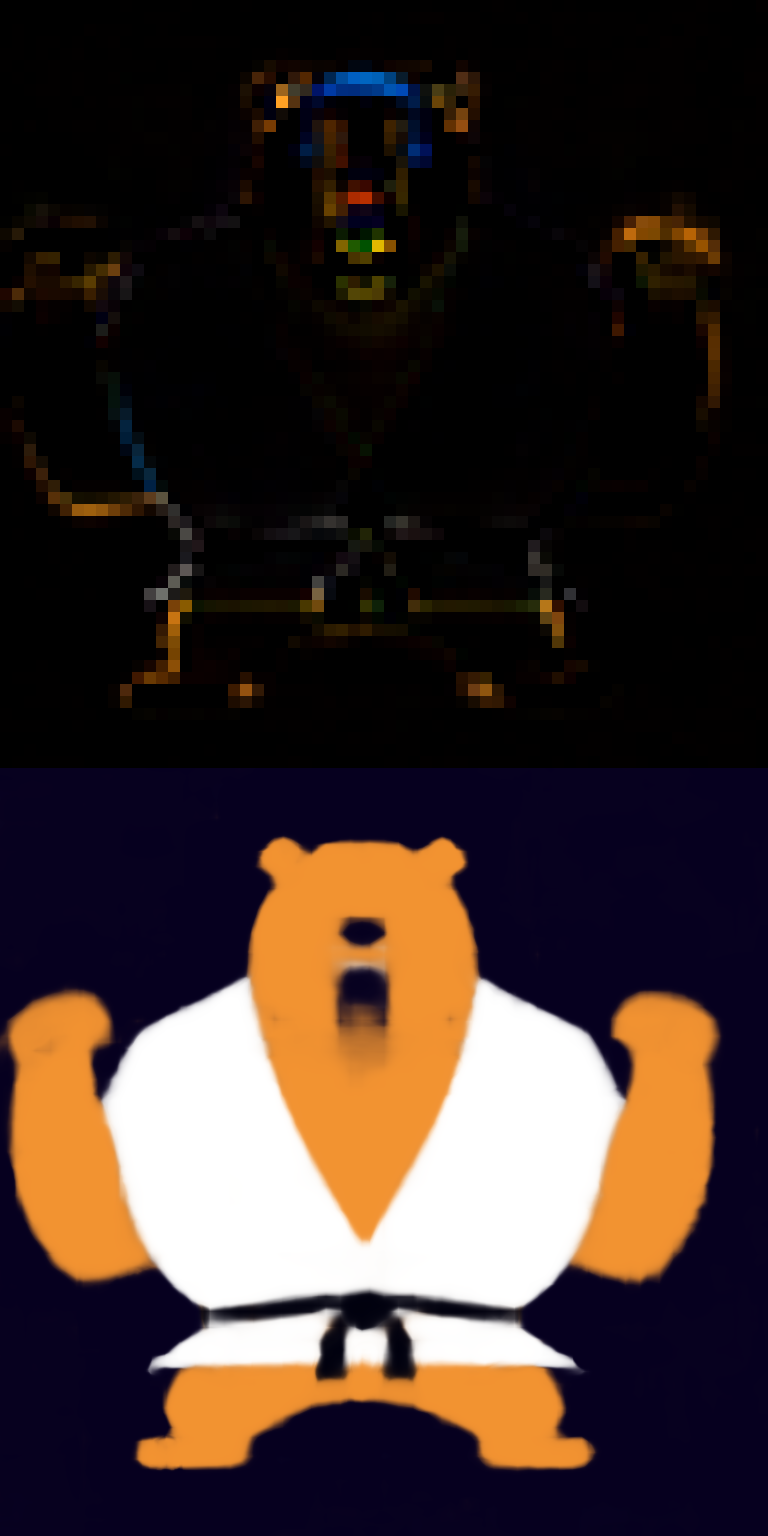} \\
    \multicolumn{16}{c}{}
    \\
    \footnotesize{\emph{Box}}&
    \multicolumn{15}{c}{\emph{\textcolor{textprompts}{"A flamingo karate master ..."}}}
    \\
    \includegraphics[width=0.06\linewidth]{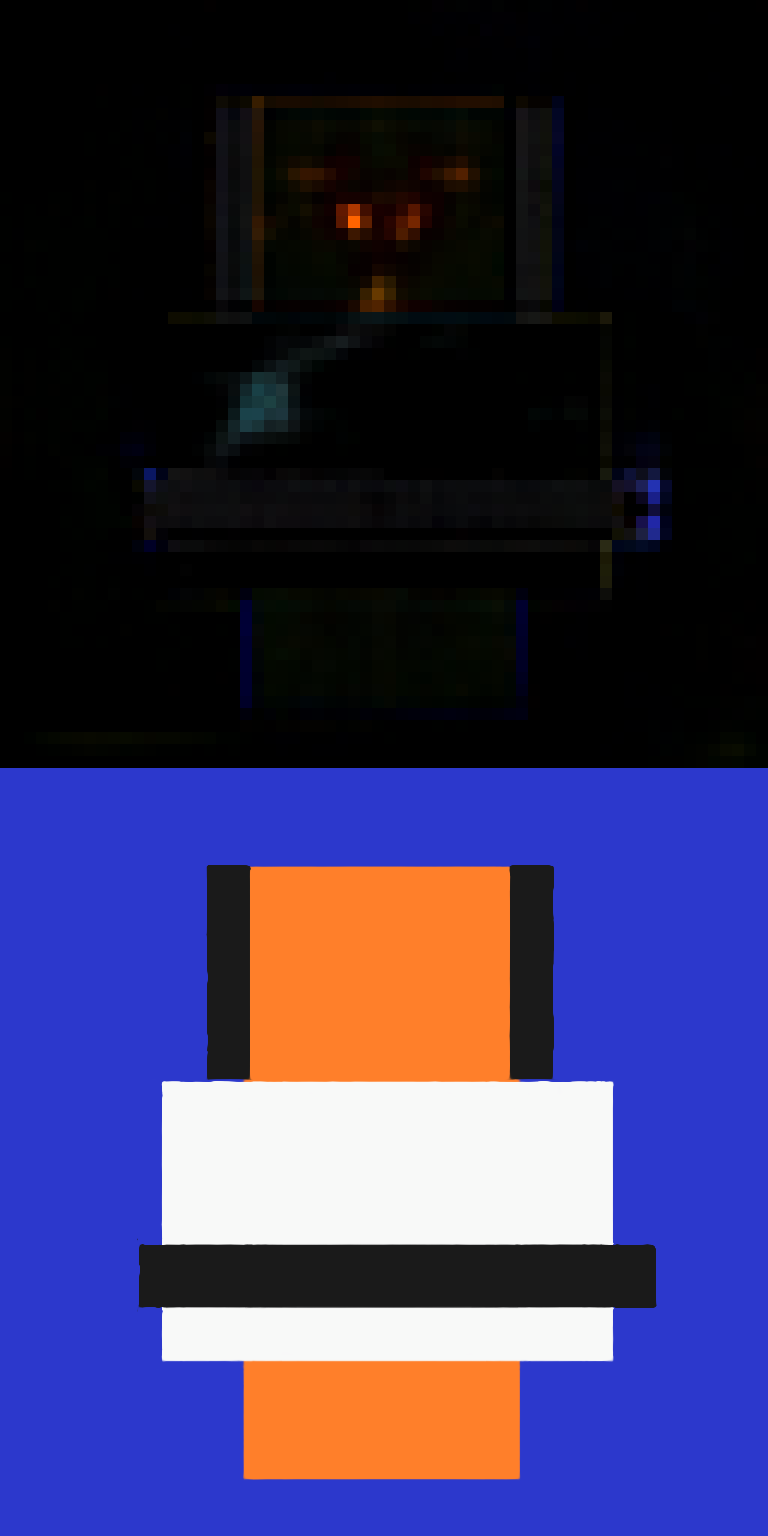} &
    \includegraphics[width=0.06\linewidth]{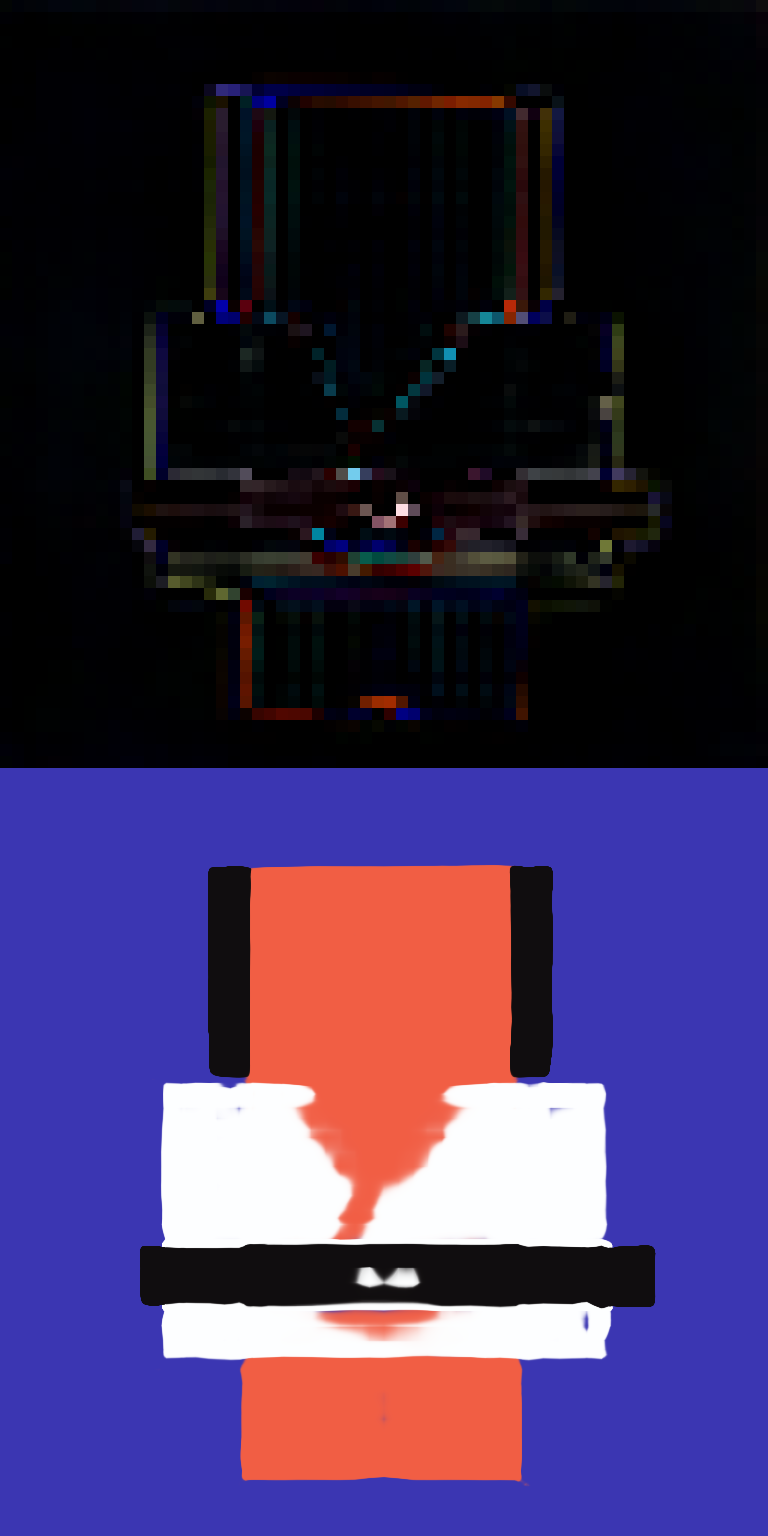} &
    \includegraphics[width=0.06\linewidth]{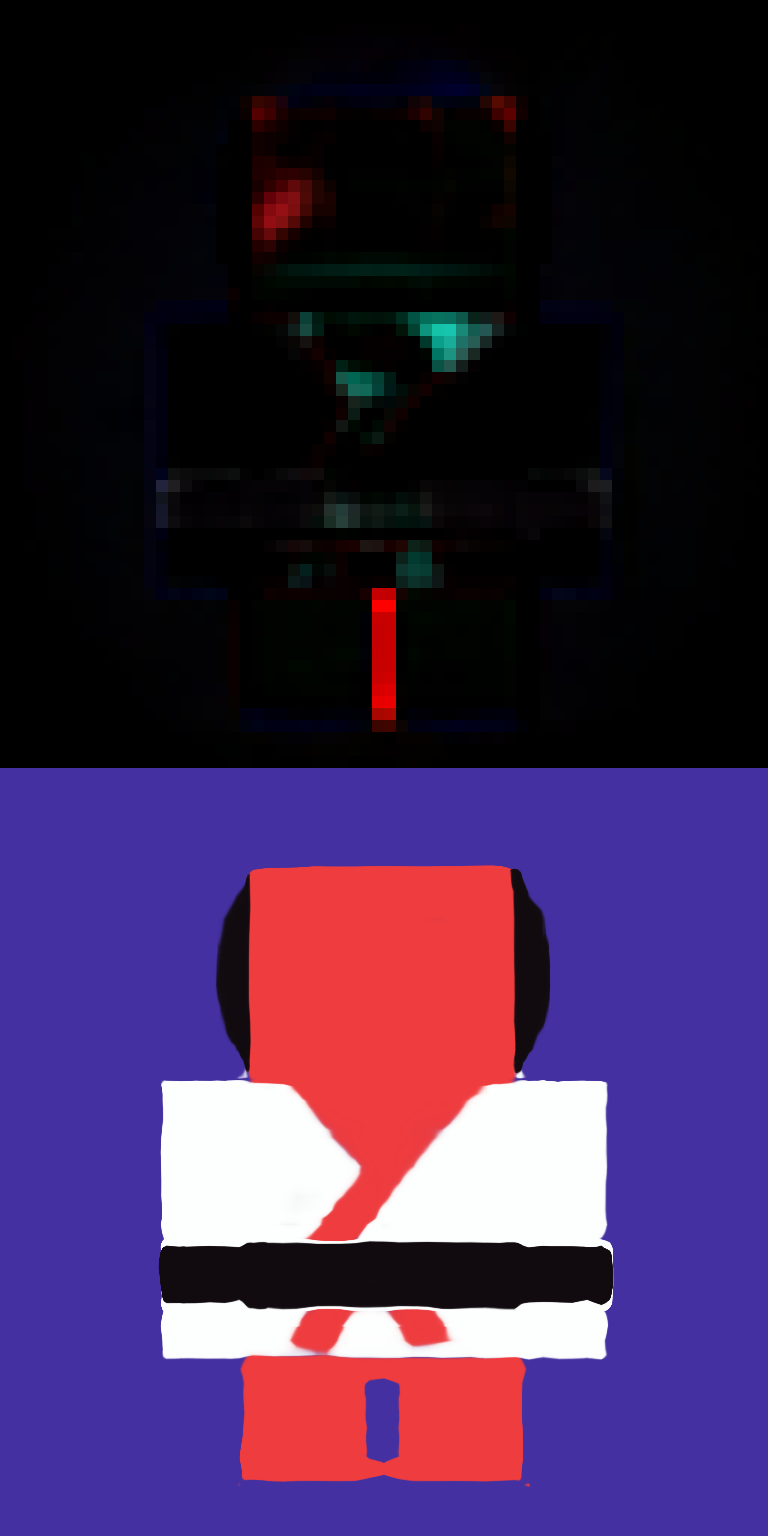} &
    \includegraphics[width=0.06\linewidth]{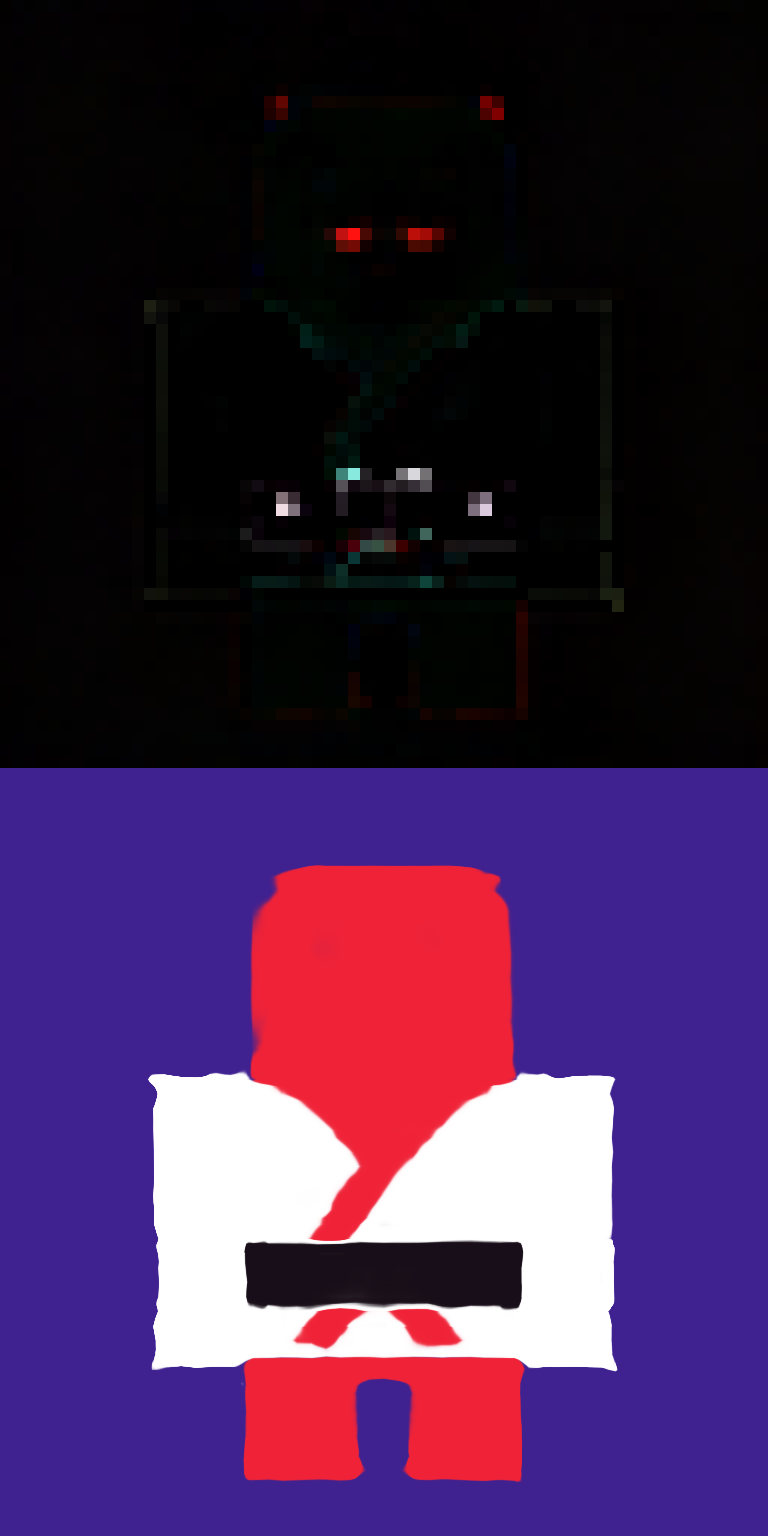} &
    \includegraphics[width=0.06\linewidth]{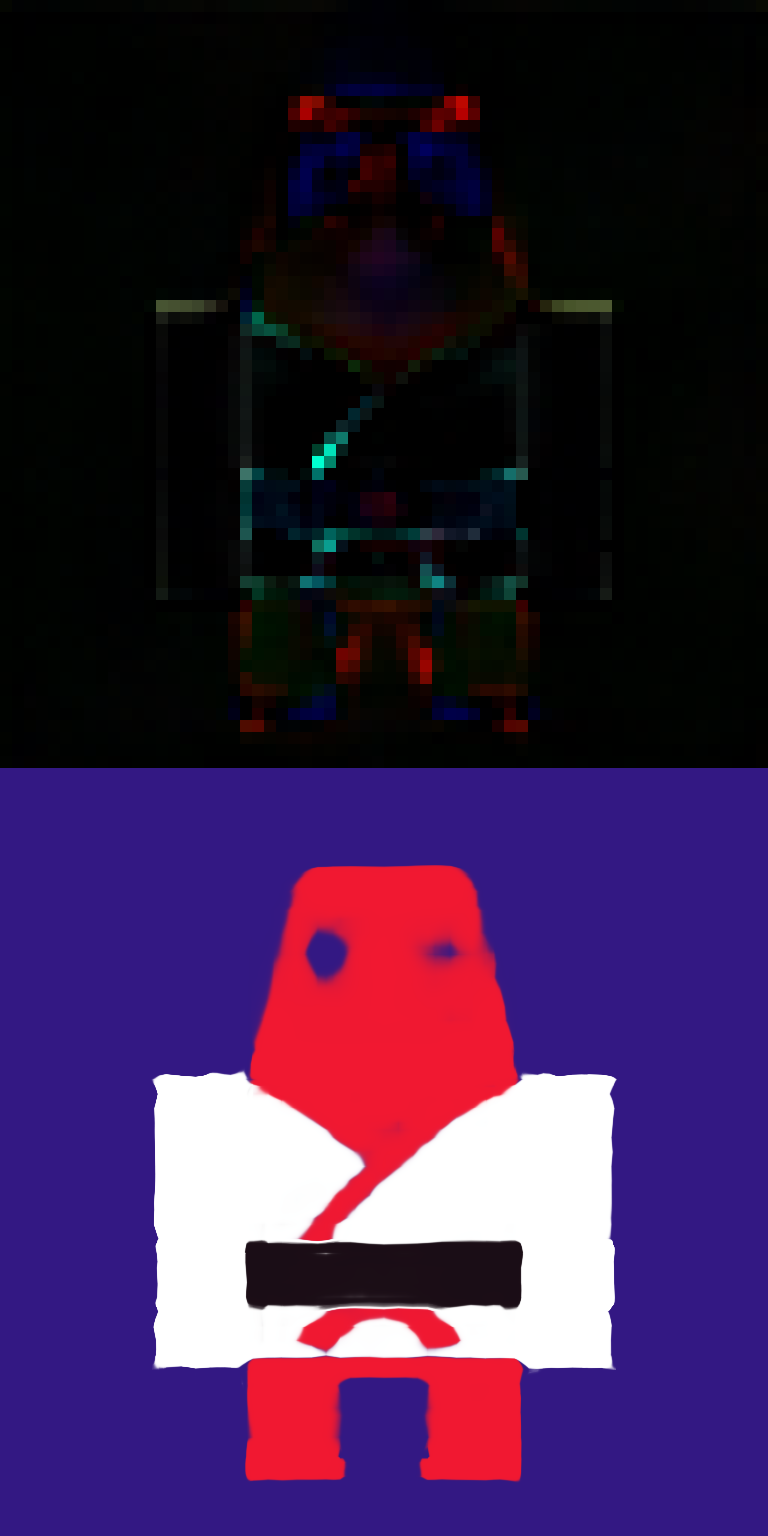} &
    \includegraphics[width=0.06\linewidth]{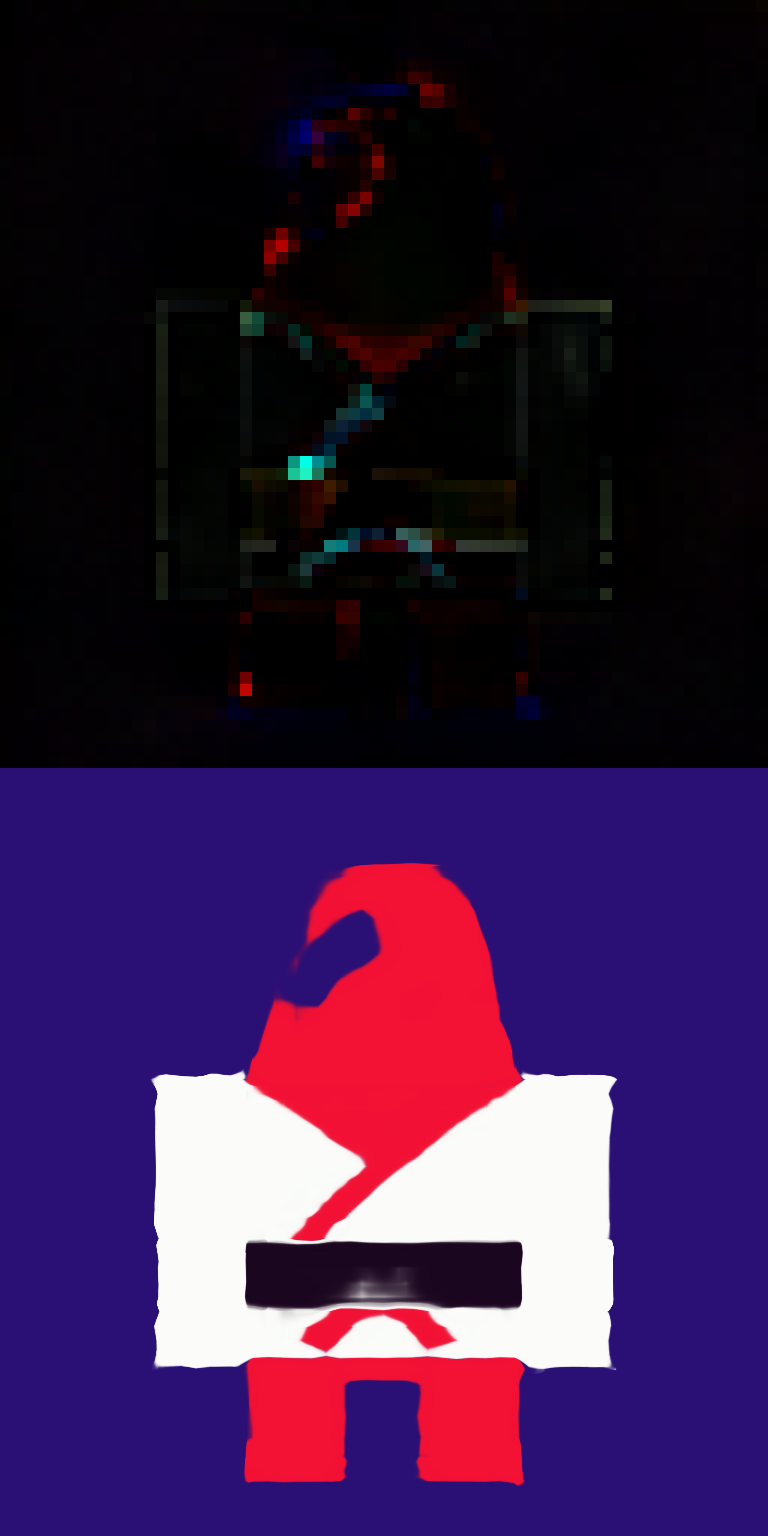} &
    \includegraphics[width=0.06\linewidth]{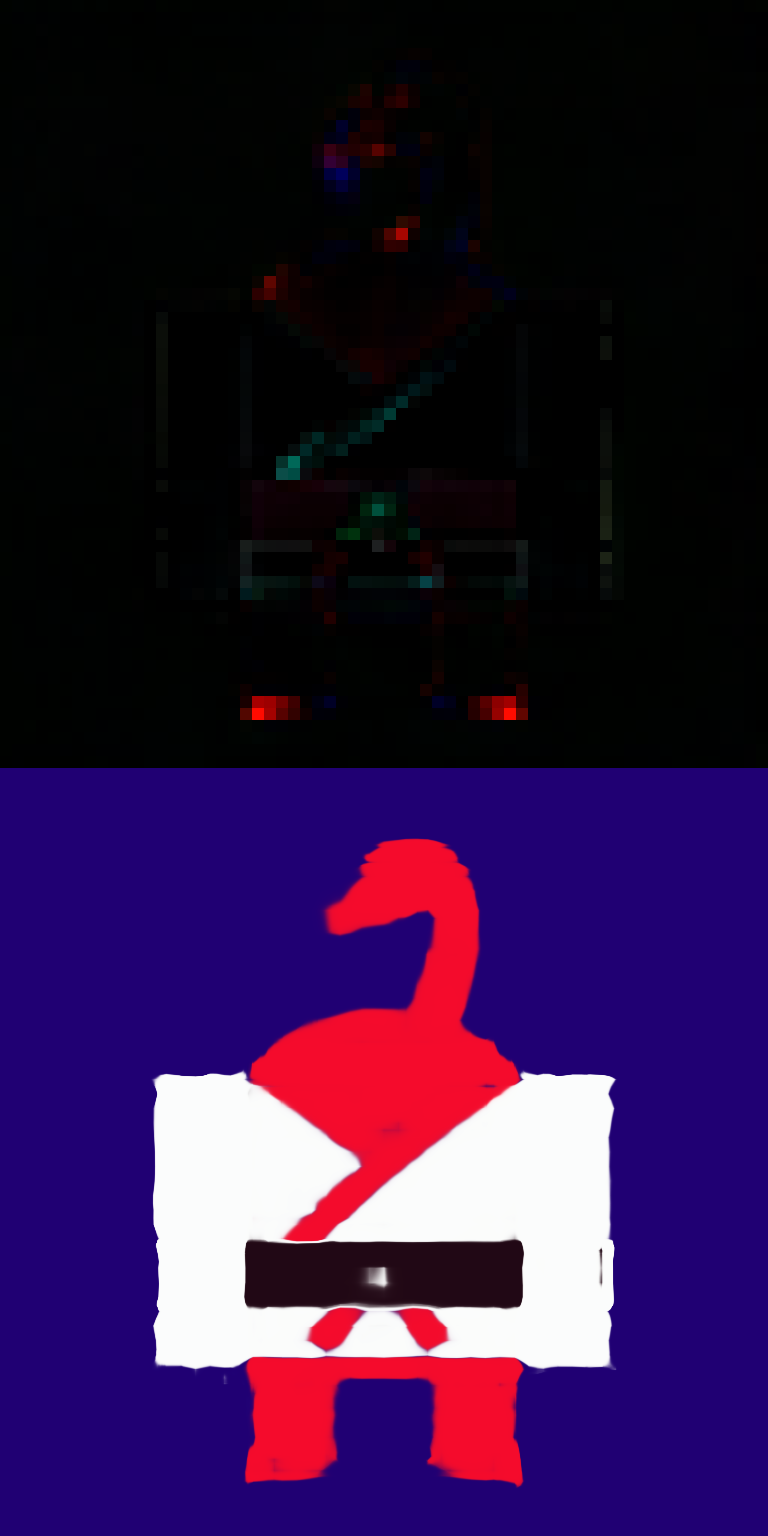} &
    \includegraphics[width=0.06\linewidth]{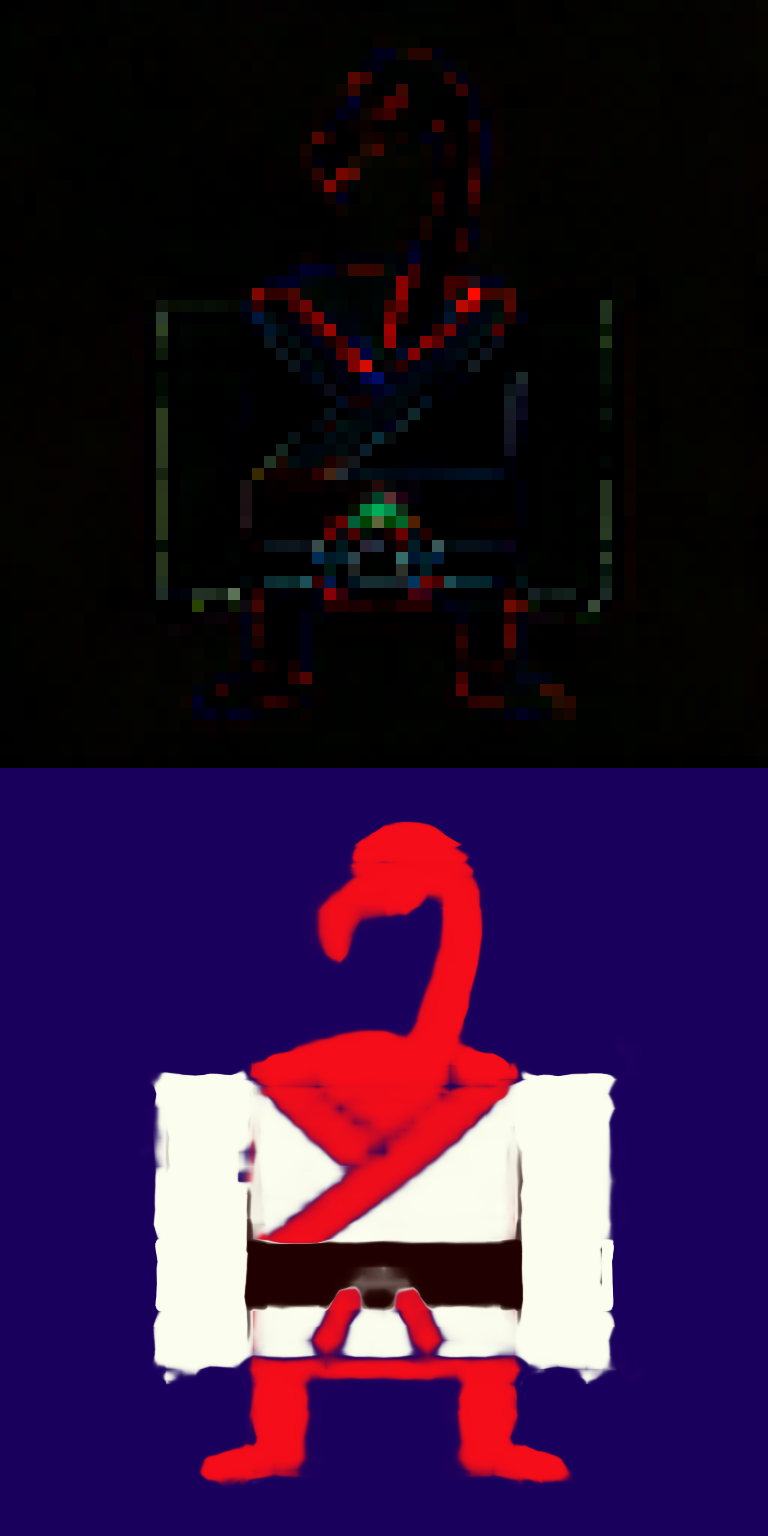} &
    \includegraphics[width=0.06\linewidth]{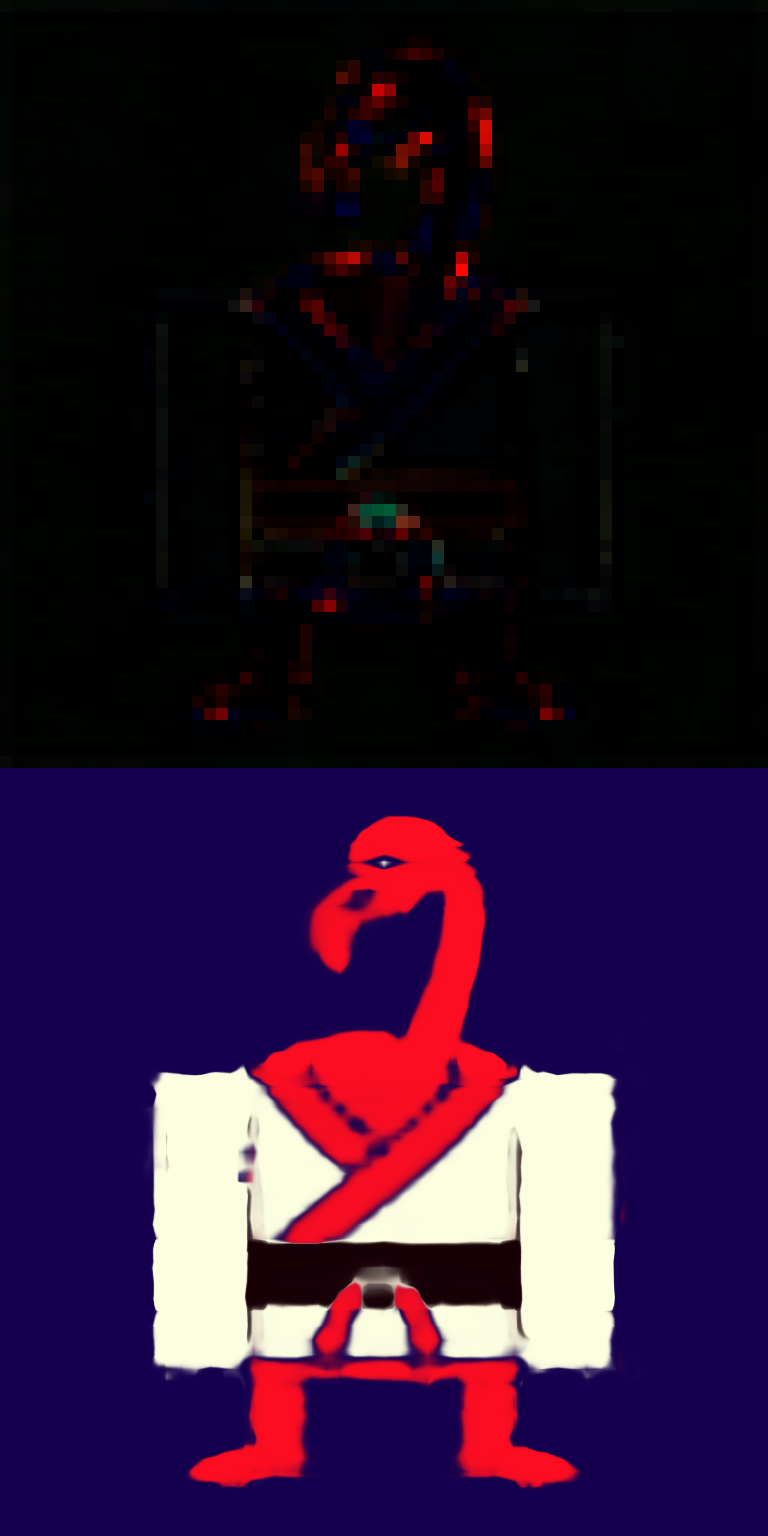} &
    \includegraphics[width=0.06\linewidth]{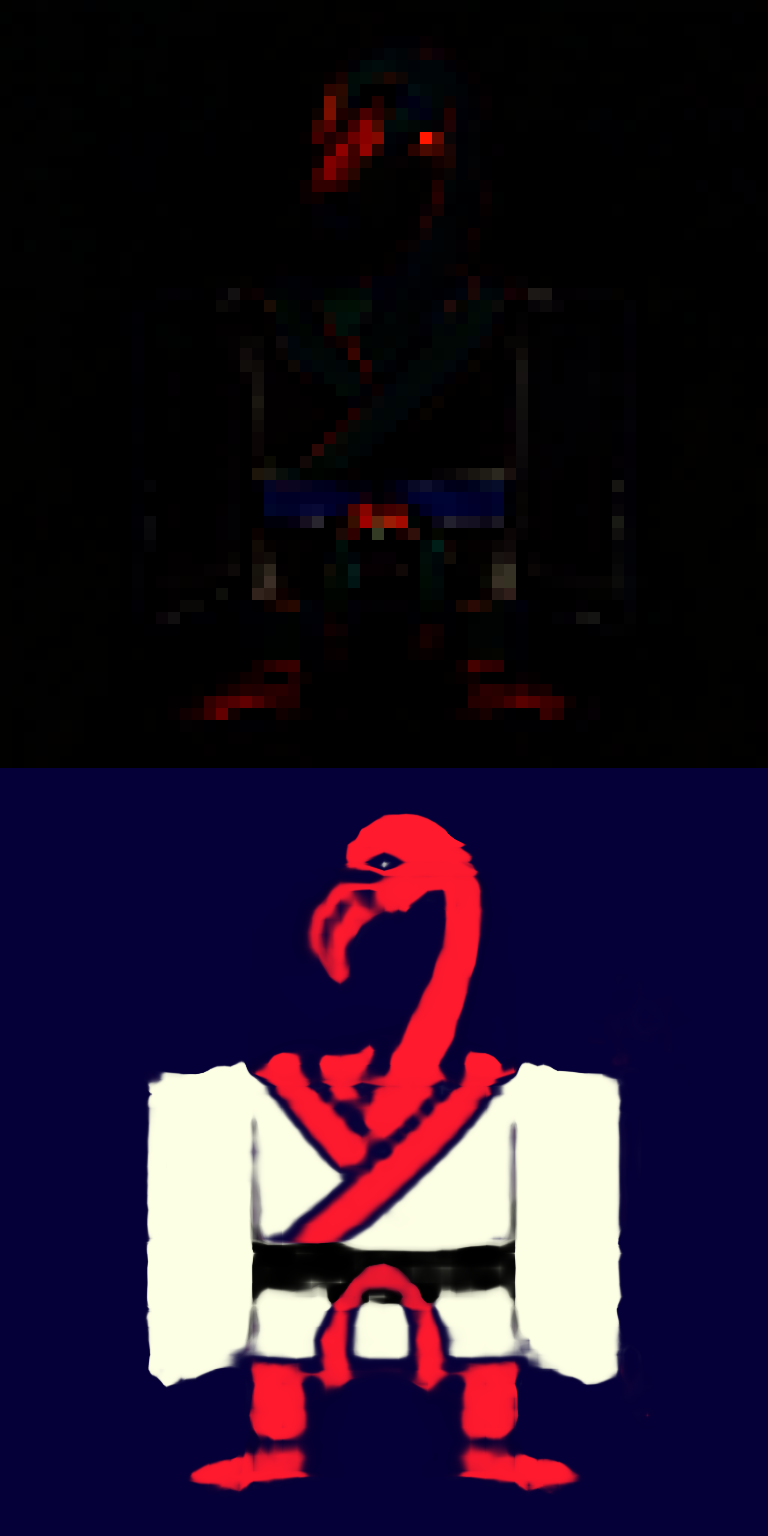} &
    \includegraphics[width=0.06\linewidth]{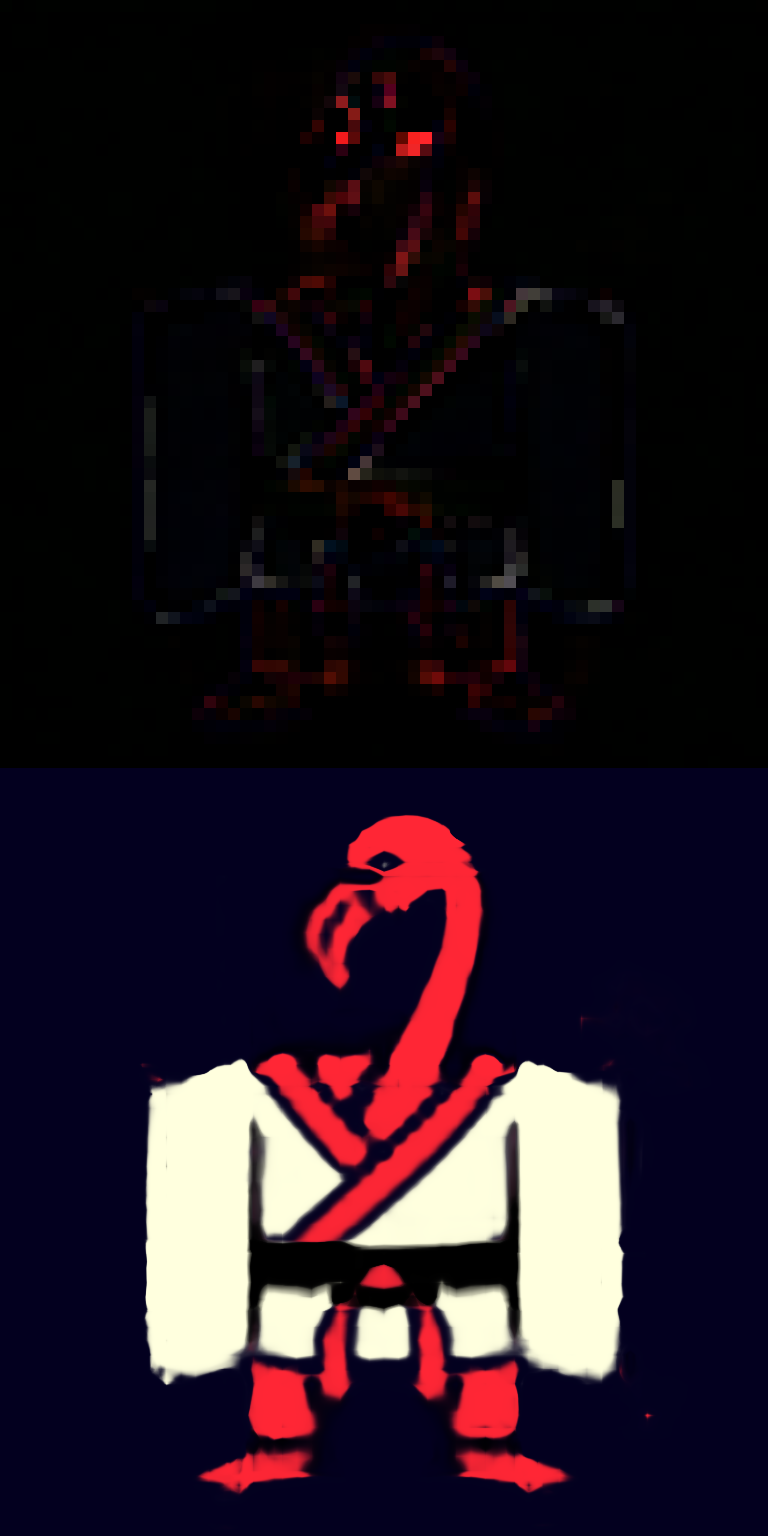} &
    \includegraphics[width=0.06\linewidth]{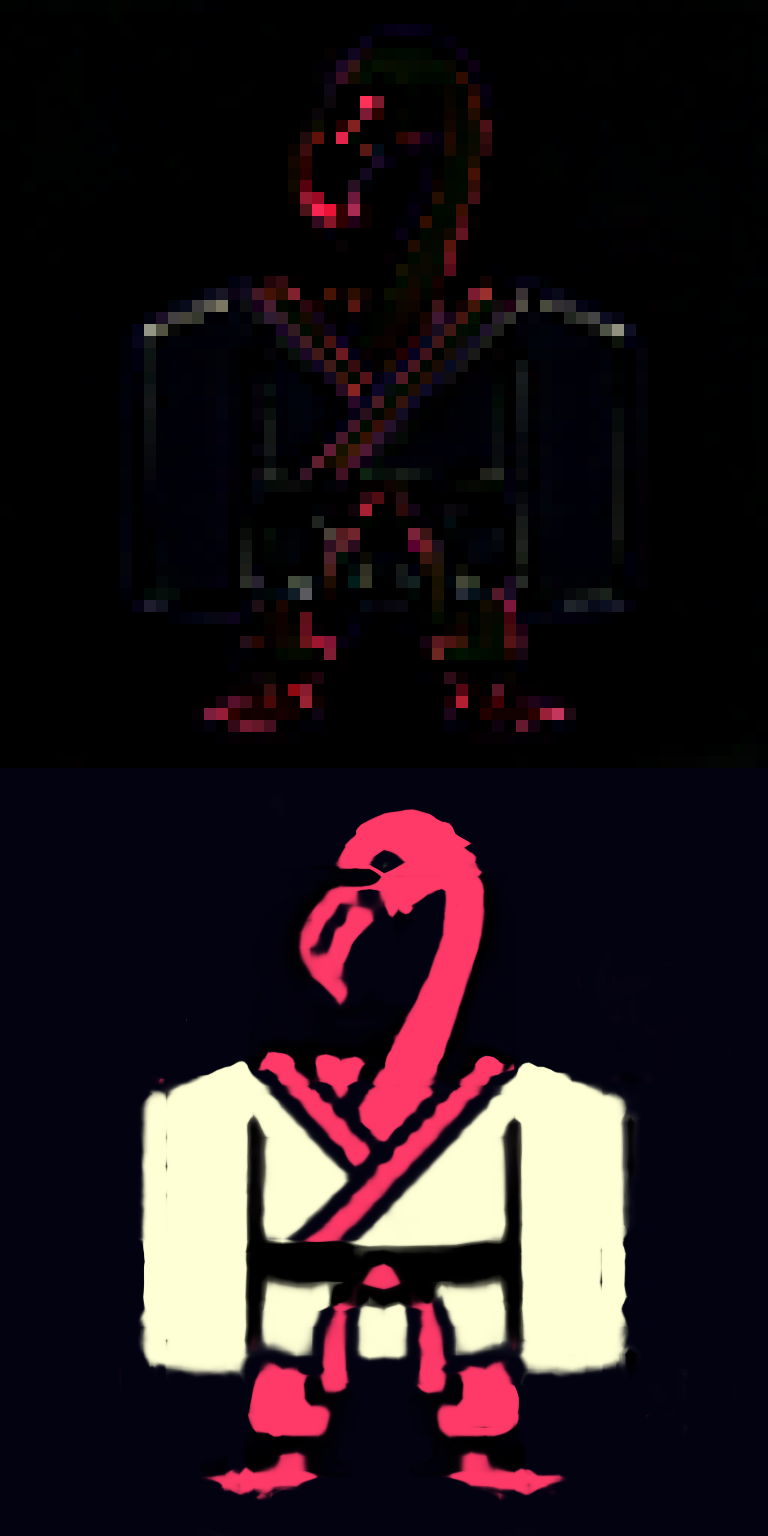} &
    \includegraphics[width=0.06\linewidth]{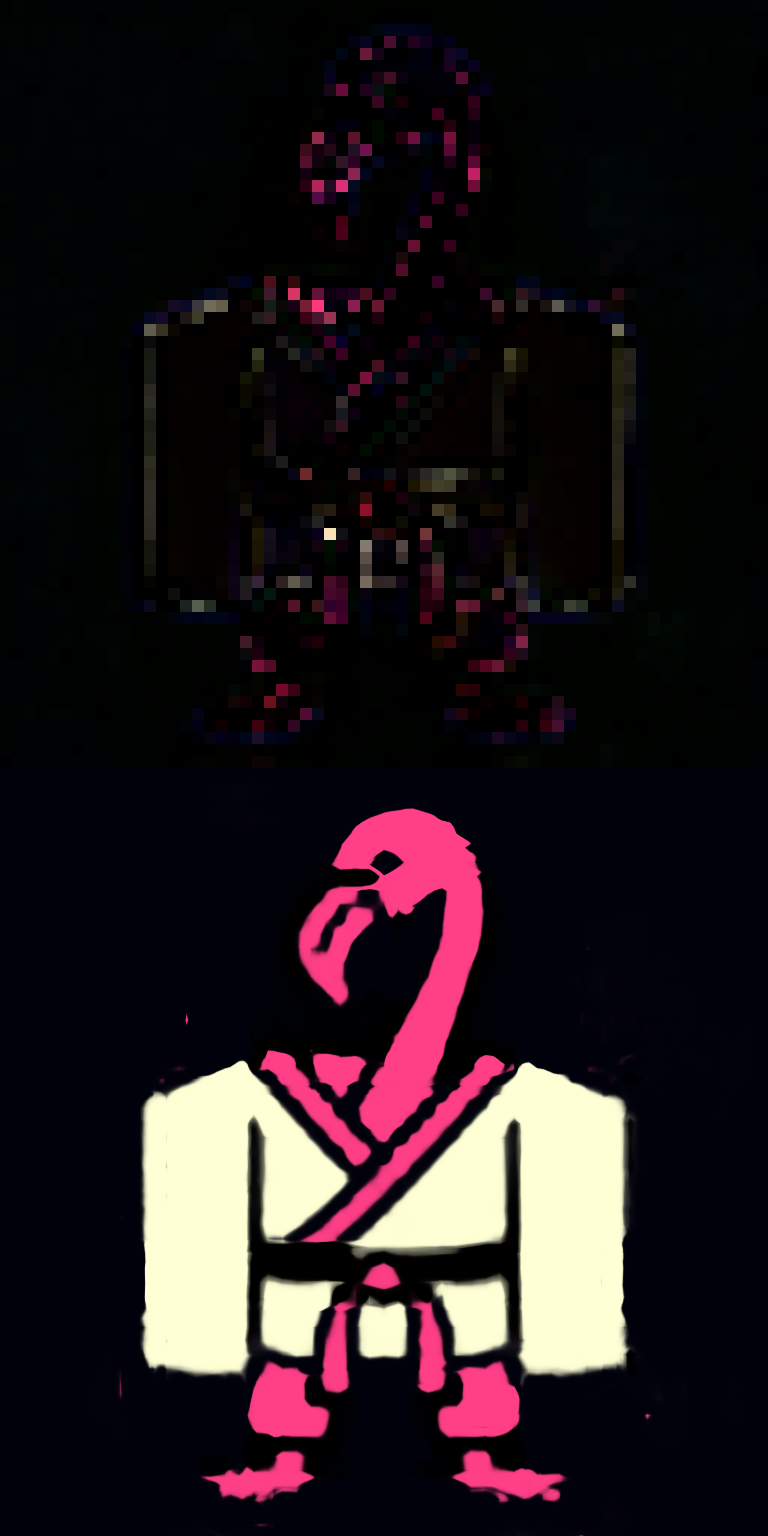} &
    \includegraphics[width=0.06\linewidth]{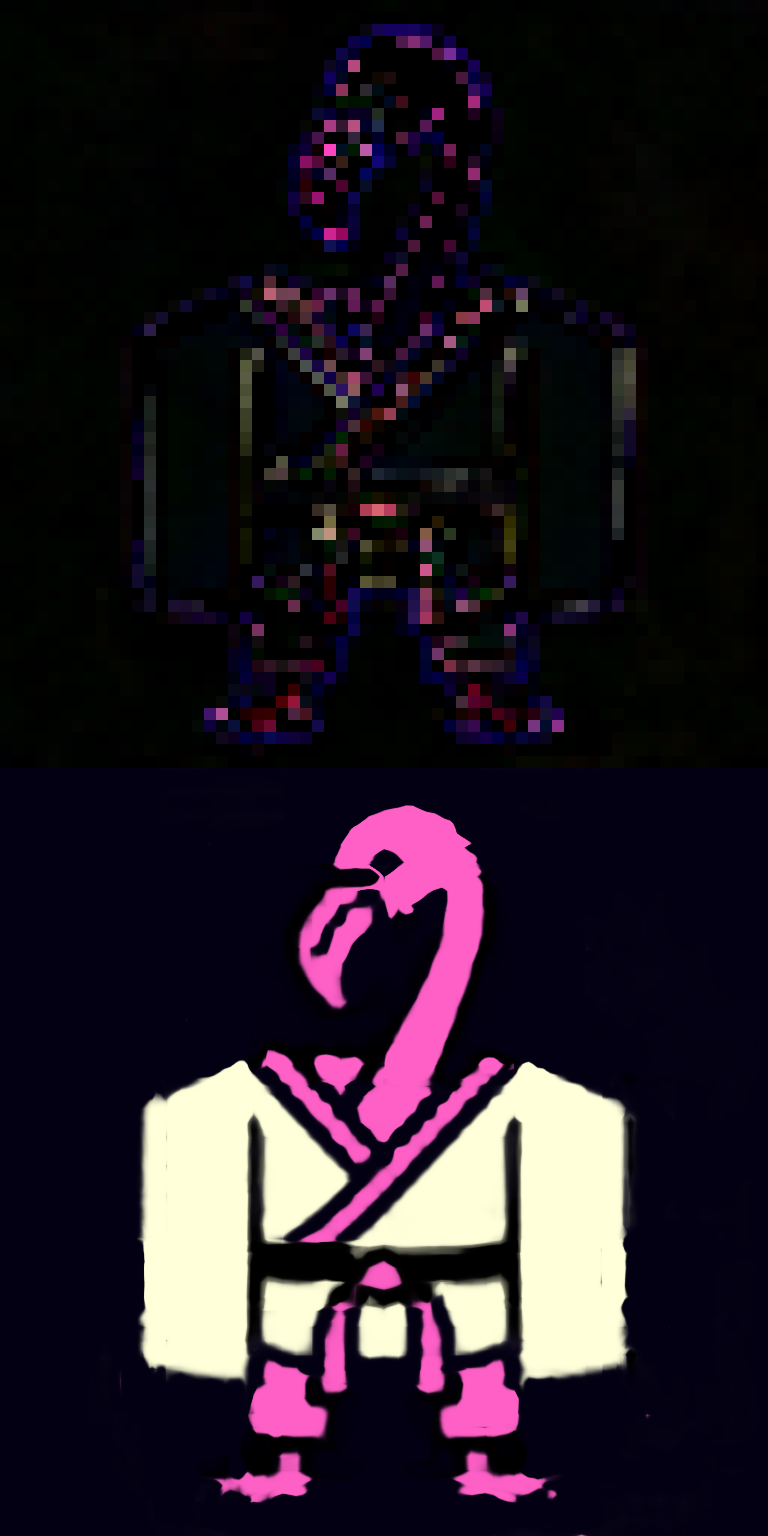} &
    \includegraphics[width=0.06\linewidth]{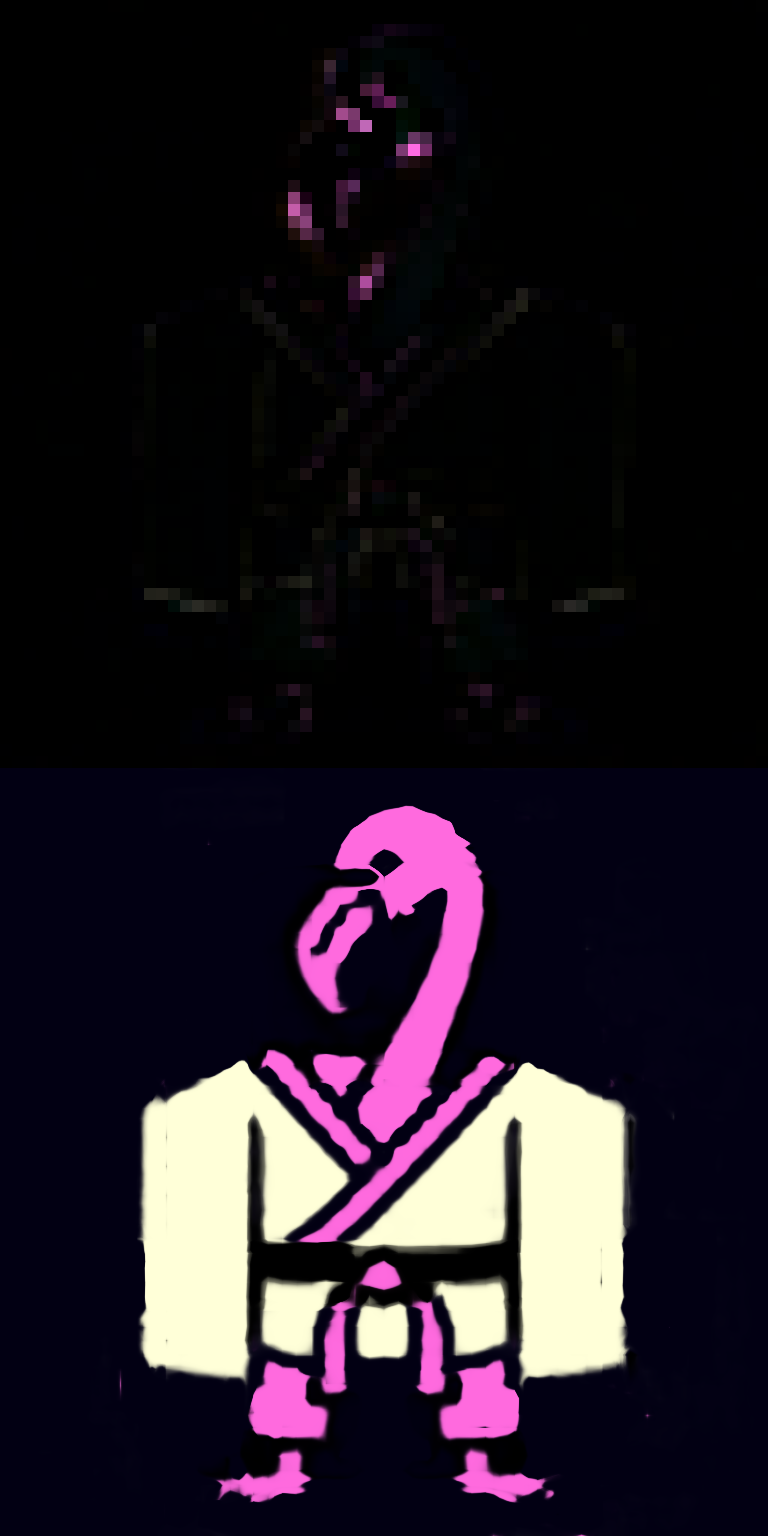} &
    \includegraphics[width=0.06\linewidth]{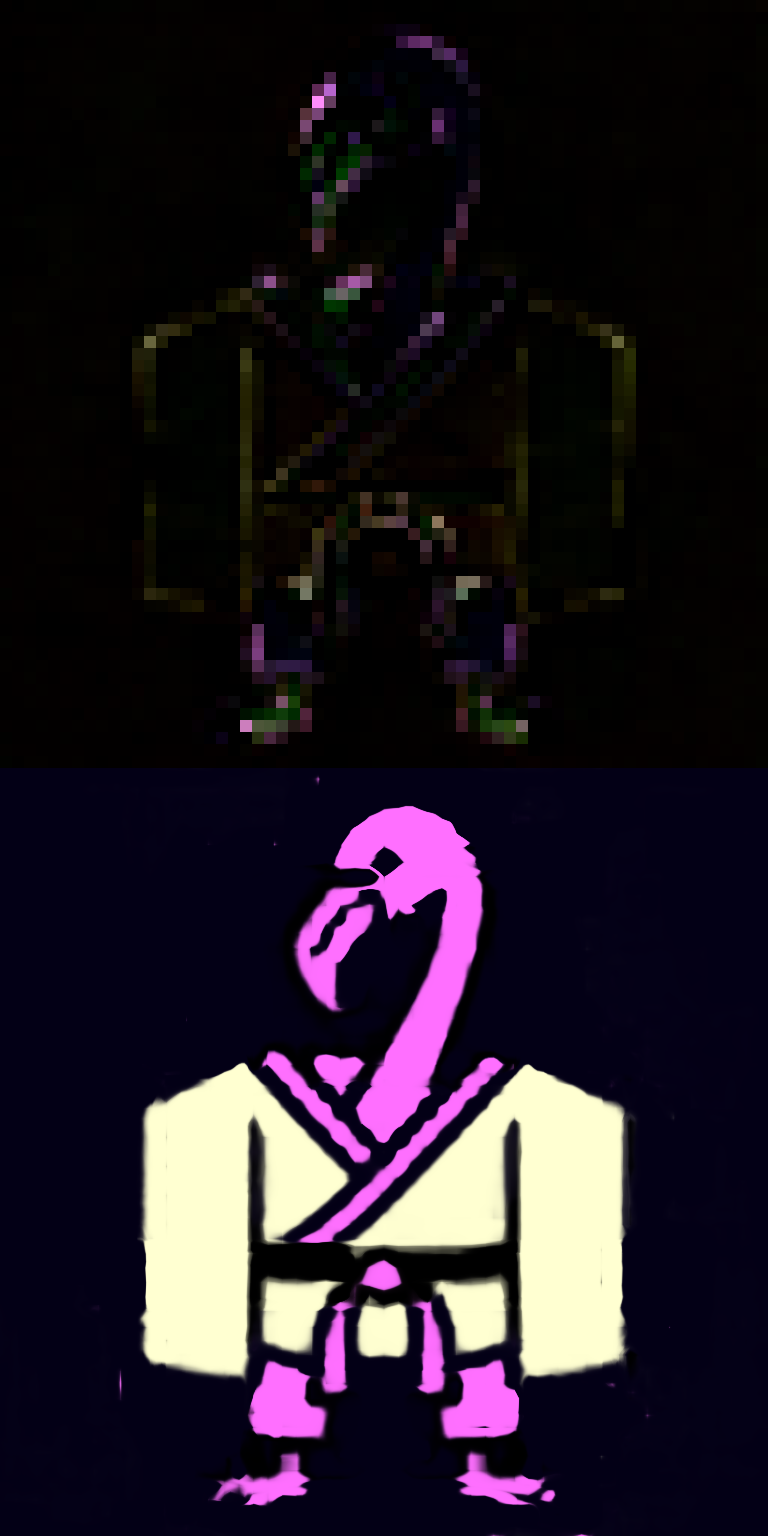} \\
    \footnotesize{\emph{Ellipse}}&
    \multicolumn{15}{c}{}
    \\
    \includegraphics[width=0.06\linewidth]{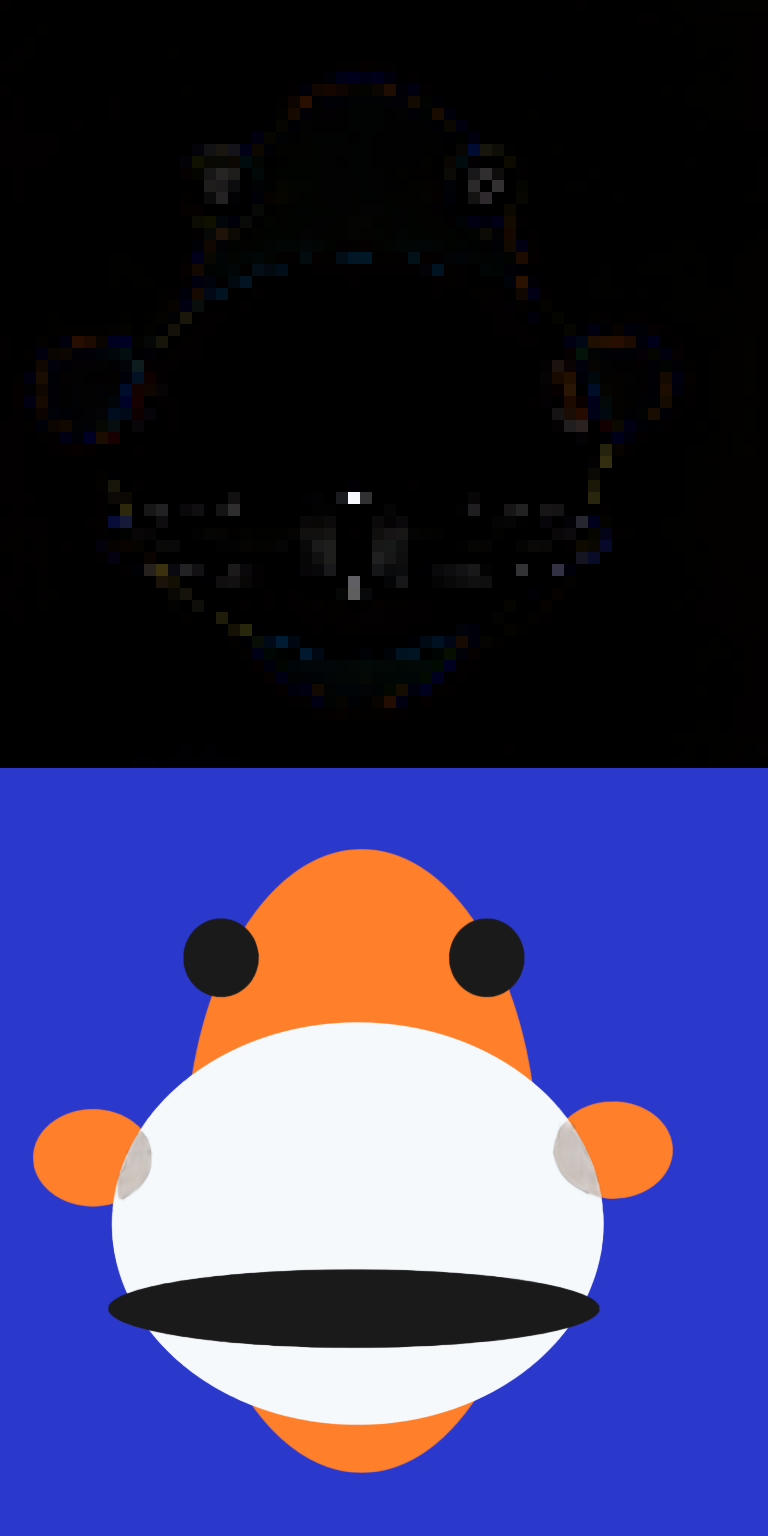} &
    \includegraphics[width=0.06\linewidth]{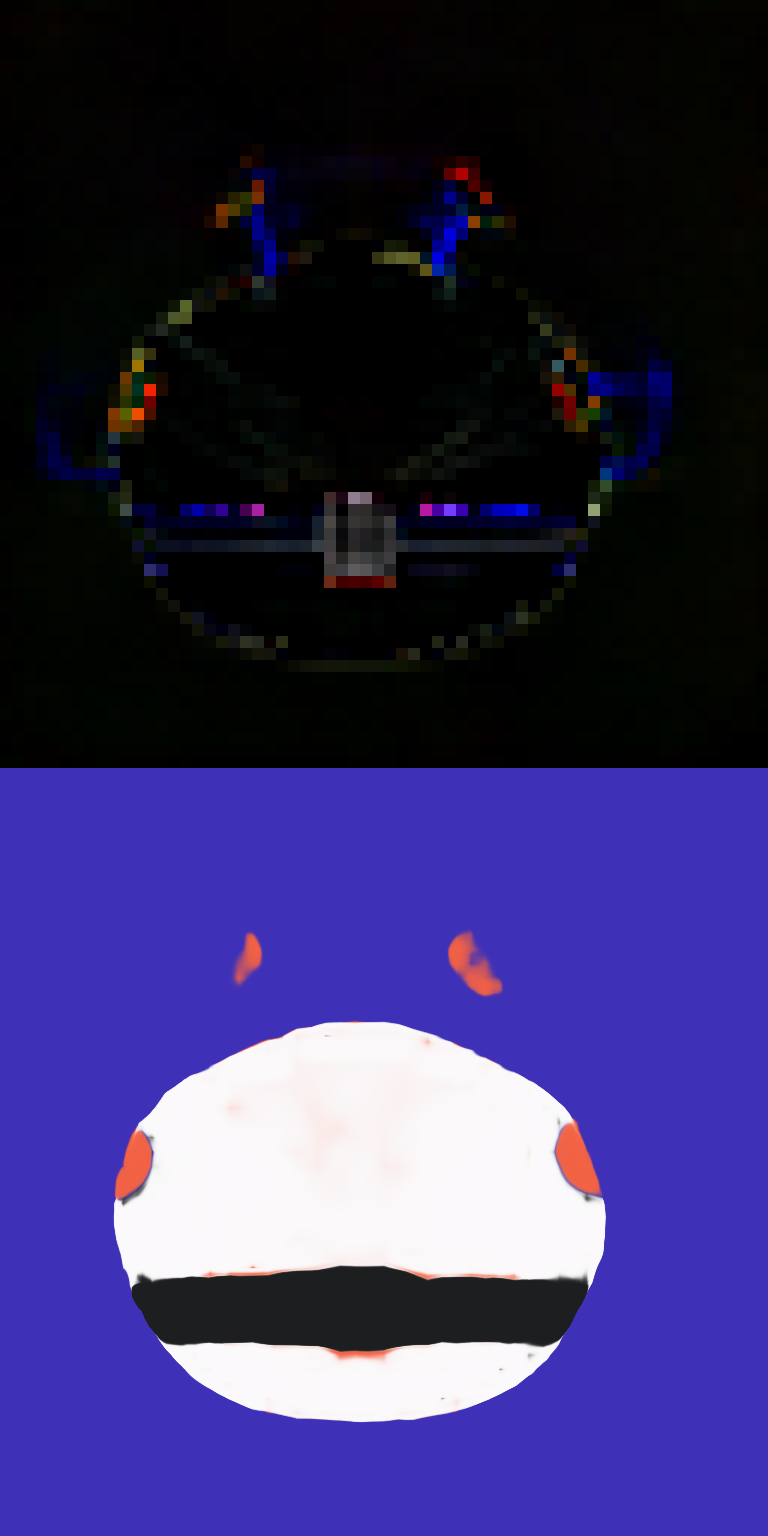} &
    \includegraphics[width=0.06\linewidth]{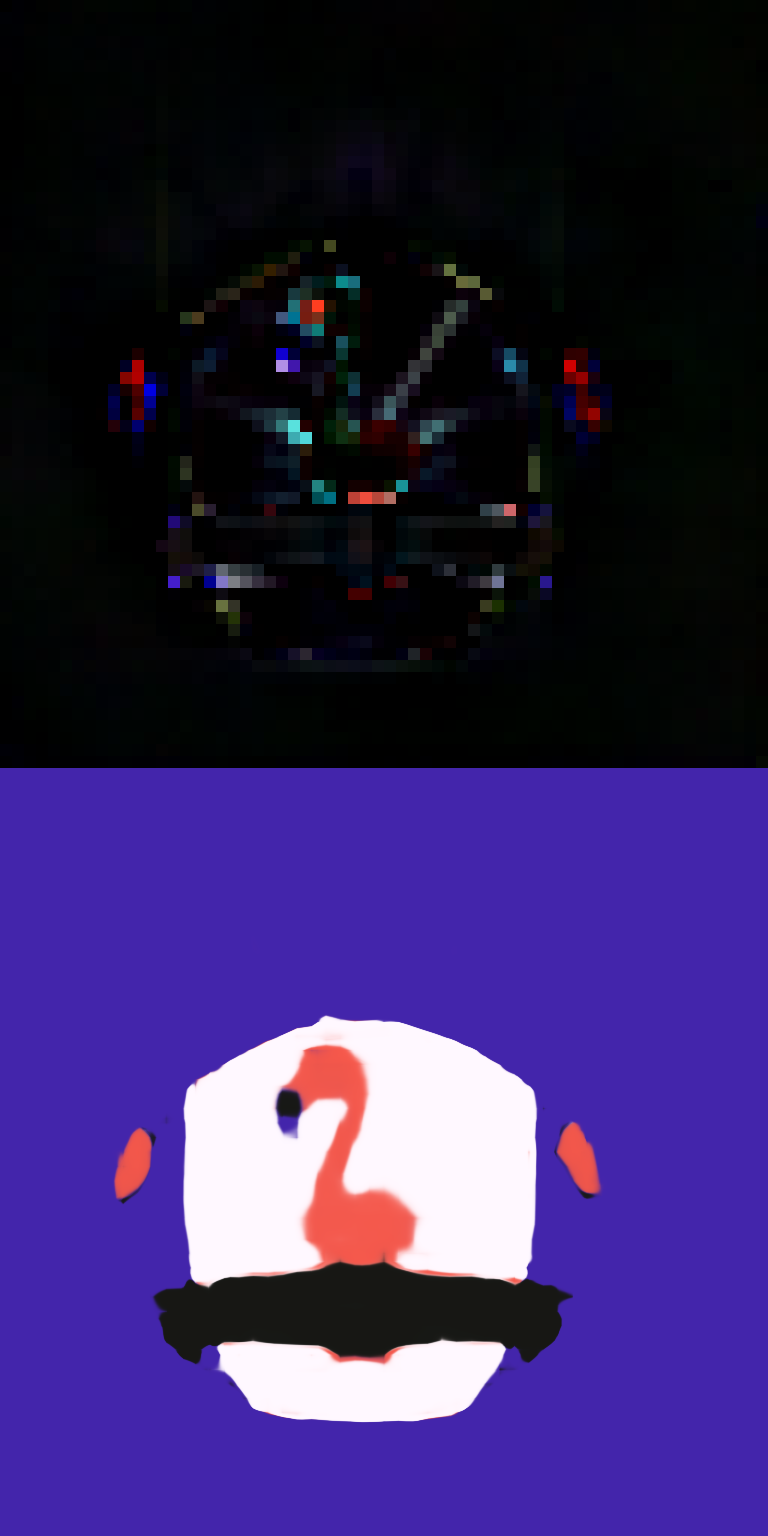} &
    \includegraphics[width=0.06\linewidth]{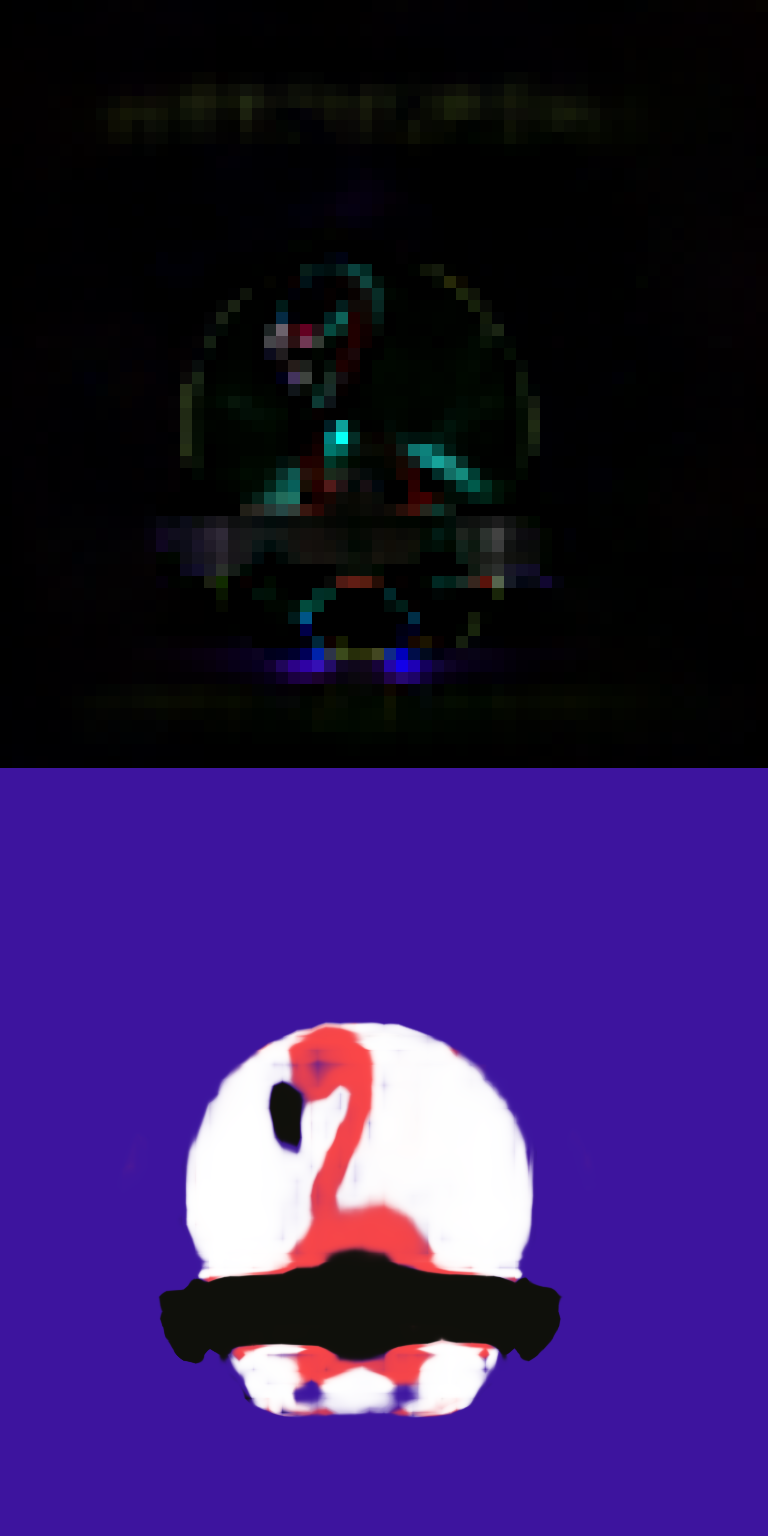} &
    \includegraphics[width=0.06\linewidth]{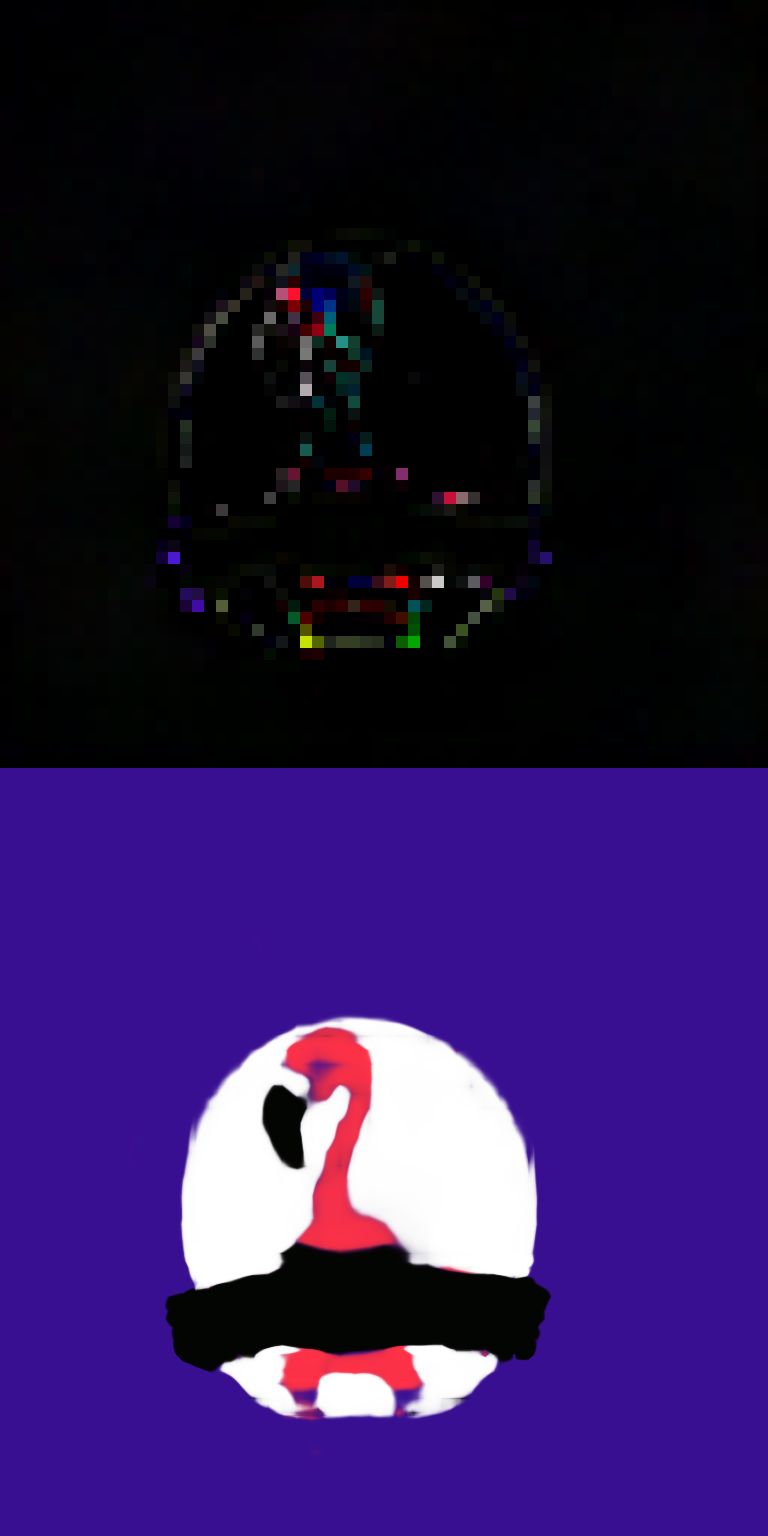} &
    \includegraphics[width=0.06\linewidth]{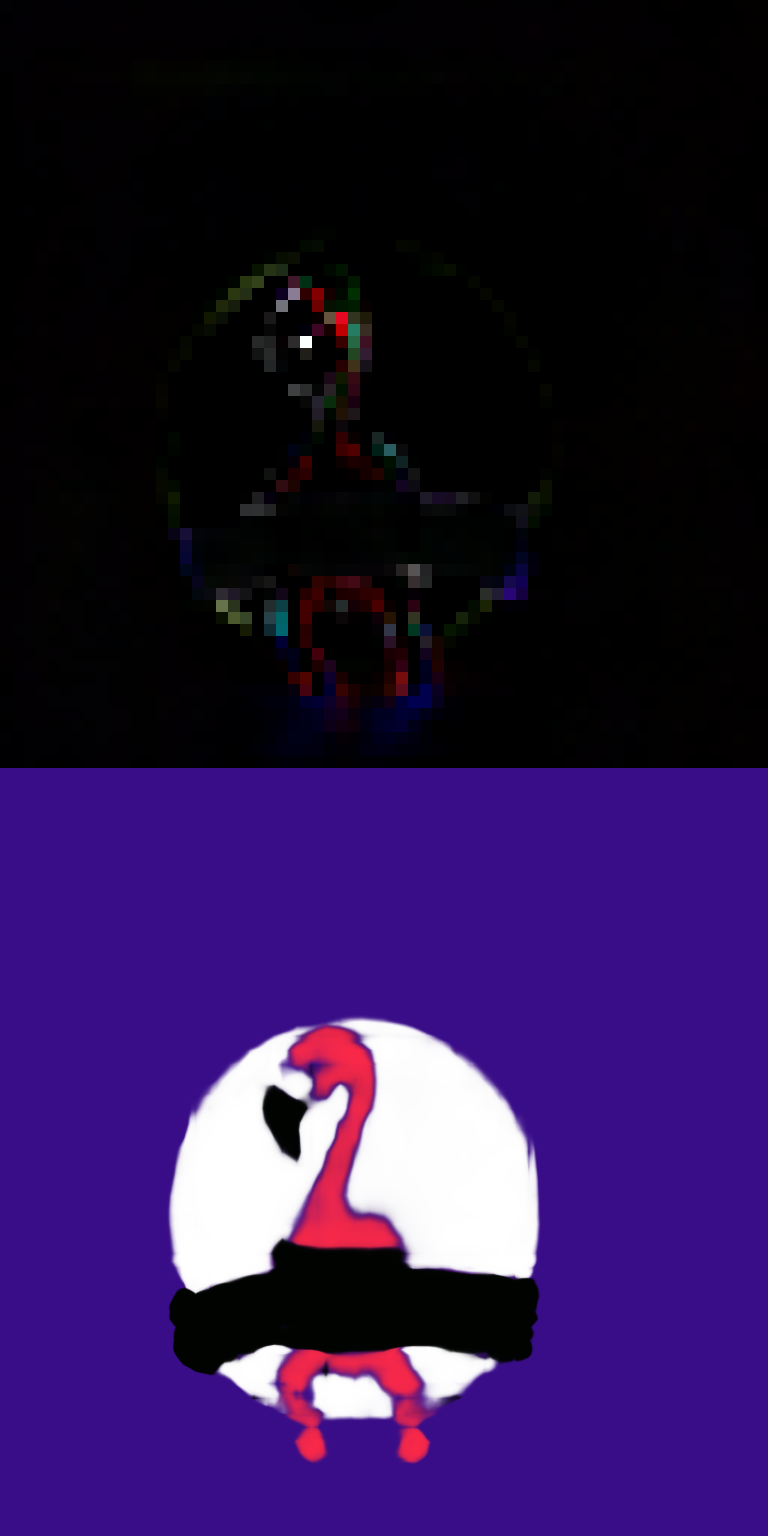} &
    \includegraphics[width=0.06\linewidth]{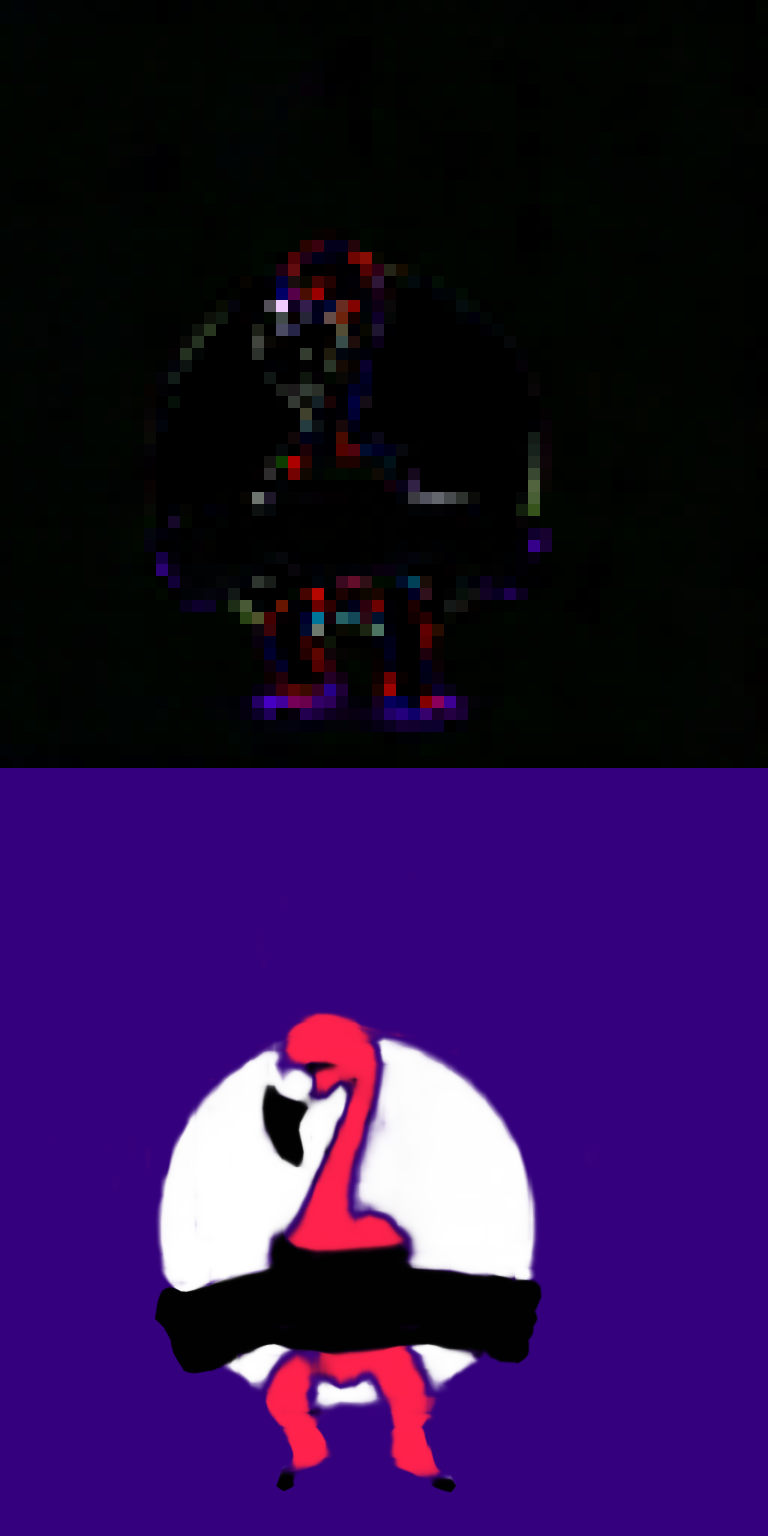} &
    \includegraphics[width=0.06\linewidth]{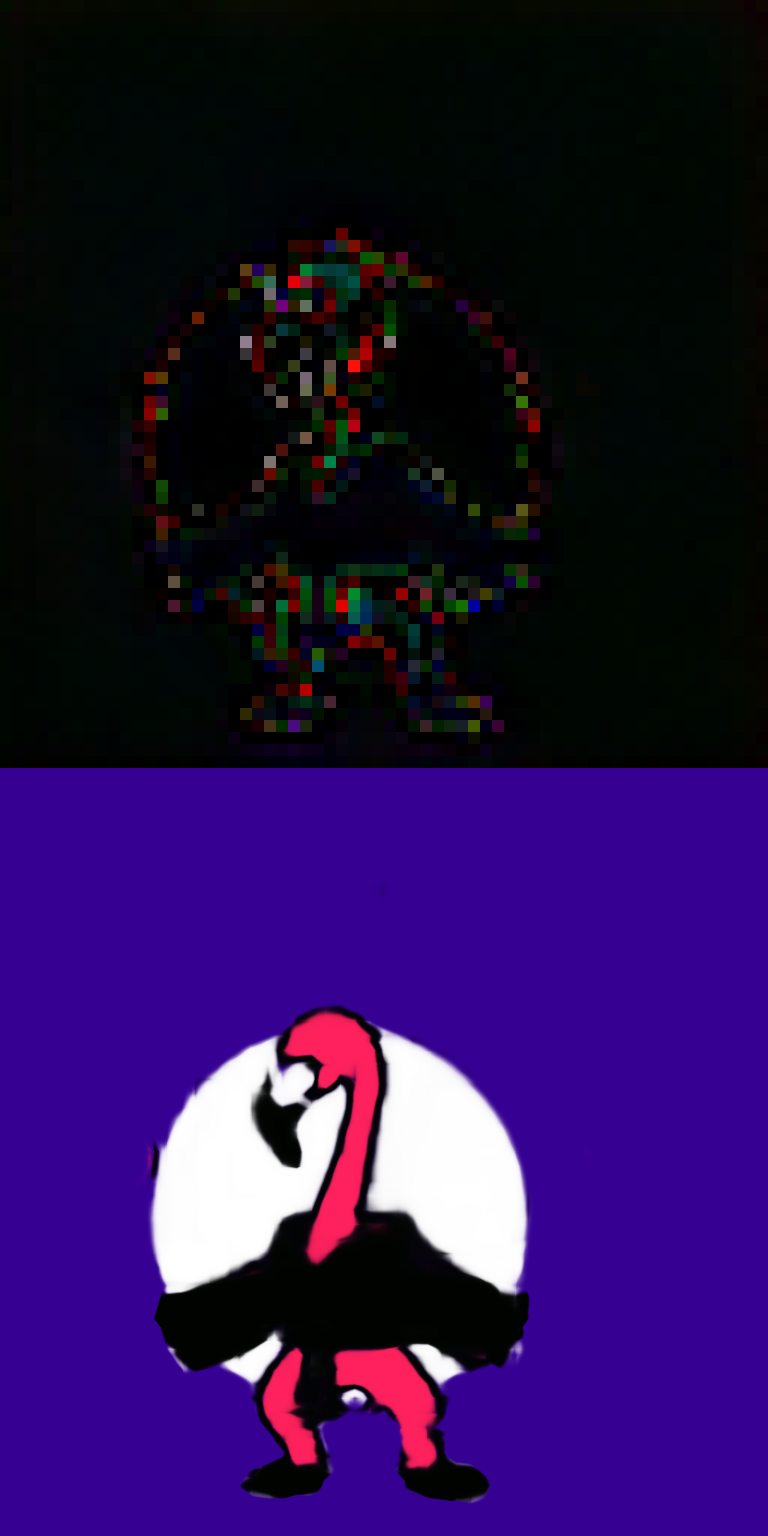} &
    \includegraphics[width=0.06\linewidth]{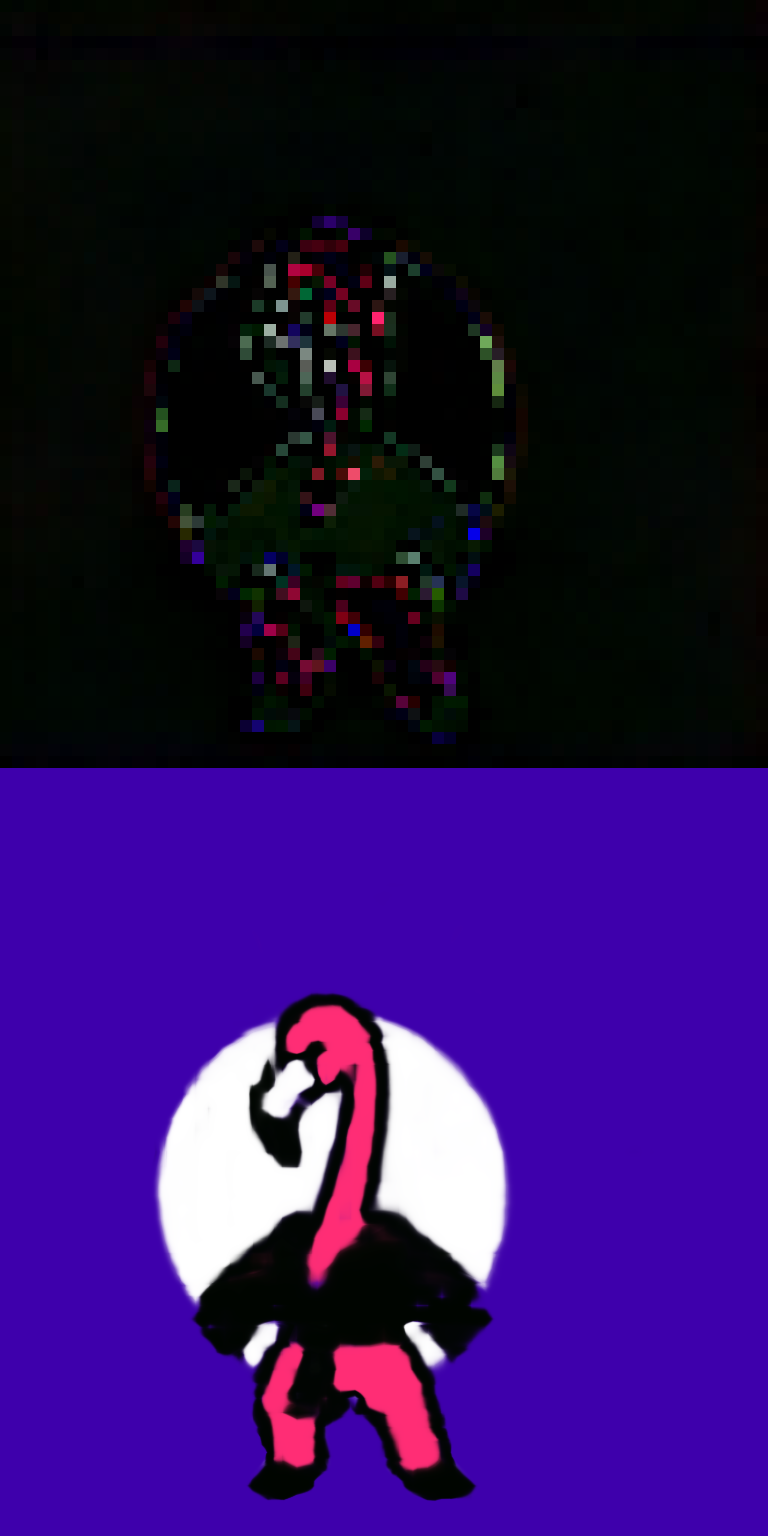} &
    \includegraphics[width=0.06\linewidth]{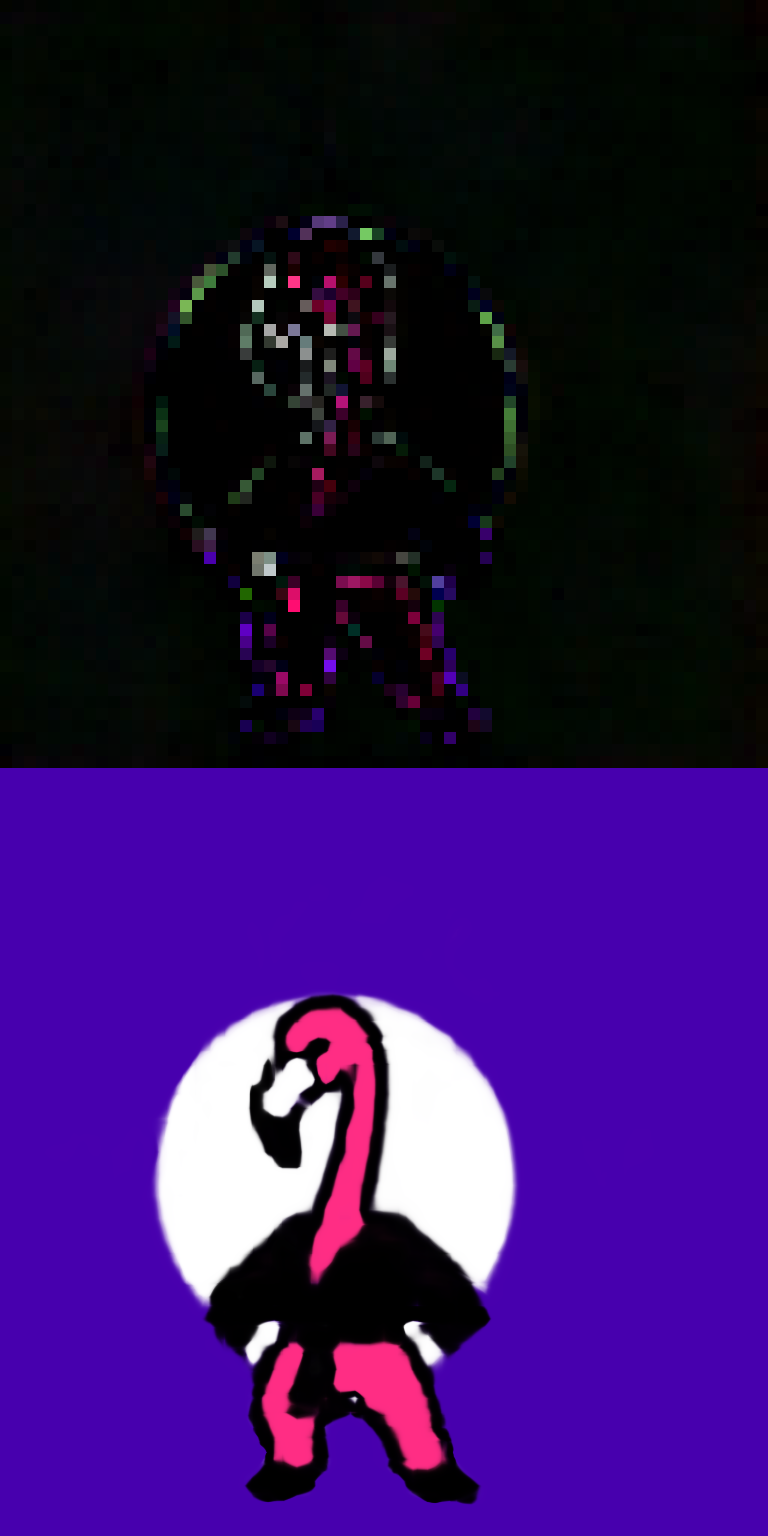} &
    \includegraphics[width=0.06\linewidth]{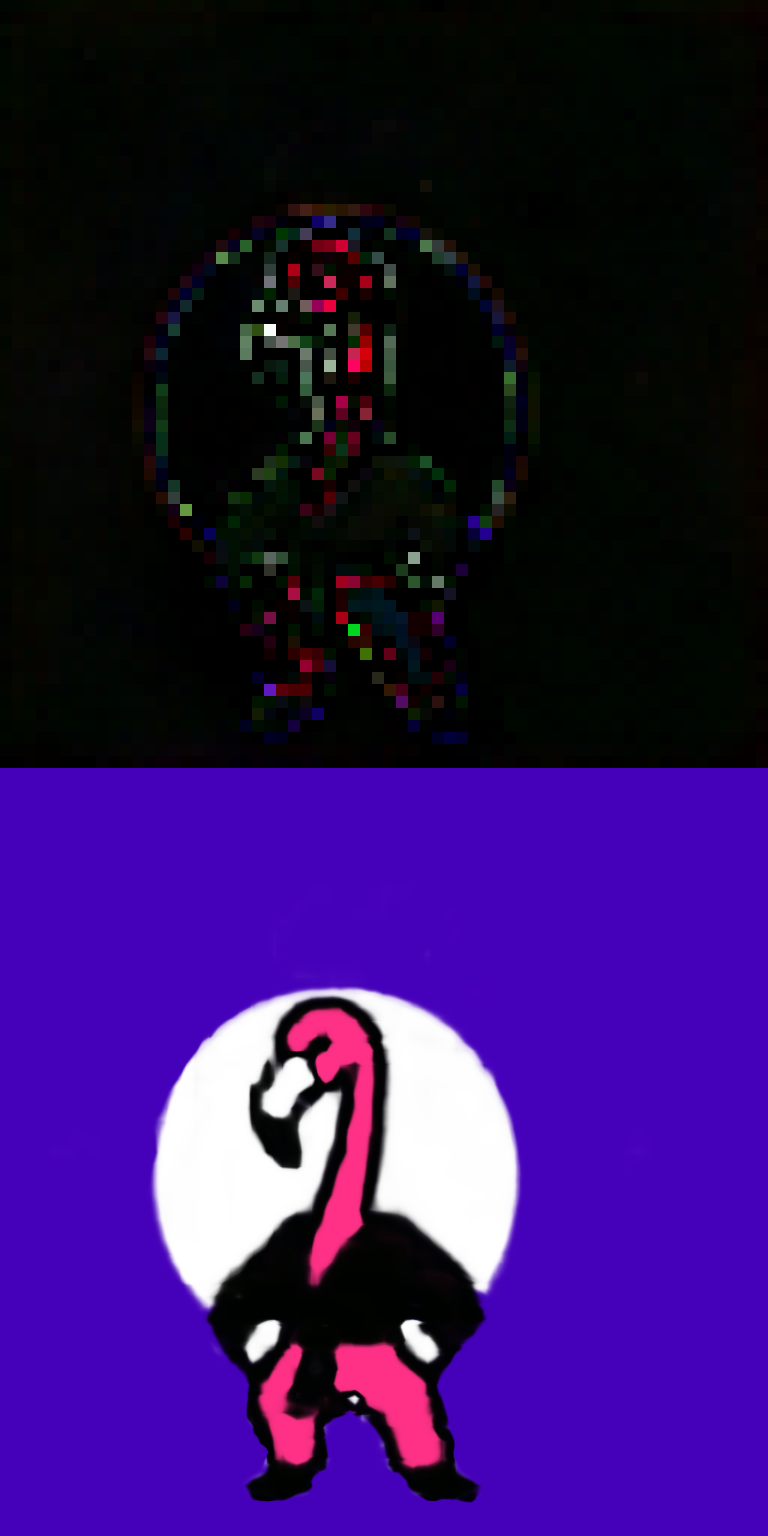} &
    \includegraphics[width=0.06\linewidth]{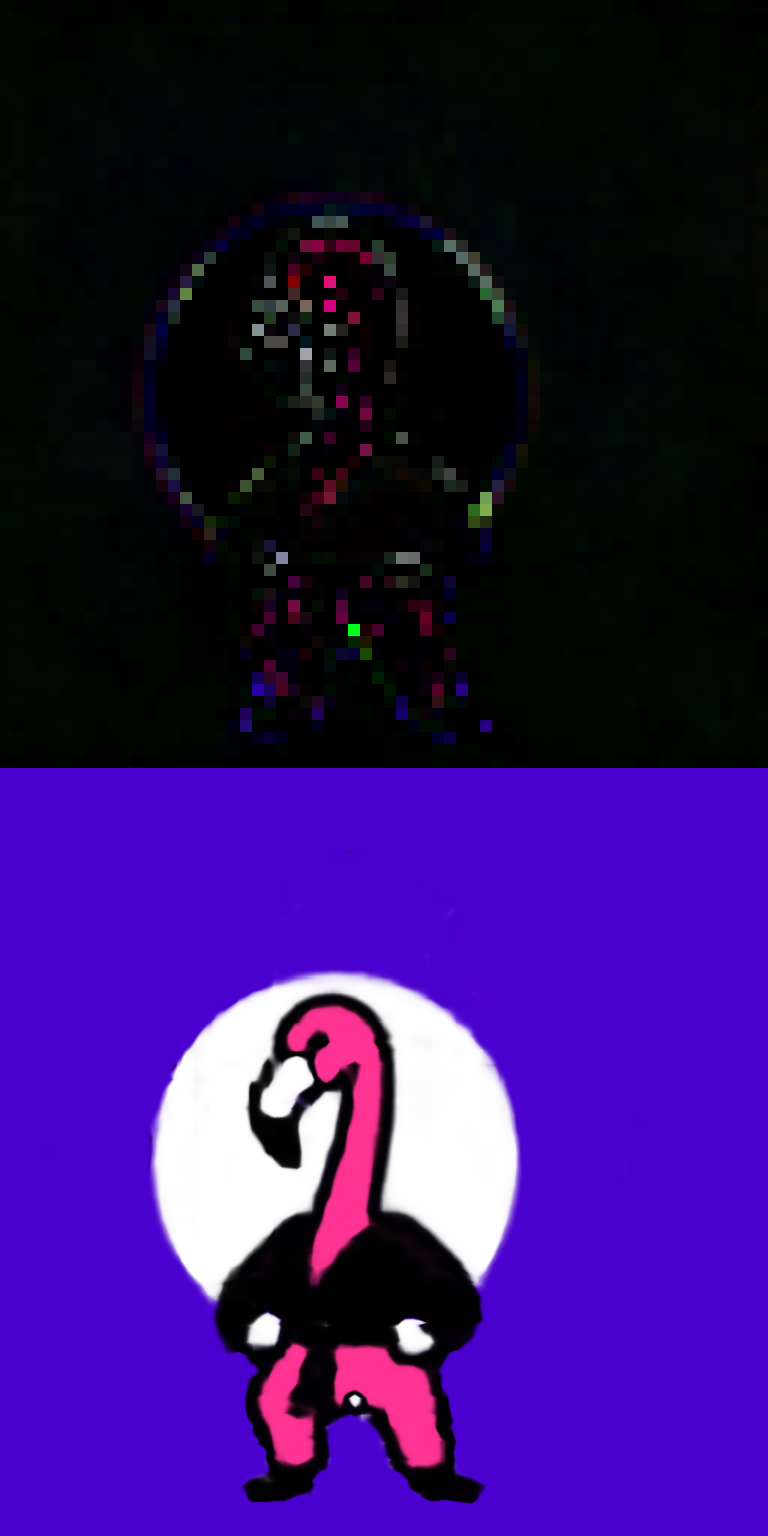} &
    \includegraphics[width=0.06\linewidth]{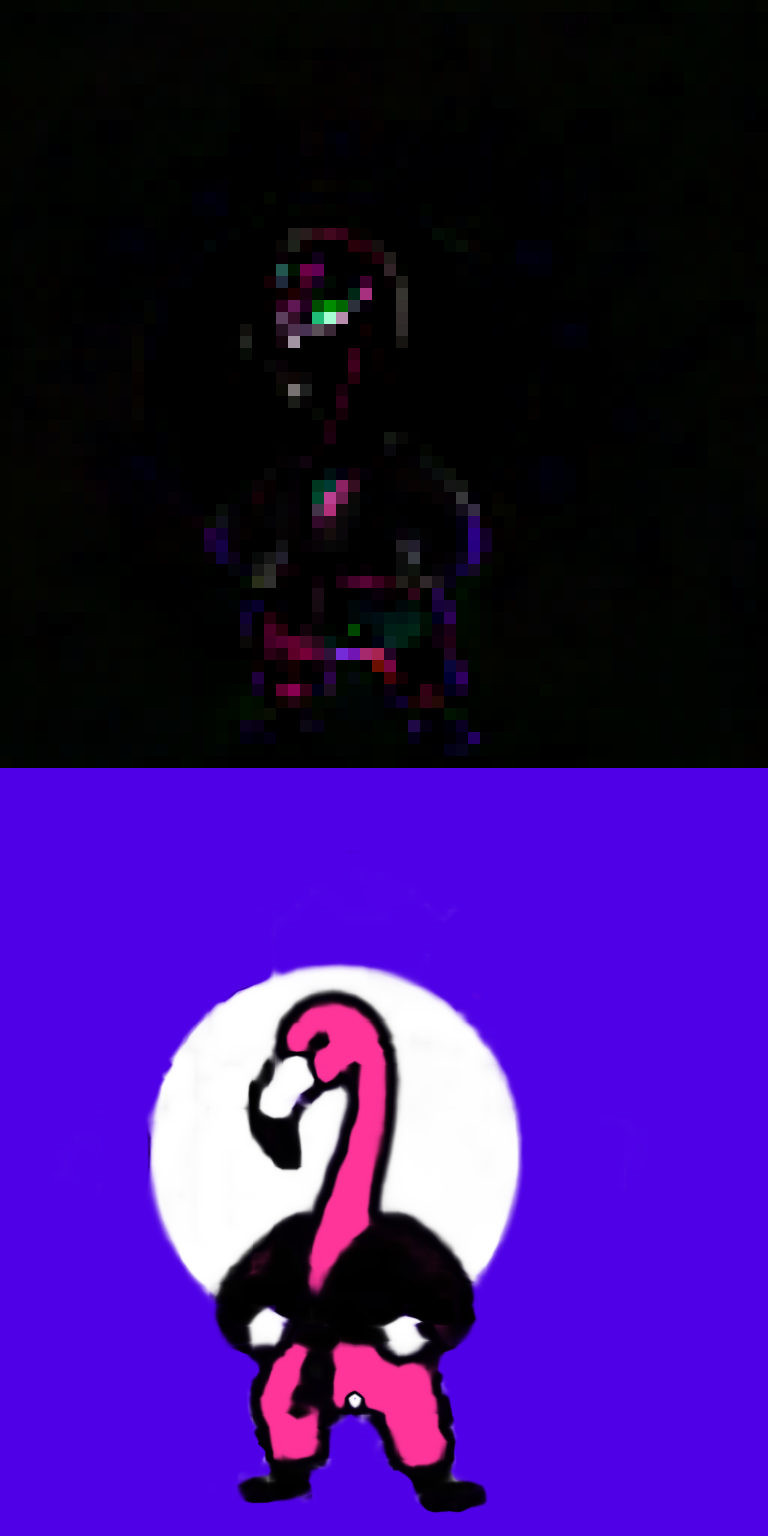} &
    \includegraphics[width=0.06\linewidth]{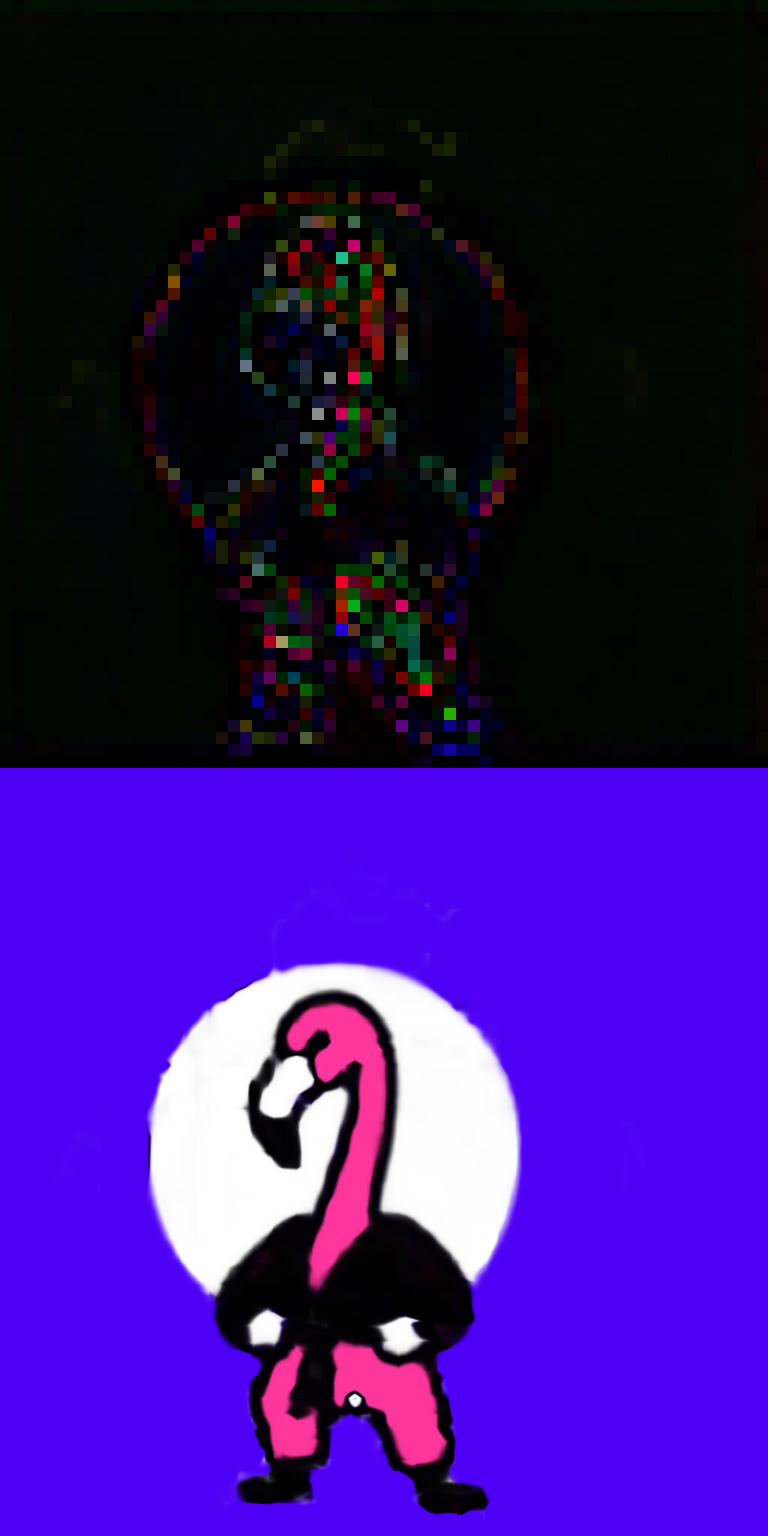} &
    \includegraphics[width=0.06\linewidth]{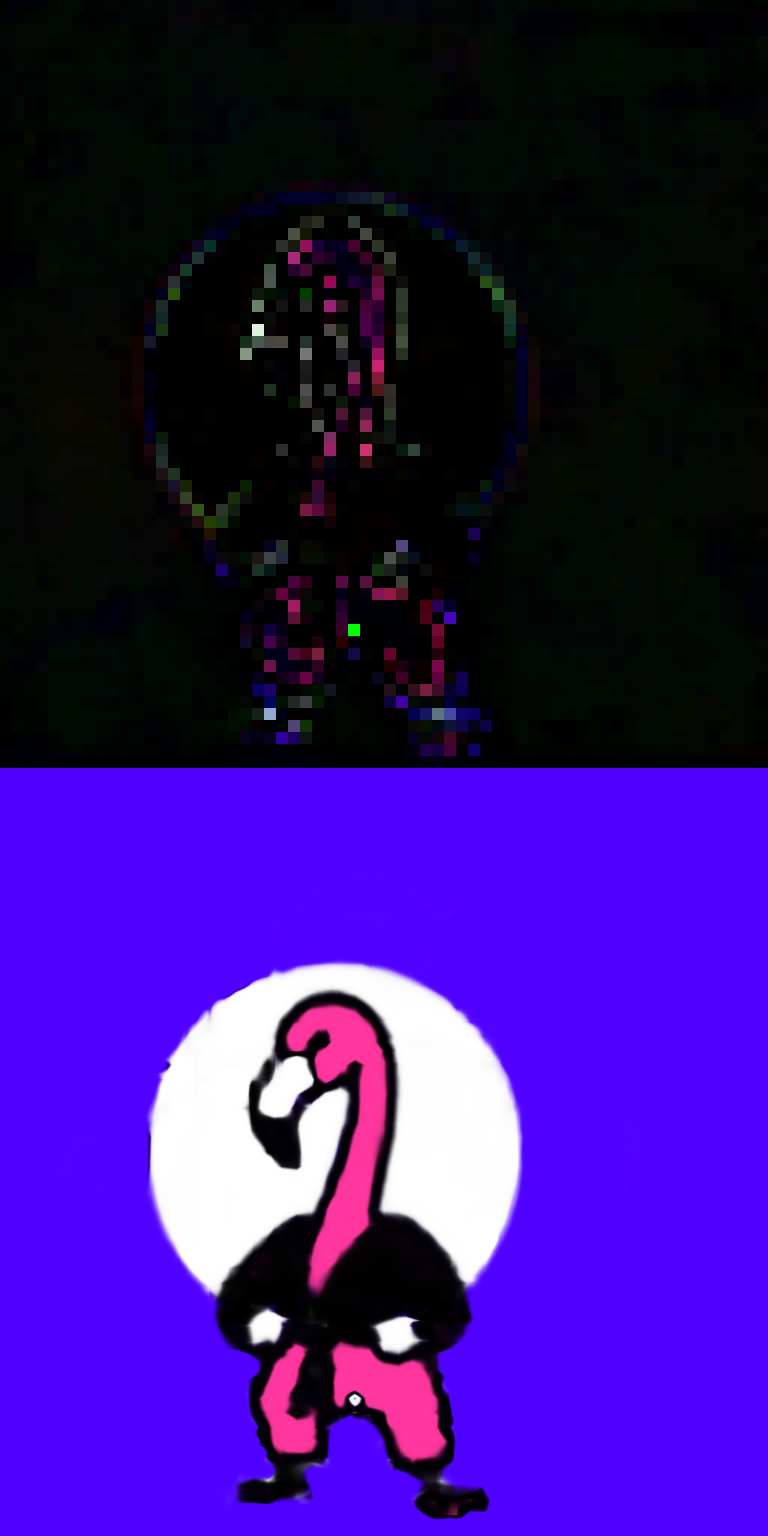} &
    \includegraphics[width=0.06\linewidth]{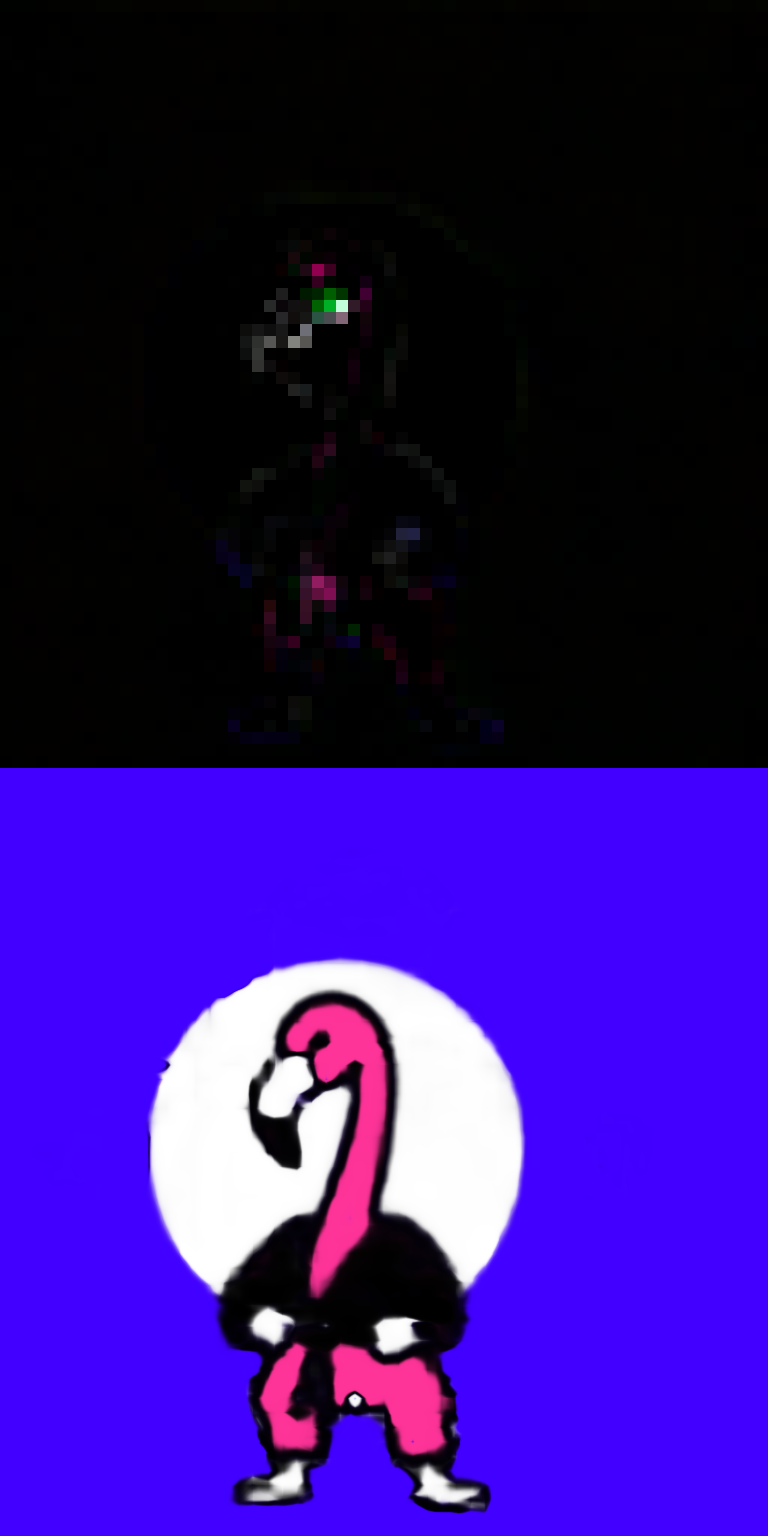} \\
    
\end{tabular}
        \vspace{-3mm}
    \caption{ We visualize the SDS gradients in rgb color space ($1^{st}$ row) and generated results ($2^{nd}$ row) during optimization for two prompts. Initialization (Init) is the leftmost column. Columns on its right show the generated result and gradients for the $i^{th}$ optimization step. The number of layers is $L=3$ here.
    \label{fig:sds_grad_viz_multi} }
    \vspace{-2mm}
\end{figure*}

\section*{C. Visualization of SDS gradients}
\label{supp:sds_gradients_visualized}
The SDS gradients provided by the frozen pre-trained diffusion model contain rich and visually interpretable signals for shape generation. In Figure \ref{fig:sds_grad_viz_single} and Figure \ref{fig:sds_grad_viz_multi}, we show the SDS gradients visualized as an RGB image ($1^{st}$ row) and the generated images ($2^{nd}$ row) across a subset of the
8000 optimization steps. In Figure \ref{fig:sds_grad_viz_single}, we initialize one layer with a Gaussian blob and show the evolution of the generated shape during the optimization. Note the ability of our representation to freely add or remove holes i.e change genus. 
Our entropy-based loss helps in producing clean boundaries. Similarly, in Figure \ref{fig:sds_grad_viz_multi}, we show these effects on multi-layer generations. We set the initialized layers to be either 2D boxes or ellipses. The choice of these two initialization procedures allows an  easier interpretation of the gradient signals. This also demonstrates that the initialization for NIVeL can be handcrafted, if desired. For the two prompts, \emph{"A grizzly bear karate master..."} and \emph{"A flamingo karate master..."}, one can observe the effect of the initialization not just on the generated shapes during the optimization but also in the SDS gradients. These two initialization show the different outcomes explored via SDS and how geometric properties of the initial shapes are preserved i.e., the presence of sharp corners with the boxes.

\begin{table}[tbp]
\renewcommand{\tabcolsep}{4.5pt}  
\centering
\begin{tabular}{lc}
\toprule
\multicolumn{2}{c}{\textbf{Hyperparameters}} \\
Guidance scale & 14.0\\
t (timestep) & $\sim \mathcal{U}(0,1)$ \\
L & 5\\
Iterations & 8000 \\
Learning rate (Color) & $5e^{-3}$ \\
Batch size & 3 \\
$\lambda$ & $1e^{-5}$ \\
$\lambda'$ & $1e^{-4}$ \\
\hline \\
\multicolumn{2}{c}{\textbf{12K Model}}  \\
F & 6 \\
Layers & 4\\
Hidden nodes & 64\\
Activation & LeakyReLU \\
Learning rate (MLP) & $1e^{-2}$ \\
\hline \\
\multicolumn{2}{c}{\textbf{1K Model}}  \\
F & 2 \\
Layers & 3\\
Hidden nodes & 32\\
Activation & LeakyReLU \\
Learning rate (MLP) & $1e^{-3}$ \\
\bottomrule
\end{tabular}
\caption{Model cards of NIVeL's variants used in our experiments. 
\vspace{-1mm}
\label{tab:model_details}}

\label{table:raster_arch}
\end{table}

\begin{figure}[t!]
\centering
\includegraphics[width=1.0\linewidth]{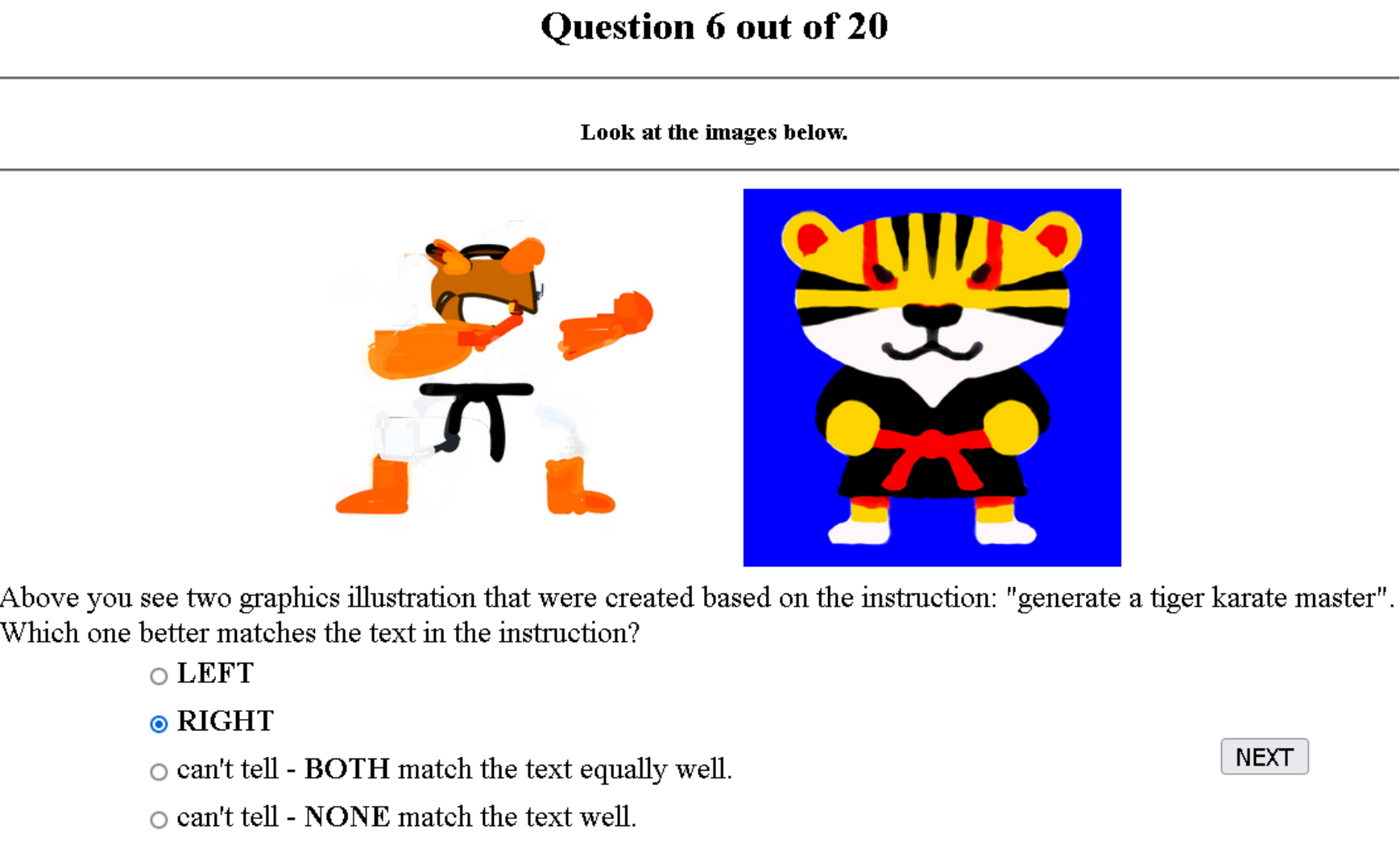}
\vspace{-7mm}
\caption{Screenshot of an example question used in our MTurk questionnaires.}
\label{fig:user_study_snapshot}
\end{figure}

\section*{D. User study details}
Figure \ref{fig:user_study_snapshot} shows a screenshot of the webpage containing an example question used in our perceptual evaluation questionnaires. 
Each questionnaire was released via the MTurk platform. It contained $10$ unique questions, each asking for one comparison between NIVel and VectorFusion.
Then these $10$ questions were repeated in the questionnaire in a random order. In these repeated questions, the order of compared illustrations was flipped. If a worker gave more than $3$ inconsistent answers for the repeated questions, then he/she was marked as ``unreliable''. Each participant was allowed to perform the questionnaire only once. The results are shown in Figure 6 of the main text.

\section*{E. List of prompts used in our experiments}
\label{supp:prompts}
Below we show all the text prompts used in our comparisons with VectorFusion (including our user study), as discussed in the main text.\\ 
\\
\emph{\footnotesize{"Line drawing of Third eye, minimal 2d line drawing, on a white background, black and white"}}\\
\emph{\footnotesize{"Line drawing of a baby penguin, minimal 2d line drawing, on a white background, black and white"}}\\
\emph{\footnotesize{"Line drawing of a ladder, minimal 2d line drawing, on a white background, black and white"}}\\
\emph{\footnotesize{"Line drawing of a crown, minimal 2d line drawing, on a white background, black and white"}}\\
\emph{\footnotesize{"Line drawing of A cat as 3D rendered in Unreal Engine, minimal 2d line drawing, on a white background, black and white"}}\\
\emph{\footnotesize{"Line drawing of Fast Food, minimal 2d line drawing, on a white background, black and white"}}\\
\emph{\footnotesize{"Line drawing of Happiness, minimal 2d line drawing, on a white background, black and white"}}\\
\emph{\footnotesize{"Line drawing of Family vacation to Walt Disney World, minimal 2d line drawing, on a white background, black and white"}}\\
\emph{\footnotesize{"Line drawing of a bottle of beer next to an ashtray with a half-smoked cigarrette, minimal 2d line drawing, on a white background, black and white"}}\\
\emph{\footnotesize{"Line drawing of A Japanese woodblock print of one cat, minimal 2d line drawing, on a white background, black and white"}}\\
\emph{\footnotesize{"Line drawing of A torii gate, minimal 2d line drawing, on a white background, black and white"}}\\
\emph{\footnotesize{"Line drawing of an elephant, minimal 2d line drawing, on a white background, black and white"}}\\
\emph{\footnotesize{"Line drawing of A spaceship flying in a starry sky, minimal 2d line drawing, on a white background, black and white"}}\\
\emph{\footnotesize{"Line drawing of Enlightenment, minimal 2d line drawing, on a white background, black and white"}}\\
\emph{\footnotesize{"Line drawing of A 3D rendering of a temple, minimal 2d line drawing, on a white background, black and white"}}\\
\emph{\footnotesize{"Line drawing of a fire-breathing dragon, minimal 2d line drawing, on a white background, black and white"}}\\
\emph{\footnotesize{"Line drawing of A realistic photograph of a cat, minimal 2d line drawing, on a white background, black and white"}}\\
\emph{\footnotesize{"Line drawing of a tree, minimal 2d line drawing, on a white background, black and white"}}\\
\emph{\footnotesize{"Line drawing of a hot air balloon with a yin-yang symbol, with the moon visible in the daytime sky, minimal 2d line drawing, on a white background, black and white"}}\\
\emph{\footnotesize{"Line drawing of A 3D wireframe model of a cat, minimal 2d line drawing, on a white background, black and white"}}\\
\emph{\footnotesize{"Line drawing of Hashtag, minimal 2d line drawing, on a white background, black and white"}}\\
\emph{\footnotesize{"Line drawing of Yeti taking a selfie, minimal 2d line drawing, on a white background, black and white"}}\\
\emph{\footnotesize{"Line drawing of A dragon-cat hybrid, minimal 2d line drawing, on a white background, black and white"}}\\
\emph{\footnotesize{"Line drawing of a tall horse next to a red car, minimal 2d line drawing, on a white background, black and white"}}\\
\emph{\footnotesize{"Line drawing of Underwater Submarine, minimal 2d line drawing, on a white background, black and white"}}\\
\emph{\footnotesize{"Line drawing of A drawing of a cat, minimal 2d line drawing, on a white background, black and white"}}\\
\emph{\footnotesize{"Line drawing of a photograph of a fiddle next to a basketball on a ping pong table, minimal 2d line drawing, on a white background, black and white"}}\\
\emph{\footnotesize{"Line drawing of Forest Temple as 3D rendered in Unreal Engine, minimal 2d line drawing, on a white background, black and white"}}\\
\emph{\footnotesize{"Line drawing of Horse eating a cupcake, minimal 2d line drawing, on a white background, black and white"}}\\
\emph{\footnotesize{"Line drawing of A realistic painting of a sailboat, minimal 2d line drawing, on a white background, black and white"}}\\
\emph{\footnotesize{"Line drawing of the Great Wall, minimal 2d line drawing, on a white background, black and white"}}\\
\emph{\footnotesize{"Line drawing of a boat, minimal 2d line drawing, on a white background, black and white"}}\\
\emph{\footnotesize{"Line drawing of A watercolor painting of a cat, minimal 2d line drawing, on a white background, black and white"}}\\
\emph{\footnotesize{"Line drawing of The space between infinity, minimal 2d line drawing, on a white background, black and white"}}\\
\emph{\footnotesize{"Line drawing of a basketball to the left of two soccer balls on a gravel driveway, minimal 2d line drawing, on a white background, black and white"}}\\
\emph{\footnotesize{"Line drawing of The Eiffel Tower, minimal 2d line drawing, on a white background, black and white"}}\\
\emph{\footnotesize{"Line drawing of A painting of a starry night sky, minimal 2d line drawing, on a white background, black and white"}}\\
\emph{\footnotesize{"Line drawing of Translation, minimal 2d line drawing, on a white background, black and white"}}\\
\emph{\footnotesize{"Line drawing of a triangle, minimal 2d line drawing, on a white background, black and white"}}\\
\emph{\footnotesize{"Line drawing of a circle, minimal 2d line drawing, on a white background, black and white"}}\\
\emph{\footnotesize{"Line drawing of a dragon breathing fire, minimal 2d line drawing, on a white background, black and white"}}\\
\emph{\footnotesize{"Line drawing of a group of squirrels rowing crew, minimal 2d line drawing, on a white background, black and white"}}\\
\emph{\footnotesize{"Line drawing of a fox and a hare tangoing together, minimal 2d line drawing, on a white background, black and white"}}\\
\emph{\footnotesize{"Line drawing of a plate piled high with chocolate chip cookies, minimal 2d line drawing, on a white background, black and white"}}\\
\emph{\footnotesize{"Line drawing of a walrus smoking a pipe, minimal 2d line drawing, on a white background, black and white"}}\\
\emph{\footnotesize{"Line drawing of a stork playing a violin, minimal 2d line drawing, on a white background, black and white"}}\\
\emph{\footnotesize{"Line drawing of a match stick on fire, minimal 2d line drawing, on a white background, black and white"}}\\
\emph{\footnotesize{"Line drawing of a friendship, minimal 2d line drawing, on a white background, black and white"}}\\
\emph{\footnotesize{"Line drawing of a banana with sun glasses, minimal 2d line drawing, on a white background, black and white"}}\\
\emph{\footnotesize{"Line drawing of a clock with dials, minimal 2d line drawing, on a white background, black and white"}}\\
\emph{\footnotesize{"Line drawing of a classic wristwatch, Minimal 2D line drawing, On a white background, black and white"}}\\
\emph{\footnotesize{"Line drawing of a vintage camera, Minimal 2D line drawing, On a white background, black and white"}}\\
\emph{\footnotesize{"Line drawing of a coffee cup and saucer, Minimal 2D line drawing, On a white background, black and white"}}\\
\emph{\footnotesize{"Line drawing of a Clown on a unicycle, Minimal 2D line drawing, On a white background, black and white"}}\\
\emph{\footnotesize{"Line drawing of a Utah teapot, Minimal 2D line drawing, On a white background, black and white"}}\\
\emph{\footnotesize{"Line drawing of a computer vision conference, Minimal 2D line drawing, On a white background, black and white"}}\\
\emph{\footnotesize{"Line drawing of a Science conference, Minimal 2D line drawing, On a white background, black and white"}}\\
\emph{\footnotesize{"Line drawing of a human hand, Minimal 2D line drawing, On a white background, black and white"}}\\
\emph{\footnotesize{"Line drawing of a still life, Minimal 2D line drawing, On a white background, black and white"}}\\
\emph{\footnotesize{"A 3D wireframe model of a cat, minimal 2D vector art, lineal color"}}\\
\emph{\footnotesize{"a bottle of beer next to an ashtray with a half-smoked cigarrette, minimal 2D vector art, lineal color"}}\\
\emph{\footnotesize{"A human hand, minimal 2D vector art, lineal color"}}\\
\emph{\footnotesize{"A dragon breathing fire, minimal 2D vector art, lineal color"}}\\
\emph{\footnotesize{"A group of squirrels rowing crew, minimal 2D vector art, lineal color"}}\\
\emph{\footnotesize{"A fox and a hare tangoing together, minimal 2D vector art, lineal color"}}\\
\emph{\footnotesize{"A plate piled high with chocolate chip cookies, minimal 2D vector art, lineal color"}}\\
\emph{\footnotesize{"A walrus smoking a pipe, minimal 2D vector art, lineal color"}}\\
\emph{\footnotesize{"A stork playing a violin, minimal 2D vector art, lineal color"}}\\
\emph{\footnotesize{"A match stick on fire, minimal 2D vector art, lineal color"}}\\
\emph{\footnotesize{"A tiger karate master, minimal 2D vector art, lineal color"}}\\
\emph{\footnotesize{"a squirrel dressed up like a victorian woman, lineal color"}}\\
\emph{\footnotesize{"A baby bunny sitting on top of a stack of pancakes, minimal 2D vector art"}}\\
\emph{\footnotesize{"A baby python sitting on top of a stack of books, minimal 2D vector art"}}\\
\emph{\footnotesize{"A vintage camera, minimal 2D vector art, lineal color"}}\\
\emph{\footnotesize{"A coffee cup and saucer, minimal 2D vector art, lineal color"}}\\
\emph{\footnotesize{"A Clown on a unicycle, minimal 2D vector art, lineal color"}}\\
\emph{\footnotesize{"A still life, minimal 2D vector art, lineal color"}}\\
\emph{\footnotesize{"A banana with sun glasses, minimal 2D vector art, lineal color"}}\\
\emph{\footnotesize{"A clock with dials, minimal 2D vector art, lineal color"}}\\
\emph{\footnotesize{"A classic wristwatch, minimal 2D vector art, lineal color"}}\\
\emph{\footnotesize{"A Utah teapot, minimal 2D vector art, lineal color"}}\\
\emph{\footnotesize{"A computer vision conference, minimal 2D vector art, lineal color"}}\\
\emph{\footnotesize{"A Science conference, minimal 2D vector art, lineal color"}}\\
\emph{\footnotesize{"A friendship, minimal 2D vector art, lineal color"}}\\
\emph{\footnotesize{"A tree, minimal 2D vector art, lineal color"}}\\
\emph{\footnotesize{"A hot air balloon with a yin-yang symbol, with the moon visible in the daytime sky, minimal 2D vector art, lineal color"}}\\
\emph{\footnotesize{"Yeti taking a selfie, minimal 2D vector art, lineal color"}}\\
\emph{\footnotesize{"A dragon-cat hybrid, minimal 2D vector art, lineal color"}}\\
\emph{\footnotesize{"A spider web, minimal 2D vector art, lineal color"}}\\
\emph{\footnotesize{"Underwater Submarine, minimal 2D vector art, lineal color"}}\\
\emph{\footnotesize{"A boiling water on a fire stove, minimal 2D vector art, lineal color"}}\\
\emph{\footnotesize{"A photograph of a fiddle next to a basketball on a ping pong table, minimal 2D vector art, lineal color"}}\\
\emph{\footnotesize{"Forest Temple as 3D rendered in Unreal Engine, minimal 2D vector art, lineal color"}}\\
\emph{\footnotesize{"Horse eating a cupcake, minimal 2D vector art, lineal color"}}\\
\emph{\footnotesize{"A sailboat, minimal 2D vector art, lineal color"}}\\
\emph{\footnotesize{"the Great Wall, minimal 2D vector art, lineal color"}}\\
\emph{\footnotesize{"A boat, minimal 2D vector art, lineal color"}}\\
\emph{\footnotesize{"A fluid simulation, minimal 2D vector art, lineal color"}}\\
\emph{\footnotesize{"The space between infinity, minimal 2D vector art, lineal color"}}\\
\emph{\footnotesize{"A basketball to the left of two soccer balls on a gravel driveway, minimal 2D vector art, lineal color"}}\\
\emph{\footnotesize{"A triangle, minimal 2D vector art, lineal color"}}\\
\emph{\footnotesize{"A circle, minimal 2D vector art, lineal color"}}\\
\emph{\footnotesize{"A Japanese woodblock print of one cat, minimal 2D vector art, lineal color"}}\\
\emph{\footnotesize{"A torii gate, minimal 2D vector art, lineal color"}}\\
\emph{\footnotesize{"An elephant, minimal 2D vector art, lineal color"}}\\
\emph{\footnotesize{"A spaceship flying in a starry sky, minimal 2D vector art, lineal color"}}\\
\emph{\footnotesize{"Enlightenment, minimal 2D vector art, lineal color"}}\\
\emph{\footnotesize{"Third eye, minimal 2D vector art, lineal color"}}\\
\emph{\footnotesize{"A baby penguin, minimal 2D vector art, lineal color"}}\\
\emph{\footnotesize{"A ladder, minimal 2D vector art, lineal color"}}\\
\emph{\footnotesize{"A crown, minimal 2D vector art, lineal color"}}\\
\emph{\footnotesize{"A cat as 3D rendered in Unreal Engine, minimal 2D vector art, lineal color"}}\\
\emph{\footnotesize{"Fast Food, minimal 2D vector art, lineal color"}}\\
\emph{\footnotesize{"Happiness, minimal 2D vector art, lineal color"}}\\
\emph{\footnotesize{"Family vacation to Walt Disney World, minimal 2D vector art, lineal color"}}\\
\emph{\footnotesize{"Line drawing of A 3D wireframe model of a Monkey,  2d vector art line drawing, black and white"}}\\
\emph{\footnotesize{"Line drawing of a bottle of beer next to an ashtray with a half-smoked cigarrette, 2d vector art line drawing, black and white"}}\\
\emph{\footnotesize{"Line drawing of A human hand showing the peace sign, 2d vector art line drawing, black and white"}}\\
\emph{\footnotesize{"Line drawing of A plate piled high with chocolate chip cookies, 2d vector art line drawing, black and white"}}\\
\emph{\footnotesize{"Line drawing of A flamingo playing a cello, 2d vector art line drawing, black and white"}}\\
\emph{\footnotesize{"Line drawing of A leopard karate master, 2d vector art line drawing, black and white"}}\\
\emph{\footnotesize{"Line drawing of a Lizard dressed up like a victorian woman, lineal color"}}\\
\emph{\footnotesize{"Line drawing of A vintage telephone, 2d vector art line drawing, black and white"}}\\
\emph{\footnotesize{"Line drawing of An orange with sun glasses, 2d vector art line drawing, black and white"}}\\
\emph{\footnotesize{"Line drawing of A duck taking a selfie, 2d vector art line drawing, black and white"}}\\
\emph{\footnotesize{"Line drawing of A dragon-corgi hybrid, 2d vector art line drawing, black and white"}}\\
\emph{\footnotesize{"Line drawing of Horse eating a hotdog, 2d vector art line drawing, black and white"}}\\
\emph{\footnotesize{"Line drawing of An elephant surfing, 2d vector art line drawing, black and white"}}\\
\emph{\footnotesize{"Line drawing of A plate of healty fast Food, 2d vector art line drawing, black and white"}}\\
\emph{\footnotesize{"A 3D wireframe model of a Monkey, minimal 2D vector art, lineal color"}}\\
\emph{\footnotesize{"a bottle of beer next to an ashtray with a half-smoked cigarrette,  lineal color"}}\\
\emph{\footnotesize{"A human hand showing the peace sign,  minimal 2D vector art, lineal color"}}\\
\emph{\footnotesize{"A dragon breathing fire,  minimal 2D vector art, lineal color"}}\\
\emph{\footnotesize{"A group of squirrels rowing crew,  minimal 2D vector art, lineal color"}}\\
\emph{\footnotesize{"A fox and a hare tangoing together,  minimal 2D vector art, lineal color"}}\\
\emph{\footnotesize{"A plate piled high with chocolate chip cookies,  minimal 2D vector art, lineal color"}}\\
\emph{\footnotesize{"A walrus smoking a pipe,  minimal 2D vector art, lineal color"}}\\
\emph{\footnotesize{"A flamingo playing a cello,  minimal 2D vector art, lineal color"}}\\
\emph{\footnotesize{"A match stick on fire,  minimal 2D vector art, lineal color"}}\\
\emph{\footnotesize{"A leopard karate master,  minimal 2D vector art, lineal color"}}\\
\emph{\footnotesize{"a Lizard dressed up like a victorian woman, minimal 2D vector art, lineal color"}}\\
\emph{\footnotesize{"A baby python sitting on top of a stack of books, minimal 2D vector art, lineal color"}}\\
\emph{\footnotesize{"A vintage telephone,  minimal 2D vector art, lineal color"}}\\
\emph{\footnotesize{"A coffee cup and saucer,  minimal 2D vector art, lineal color"}}\\
\emph{\footnotesize{"A Clown juggling balls,  minimal 2D vector art, lineal color"}}\\
\emph{\footnotesize{"A still life painting,  minimal 2D vector art, lineal color"}}\\
\emph{\footnotesize{"An orange with sun glasses,  minimal 2D vector art, lineal color"}}\\
\emph{\footnotesize{"A symbol of friendship,  minimal 2D vector art, lineal color"}}\\
\emph{\footnotesize{"A banyan tree,  minimal 2D vector art, lineal color"}}\\
\emph{\footnotesize{"A duck taking a selfie,  minimal 2D vector art, lineal color"}}\\
\emph{\footnotesize{"A dragon-corgi hybrid,  minimal 2D vector art, lineal color"}}\\
\emph{\footnotesize{"A spider web,  minimal 2D vector art, lineal color"}}\\
\emph{\footnotesize{"A Temple as 3D rendered in Unreal Engine,  minimal 2D vector art, lineal color"}}\\
\emph{\footnotesize{"Horse eating a hotdog,  minimal 2D vector art, lineal color"}}\\
\emph{\footnotesize{"A luxury boat,  minimal 2D vector art, lineal color"}}\\
\emph{\footnotesize{"A fluid simulation,  minimal 2D vector art, lineal color"}}\\
\emph{\footnotesize{"The space between infinity,  minimal 2D vector art, lineal color"}}\\
\emph{\footnotesize{"A torii gate,  minimal 2D vector art, lineal color"}}\\
\emph{\footnotesize{"An elephant surfing,  minimal 2D vector art, lineal color"}}\\
\emph{\footnotesize{"A spaceship flying in a starry sky,  minimal 2D vector art, lineal color"}}\\
\emph{\footnotesize{"Enlightenment,  minimal 2D vector art, lineal color"}}\\
\emph{\footnotesize{"Third eye,  minimal 2D vector art, lineal color"}}\\
\emph{\footnotesize{"A baby penguin and a polar bear taking an impossible selfie,  minimal 2D vector art, lineal color"}}\\
\emph{\footnotesize{"A plate of healty fast Food,  minimal 2D vector art, lineal color"}}

\end{document}